\documentclass[11pt]{article}
\usepackage{booktabs}       
\usepackage{amsfonts}       
\usepackage{amsmath}
\usepackage{graphicx}
\usepackage{amsthm}
\usepackage{amssymb}
\usepackage{multirow}
\usepackage{makecell}
\usepackage[vlined,linesnumbered,ruled,resetcount]{algorithm2e}
\usepackage[round]{natbib}
\usepackage[colorlinks,linkcolor=magenta,filecolor=blue,citecolor=blue,urlcolor=blue]{hyperref}%
\usepackage[top=1in, left=1in, right=1in, bottom=1in]{geometry}
\usepackage{subfigure}
\usepackage{amssymb}
\usepackage{pifont}
\usepackage{mathtools}
\usepackage{stmaryrd}
\usepackage[T1]{fontenc}

\begin{document}

\title{Fast ABC-Boost: A Unified Framework for Selecting the Base Class in Multi-Class Classification}

\author{\textbf{Ping Li} and \textbf{Weijie Zhao}\\
Cognitive Computing Lab\\
Baidu Research\\
10900 NE 8th St. Bellevue, WA 98004, USA\\
\texttt{\{pingli98, zhaoweijie12\}@gmail.com}
}

\date{}

\maketitle              
\begin{abstract}

\noindent\footnote{This line of work has gone through a long history of development. The original idea of using the ``worst class'' as the base class was initially proposed in 2008 in~\cite{li2008adaptive} but it was not officially published. Instead, a computationally expensive exhaustive ``best class'' strategy was presented in ICML'09~\citep{Proc:ABC_ICML09}. The idea of introducing the ``gap'' in the search process for the bas class was proposed in~\cite{li2010fast}, which was not formally published either. Some of the original ideas were also used in the 2010 Yahoo! Learning to Rank Grand Challenge.  In this work, we introduce several new ideas integrated with existing ones to achieve ``fast ABC-Boost'' in a unified robust framework.}The work of~\citet{Proc:ABC_ICML09} in ICML'09 showed that the derivatives of the classical multi-class logistic regression loss function could be re-written in terms of a pre-chosen ``base class'' and \citet{Proc:ABC_ICML09} applied the new derivatives in the popular boosting framework. In order to make use of the new derivatives, one must have a strategy to identify/choose the base class at each boosting iteration. The idea of ``adaptive base class boost'' (ABC-Boost) initially published in ICML'09, adopted a computationally expensive ``exhaustive search'' strategy for the base class at each  iteration. It has been well demonstrated that ABC-Boost, when integrated with trees,  can achieve substantial improvements in many multi-class classification tasks. Furthermore, \cite{Proc:ABC_UAI10} in UAI'10 derived the explicit second-order tree split gain formula which typically improved the classification accuracy considerably, compared with using only the fist-order information for tree-splitting, for both multi-class and binary-class classification tasks. Nevertheless, deploying ABC-Boost as published in~\citet{Proc:ABC_ICML09,Proc:ABC_UAI10} has encountered severe computational difficulties due to its high computational cost of the exhaustive search strategy for the base class.

\vspace{0.1in}

\noindent In this paper, we develop a unified framework for effectively selecting the base class by introducing a series of ideas to improve the computational efficiency of ABC-Boost. The first (and also the simplest) proposal is the ``worst class'' strategy. That is, at each boosting iteration, we choose the class which is the ``worst'' in some criterion (e.g., the largest training loss) for the   next iteration. This strategy often works well but the accuracy is typically slightly inferior to that of the ``exhaustive search'' strategy. More importantly, the ``worst class'' strategy can be in some cases numerical unstable and may lead to catastrophic failures.

\vspace{0.1in}

\noindent Our unified framework has several additional parameters $(s,g,w)$. At each boosting iteration, we only search for the ``$s$-worst classes'' (instead of all classes) to determine the base class. When $s=1$ this becomes the ``worst class'' strategy, and when $s=K$ (where $K$ is the number of classes), we recover the ``exhaustive search'' strategy.  We also allow a ``gap'' $g$ when conducting the search. That is, we only search for the base class at every $g+1$ iterations. We furthermore  allow a ``warm up'' stage by only starting the search after $w$ boosting iterations. The parameters $s$, $g$, $w$, can be viewed as tunable parameters and certain combinations of $(s,g,w)$ may even lead to better test accuracy than the ``exhaustive search'' strategy. Overall, our proposed framework provides a robust and reliable scheme for implementing ABC-Boost in practice.

\end{abstract}

\newpage\clearpage

\section{Introduction}
Boosting algorithms \citep{Article:Schapire_ML90,Article:Freund_95,Article:Freund_JCSS97,Article:Bartlett_AS98,Article:Schapire_ML99,Article:FHT_AS00,Article:Friedman_AS01} have become very successful in machine learning theory and applications. It is also the standard practice to integrate boosting with trees~\citep{Book:Friedman_83} to produce accurate and robust prediction results. In the past decade or longer, multiple practical developments have enhanced the performance as well as the efficiency of boosted tree algorithms, including
\begin{itemize}
    \item (i) the explicit (and robust) formula for tree-split criterion using the second-order gain information~\citep{Proc:ABC_UAI10} (i.e., the so-called ``Robust LogitBoost''), which typically improves the accuracy, compared to the implementation based on the criterion of using only the first-order gain information~\citep{Article:Friedman_AS01};

\item (ii) the adaptive binning strategy developed in~\cite{Proc:McRank_NIPS07} which effectively transformed general features to integer values and substantially simplified the implementation and improved the efficiency of trees as well;

\item (iii) the ``adaptive base class boost'' (ABC-Boost) scheme~\citep{Proc:ABC_ICML09,Proc:ABC_UAI10} for multi-class classification by re-writing the derivatives of the classical multi-class logistic regression loss function, by enforcing the ``sum-to-zero'' constraint. ABC-Boost often improves the accuracy of multi-class classification tasks, in many cases substantially so.
\end{itemize}
The above developments were  summarized in a recent paper on trees~\citep{fan2020classification}. Readers are also referred to some interesting discussions in 2010~\url{https://hunch.net/?p=1467}. Since 2011, the author Ping Li had taught materials on ``ABC-Boost'' and ``Robust LogitBoost'' at Cornell University and Rutgers University; see for example the following lecture notes and tutorial\\~\url{http://statistics.rutgers.edu/home/pingli/STSCI6520/Lecture/ABC-LogitBoost.pdf}.\\
\url{http://www.stat.rutgers.edu/home/pingli/doc/PingLiTutorial.pdf} (pages 15--77).

\subsection{The Exhaustive Search (or the ``Best Class'') Strategy in ABC-Boost}

In ABC-Boost (adaptive base class boost) for ($K$-class) multi-class classification, a crucial step is to select one from $K$ classes as the base class, at every boosting iteration. As presented,  both ``ABC-MART''~\citep{Proc:ABC_ICML09} and ``ABC-RobustLogitBoost''~\citep{Proc:ABC_UAI10} adopted the ``exhaustive search'' strategy. That is, at each boosting iteration, every class will be tested as a candidate for the base class, and the ``best class'' (for example, the choice which minimizes the training loss) is selected as the base class for the \textbf{current iteration}. Obviously, this strategy is  computationally very expensive, especially when $K$, the number of classes, is large.  Assume $M$ boosting iterations. While the computational cost of the standard multi-class boosting would be $O(KM)$, the cost becomes $O(K(K-1)M)$ for ABC-Boost with the exhaustive search strategy.

\subsection{New Strategies for Fast ABC-Boost}

It is the focus of this paper to find more effective strategies to achieve ``fast ABC-Boost'', compared to the ``exhaustive search'' method~\citep{Proc:ABC_ICML09,Proc:ABC_UAI10}. The first idea is the ``worst class'' strategy, which often works well but not always. This idea has also been made public through technical reports, lectures, and seminar talks, e.g.,~\cite{li2008adaptive}, but it was not formally published. Assume the total number of boosting iterations is $M$. Then the computational cost of the ``worst class'' strategy would be $((K-1)M)$, which is actually more efficient than $O(KM)$, the cost of the classical boosting methods~\citep{Article:FHT_AS00,Article:Friedman_AS01}.

Let's elaborate more on the ``worst class'' strategy. That is, at each iteration, we choose the ``worst class'' according to some measure, for example, the class which has the largest training loss, for the \textbf{next iteration}. Needless to say, this ``worst class'' strategy is very efficient, actually more efficient than the classical multi-class boosting because one less class would need to be trained. The rational behind this strategy is also intuitive, to let the next boosting iteration focus on the most challenge task. We will show that the ``worst class'' strategy works well in general but the accuracy is usually somewhat worse than that of the ``exhaustive search'' strategy. In some datasets (and for some parameters), we even observe ``catastrophic failures'' when using the ``worst class'' method. This motivates us to develop better strategies to search for the base class.

Instead of simply using the ``worst class'' as the base class for the next iteration, we propose to search for the ``$s$-worst classes'', where $1\leq s\leq K$. When $s=1$, this becomes the ``worst class'' strategy, and when $s=K$, it becomes the ``exhaustive search''. In our experiments, we find that typically $s$ does not need to be large. And, of course, as a tuning parameter,  some choice of $s$ may even lead to better test accuracy than using $s=K$. With the ``$s$-worst classes'' strategy, the computational cost becomes $O(s(K-1)M)$.

Our next idea is to introduce a ``gap'' parameter $g$. That is, we only conduct the search (with parameter $s$) at every $g+1$ iterations. Here $g=0$  means that there is no gap, i.e., the search is conducted at every iteration. This idea appeared in an unpublished technical report~\citep{li2010fast}. The cost of ABC-Boost with parameters $g$ and $s$ becomes $O\left(s(K-1)\frac{1}{g+1}M + (K-1)\frac{g}{g+1}M\right)$.

In addition to the two parameters $s$ and $g$, we also allow a ``warm-up'' parameter $w$. That is, we only start using ABC-Boost after $w$ iterations. With parameters $w$, $g$, and $w$, the computational cost becomes $O\left(Kw+s(K-1)\frac{1}{g+1}(M-w) + (K-1)\frac{g}{g+1}(M-w)\right)$.

All these efforts aim at reducing the cost of searching for the base class. Suppose we choose $g\approx K$ and assume $K\approx K-1$, the computational cost essentially becomes $O\left(s(M-w)+KM\right)$, which can be fairly close to  $O(KM)$ the original cost of classical multi-class boosting.

Here we would like to emphasize that, although we have introduced additional parameters $(s, g, w)$, the test accuracy is in general not sensitive to these parameters. In fact, the ``worst class'' strategy, which is the most efficient, already works pretty well in most cases. In a sense, those parameters  $(s, g, w)$ are proposed to improve the robustness of the ``worst class'' strategy without increasing much the computational cost.

\section{LogitBoost, MART, and Robust LogitBoost}

We denote a training dataset by $\{y_i,\mathbf{x}_i\}_{i=1}^N$, where $N$ is the number of training samples, $\mathbf{x}_i$ is the $i$-th feature vector, and  $y_i \in \{0, 1, 2, ..., K-1\}$ is the $i$-th class label, where $K\geq 3$ in multi-class classification. Both LogitBoost~\citep{Article:FHT_AS00} and MART (multiple additive regression trees)~\citep{Article:Friedman_AS01} algorithms can be viewed as generalizations to logistic regression, which assumes the class probabilities $p_{i,k}$  to be
\begin{align}\label{eqn_logit}
p_{i,k} = \mathbf{Pr}\left(y_i = k|\mathbf{x}_i\right) = \frac{e^{F_{i,k}(\mathbf{x_i})}}{\sum_{s=0}^{K-1} e^{F_{i,s}(\mathbf{x_i})}},\hspace{0.2in} i = 1, 2, ..., N,
\end{align}
While the traditional logistic regression assumes $F_{i,k}(\mathbf{x}_i) = \beta^\text{T}\mathbf{x}_i$, LogitBoost and MART adopt the flexible ``additive model'',  which is a function of $M$ terms:
\begin{align}\label{eqn_F_M}
F^{(M)}(\mathbf{x}) = \sum_{m=1}^M \rho_m h(\mathbf{x};\mathbf{a}_m),
\end{align}
where  $h(\mathbf{x};\mathbf{a}_m)$, the base learner, is typically a regression tree. The parameters, $\rho_m$ and $\mathbf{a}_m$, are learned from the data, by maximum likelihood, which is equivalent to minimizing the {\em negative log-likelihood loss}
\begin{align}\label{eqn_loss}
L = \sum_{i=1}^N L_i, \hspace{0.4in} L_i = - \sum_{k=0}^{K-1}r_{i,k}  \log p_{i,k}
\end{align}
where $r_{i,k} = 1$ if $y_i = k$ and $r_{i,k} = 0$ otherwise. Note that because the class probabilities have to sum to one, there are basically only $K-1$ degrees of freedom. For identifiability, the ``sum-to-zero'' constraint, i.e., $\sum_{k=0}^{K-1}F_{i,k} = 0$,  is routinely assumed~\citep{Article:FHT_AS00,Article:Friedman_AS01,Article:Lee_JASA04,Article:Tewari_JMLR07,Article:Zou_AOAS08,Article:Zhu_Adaboost09}.

\subsection{The Original LogitBoost Algorithm}

\begin{algorithm}[h]{\small
$r_{i,k} = 1$, if $y_{i} = k$, $r_{i,k} =0$ otherwise.\\
$F_{i,k} = 0$,\ \  $p_{i,k} = \frac{1}{K}$, \ \ \ $k = 0$ to  $K-1$, \ $i = 1$ to $N$ \\
For $m=1$ to $M$ Do\\
\hspace{0.2in}    For $k=0$ to $K-1$, Do\\
\hspace{0.4in}    Compute $w_{i,k} = p_{i,k}\left(1-p_{i,k}\right)$.\\
\hspace{0.4in}    Compute $z_{i,k} = \frac{r_{i,k} - p_{i,k}}{p_{i,k}\left(1-p_{i,k}\right) }$.\\
\hspace{0.4in}   Fit the function $f_{i,k}$  by a weighted  least-square of $z_{i,k}$ to $\mathbf{x}_i$ with weights $w_{i,k}$.\\
 \hspace{0.4in}  $F_{i,k} = F_{i,k} + \nu \frac{K-1}{K}\left( f_{i,k} - \frac{1}{K}\sum_{k=0}^{K-1}f_{i,k}\right)$\\
   \hspace{0.2in} End\\
 \hspace{0.2in}  $p_{i,k} = \exp(F_{i,k})/\sum_{s=0}^{K-1}\exp(F_{i,s})$\\
End
\caption{\small LogitBoost~\citep[Algorithm 6]{Article:FHT_AS00}. $\nu$ is the shrinkage. }\label{alg:LogitBoost}}
\end{algorithm}

As described in Algorithm~\ref{alg:LogitBoost}, \citet{Article:FHT_AS00} built the additive model (\ref{eqn_F_M}) by a greedy stage-wise  procedure, using a second-order (diagonal) approximation, which requires computing the first two derivatives of the loss function (\ref{eqn_loss}) with respective to the function values $F_{i,k}$ as follows:
\begin{align}\label{eqn:logit_d1d2}
&\frac{\partial L_i}{\partial F_{i,k}} = - \left(r_{i,k} - p_{i,k}\right),
\hspace{0.5in}
\frac{\partial^2 L_i}{\partial F_{i,k}^2} = p_{i,k}\left(1-p_{i,k}\right),
\end{align}
which are standard results in textbooks.

At each stage, LogitBoost fits an individual regression function separately for each class. This is analogous to the popular {\em individualized regression} approach in multinomial logistic regression, which is known~\citep{Article:Begg_84,Book:Agresti} to result in  loss of statistical efficiency, compared to the full (conditional) maximum likelihood approach. On the other hand, in order to use trees as base learner, the diagonal approximation appears to be a must, at least from the practical perspective.

\newpage

\subsection{Robust LogitBoost}

The MART paper~\citep{Article:Friedman_AS01} and a later (2008) discussion paper~\citep{Article:FHT_JMLR08} commented that the LogitBoost (Algorithm~\ref{alg:LogitBoost}) can be numerically unstable. In fact, the  LogitBoost paper suggested some ``crucial implementation protections'' on page 17 of~\citet{Article:FHT_AS00}:

\begin{itemize}
\item In Line 5 of Algorithm~\ref{alg:LogitBoost}, compute the response $z_{i,k}$ by $\frac{1}{p_{i,k}}$ (if $r_{i,k}=1$) or $\frac{-1}{1-p_{i,k}}$ (if $r_{i,k}=0$).
\item Bound the response $|z_{i,k}|$ by $z_{max}\in[2,4]$. The value of $z_{max}$ is not sensitive as long as in $[2,4]$
\end{itemize}
Note that the above operations were applied to each individual sample. The goal was to ensure that the response $|z_{i,k}|$ should  not be too large. On the other hand, with more boosting iterations, the fitted class probabilities $p_{i,k}$ are expected to approach 0 or 1, which means we must have large $|z_{i,k}|$ values. Therefore, limiting the values of $|z_{i,k}|$ (e.g., by 2 or 4) would hurt the performance.

The next subsection explains that, if implemented as in~\citet{Proc:ABC_UAI10}, there is no need to restrict the values of $|z_{i,k}|$, and LogitBoost would be almost identical to MART, with the only difference being the tree-splitting criterion.

\subsection{Tree-Splitting Criterion Using Second-Order Information}\label{sec_split}

Consider $N$ weights $w_i$, and $N$ response values $z_i$, $i=1$ to $N$, which are assumed to be ordered according to the sorted order of the corresponding feature values. The tree-splitting procedure is to find the index $t$, $1\leq t<N$, such that the weighted  square error (SE) is reduced the most if split at $t$.  That is, we seek the $t$ to maximize
\begin{align}\notag
Gain(t) = &SE_{total} - (SE_{left} + SE_{right})\\\notag
=&\sum_{i=1}^N (z_i - \bar{z})^2w_i - \left[
\sum_{i=1}^t (z_i - \bar{z}_L)^2w_i + \sum_{i=t+1}^N (z_i - \bar{z}_R)^2w_i\right]
\end{align}
where $\bar{z} = \frac{\sum_{i=1}^N z_iw_i}{\sum_{i=1}^N w_i}$,
$\bar{z}_{left} = \frac{\sum_{i=1}^t z_iw_i}{\sum_{i=1}^t w_i}$,
$\bar{z}_{right} = \frac{\sum_{i=t+1}^N z_iw_i}{\sum_{i=t+1}^{N} w_i}$.  With some algebra, one can obtain
\begin{align}\notag
\sum_{i=1}^N (z_i - \bar{z})^2w_i =& \sum_{i=1}^N z_i^2w_i + \bar{z}^2w_i-2z_i\bar{z}w_i\\\notag
=&\sum_{i=1}^N z_i^2w_i + \left(\frac{\sum_{i=1}^N z_iw_i}{\sum_{i=1}^N w_i}\right)^2\sum_{i=1}^N w_i - 2\frac{\sum_{i=1}^N z_iw_i}{\sum_{i=1}^N w_i}\sum_{i=1}^N z_iw_i\\\notag
=&\sum_{i=1}^N z_i^2w_i - \frac{\left(\sum_{i=1}^N z_iw_i\right)^2}{\sum_{i=1}^N w_i}\\\notag
\sum_{i=1}^t (z_i - \bar{z}_{left})^2w_i
=&\sum_{i=1}^t z_i^2w_i - \frac{\left(\sum_{i=1}^t z_iw_i\right)^2}{\sum_{i=1}^t w_i}\\\notag
\sum_{i=t+1}^N (z_i - \bar{z}_{right})^2w_i
=&\sum_{i=t+1}^N z_i^2w_i - \frac{\left(\sum_{i=t+1}^N z_iw_i\right)^2}{\sum_{i=t+1}^N w_i}
\end{align}
Therefore, after simplification, we obtain
\begin{align}\notag
Gain(t) =& \frac{\left[\sum_{i=1}^t z_iw_i\right]^2}{\sum_{i=1}^t w_i}+\frac{\left[\sum_{i=t+1}^N z_iw_i\right]^2}{\sum_{i=t+1}^{N} w_i}- \frac{\left[\sum_{i=1}^N z_iw_i\right]^2}{\sum_{i=1}^N w_i}
\end{align}

Plugging in  $w_i = p_{i,k}(1-p_{i,k})$, $z_i = \frac{r_{i,k}-p_{i,k}}{p_{i,k}(1-p_{i,k})}$  yields,
\begin{align}\label{eqn:logit_gain}
Gain(t) =&  \frac{\left[\sum_{i=1}^t \left(r_{i,k} - p_{i,k}\right) \right]^2}{\sum_{i=1}^t p_{i,k}(1-p_{i,k})}+\frac{\left[\sum_{i=t+1}^N \left(r_{i,k}- p_{i,k}\right) \right]^2}{\sum_{i=t+1}^{N} p_{i,k}(1-p_{i,k})}- \frac{\left[\sum_{i=1}^N \left(r_{i,k} - p_{i,k}\right) \right]^2}{\sum_{i=1}^N p_{i,k}(1-p_{i,k})}.
\end{align}
Because the computations involve $\sum p_{i,k}(1-p_{i,k})$ as a group, this procedure is actually numerically stable.\\

In comparison, MART~\citep{Article:Friedman_AS01}  used the first order information to construct the trees, i.e.,
\begin{align}\label{eqn:mart_gain}
MartGain(t) =&  \frac{1}{t}\left[\sum_{i=1}^t \left(r_{i,k} - p_{i,k}\right) \right]^2+
\frac{1}{N-t}\left[\sum_{i=t+1}^N \left(r_{i,k} - p_{i,k}\right) \right]^2-\frac{1}{N}
\left[\sum_{i=1}^N \left(r_{i,k} - p_{i,k}\right) \right]^2.
\end{align}

{\scriptsize\begin{algorithm}{\small
$F_{i,k} = 0$, $p_{i,k} = \frac{1}{K}$, $k = 0$ to  $K-1$, $i = 1$ to $N$ \\
For $m=1$ to $M$ Do\\
\hspace{0.1in}    For $k=0$ to $K-1$ Do\\
\hspace{0.2in}  $\left\{R_{j,k,m}\right\}_{j=1}^J = J$-terminal node regression tree from
 $\{r_{i,k} - p_{i,k}, \ \ \mathbf{x}_{i}\}_{i=1}^N$,  with weights $p_{i,k}(1-p_{i,k})$, using the tree split gain formula Eq.~\eqref{eqn:logit_gain}.\\
 \hspace{0.2in}  $\beta_{j,k,m} = \frac{K-1}{K}\frac{ \sum_{\mathbf{x}_i \in
  R_{j,k,m}} r_{i,k} - p_{i,k}}{ \sum_{\mathbf{x}_i\in
  R_{j,k,m}}\left(1-p_{i,k}\right)p_{i,k} }$ \\
\hspace{0.2in}  $F_{i,k} = F_{i,k} +
\nu\sum_{j=1}^J\beta_{j,k,m}1_{\mathbf{x}_i\in R_{j,k,m}}$ \\
 \hspace{0.1in} End\\
\hspace{0.12in} $p_{i,k} = \exp(F_{i,k})/\sum_{s=0}^{K-1}\exp(F_{i,s})$\\
End
\caption{Robust LogitBoost. MART is similar, with the only difference in Line 4.  }
\label{alg:robust_LogitBoost}}
\end{algorithm}}

Algorithm~\ref{alg:robust_LogitBoost} describes Robust LogitBoost using the tree split gain formula in Eq.~\eqref{eqn:logit_gain}. Note that after trees are constructed, the values of the terminal nodes are computed by
\begin{align}\notag
\frac{\sum_{node} z_{i,k} w_{i,k}}{\sum_{node} w_{i,k}} =
\frac{\sum_{node} \left(r_{i,k} - p_{i,k}\right)}{\sum_{node} p_{i,k}(1-p_{i,k})},
\end{align}
which explains Line 5 of Algorithm~\ref{alg:robust_LogitBoost}. In the implementation, to avoid occasional numerical issues, a very small ``damping'' could be added to the denominator, e.g.,  $\left\{\epsilon+\sum_{node} p_{i,k}(1-p_{i,k})\right\}$.

\vspace{0.1in}

To avoid repetition, we do not provide the pseudo code for MART~\citep{Article:Friedman_AS01}, which is in fact almost identical to Algorithm~\ref{alg:robust_LogitBoost}. The only difference is in Line 4, which for MART becomes \\

$\left\{R_{j,k,m}\right\}_{j=1}^J = J$-terminal node regression tree from
 $\{r_{i,k} - p_{i,k}, \ \ \mathbf{x}_{i}\}_{i=1}^N$,\\
 $~\ $\hspace{3in} using the tree split gain formula Eq.~\eqref{eqn:mart_gain}.


In retrospect, deriving Eq.~\eqref{eqn:logit_gain}, the explicit  tree split gain formula using the second-order information, is easy. Nevertheless, until the formula was first presented by~\cite{Proc:ABC_UAI10}, the original LogitBoost was believed to have numerical issues and the more robust MART algorithm was a  popular alternative~\citep{Article:Friedman_AS01,Article:FHT_JMLR08}. As shown in~\cite{Proc:ABC_UAI10}, in many datasets, because it used only the first derivative information for tree split, MART typically did not achieve the same accuracy as Robust LogitBoost. Nowadays, the split gain formula in Eq.~\eqref{eqn:logit_gain} is the standard implementation in popular tree platforms.


\section{ABC-Boost: Adaptive Base Class Boost}

In this section, we review ABC-Boost (adaptive base class boost) as presented in~\citet{Proc:ABC_ICML09,Proc:ABC_UAI10}. The development of ABC-Boost began with the finding that the derivatives of the classical logistic regression loss function can be re-written in a new format.

\subsection{The New Derivatives of Logistic Loss Function}

Recall that, in the $K$-class  multi-class classification, because the class probabilities must sum to one: $\sum_{k=0}^{K-1} p_{i,k} = 1$, there are only $K-1$ degrees of freedom. Typically, the ``sum-to-zero'' assumption is made on the function values: $\sum_{k=0}^{K-1} F_{i,k} = 0$. Instead of using the classical derivatives $\frac{\partial L_i}{\partial F_{i,k}}$ and $\frac{\partial^2 L_i}{\partial F_{i,k}^2}$ as in Eq.~\eqref{eqn:logit_d1d2}, \citet{Proc:ABC_ICML09} derived the derivatives of the loss function (\ref{eqn_loss}) under this sum-to-zero constraint. Without loss of generality, we can assume that class 0 is the base class. For any $k\neq 0$, the derivatives can be expressed differently from the classical derivatives in Eq.~\eqref{eqn:logit_d1d2}, as follows:
\begin{align}\label{eqn:abc_d1}
&\frac{\partial L_i}{\partial F_{i,k}}  = \left(r_{i,0} - p_{i,0}\right) - \left(r_{i,k} - p_{i,k}\right),\\\label{eqn:abc_d2}
&\frac{\partial^2 L_i}{\partial F_{i,k}^2} = p_{i,0}(1-p_{i,0}) + p_{i,k}(1-p_{i,k}) + 2p_{i,0}p_{i,k}.
\end{align}

In case some Readers are curious, here we would like to repeat the derivation of the above two derivatives. For any $k\neq 0$, we have
\begin{align}\notag
\frac{\partial L_i}{\partial F_{i,k}} =& -\frac{\partial }{\partial F_{i,k}}  \sum_{k=0}^{K-1}r_{i,k}  \log p_{i,k}, \hspace{0.2in} \text{where } p_{i,k} = \frac{e^{F_{i,k}}}{\sum_{t=0}^{K-1} e^{F_{i,t}}},\  F_{i,0} = -\sum_{k=1}^{K-1}F_{i,k}\\\notag
=&-\sum_{t=1,t\neq k} \frac{r_{i,t}}{p_{i,t}} \frac{\partial p_{i,t}}{\partial F_{i,k}} - \frac{r_{i,k}}{p_{i,k}} \frac{\partial p_{i,k}}{\partial F_{i,k}} - \frac{r_{i,0}}{p_{i,0}} \frac{\partial p_{i,0}}{\partial F_{i,k}}
\end{align}
which requires the following three derivatives
\begin{align}\notag
&\frac{\partial p_{i,k}}{\partial F_{i,k}} =     \frac{e^{F_{i,k}}\left[\sum_{t=0}^{K-1} e^{F_{i,t}}\right] -e^{F_{i,k}}\left[e^{F_{i,k}} - e^{F_{i,0}}\right] }{\left[\sum_{t=0}^{K-1} e^{F_{i,t}}\right]^2} = p_{i,k}\left(1+p_{i,0}-p_{i,k}\right)\\\notag
&\frac{\partial p_{i,0}}{\partial F_{i,k}} =     \frac{-e^{F_{i,0}}\left[\sum_{t=0}^{K-1} e^{F_{i,t}}\right] -e^{F_{i,0}}\left[e^{F_{i,k}} - e^{F_{i,0}}\right] }{\left[\sum_{t=0}^{K-1} e^{F_{i,t}}\right]^2} = p_{i,0}\left(-1+p_{i,0}-p_{i,k}\right), \ \ k\neq 0\\\notag
&\frac{\partial p_{i,k}}{\partial F_{i,s}} =     \frac{ -e^{F_{i,k}}\left[e^{F_{i,s}} - e^{F_{i,0}}\right] }{\left[\sum_{t=0}^{K-1} e^{F_{i,t}}\right]^2} = p_{i,k}\left(p_{i,0}-p_{i,s}\right), \ \ k\neq s\neq 0
\end{align}
Therefore, we have
\begin{align}\notag
\frac{\partial L_i}{\partial F_{i,k}}
=&-\sum_{t=1,t\neq k} \frac{r_{i,t}}{p_{i,t}} \frac{\partial p_{i,t}}{\partial F_{i,k}} - \frac{r_{i,k}}{p_{i,k}} \frac{\partial p_{i,k}}{\partial F_{i,k}} - \frac{r_{i,0}}{p_{i,0}} \frac{\partial p_{i,0}}{\partial F_{i,k}}\\\notag
=&-\sum_{t=1,t\neq k}\frac{r_{i,t}}{p_{i,t}}  p_{i,t}\left(p_{i,0}-p_{i,k}\right) - \frac{r_{i,k}}{p_{i,k}}p_{i,k}\left(1+p_{i,0}-p_{i,k}\right) - \frac{r_{i,0}}{p_{i,0}} p_{i,0}\left(-1+p_{i,0}-p_{i,k}\right)\\\notag
=&-\sum_{t=1,t\neq k}r_{i,t} \left(p_{i,0}-p_{i,k}\right) - r_{i,k}\left(1+p_{i,0}-p_{i,k}\right) - r_{i,0}\left(-1+p_{i,0}-p_{i,k}\right)\\\notag
=&- \left(p_{i,0}-p_{i,k}\right)\sum_{t=0}^{K-1}r_{i,t} - r_{i,k}+r_{i,0}
=- \left(p_{i,0}-p_{i,k}\right) - r_{i,k}+r_{i,0}
=\left(r_{i,0} - p_{i,0}\right) - \left(r_{i,k} - p_{i,k}\right)
\end{align}
Note that $\sum_{t=0}^{K-1}r_{i,t}=1$ because only one of the $r_{i,t}$ can be 1.

Deriving the second derivative is then straightforward:
\begin{align}\notag
\frac{\partial^2 L_i}{\partial F_{i,k}^2} =&-\frac{\partial L_i}{\partial F_{i,k}} \left(p_{i,0}-p_{i,k}\right) \\\notag
=&p_{i,k}\left(1+p_{i,0}-p_{i,k}\right) - p_{i,0}\left(-1+p_{i,0}-p_{i,k}\right)\\\notag
=&p_{i,0}(1-p_{i,0}) + p_{i,k}(1-p_{i,k}) + 2p_{i,0}p_{i,k}
\end{align}

Because it is the same logistic loss function, the expressions of derivatives may look different but readers would expect that they should be just a rewrite and should not affect the training/test accuracy. This is a  valid question. In fact, if readers check the determinant of the Hessian matrix using the classical derivatives in Eq.~\eqref{eqn:logit_d1d2}, they will quickly realize that the determinant if always zero, which is expected because there are only $K-1$ degrees of freedom. However, the determinant of the $(K-1)\times(K-1)$ Hessian matrix using the new derivatives would be non-zero and is in fact independent of the choice of the base class. In other words, if we use the full Hessian matrix, then it would not make a difference whether we use the classical derivatives or the new derivatives. It makes the difference because boosted trees are implemented using ``diagonal approximation''~\citep{Article:FHT_AS00}.

In the implementation of boosting with trees, for example, Algorithm~\ref{alg:LogitBoost} and Algorithm~\ref{alg:robust_LogitBoost}, at each iteration, each class is trained separately. This  ``diagonal approximation'' is really convenient. In fact, the authors do not know a better way to simultaneously train $K$ trees.

\subsection{ABC-Boost with the ``Exhaustive Search'' Strategy}

The base class must be identified at each boosting iteration during training. \citet{Proc:ABC_ICML09} suggested an exhaustive procedure to adaptively find the best base class, hence the name ABC-Boost: adaptive base class boost. ABC-Boost consists of two key components:
\begin{enumerate}
\item Using the {\em sum-to-zero} constraint on the loss function,  one can formulate boosting algorithms only for $K-1$ classes, by treating one class as the \textbf{base class}.
\item At each boosting iteration, \textbf{adaptively} select the base class according to the training loss.
\end{enumerate}

\citet{Proc:ABC_ICML09,Proc:ABC_UAI10} adopted the ``exhaustive search'' strategy for the base class. \citet{Proc:ABC_ICML09} combined ABC-Boost with MART to develop ABC-MART. Then \citet{Proc:ABC_UAI10} developed ABC-RobustLogitBoost, the combination of ABC-Boost with Robust LogitBoost.

\begin{algorithm}[h]{\small
$F_{i,k} = 0$,\ \  $p_{i,k} = \frac{1}{K}$, \ \ \ $k = 0$ to  $K-1$, \ $i = 1$ to $N$ \\
For $m=1$ to $M$ Do\\
\hspace{0.1in}    For $b=0$ to $K-1$, Do\\
\hspace{0.2in}    For $k=0$ to $K-1$, $k\neq b$, Do\\
\hspace{0.3in}  $\left\{R_{j,k,m}\right\}_{j=1}^J = J$-terminal
node regression tree from  $\{-(r_{i,b} - p_{i,b}) +  (r_{i,k} - p_{i,k}), \ \ \mathbf{x}_{i}\}_{i=1}^N$  with weights $p_{i,b}(1-p_{i,b})+p_{i,k}(1-p_{i,k})+2p_{i,b}p_{i,k}$, using the tree split gain formula~\eqref{eqn:abclogit_gain}.
 \\
 \hspace{0.3in}  $\beta_{j,k,m} = \frac{ \sum_{\mathbf{x}_i \in
  R_{j,k,m}} -(r_{i,b} - p_{i,b}) + (r_{i,k} - p_{i,k})  }{ \sum_{\mathbf{x}_i\in
  R_{j,k,m}} p_{i,b}(1-p_{i,b})+ p_{i,k}\left(1-p_{i,k}\right) + 2p_{i,b}p_{i,k} }$ \\
\hspace{0.3in}  $G_{i,k,b} = F_{i,k} +
\nu\sum_{j=1}^J\beta_{j,k,m}1_{\mathbf{x}_i\in R_{j,k,m}}$ \\
 \hspace{0.2in} End\\
\hspace{0.2in} $G_{i,b,b} = - \sum_{k\neq b} G_{i,k,b}$ \\
\hspace{0.2in}  $q_{i,k} = \exp(G_{i,k,b})/\sum_{s=0}^{K-1}\exp(G_{i,s,b})$ \\
\hspace{0.2in} $L^{(b)} = -\sum_{i=1}^N \sum_{k=0}^{K-1} r_{i,k}\log\left(q_{i,k}\right)$\\
\hspace{0.1in} End\\
\hspace{0.1in} $B(m) = \underset{b}{\text{argmin}} \  \ L^{(b)}$\\
\hspace{0.1in} $F_{i,k} = G_{i,k,B(m)}$\\
\hspace{0.1in}  $p_{i,k} = \exp(F_{i,k})/\sum_{s=0}^{K-1}\exp(F_{i,s})$ \\
End}
\caption{ABC-RobustLogitBoost using the exhaustive search strategy for the base class as in~\cite{Proc:ABC_ICML09,Proc:ABC_UAI10}.  The vector $B$ stores the base class numbers. }
\label{alg:abc-LogitBoost}
\end{algorithm}%

Algorithm~\ref{alg:abc-LogitBoost} presents  ABC-RobustLogitBoost using the derivatives in Eq.~\eqref{eqn:abc_d1} and Eq.~\eqref{eqn:abc_d2} and the  exhaustive search strategy. Again, ABC-RobustLogitBoost differs from ABC-MART only in the tree split procedure (Line 5).

In Algorithm~\ref{alg:abc-LogitBoost}, the gain formula for tree split in ABC-RobustLogitBoost is similar to~\eqref{eqn:logit_gain}:
\begin{align}\label{eqn:abclogit_gain}
ABCGain(t,b) =&  \frac{\left[\sum_{i=1}^t \left(r_{i,k} - p_{i,k}\right) -\left(r_{i,b} - p_{i,b}\right)\right]^2}{\sum_{i=1}^t p_{i,b}(1-p_{i,b})+p_{i,k}(1-p_{i,k})+2p_{i,b}p_{i,k}}\\\notag
&+\frac{\left[\sum_{i=t+1}^N \left(r_{i,k} - p_{i,k}\right) -\left(r_{i,b} - p_{i,b}\right)\right]^2}{\sum_{i=t+1}^N p_{i,b}(1-p_{i,b})+p_{i,k}(1-p_{i,k})+2p_{i,b}p_{i,k}}\\\notag
&-\frac{\left[\sum_{i=1}^N \left(r_{i,k} - p_{i,k}\right) -\left(r_{i,b} - p_{i,b}\right)\right]^2}{\sum_{i=1}^N p_{i,b}(1-p_{i,b})+p_{i,k}(1-p_{i,k})+2p_{i,b}p_{i,k}}
\end{align}

Algorithm~\ref{alg:robust_LogitBoost} and Algorithm~\ref{alg:abc-LogitBoost}  have three parameters ($J$, $\nu$ and $M$), to which the performance is in general not very sensitive, as long as they fall in some reasonable range. This is a significant advantage in practice.

The number of terminal nodes, $J$,  determines the capacity of the base learner. In our experiments and in~\citet{Proc:McRank_NIPS07,Proc:ABC_ICML09,Proc:ABC_UAI10}, we have found that using $J=20$ is often a reasonable starting point. The shrinkage, $\nu$, should be large enough to make sufficient progress at each step and  small enough to avoid over-fitting.  In general $\nu\leq 0.1$ and typically $\nu=0.1$ might be a good choice.  The number of boosting iterations, $M$, is to an extent determined  by the affordable computing time. A commonly-regarded merit of boosting is that, on many datasets, over-fitting can be largely avoided for reasonable $J$, and $\nu$.

\newpage

\subsection{An Evaluation Study}

We would like to provide an evaluation study for comparing MART, Robust LogitBoost, ABC-MART and ABC-RobustLogitBoost, with the ``exhaustive search'' strategy. Even though some of the results were already presented in~\citet{Proc:ABC_ICML09,Proc:ABC_UAI10}, we would like to include them for the convenience of comparisons with the results of other search strategies which will soon be presented in this paper.

\begin{table}[h]
\caption{Datasets}
\begin{center}{
\begin{tabular}{l r r r r}
\hline \hline
dataset &$K$ & \# training & \# test &\# features\\
\hline
Covertype &7 & 290506 & 290506 & 54\\
Mnist10k &10 &10000 &60000&784\\
M-Image &10 &12000 &50000&784\\
M-Rand &10 &12000 &50000&784\\
M-Noise1 &10 &10000 &2000&784\\
Letter &26 &10000 &10000 &16\\
\hline\hline
\end{tabular}
}
\end{center}
\label{tab:data}
\end{table}

Table~\ref{tab:data} lists the datasets used in our empirical study. They are common public datasets.  The {\em Covertype} and {\em Letter} datasets are from the UCI repository.  While the original {\em Mnist} dataset is extremely popular, that dataset is known to be too easy~\citep{Proc:Larochelle_ICML07}. Originally, {\em Mnist} used 60000 samples for training and 10000 samples for testing. {\em Mnist10k} uses the original (10000) test set for training and the original (60000) training set for testing, creating a more challenging task. \citet{Proc:Larochelle_ICML07} created a variety of much more difficult datasets by adding various background (correlated) noise, background images,  rotations, etc, to the original {\em Mnist} dataset. We shortened the notations of the generated datasets  and only used three of those datasets in~\citet{Proc:Larochelle_ICML07}:  {\em M-Image}, {\em M-Rand}, and {\em M-Noise1}, because the results in other datasets pretty much show the same patterns.\footnote{We would like to thank Dumitru Erhan, one of the authors of~\citet{Proc:Larochelle_ICML07}, for the private communications around 2009, about the datasets. }

\vspace{0.1in}

\begin{figure}[h]
\begin{center}
\mbox{
    \includegraphics[width=2.4in]{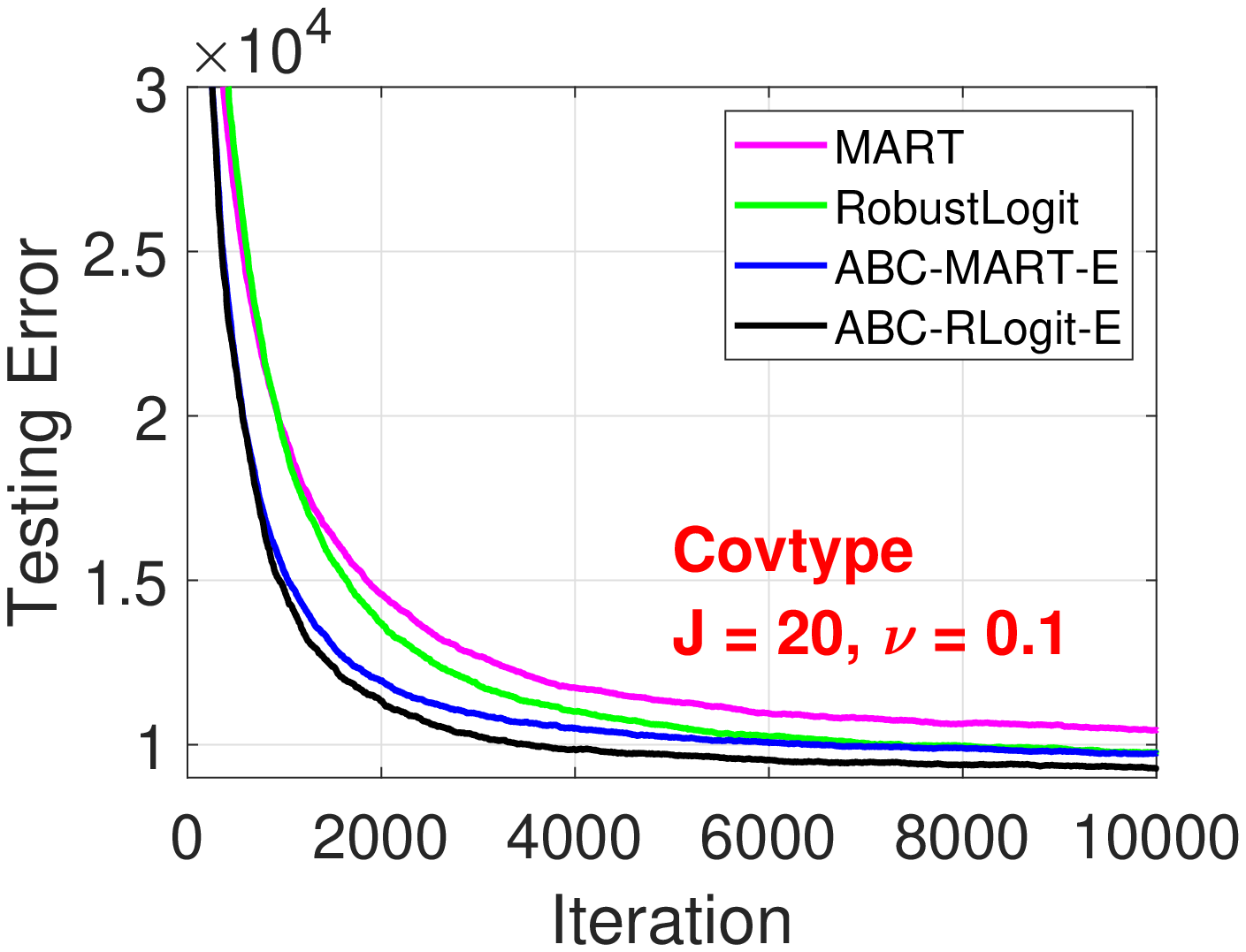}
    \includegraphics[width=2.4in]{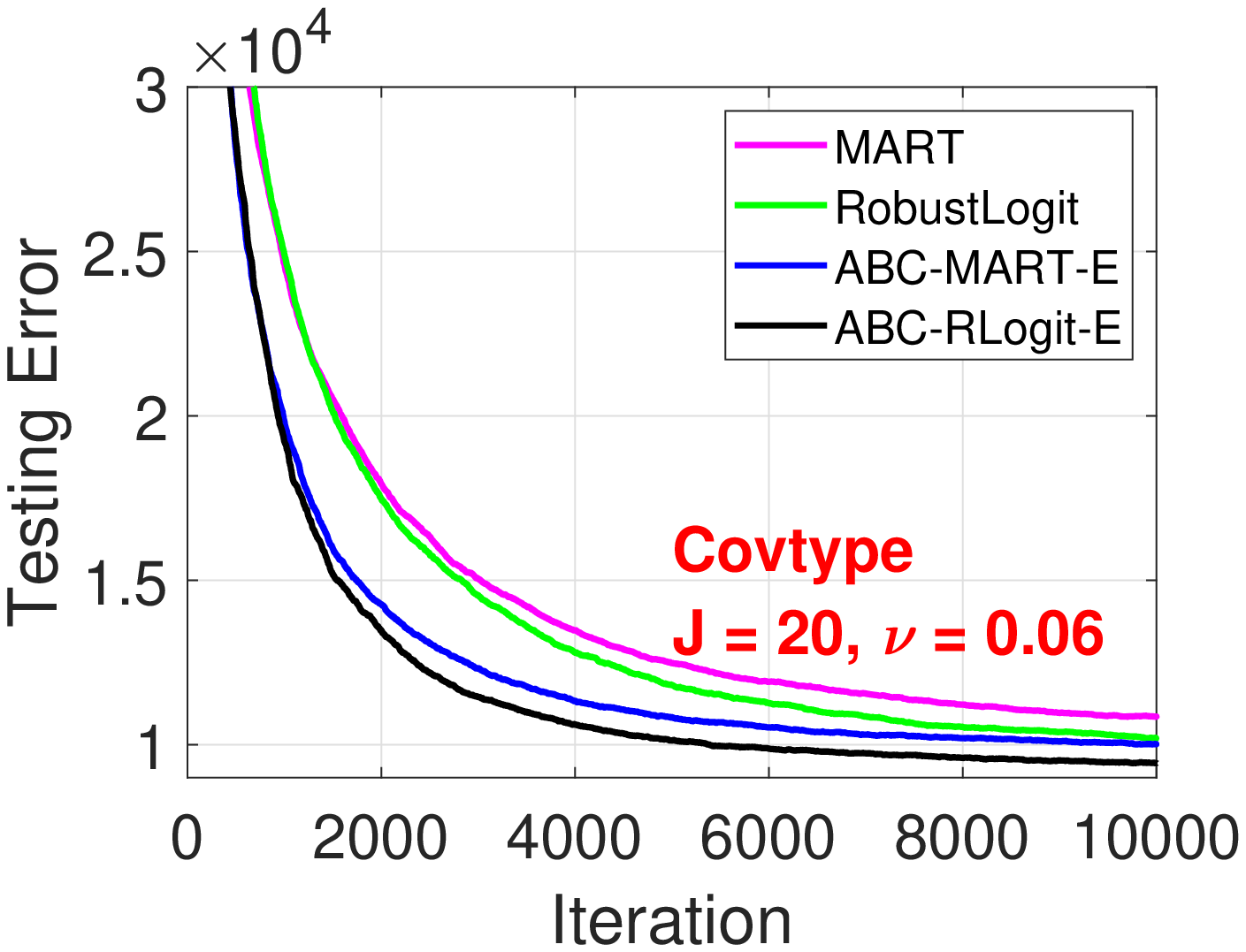}
}
\mbox{
    \includegraphics[width=2.4in]{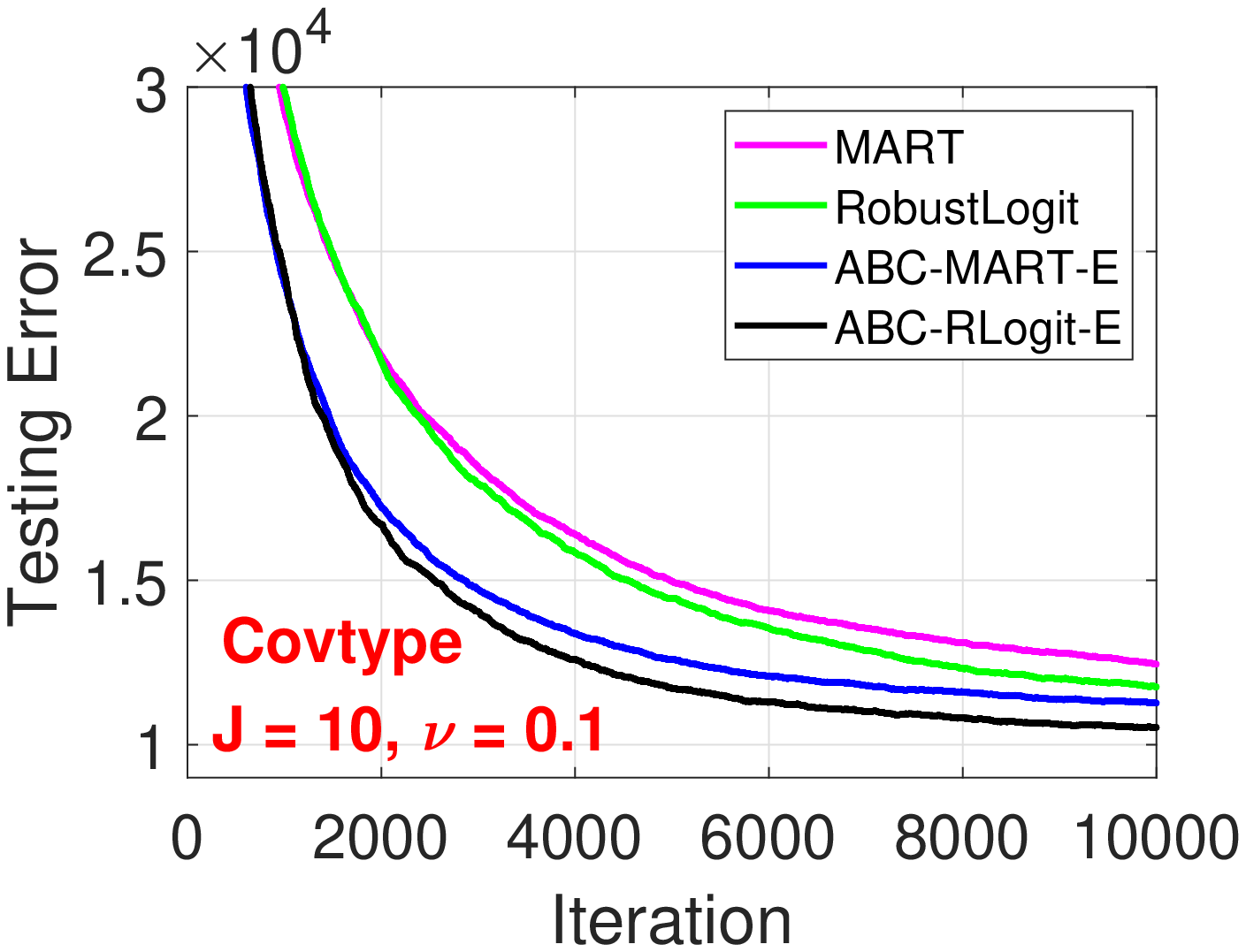}
    \includegraphics[width=2.4in]{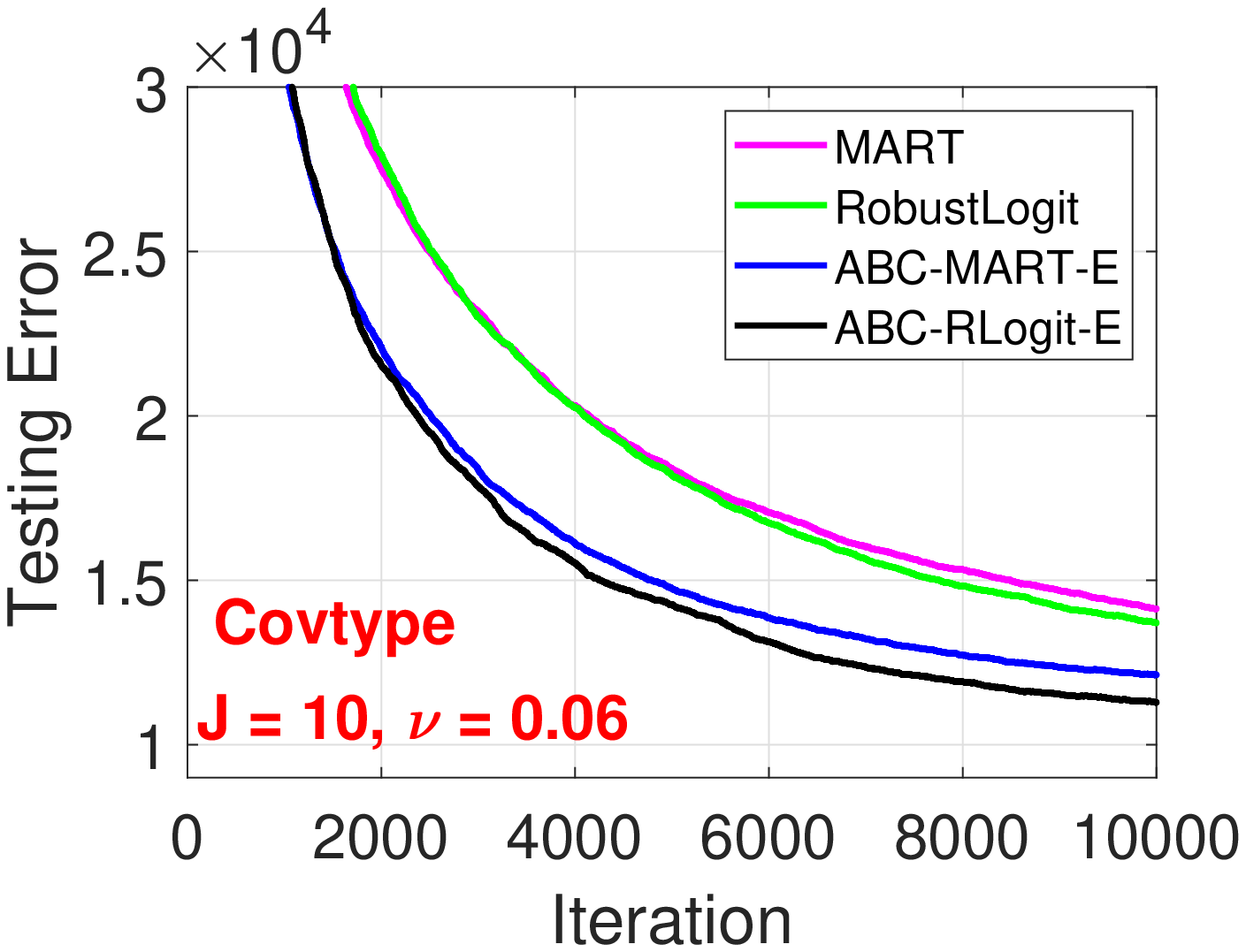}
}

\end{center}

\vspace{-0.1in}

\caption{{\em Covertype} dataset. We compare the test classification errors for four methods: MART, Robust LogitBoost, ABC-MART, and ABC-RobustLogitBoost, for $J\in \{10, 20\}$ and $\nu \in \{0.06, 0.1\}$.  Both ABC methods used the ``exhaustive search'' strategy.  }\label{fig:Covertype}
\end{figure}

\begin{figure}[h]
\begin{center}
\mbox{
    \includegraphics[width=2.4in]{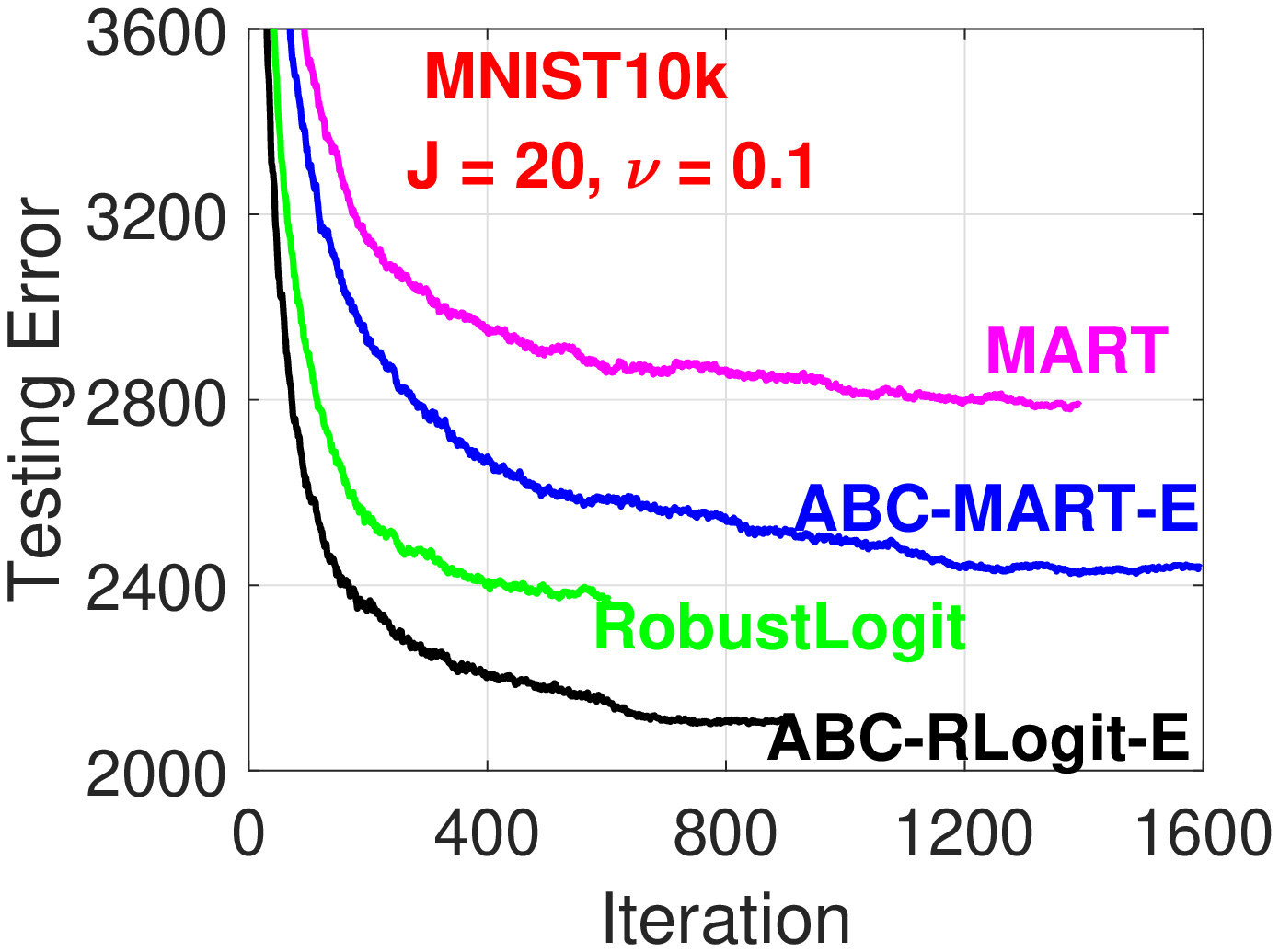}
    \includegraphics[width=2.4in]{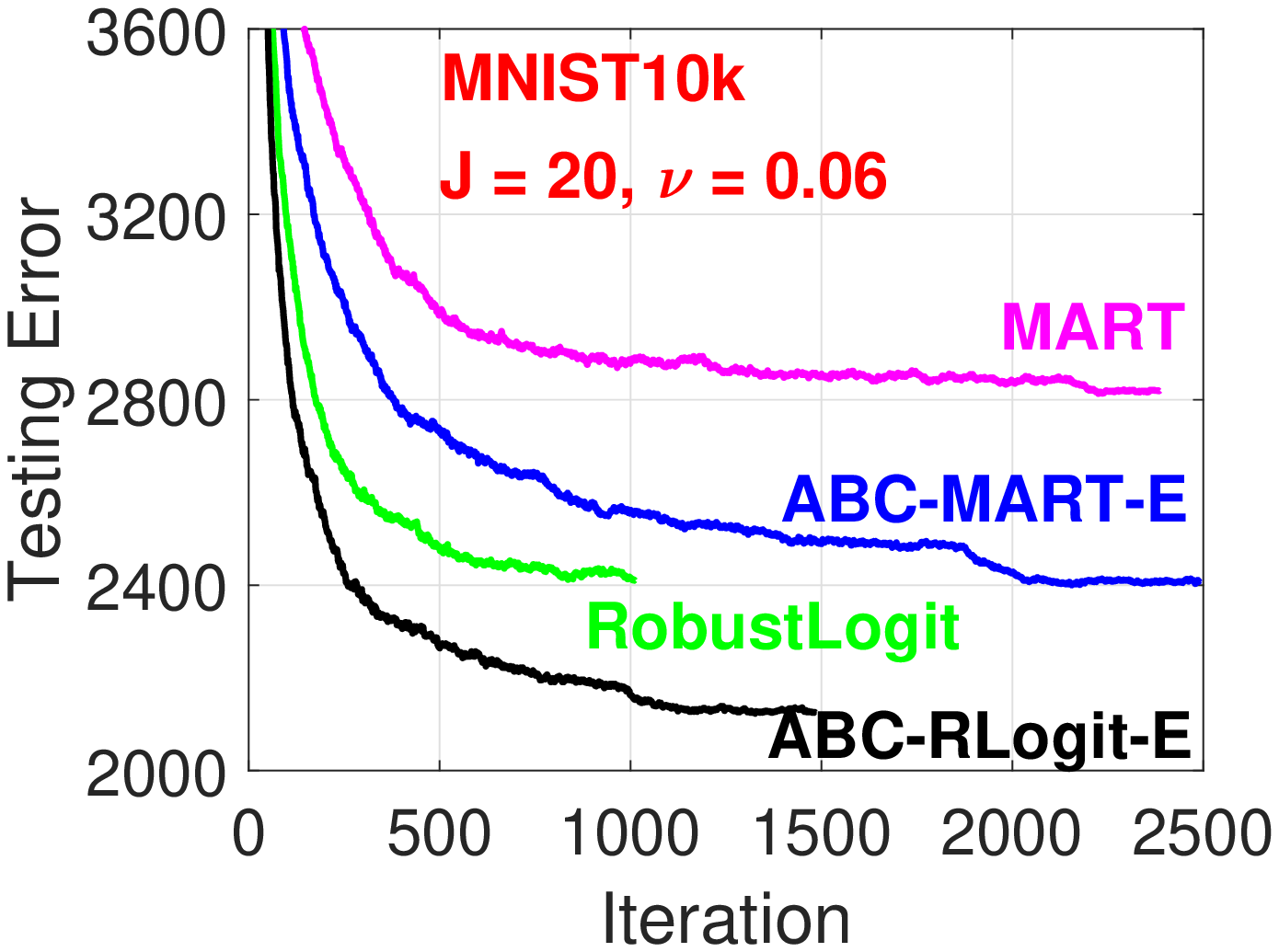}
}
\mbox{
    \includegraphics[width=2.4in]{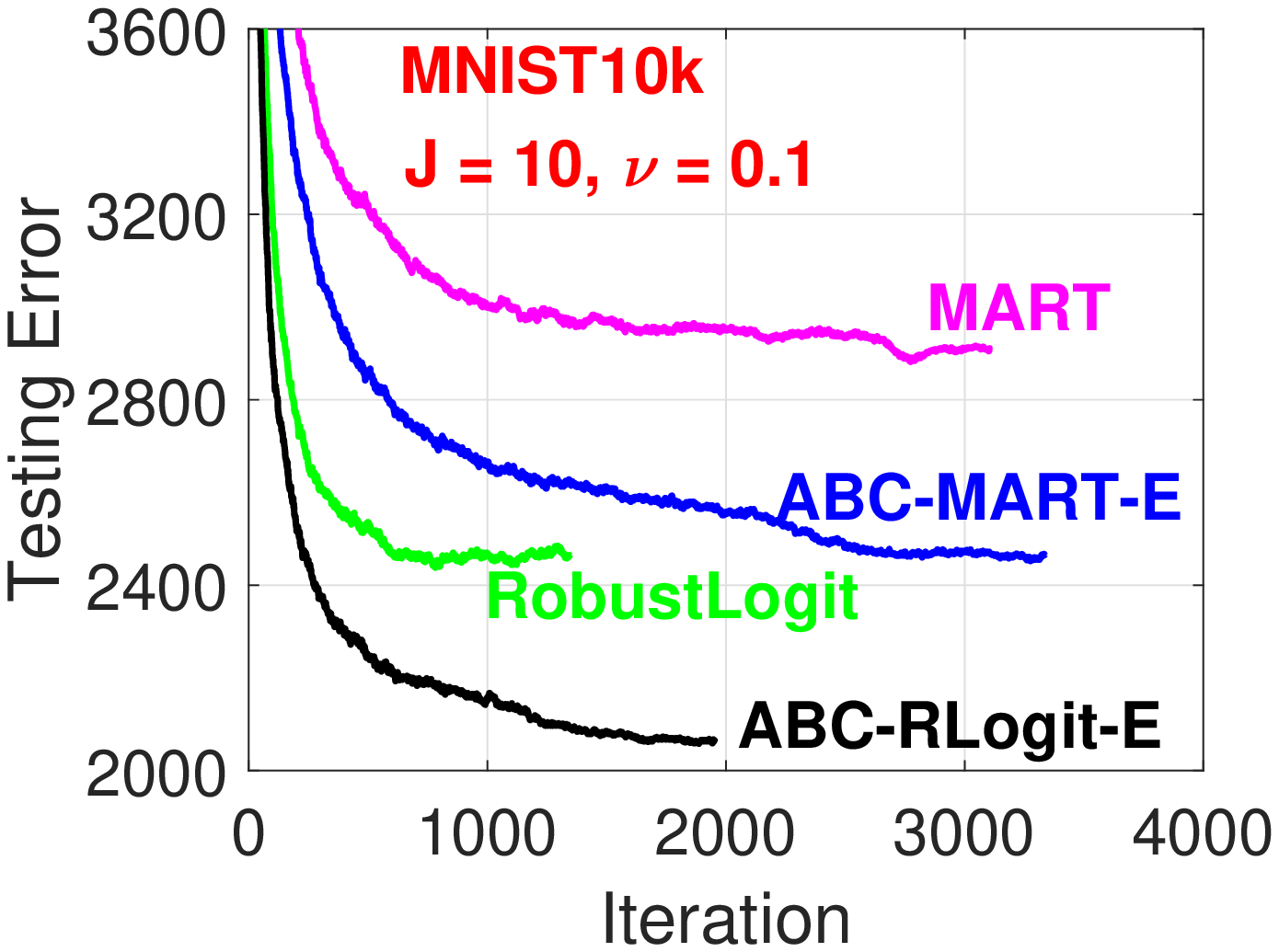}
    \includegraphics[width=2.4in]{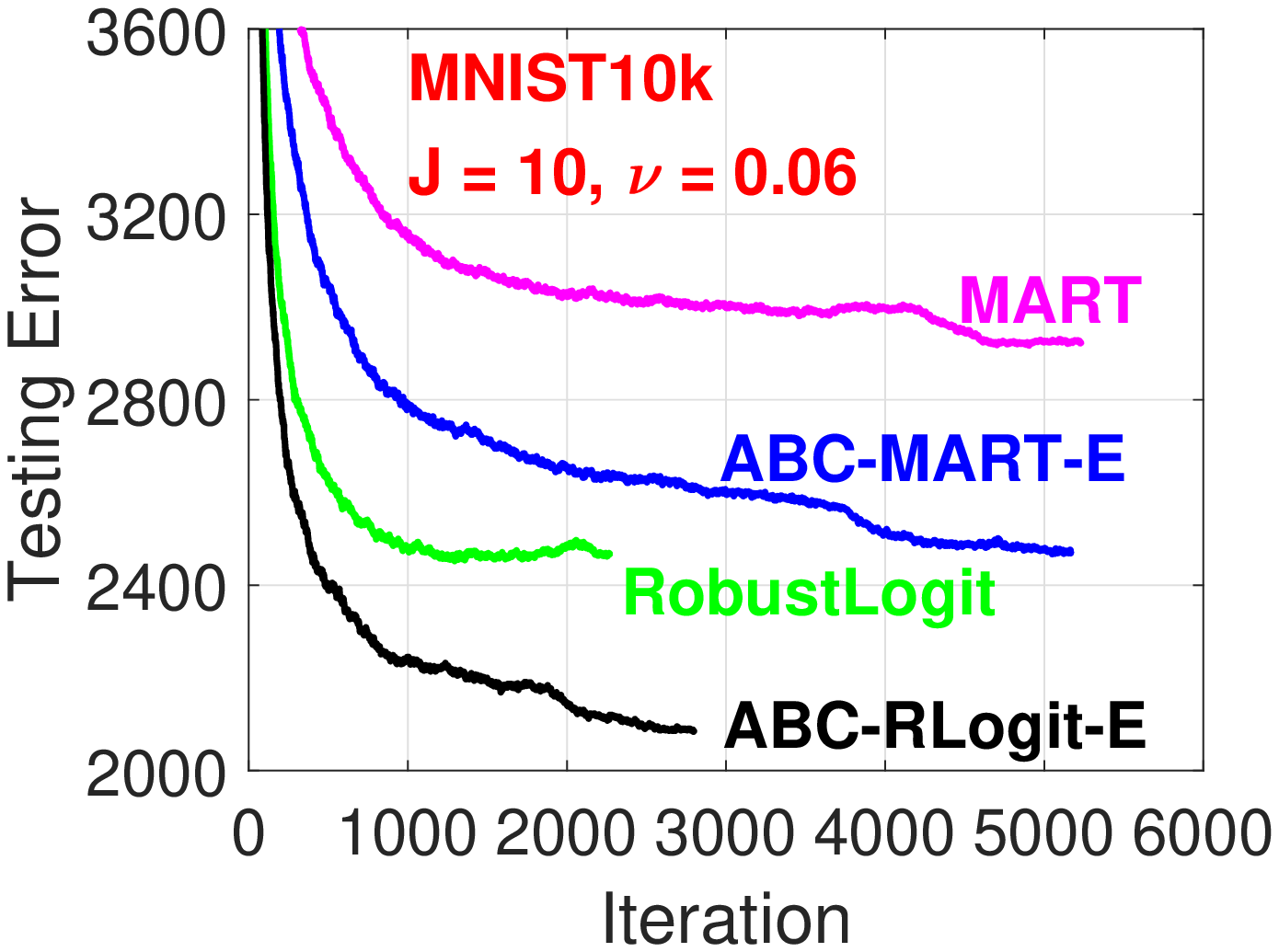}
}

\end{center}

\vspace{-0.1in}

\caption{{\em Mnist10k} dataset. We compare the test classification errors for four methods: MART, Robust LogitBoost, ABC-MART, and ABC-RobustLogitBoost, for $J\in \{10, 20\}$ and $\nu \in \{0.06, 0.1\}$. Both ABC methods used the ``exhaustive search'' strategy. }\label{fig:Mnist10k}
\end{figure}

\begin{figure}[h]
\begin{center}
\mbox{
    \includegraphics[width=2.4in]{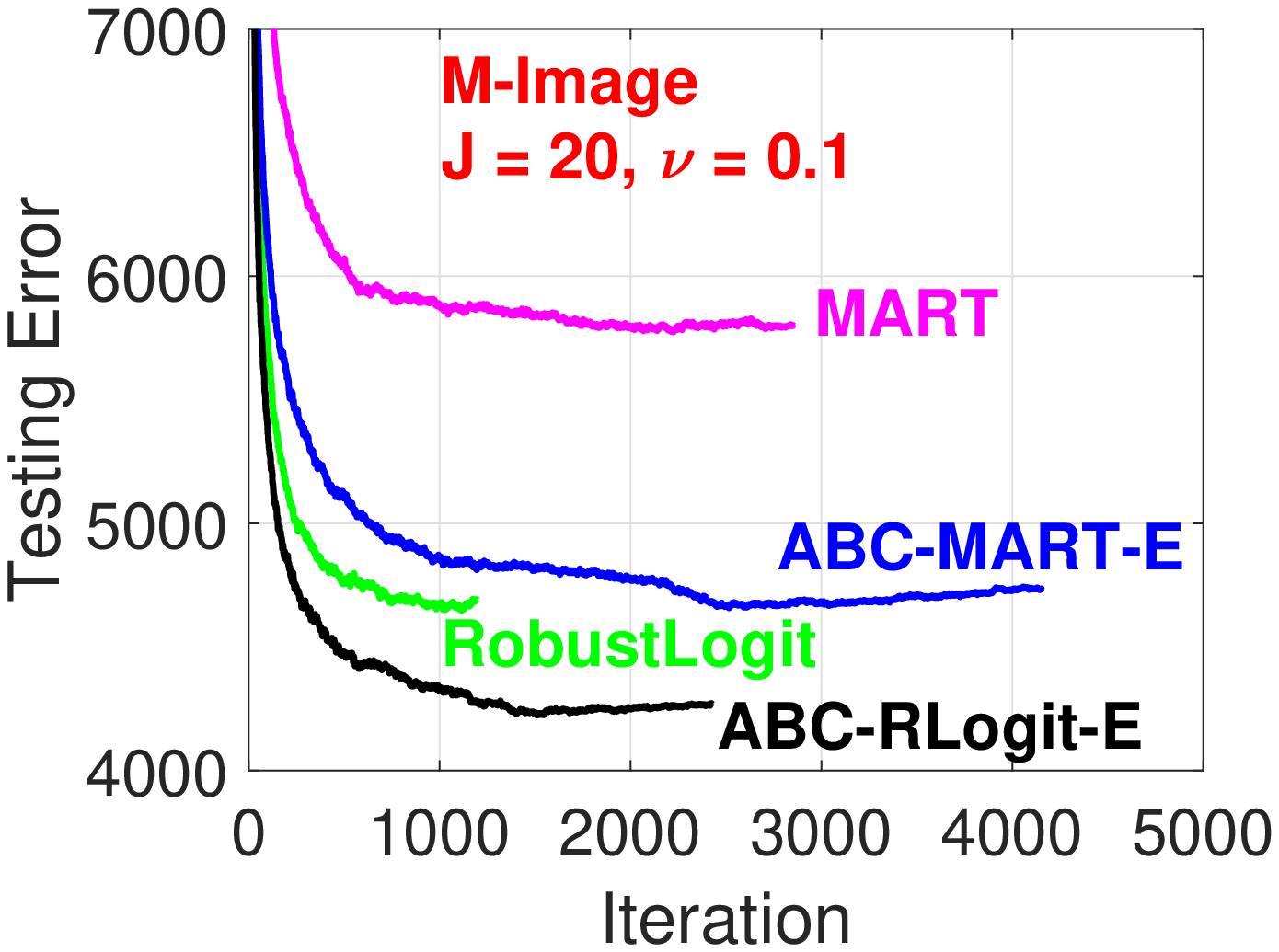}
    \includegraphics[width=2.4in]{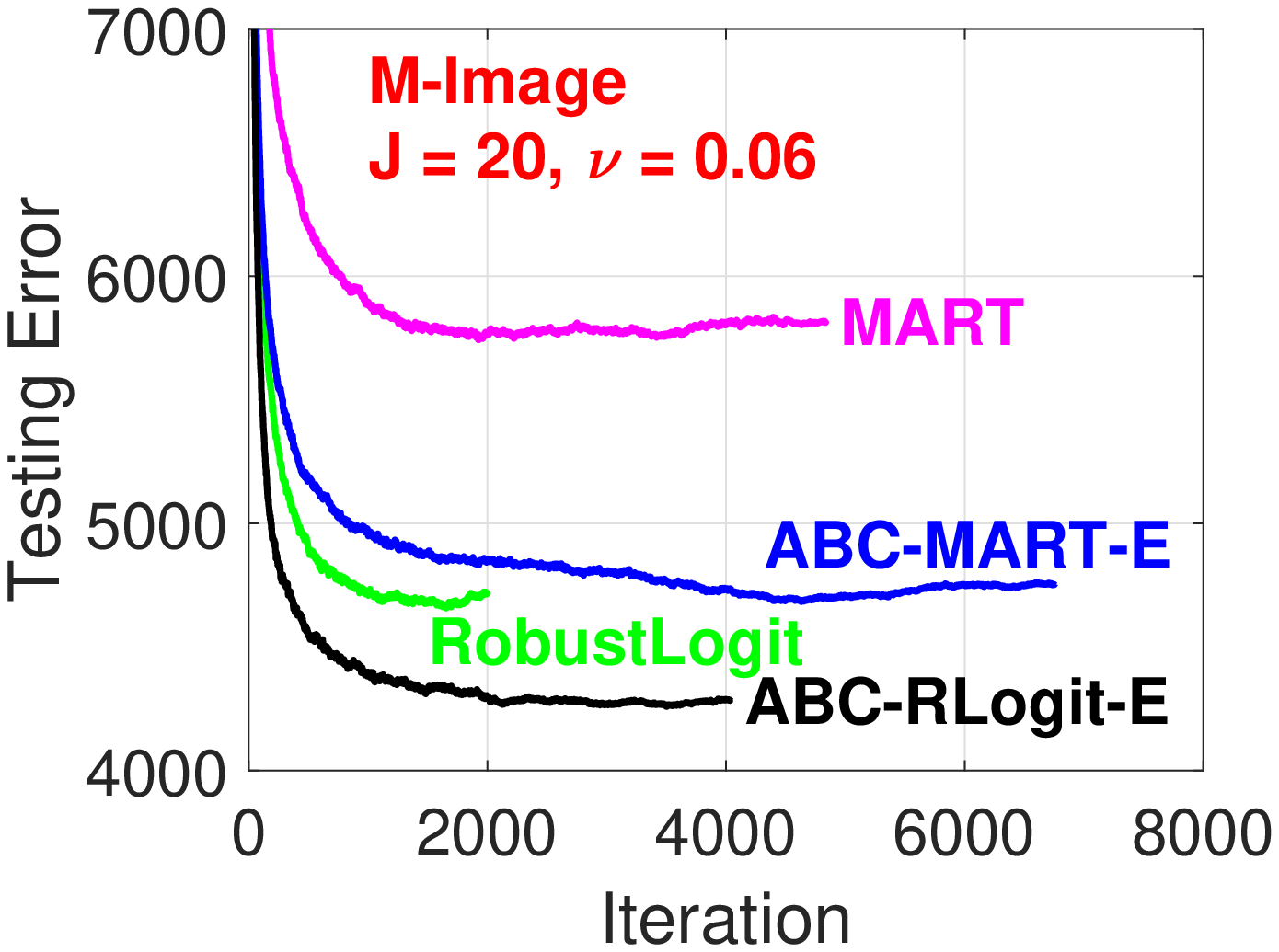}
}
\mbox{
    \includegraphics[width=2.4in]{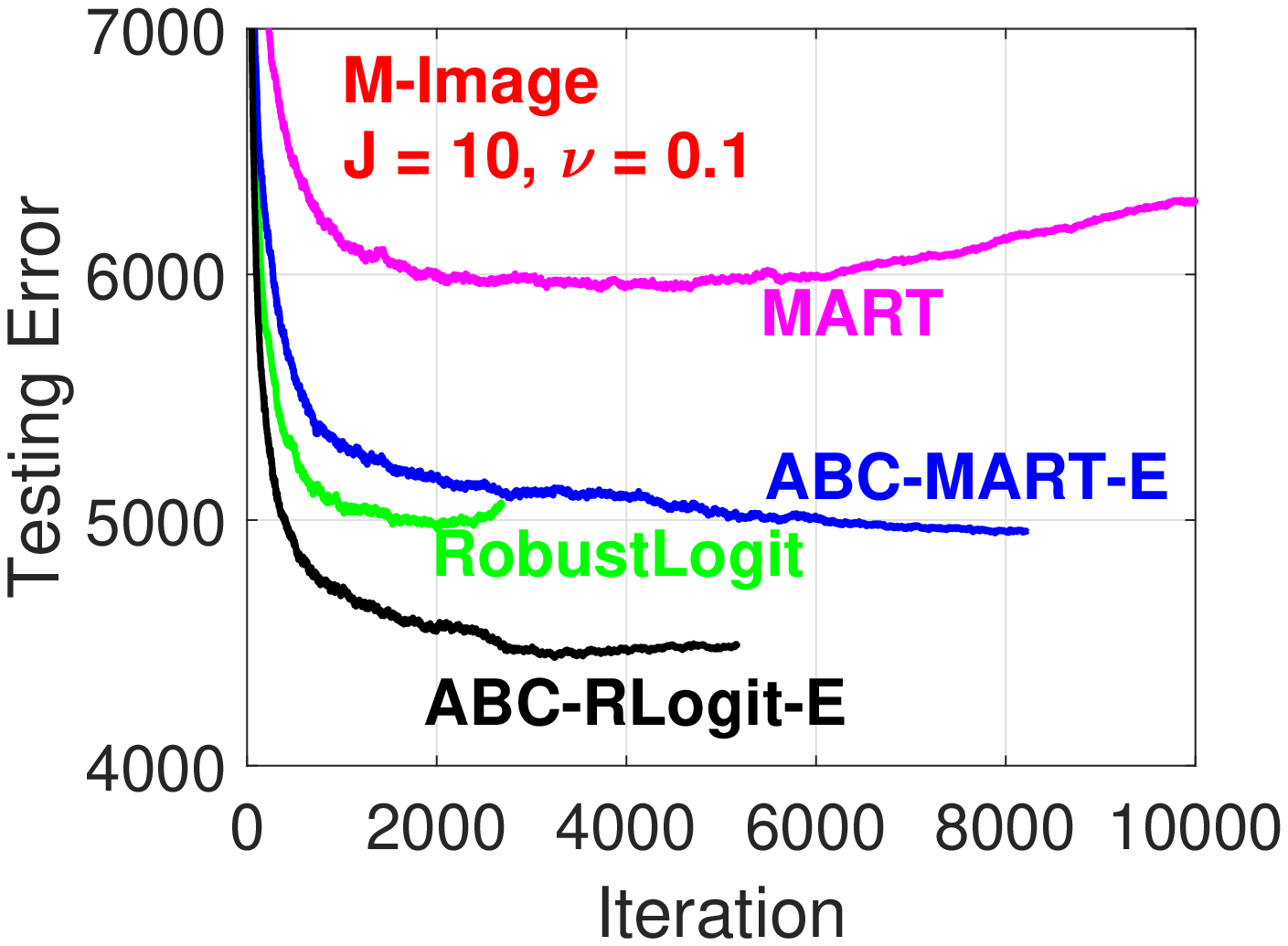}
    \includegraphics[width=2.4in]{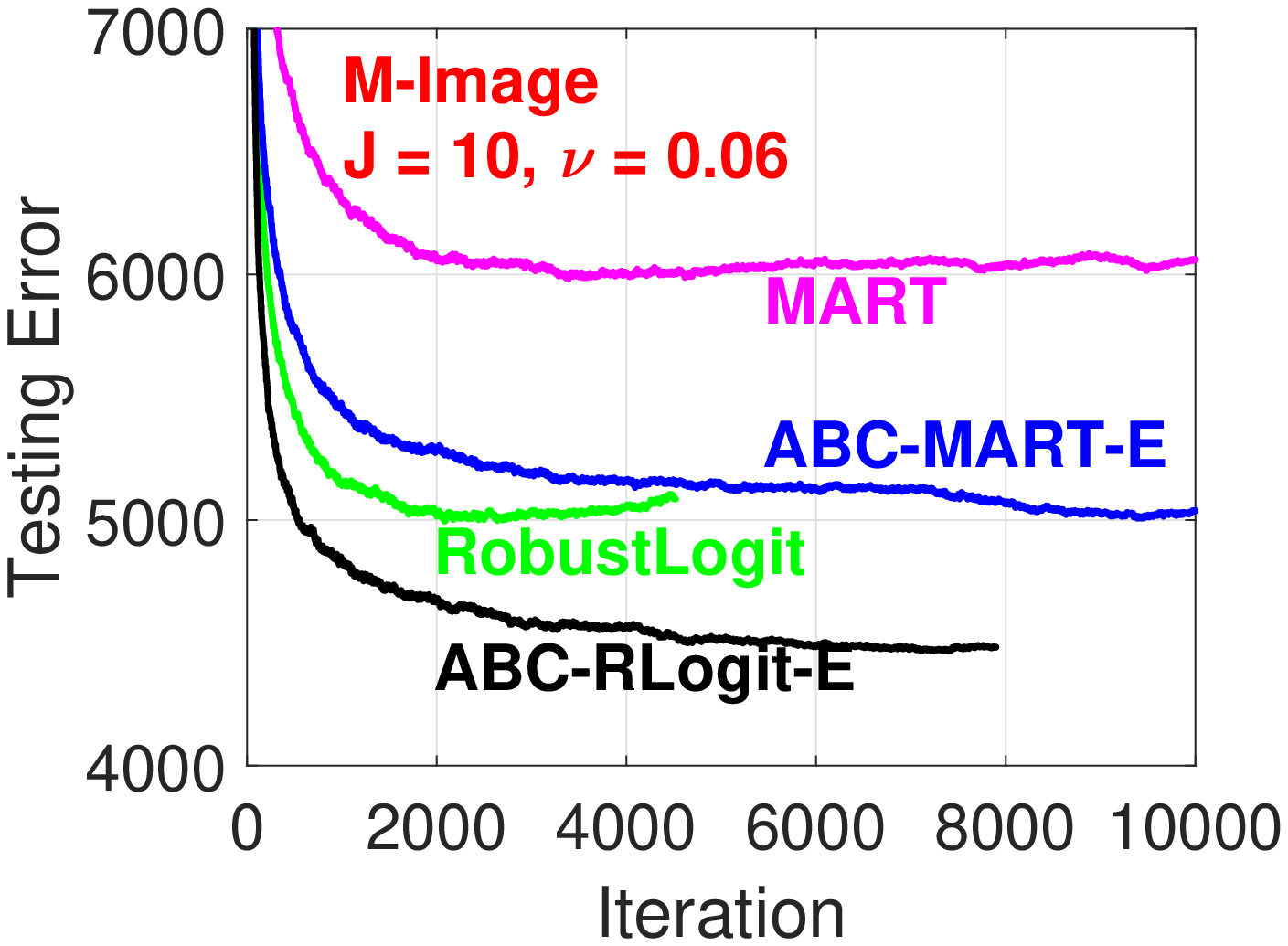}
}

\end{center}

\vspace{-0.1in}

\caption{{\em M-Image} dataset. We compare the test classification errors for four methods: MART, Robust LogitBoost, ABC-MART, and ABC-RobustLogitBoost, for $J\in \{10, 20\}$ and $\nu \in \{0.06, 0.1\}$. Both ABC methods used the ``exhaustive search'' strategy. }\label{fig:M-Image}
\end{figure}

\begin{figure}[h]
\begin{center}
\mbox{
    \includegraphics[width=2.4in]{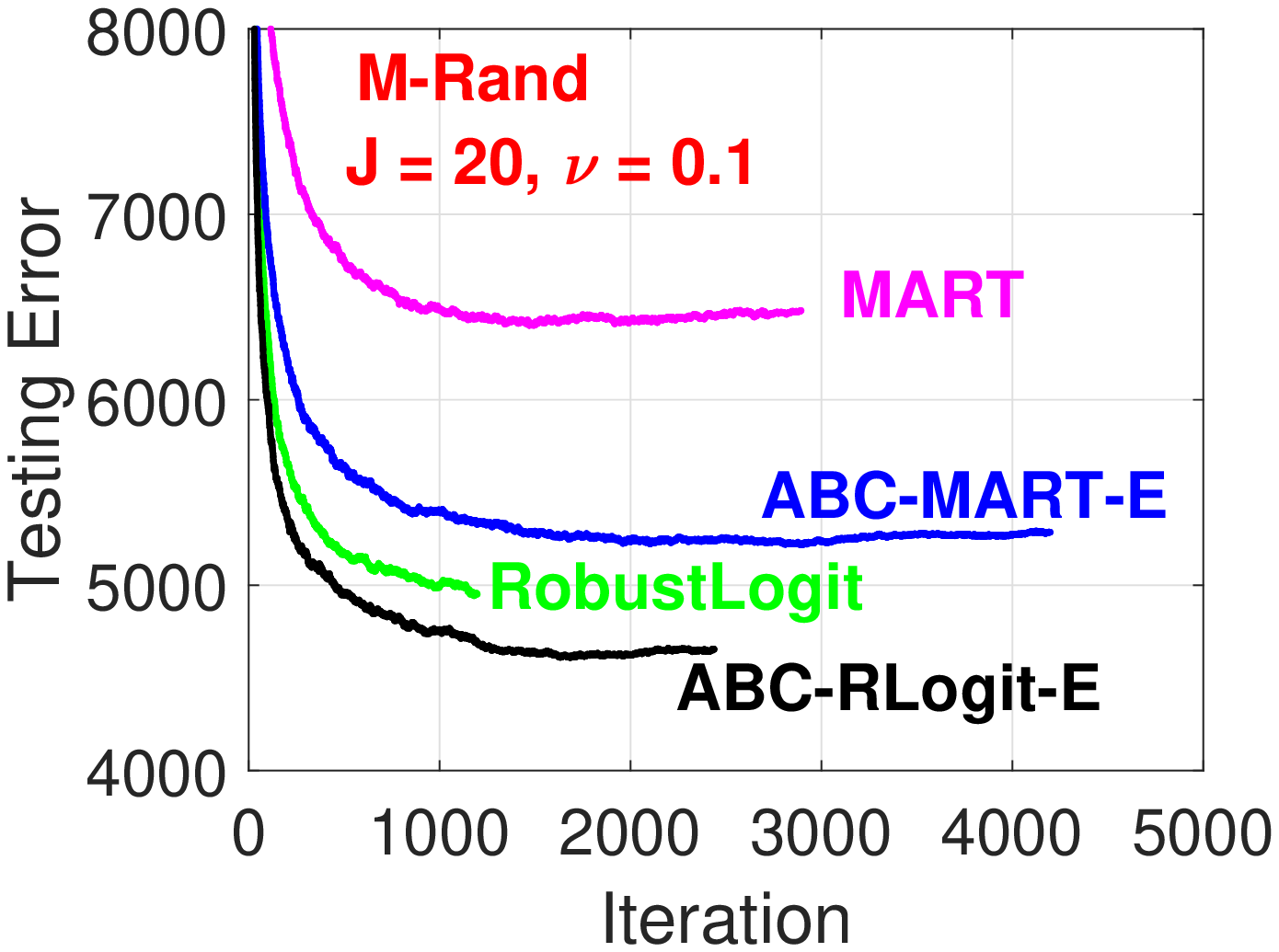}
    \includegraphics[width=2.4in]{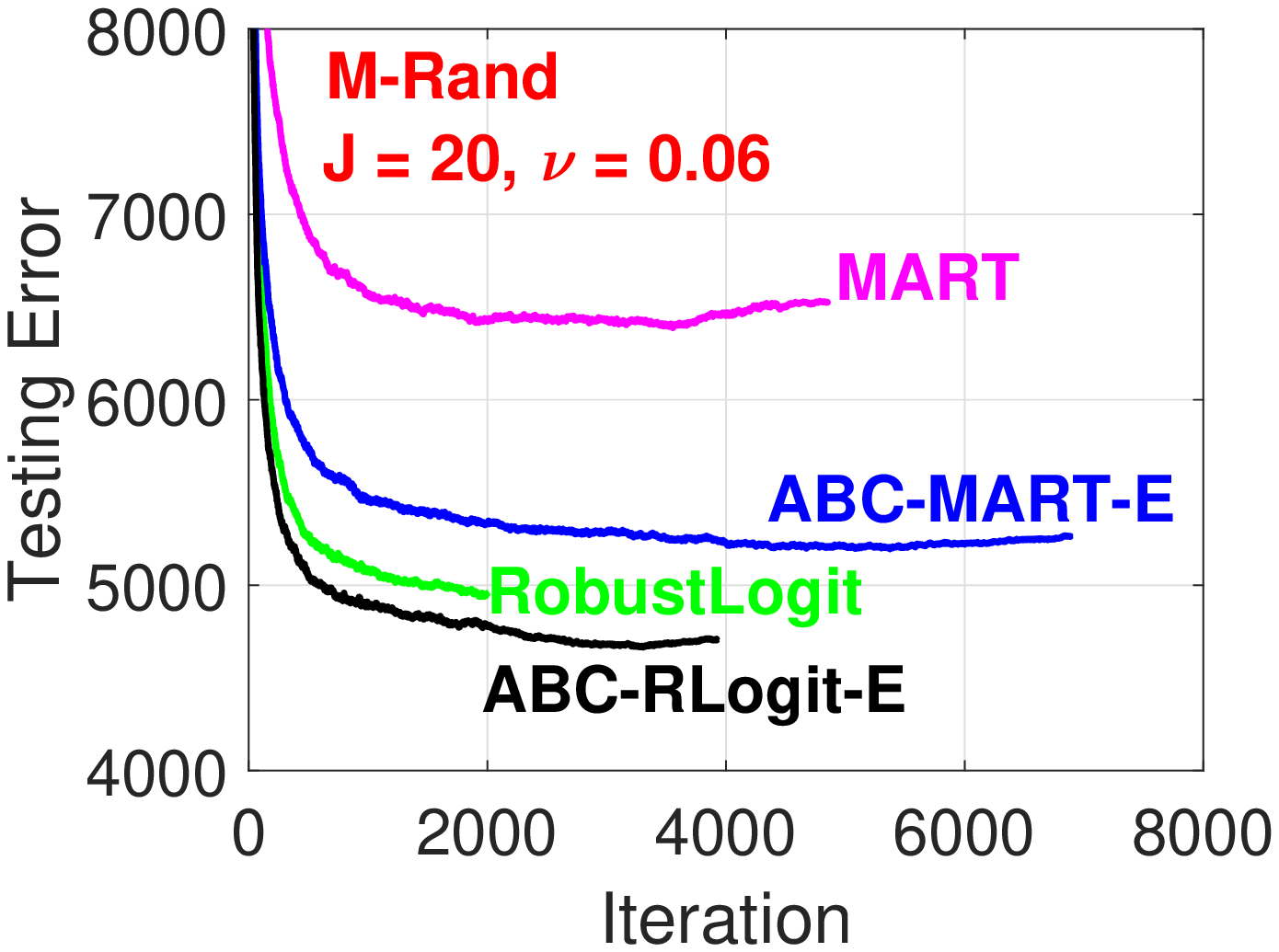}
}
\mbox{
    \includegraphics[width=2.4in]{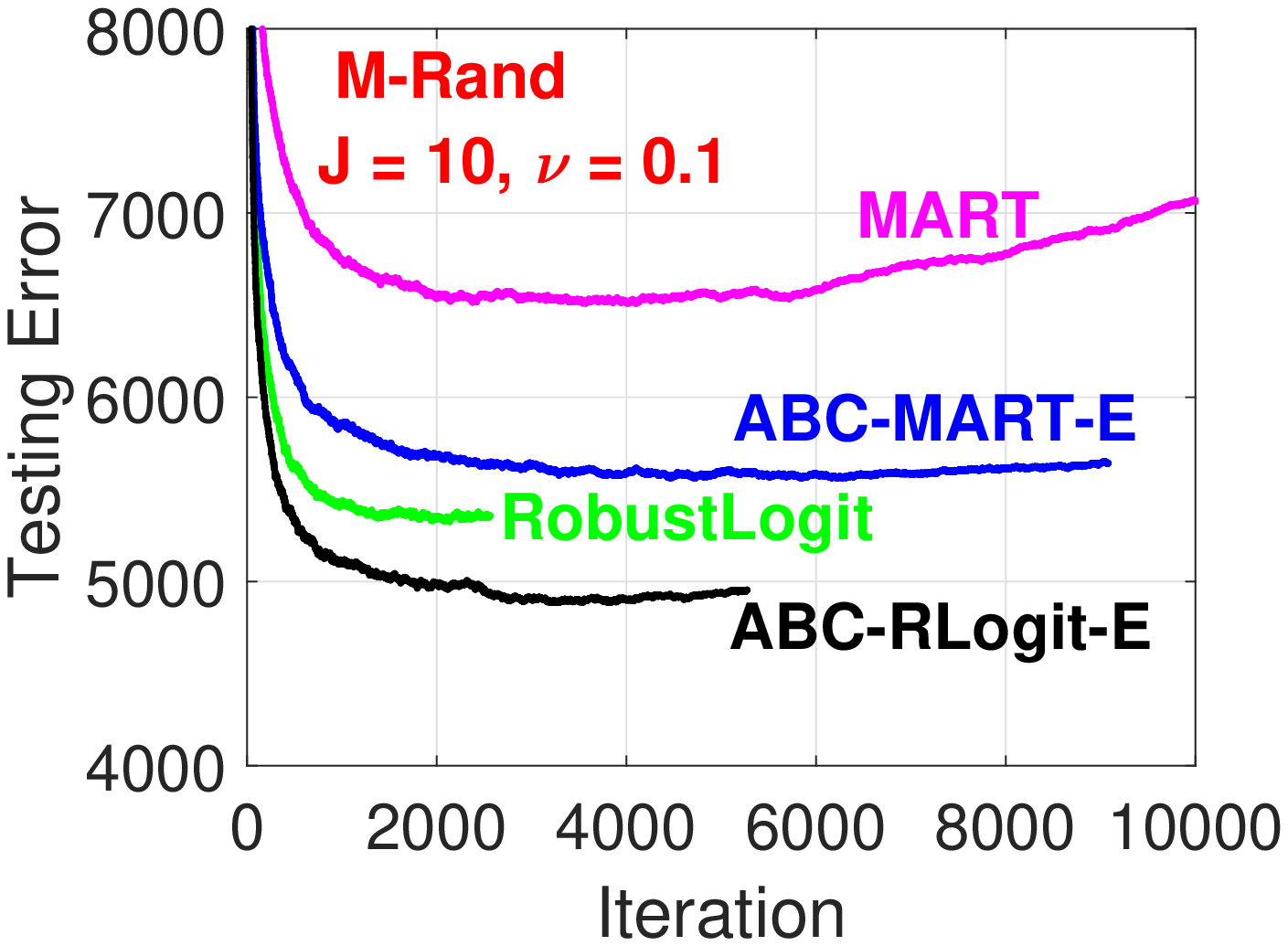}
    \includegraphics[width=2.4in]{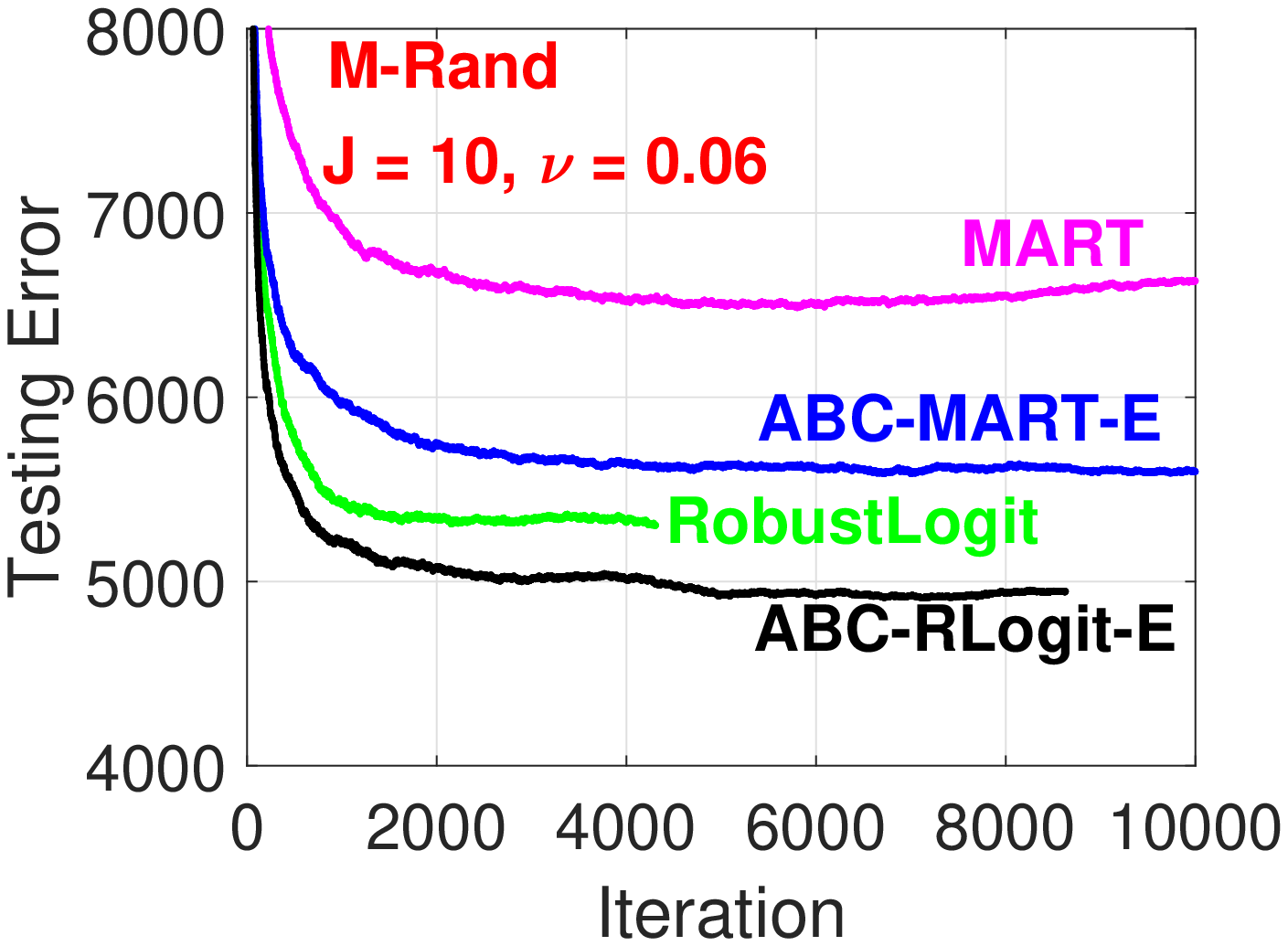}
}

\end{center}

\vspace{-0.1in}

\caption{{\em M-Rand} dataset. We compare the test classification errors for four methods: MART, Robust LogitBoost, ABC-MART, and ABC-RobustLogitBoost, for $J\in \{10, 20\}$ and $\nu \in \{0.06, 0.1\}$. Both ABC methods used the ``exhaustive search'' strategy. }\label{fig:M-Rand}
\end{figure}

\begin{figure}[h]
\begin{center}
\mbox{
    \includegraphics[width=2.4in]{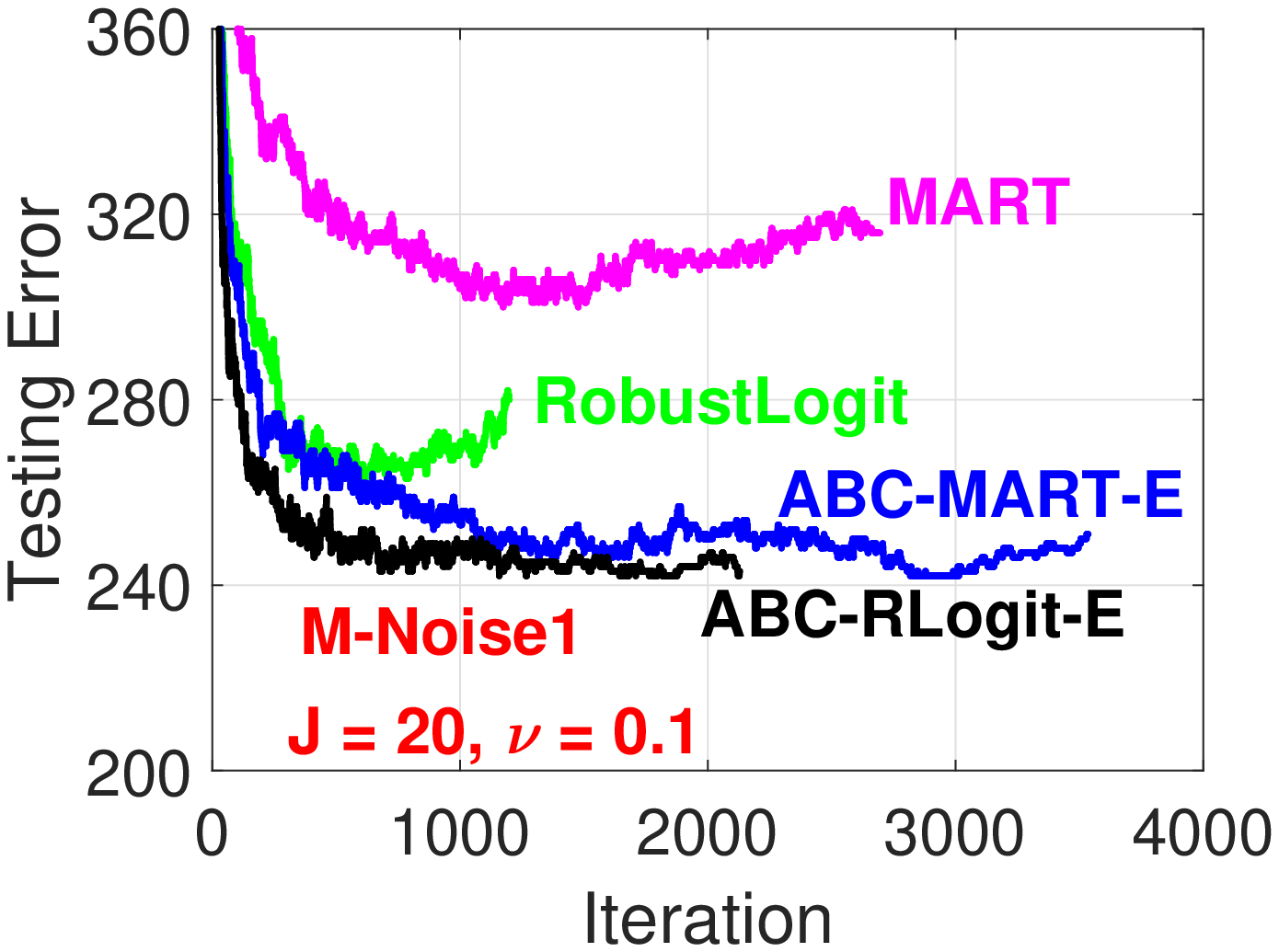}
    \includegraphics[width=2.4in]{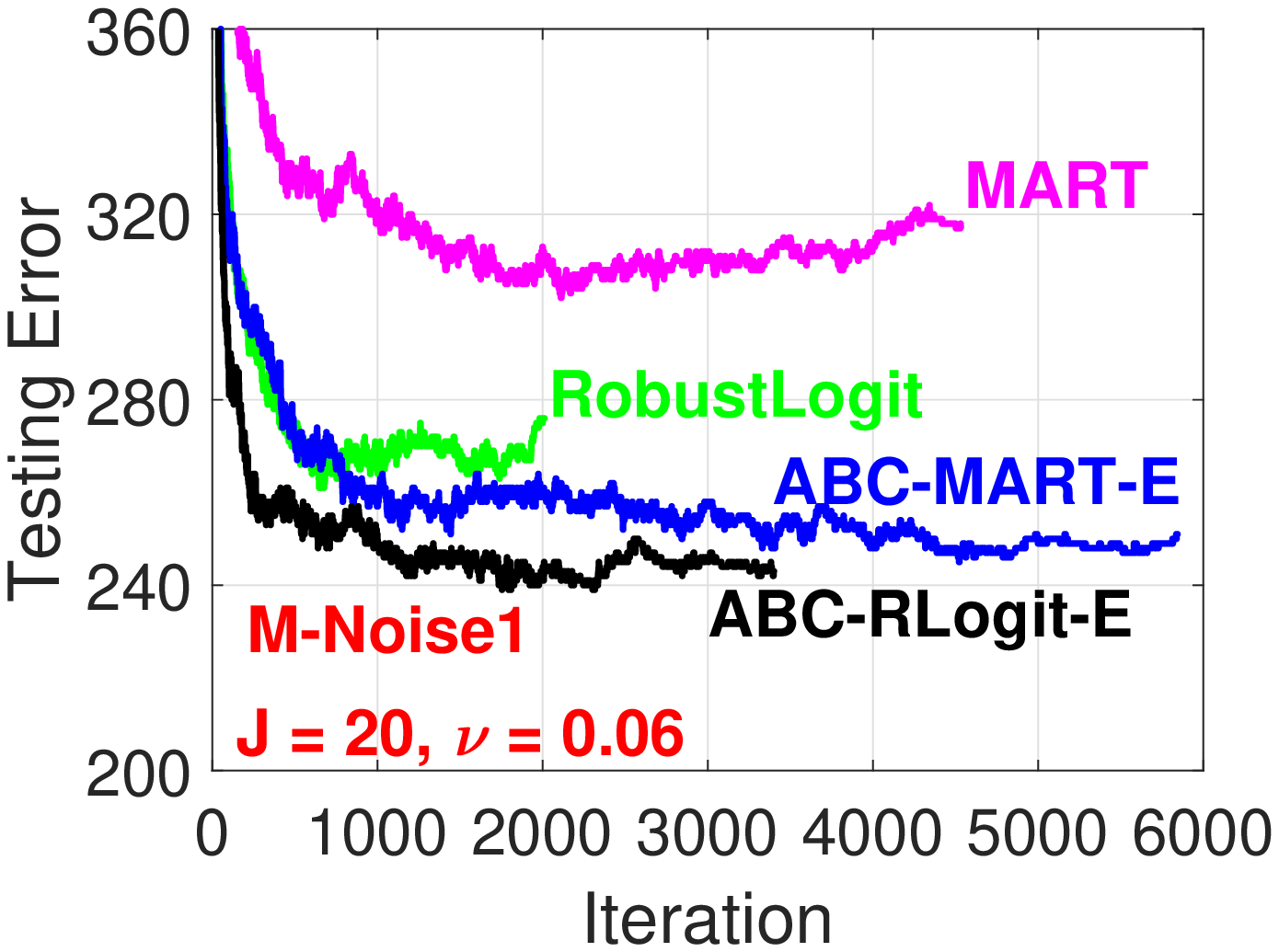}
}
\mbox{
    \includegraphics[width=2.4in]{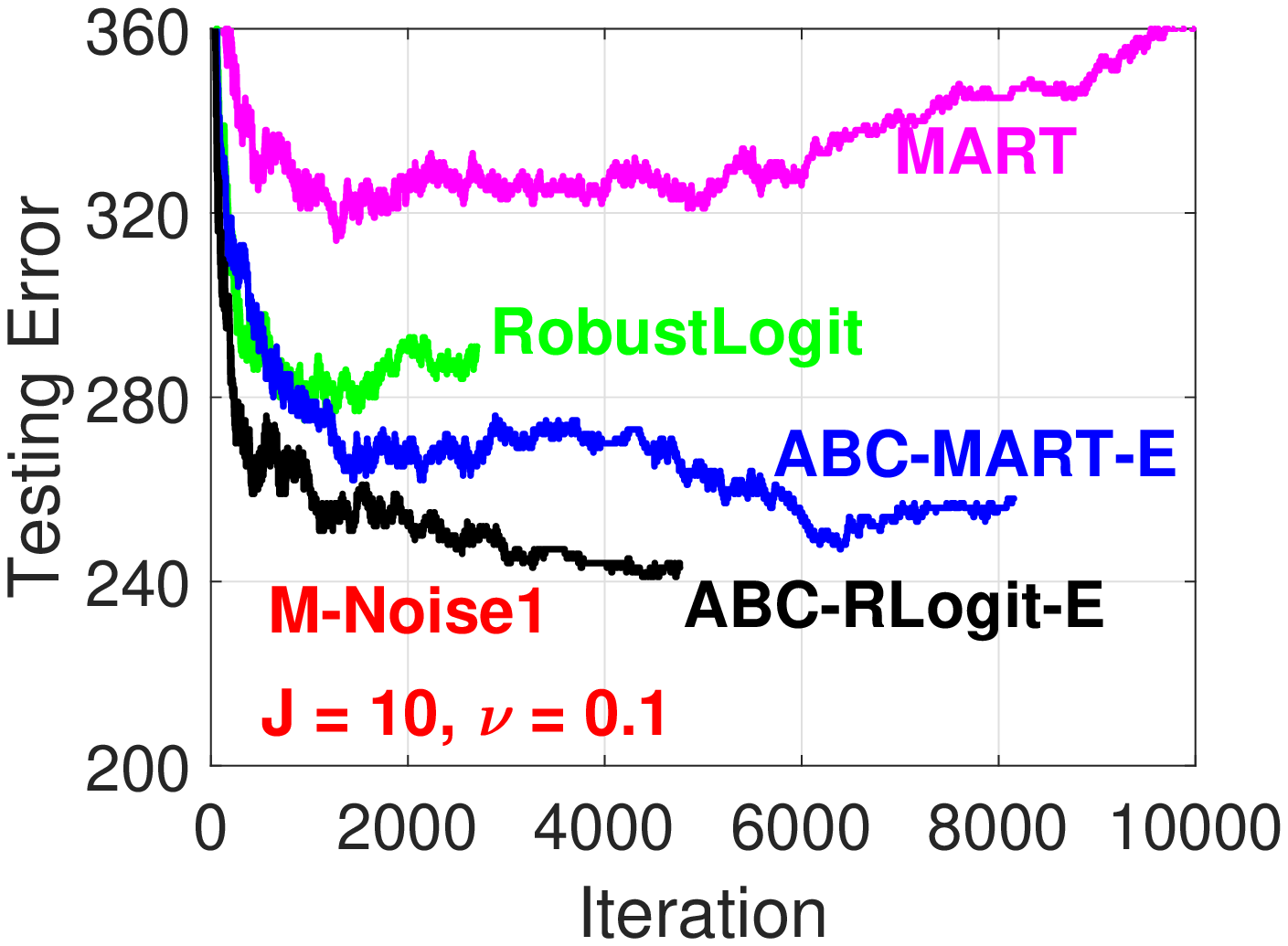}
    \includegraphics[width=2.4in]{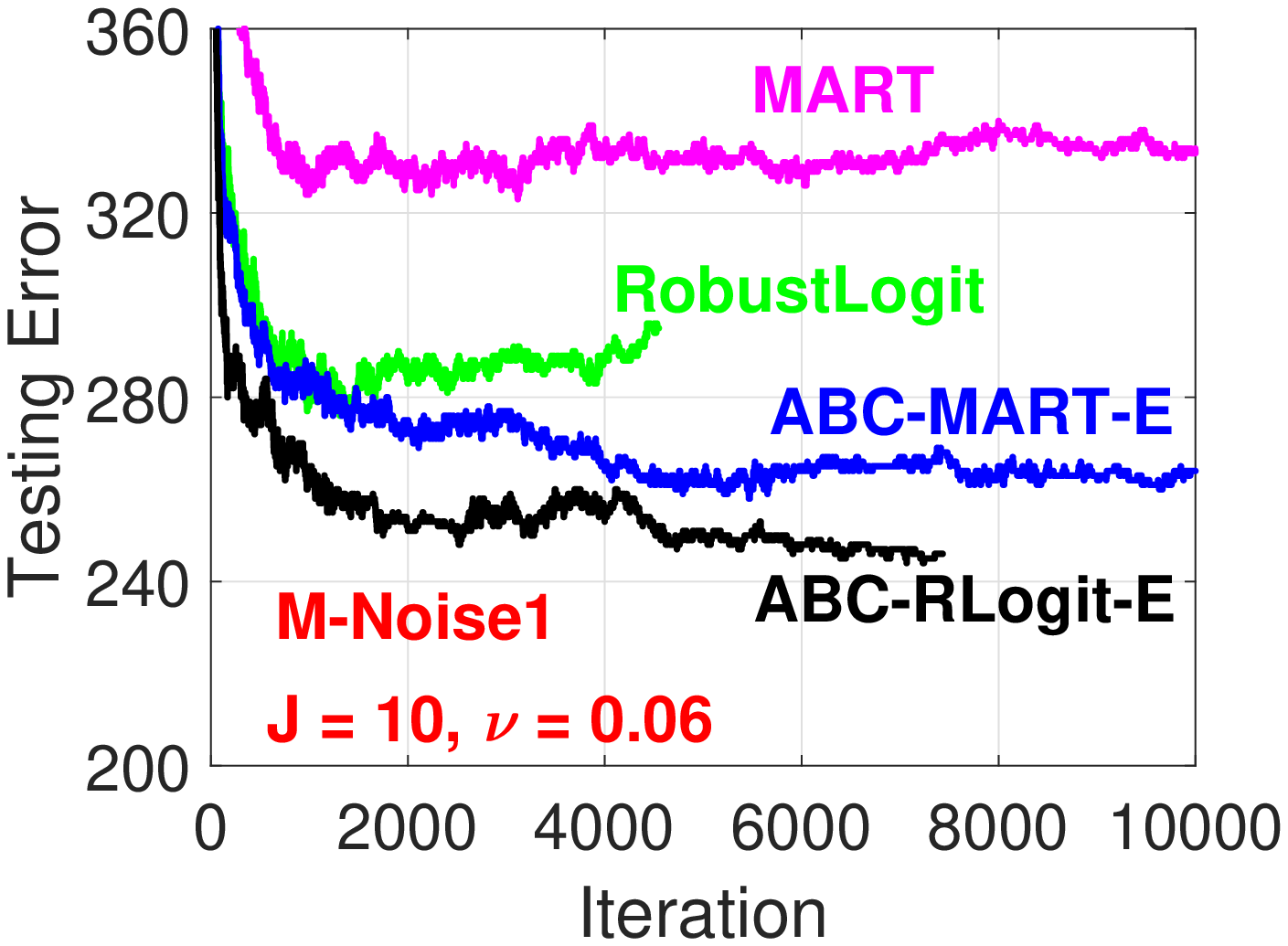}
}

\end{center}

\vspace{-0.1in}

\caption{{\em M-Noise1} dataset. We compare the test classification errors for four methods: MART, Robust LogitBoost, ABC-MART, and ABC-RobustLogitBoost, for $J\in \{10, 20\}$ and $\nu \in \{0.06, 0.1\}$. Both ABC methods used the ``exhaustive search'' strategy. }\label{fig:M-Noise1}
\end{figure}

\begin{figure}[h]
\begin{center}
\mbox{
    \includegraphics[width=2.4in]{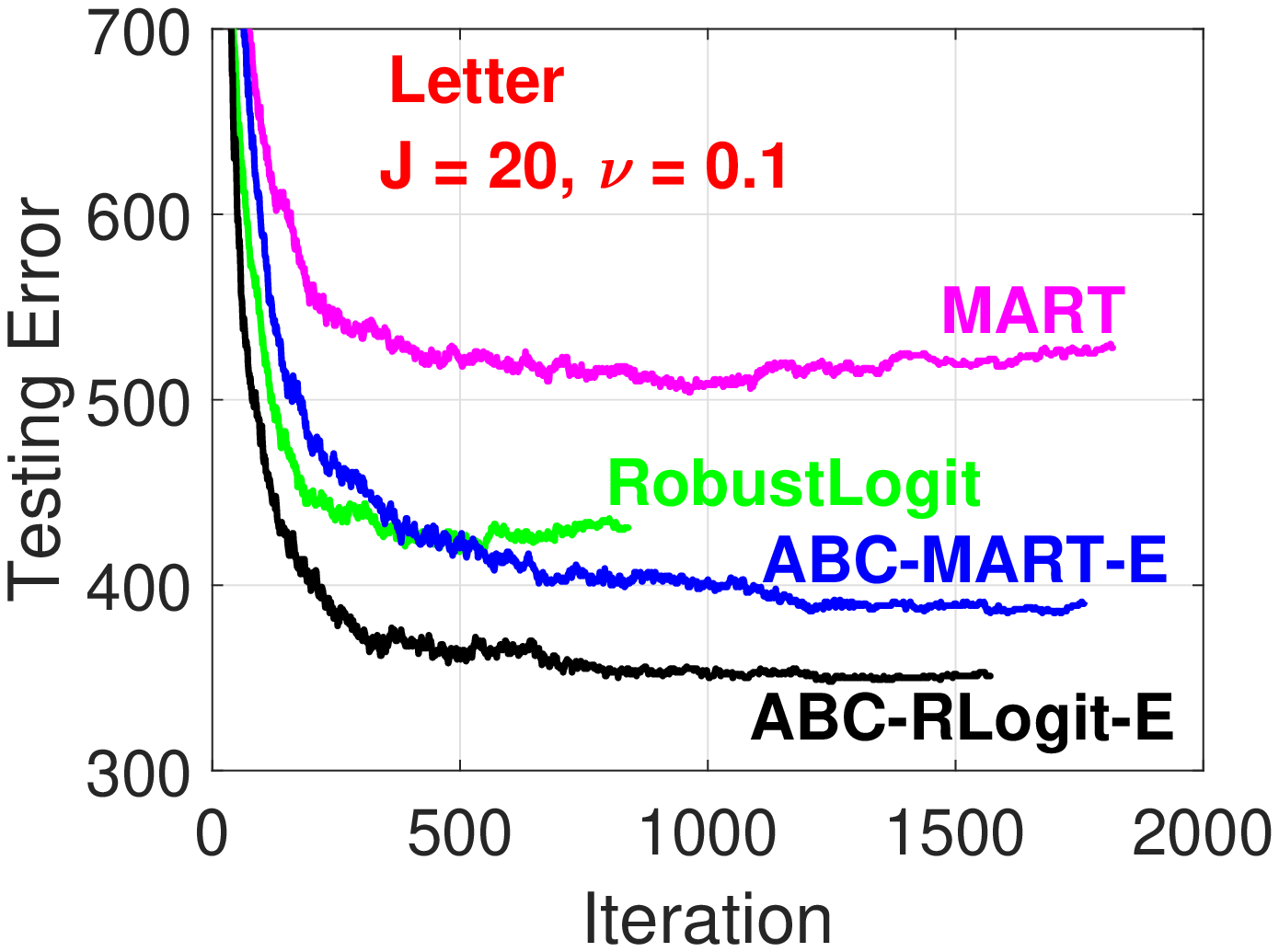}
    \includegraphics[width=2.4in]{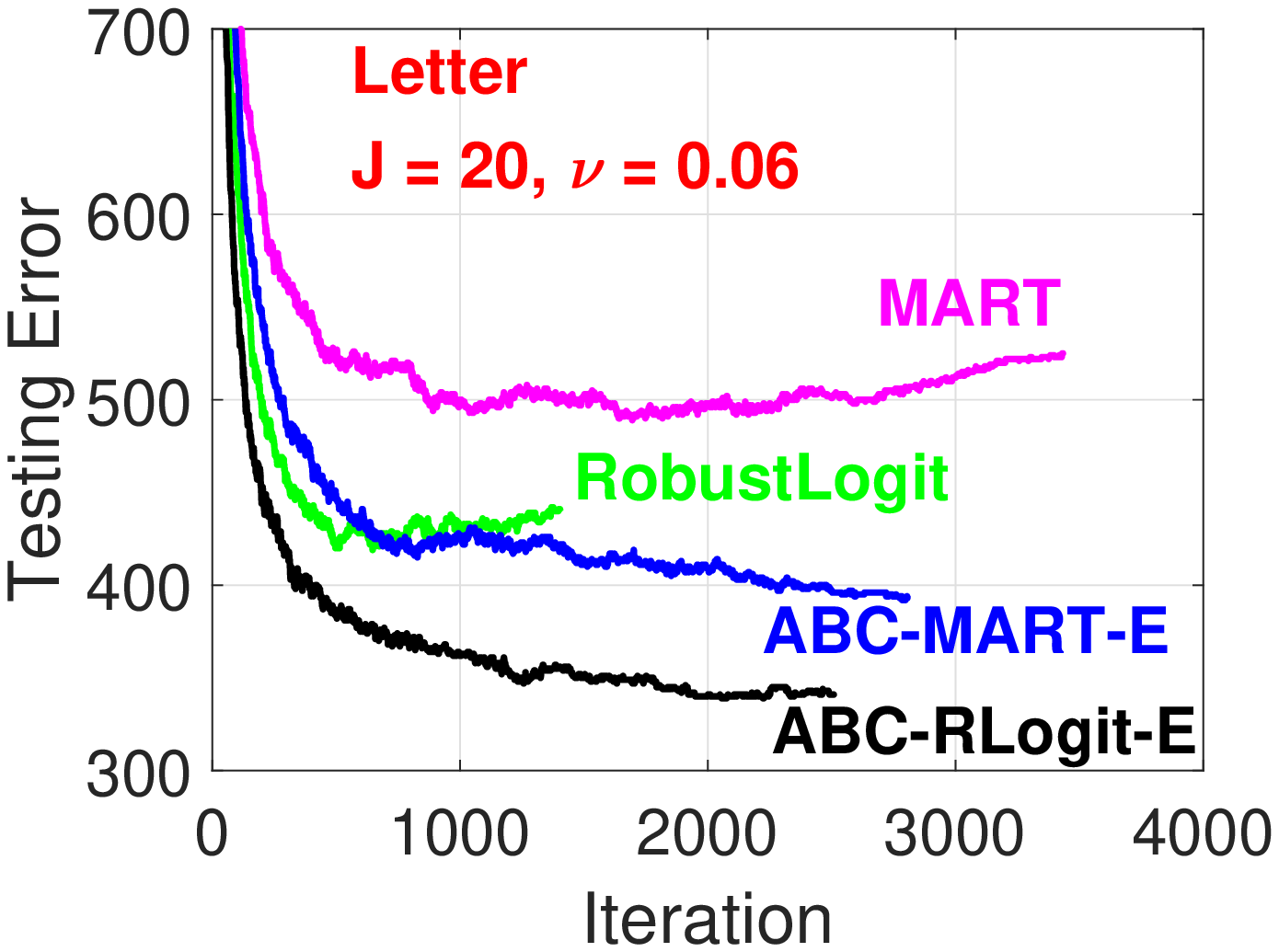}
}
\mbox{
    \includegraphics[width=2.4in]{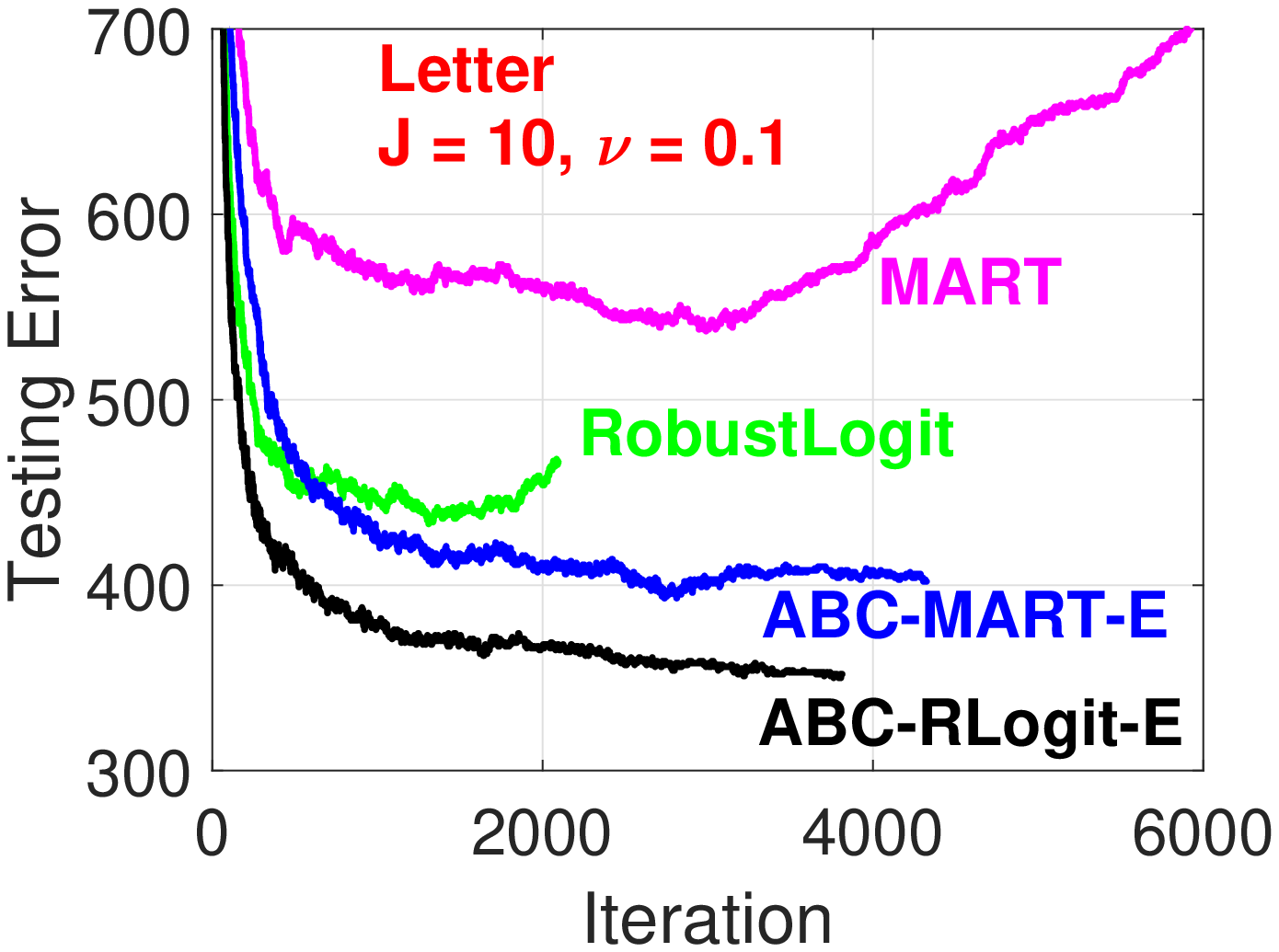}
    \includegraphics[width=2.4in]{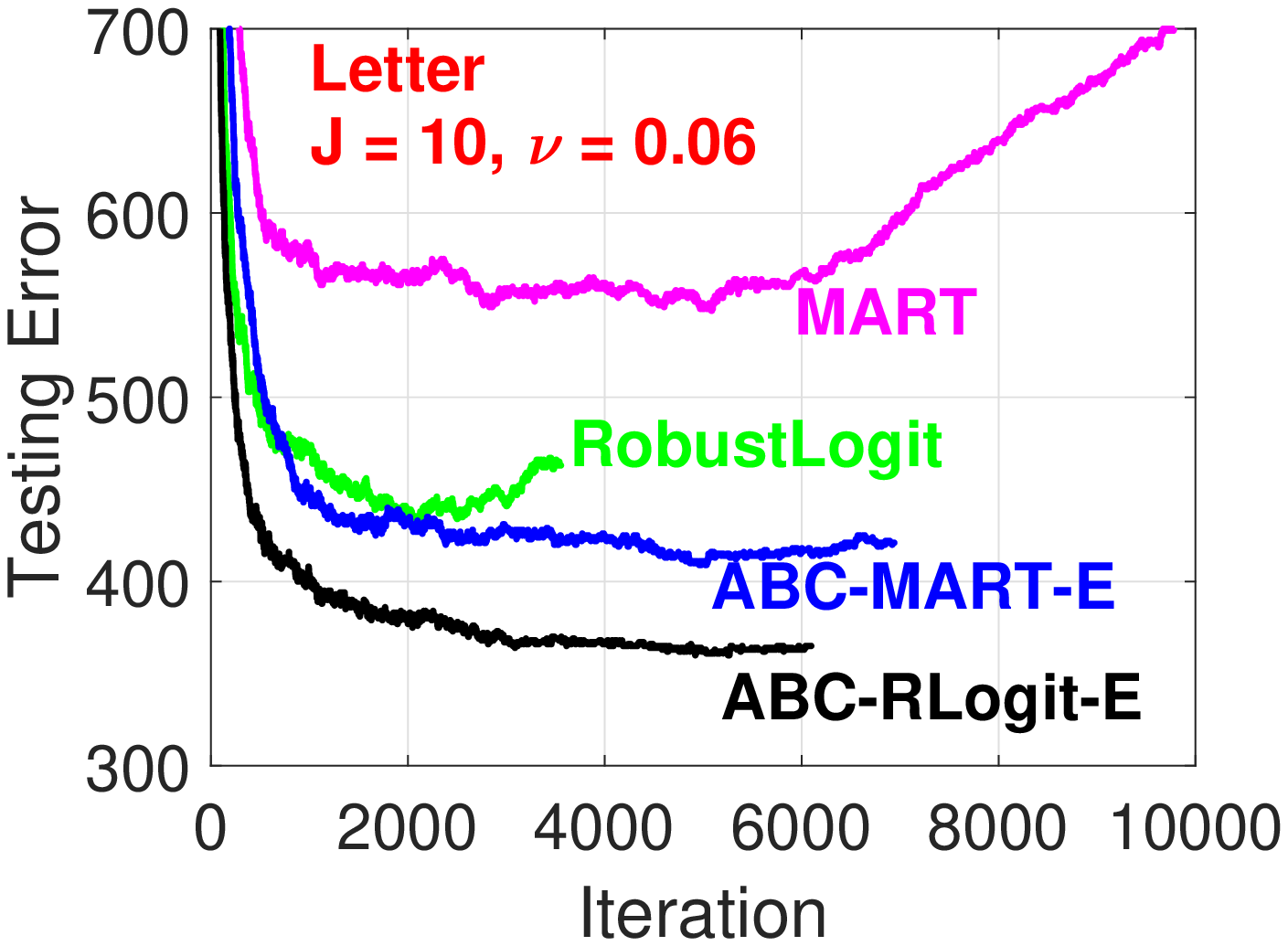}
}

\end{center}

\vspace{-0.1in}

\caption{{\em Letter} dataset. We compare the test classification errors for four methods: MART, Robust LogitBoost, ABC-MART, and ABC-RobustLogitBoost, for $J\in \{10, 20\}$ and $\nu \in \{0.06, 0.1\}$. Both ABC methods used the ``exhaustive search'' strategy. }\label{fig:Letter10k}
\end{figure}

Figures~\ref{fig:Covertype} to~\ref{fig:Letter10k} present the test classification errors (smaller/lower the better) for the 6 datasets, for $J\in \{10, 20\}$ and $\nu\in\{0.06, 0.1\}$. The results confirm that
\begin{itemize}
\item Robust LogitBoost considerably improves MART, due to the use of the second-order information when computing the gains for tree split.
\item ABC-RobustLogitBoost considerably improves the ABC-MART, again due to the use of the second-order information.
\item The exhaustive search strategy works very well in terms of the test accuracy, although it is computationally expensive.
\end{itemize}

\newpage\clearpage

\section{Strategies for Fast ABC-Boost}

Figures~\ref{fig:Covertype} to~\ref{fig:Letter10k} have  demonstrated that the ABC-Boost works well with the ``exhaustive search'' strategy for the base class, which is computationally very expensive. Assume $K$ classes and $M$ iterations, the training cost of the original MART and RobustLogitBoost would be just $O(KM)$. However, the cost of ABC-Boost with the ``exhaustive search'' strategy becomes $O(K(K-1)M)$. We need better (more efficient) methods to search for the base class.

\subsection{The ``Worst Class'' Strategy and the ``$s$-Worst Classes'' Strategy}

The ``exhaustive search'' strategy, i.e., trying every class as the base class candidate and using the best one as the base class for the \textbf{current iteration}, is an obvious idea and intuitively should work well even though it is very expensive.  In comparison, the ``worst class'' idea might be less intuitive. That is, at each iteration, we identify the ``worst class'' in some measure (such as the one with the largest training loss) and use it for the \textbf{next iteration}. This basically means that we assign the hardest work to the next iteration.  The ``worst class'' idea has been made public through technical reports, lecture notes, and talks, e.g.,~\cite{li2008adaptive}, but it was not formally published.

Recall that, we assume $N$ total number of training samples $\{x_i, y_i\}_{i=1}^N$, where the labels $y_i\in\{0, 1, ..., K-1\}$. We let the indicator $r_{i,k} = 1$ if $y_i = k$, and $r_{i,k} = 0$ if $y_i \neq k$. At the end of each boosting iteration, we compute the current class probabilities, $p_{i,k}$, for each data point, and then compute the total loss for each class, i.e.,
\begin{align}
L^{(k)} = -\sum_{i=1}^N \sum_{k=0}^{K-1} r_{i,k}\log\left(p_{i,k}\right),\hspace{0.3in} k = 0, 1, ..., K-1.
\end{align}
With the ``worst class'' strategy, we choose the class $b$ which  maximizes the loss, i.e.,
\begin{align}
b = \underset{0\leq k \leq K-1}{\text{argmin}} L^{(k)}
\end{align}

Again, assume the total number of boosting iterations is $M$. Then the computational cost of the ``worst class'' strategy would be $((K-1)M)$, which is actually more efficient than $O(KM)$, the cost of MART and Robust LogitBoost.

\vspace{0.1in}

The ``worst class'' strategy works pretty well in most cases but in general its test accuracy would be somewhat worse than that of the ``exhaustive search'' strategy. Also, in some cases we observe ``catastrophic failures'' when using the ``worst class'' strategy. This means we cannot really recommend the ``worst class'' strategy to the users. In comparisons, the ``exhaustive search'' strategy is very robust (albeit very slow) and we have not observed any ``catastrophic failures'' after using it for over 10 years. Figure~\ref{fig:Letter10k_s} provides an example of ``catastrophic failure'' on the {\em Letter} dataset when using the ``worst class'' strategy, while the ``exhaustive search'' strategy still works well on that dataset.

\vspace{0.1in}

Therefore, we have two strategies at the extremes: one  is very fast and not always reliable, the other  is very slow and robust. This motivates us to develop the ``$s$-worst classes'' strategy. That is, at the beginning of every iteration, we sort the losses of all classes: $L^{(k)}$, $k = 0, 1, ...., K-1$ and identify the $s$ classes with the largest losses. With each of the chosen $s$ classes, we use it as the base class and conduct the training. After all $s$ classes have been tried, we choose the one with the smallest training loss as the base class for the \textbf{current iteration}.

\newpage\clearpage

Obviously, when $s=K$, the ``$s$-worst classes'' strategy becomes the ``exhaustive search'' strategy. If $s=1$, then at the beginning of some iteration, it can only use the ``worst class'' from the previous iteration as the base class to conduct the training for that iteration. This means that,  when $s=1$, it actually recovers the ``worst class'' strategy.

Using the ``$s$-worst classes'' strategy, the computational cost would be $O(s(K-1)M)$. If $s$ is not large, for example, $s=2$ or $s=3$, then the cost would not increase much compared with MART and Robust LogitBoost. As we can see from  Figures~\ref{fig:Covertype_s} to~\ref{fig:Letter10k_s} which present the experimental results on the 6 datasets in Table~\ref{tab:data}, in general using $s=K$ (i.e., ``exhaustive search'') leads to better test accuracy than using $s=1$ (i.e., ``worst class'') but the difference is usually not too big. However, in some cases (e.g., the {\em Letter} dataset) using $s=1$ may lead to ``catastrophic failure'' for some boosting parameters.  Interestingly, on these datasets, if we let $s= 2\sim 4$, the results are pretty close to the best accuracy. Another interesting phenomenon is that the best test accuracy does not necessarily occur at $s=K$. Hence, a good practice may be treating $s$ as a tuning parameter and start with $s=2\sim 4$.

\subsection{The Strategy of Introducing the ``Gap'' Parameter}

The cost of the ``$s$-worst classes'' search strategy would be $O(s(K-1)M)$, which is still roughly $s$ times larger than the cost of the original MART or Robust LogitiBoost. Even if $s$ is just as small as $2\sim 4$, it is still a non-negligible increase of the training cost. This motivates us to introduce the ``gap'' parameter $g$. That is, we only conduct the search at every $(g+1)$ iterations. $g=0$ means no gap. The cost of ABC-Boost with parameters $g$ and $s$ becomes $O\left(s(K-1)\frac{1}{g+1}M + (K-1)\frac{g}{g+1}M\right)$.

The idea of using ``gap'' is also quite natural. Intuitively, once a base class is chosen and used for training one iteration, it is probably still a pretty good choice as the base class (not necessarily the most ``optimal'') for the next a few iterations. If the cost of searching for the base class is expensive, we might as well delay the search until the previously chosen base class is no longer helpful.  Algorithm~\ref{alg:fast-abc-LogitBoost} summarizes the proposed Fast ABC-RobustLogitBoost method with parameters $s$ and $g$. See the implementation details in~\cite{li2022package}. We also provide a report for using this package on regression tasks~\citep{li2022pGMM}.

\vspace{0.1in}

\begin{figure}[h]
\begin{center}
\mbox{
    \includegraphics[width=2.2in]{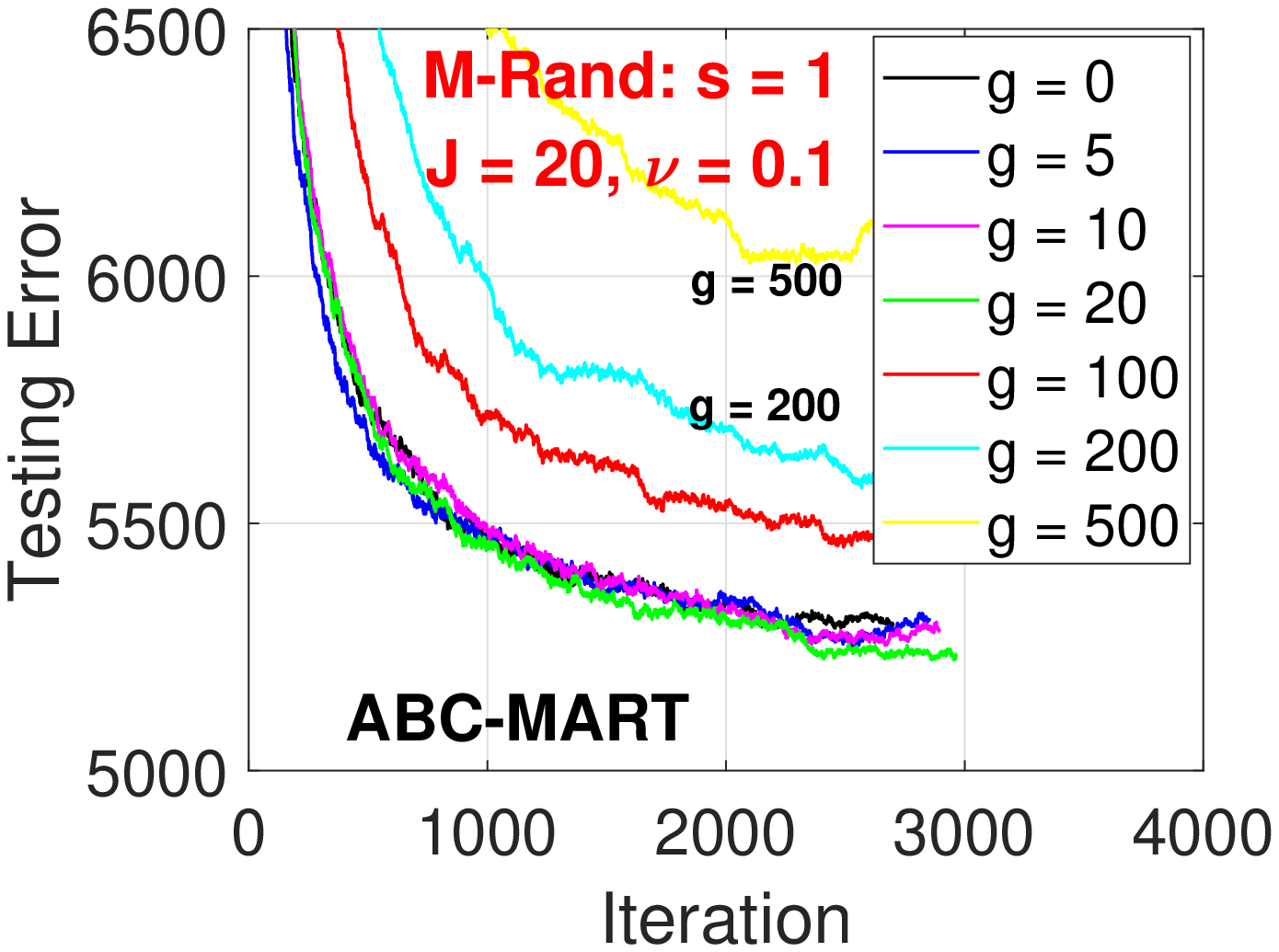}
    \includegraphics[width=2.2in]{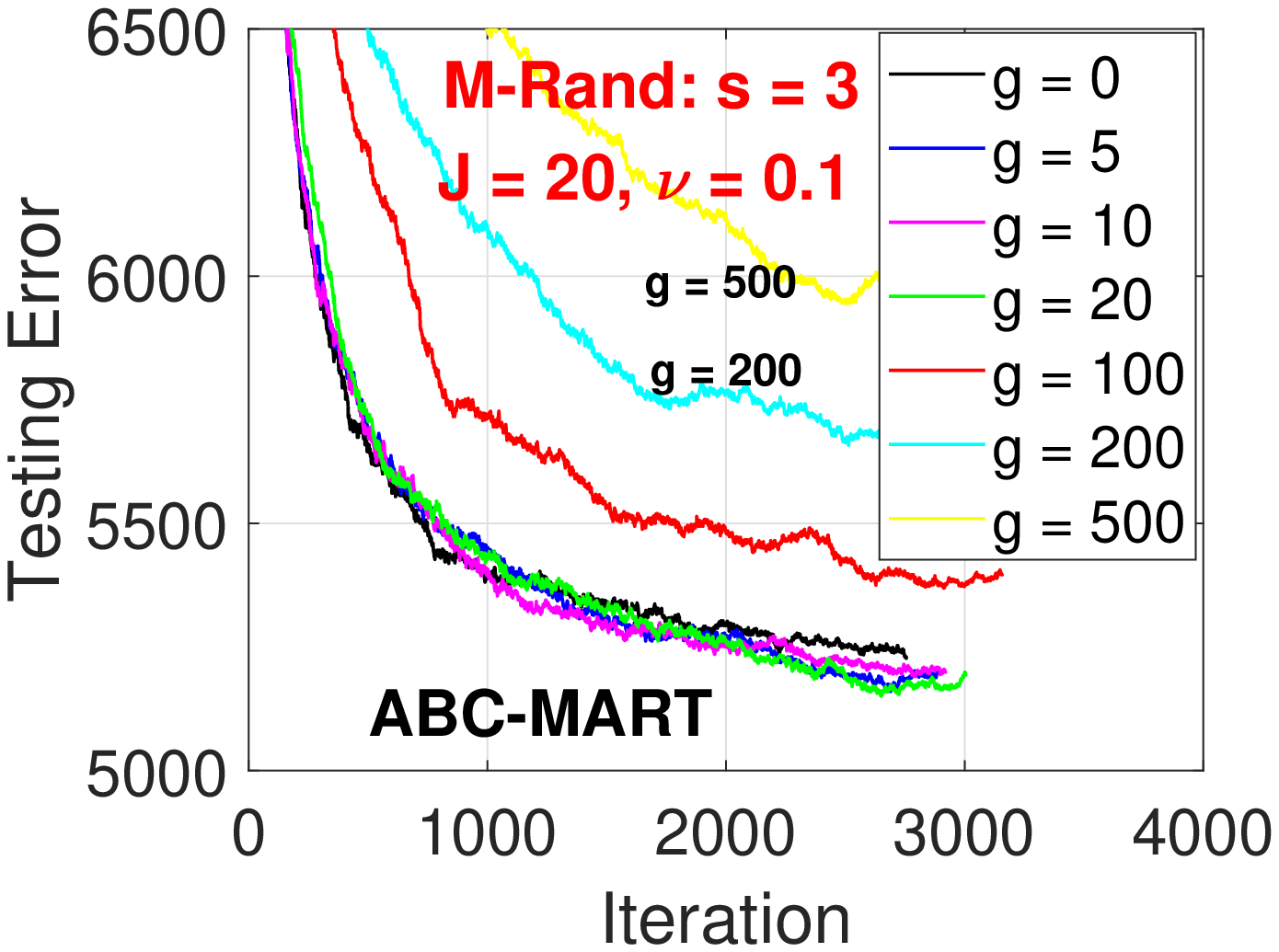}
    \includegraphics[width=2.2in]{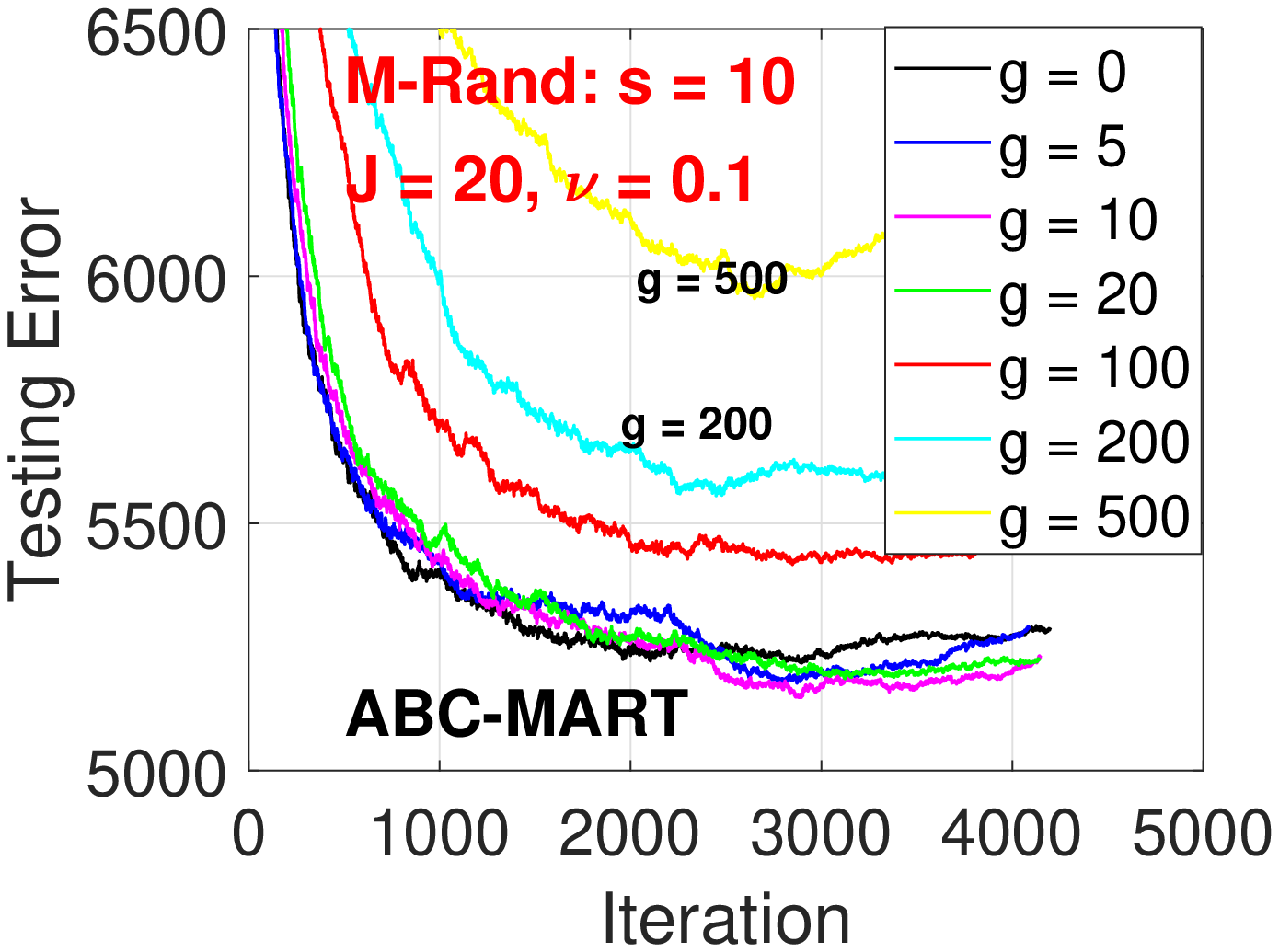}
}

\mbox{
    \includegraphics[width=2.2in]{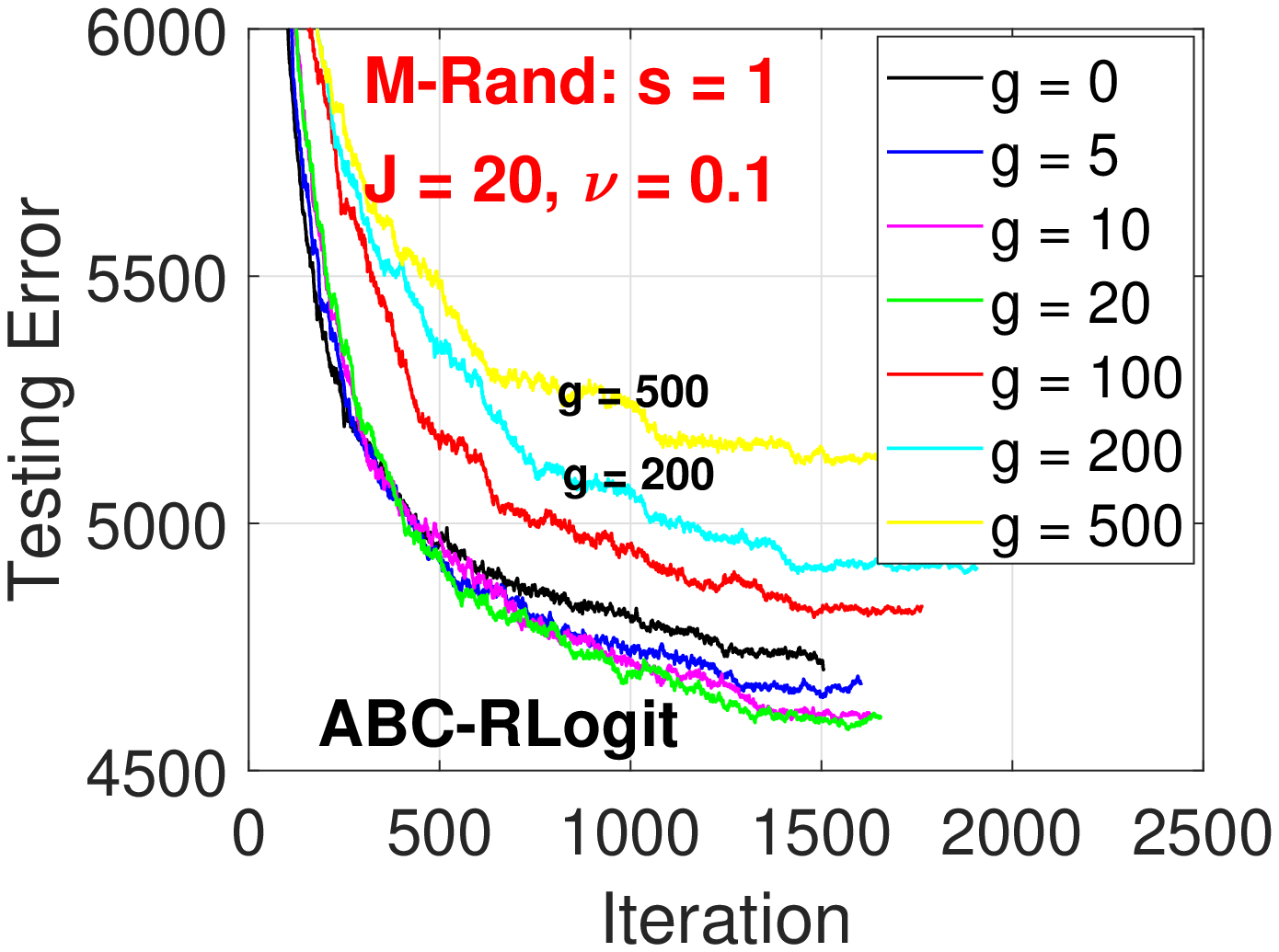}
    \includegraphics[width=2.2in]{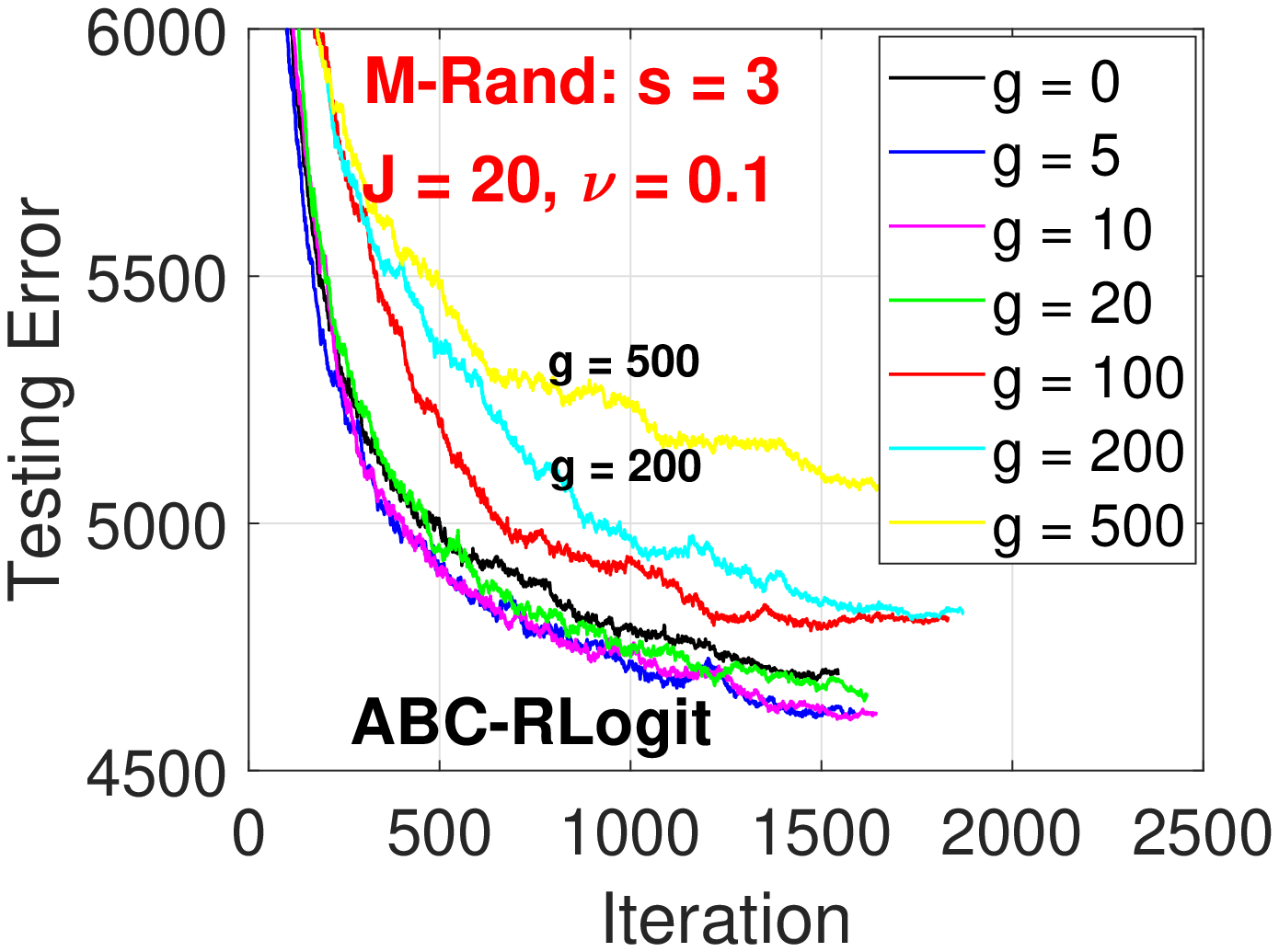}
    \includegraphics[width=2.2in]{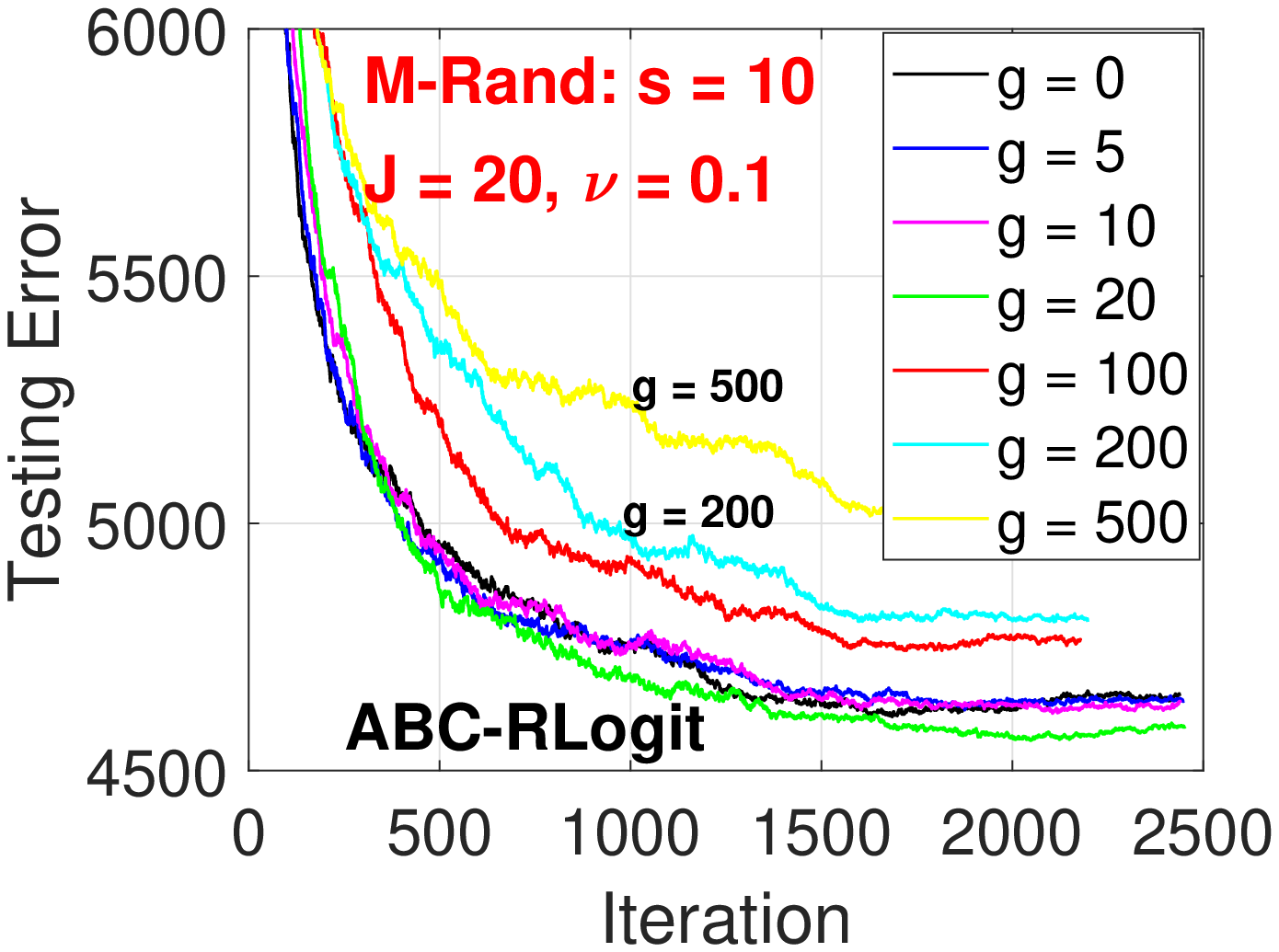}
}

\mbox{
    \includegraphics[width=2.2in]{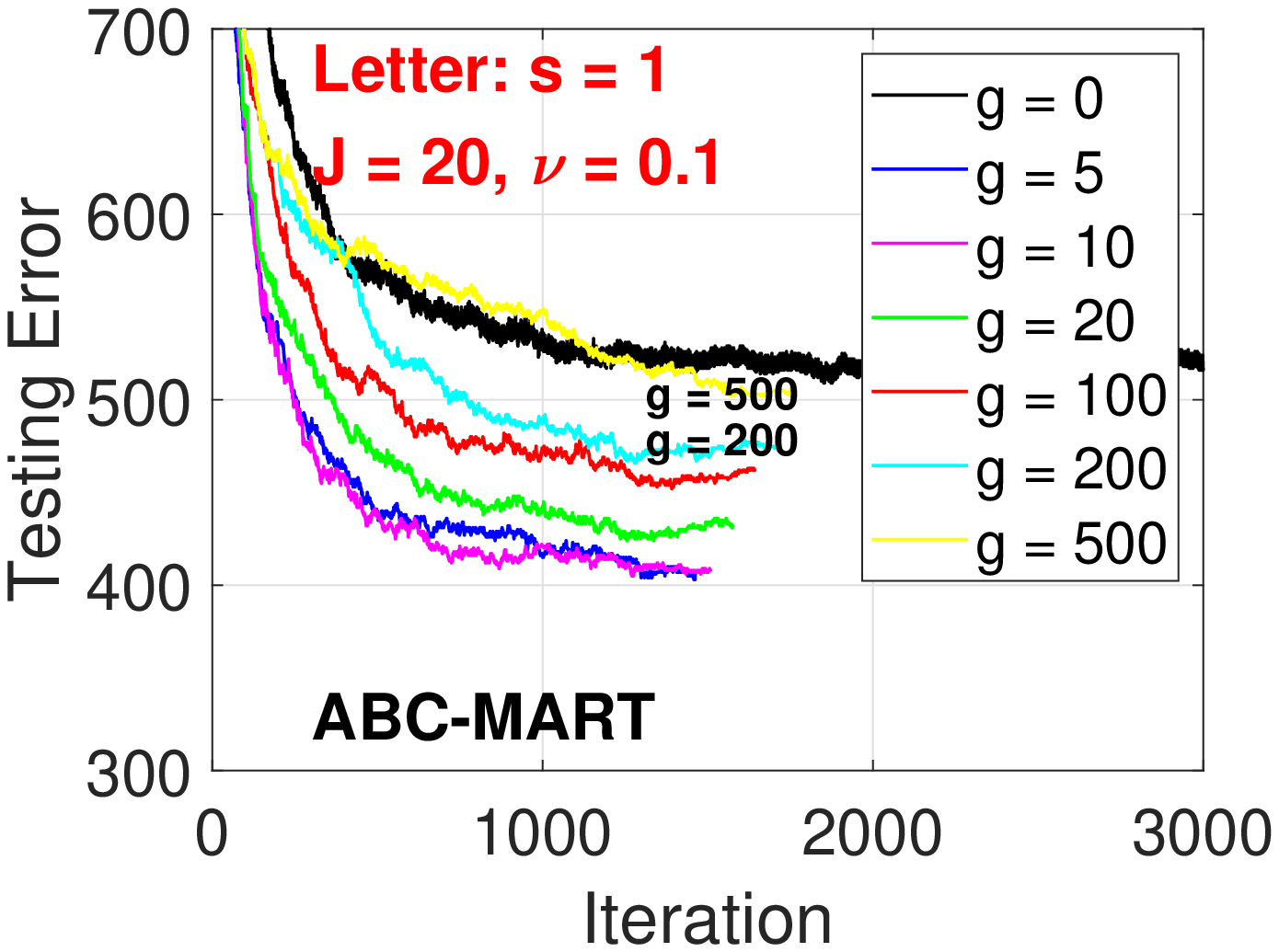}
    \includegraphics[width=2.2in]{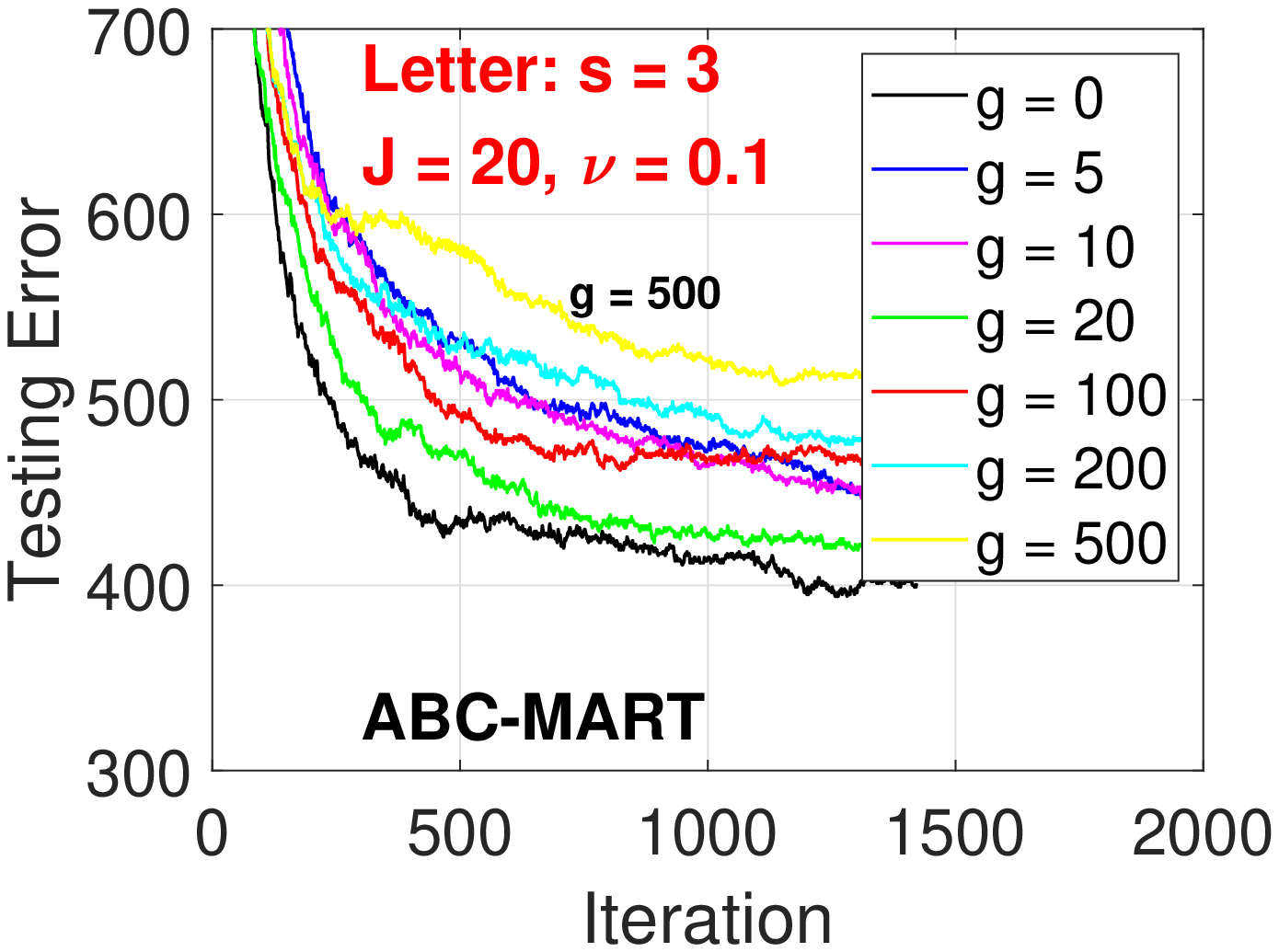}
    \includegraphics[width=2.2in]{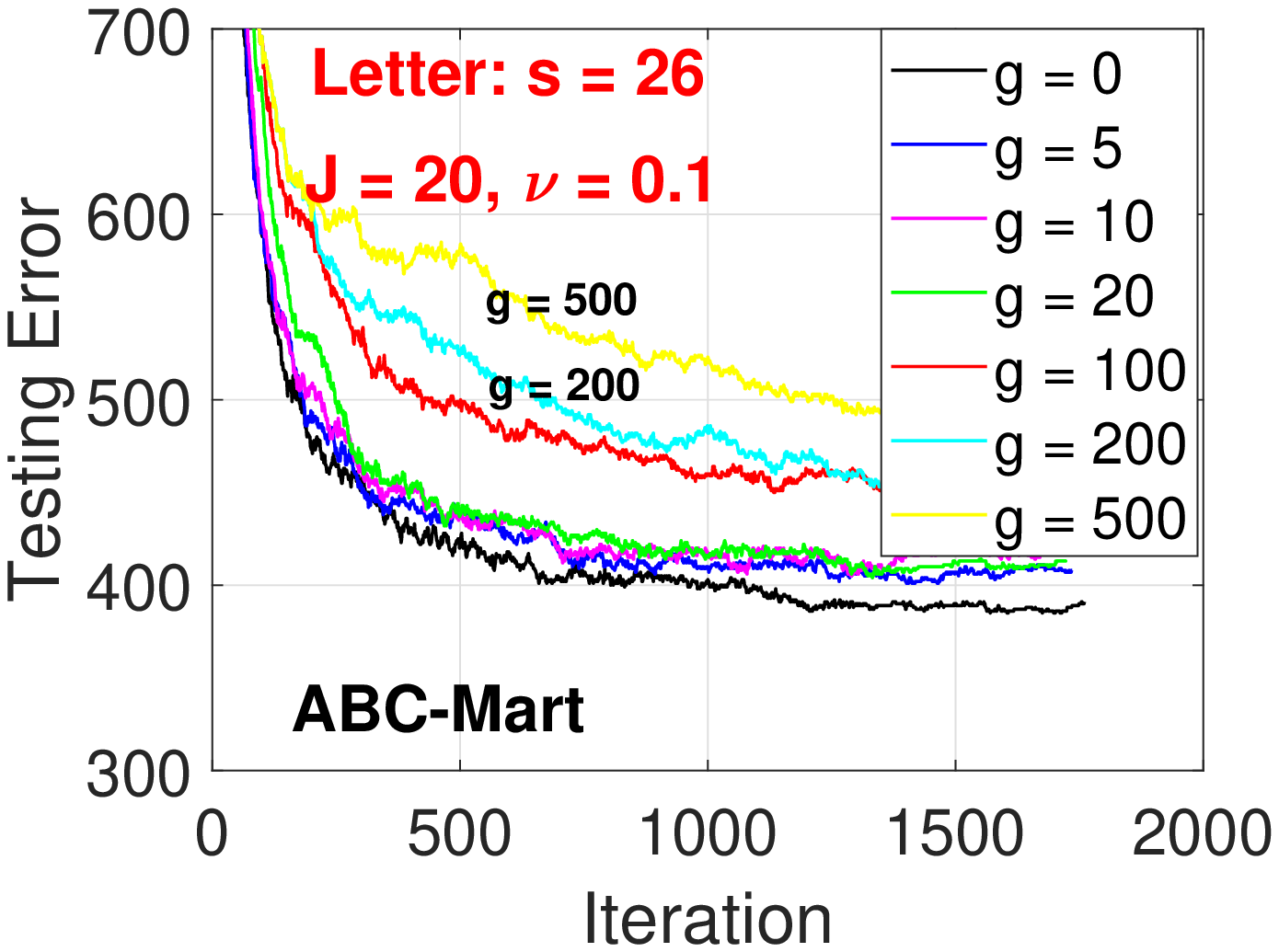}
}

\mbox{
    \includegraphics[width=2.2in]{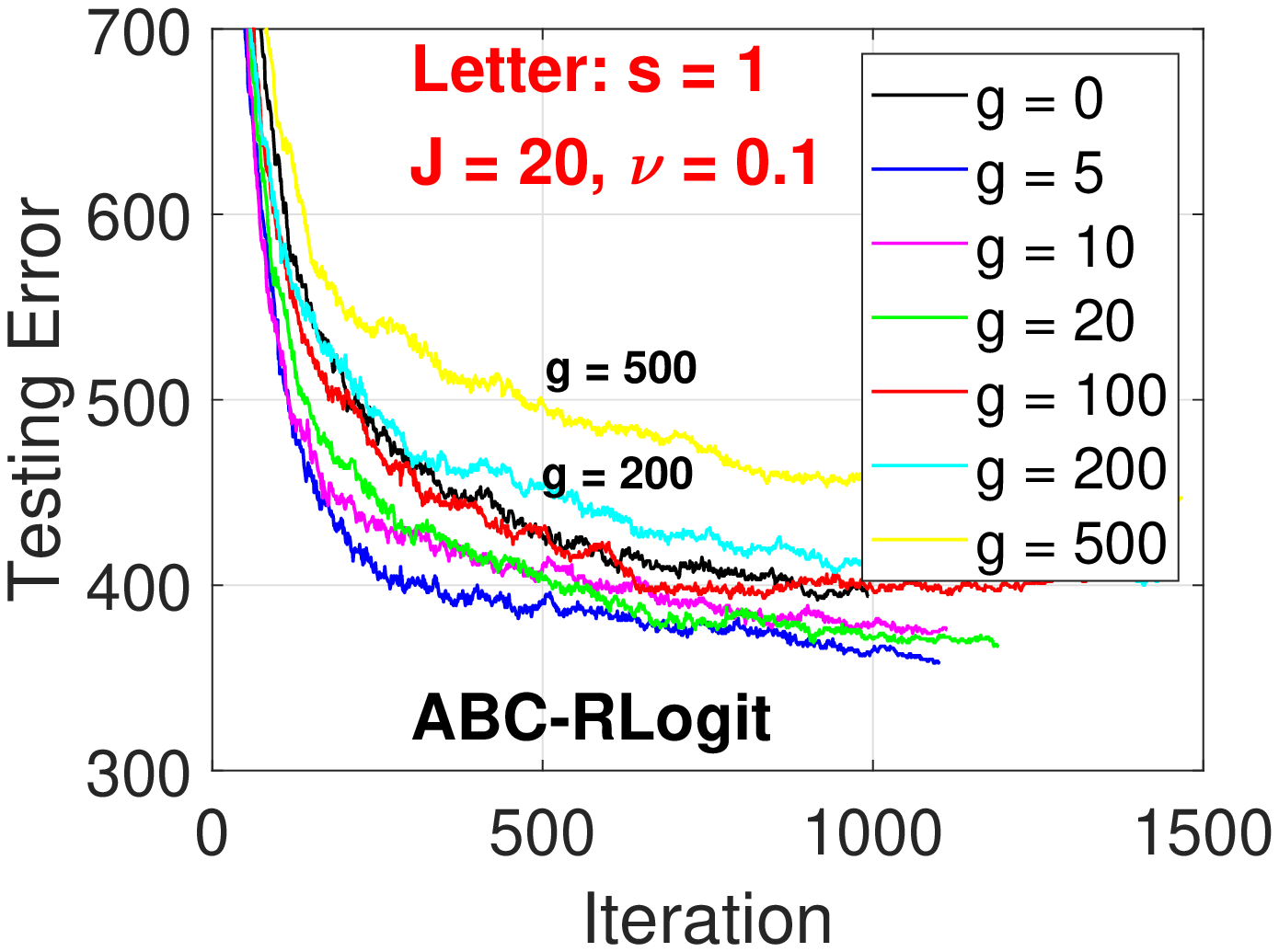}
    \includegraphics[width=2.2in]{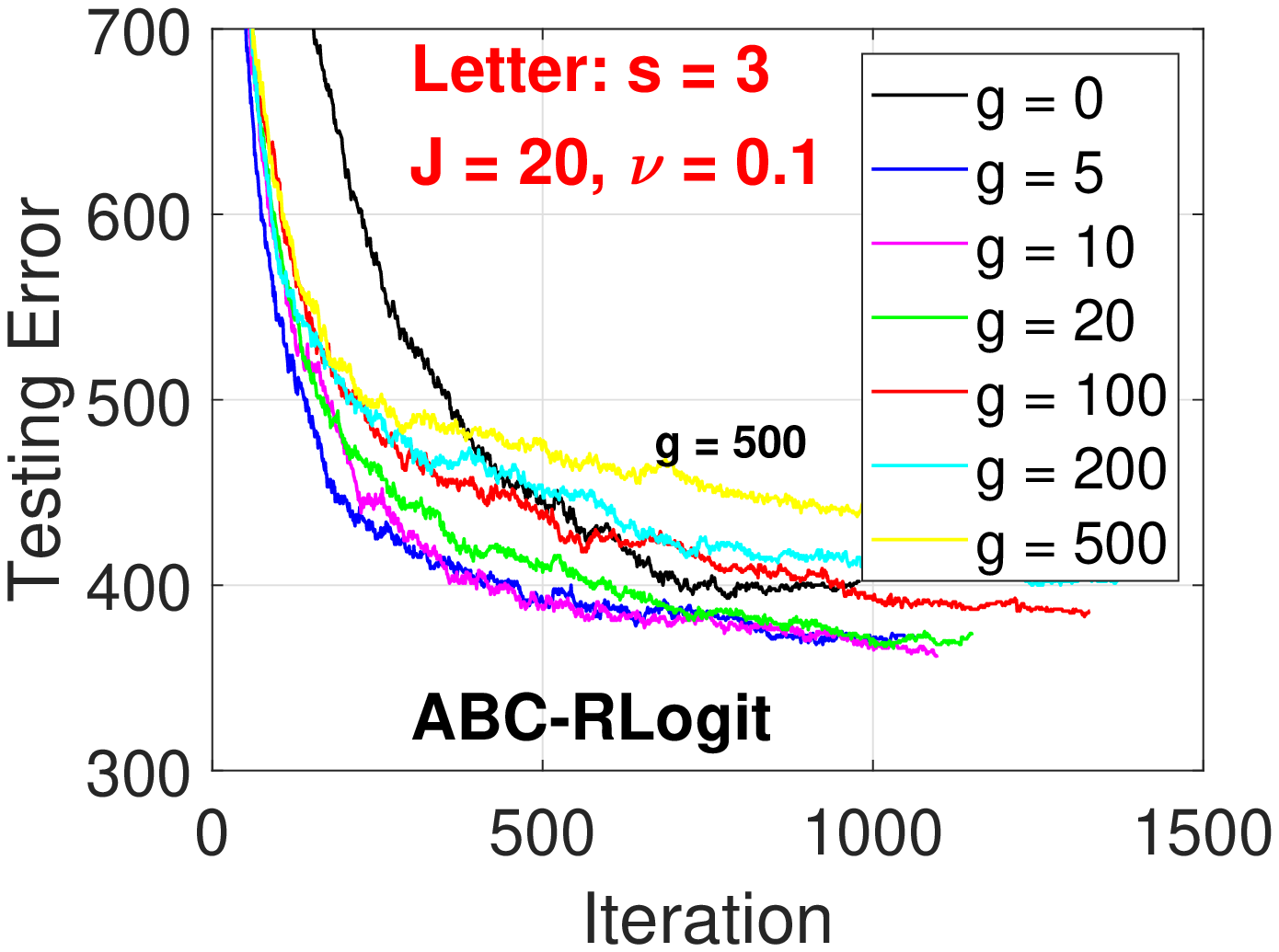}
    \includegraphics[width=2.2in]{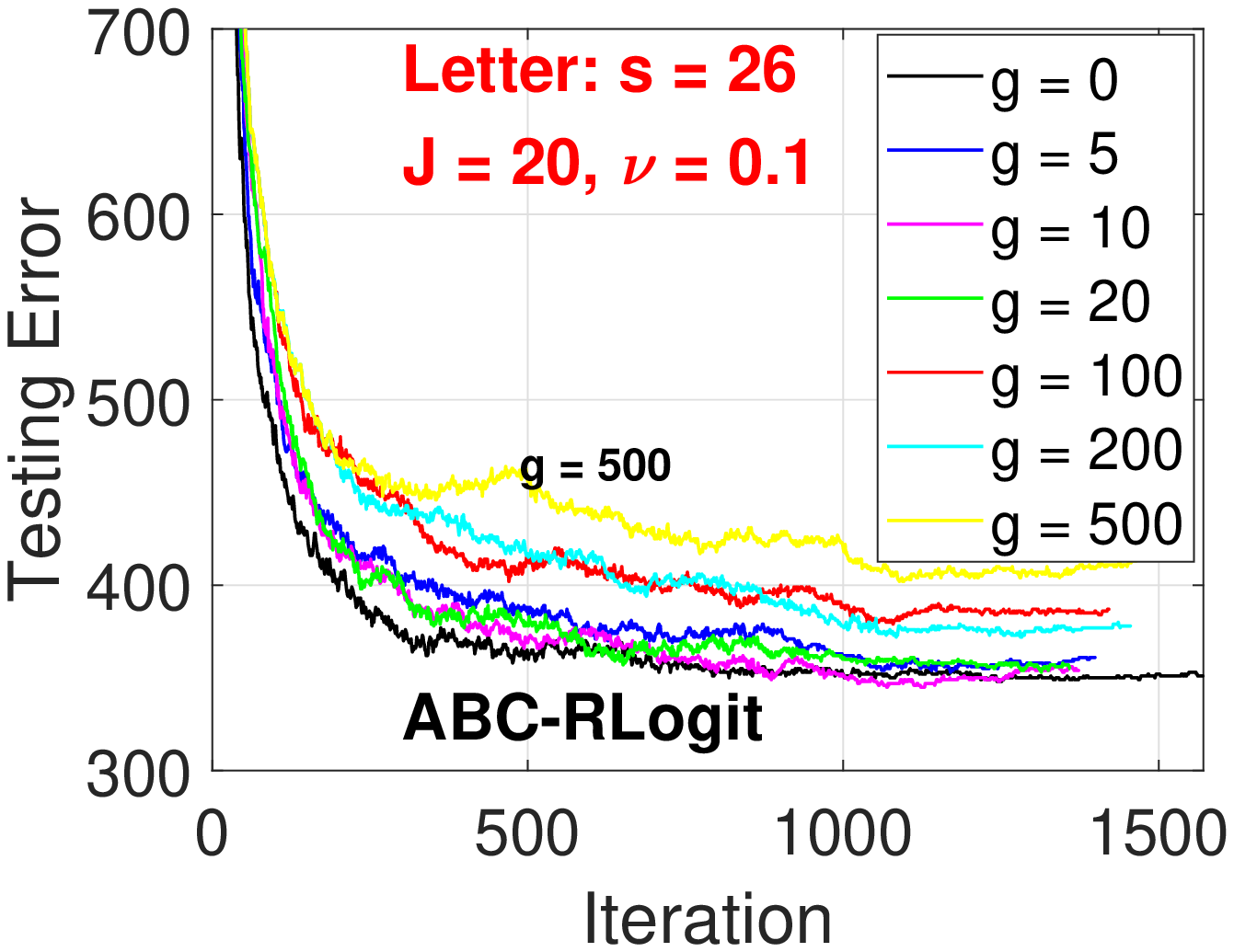}
}

\end{center}

\vspace{-0.1in}

\caption{{\em M-Rand and Letter}  datasets. Test classification errors of ABC-Boost based on the ``$s$-worst classes''  search strategy with a gap parameter $g$, for both ABC-MART and ABC-RobustLogitBoost, for $J=20$ and $\nu=0.1$. We present results for $s=1$ (``worst class''), $s=K$ (``exhaustive search''), as well as $s=3$. When $g=0$ (i.e., no gap), we can see that using $s=1$ typically does not perform as well as using $s=K$, although the difference is often not too big. However, for the {\em Letter} dataset, ABC-MART with $s=1$ and $g=0$ had the ``catastrophic failure'' (i.e., very large test errors).  This means that the ``worst class'' strategy may not be reliable.  On the other hand, the ``gap'' strategy performs well. For the {\em M-Rand} dataset, using $g=5$, $10$, or $20$  reaches similar (or even better) accuracy compared to using $g=0$. For the {\em  Letter} dataset, using $g=20$ achieves close-to-the-best accuracy. Interestingly, for the {\em Letter} dataset, the ``gap'' strategy appears to provide one solution to avoid ``catastrophic failures''. }\label{fig:M-Rand-Letter10k}
\end{figure}

\begin{figure}[h]
\begin{center}
\mbox{
    \includegraphics[width=2.4in]{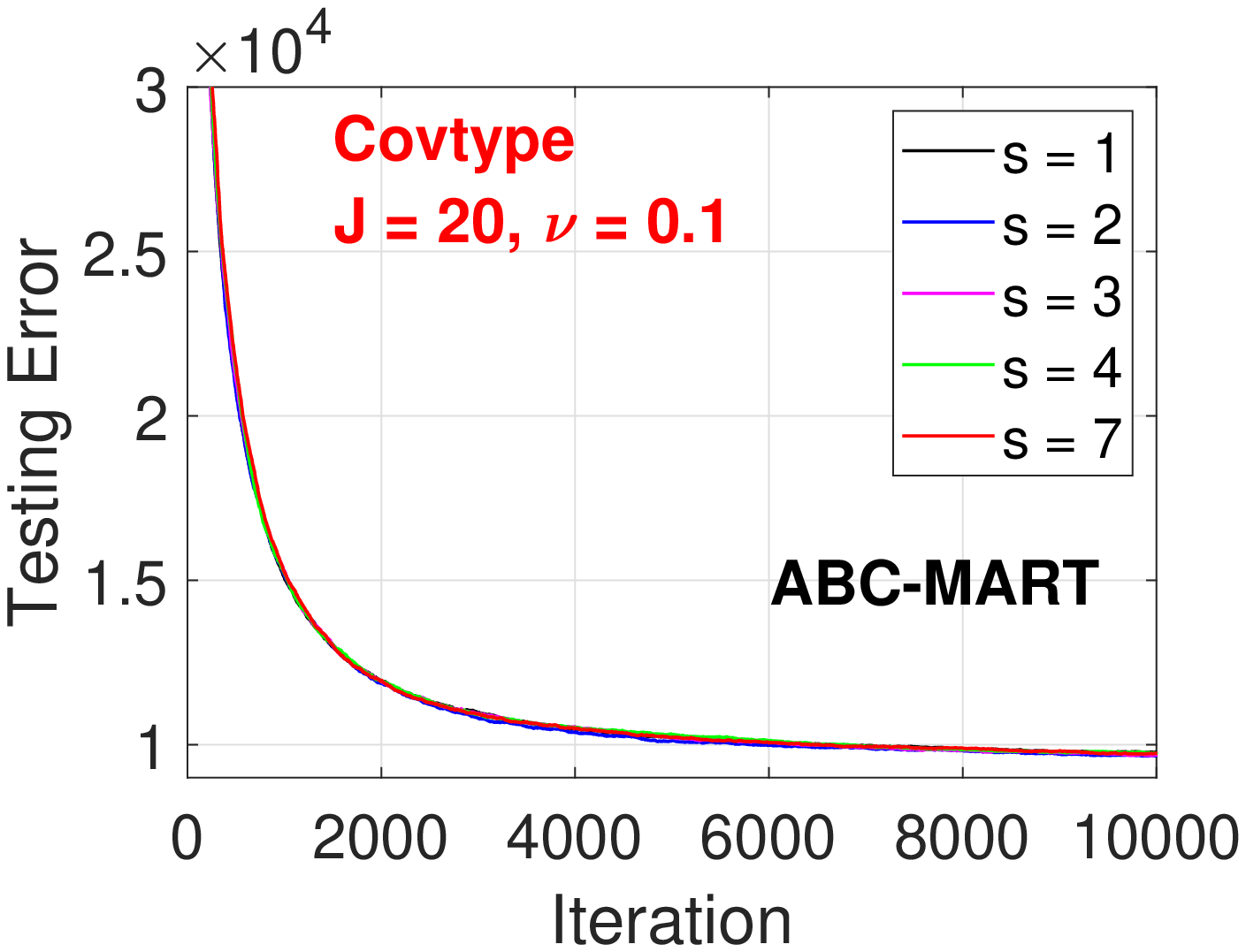}
    \includegraphics[width=2.4in]{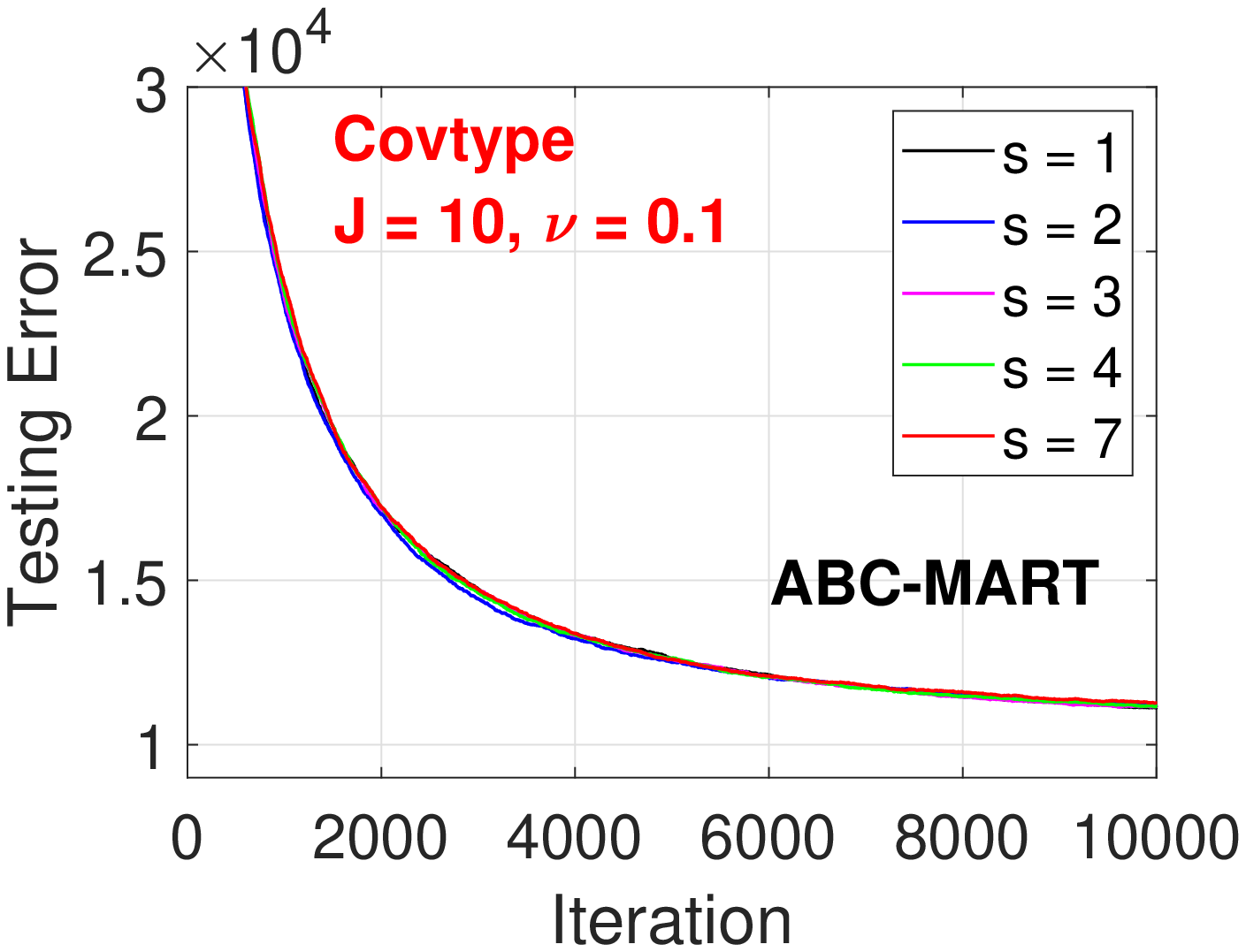}
}

\mbox{
    \includegraphics[width=2.4in]{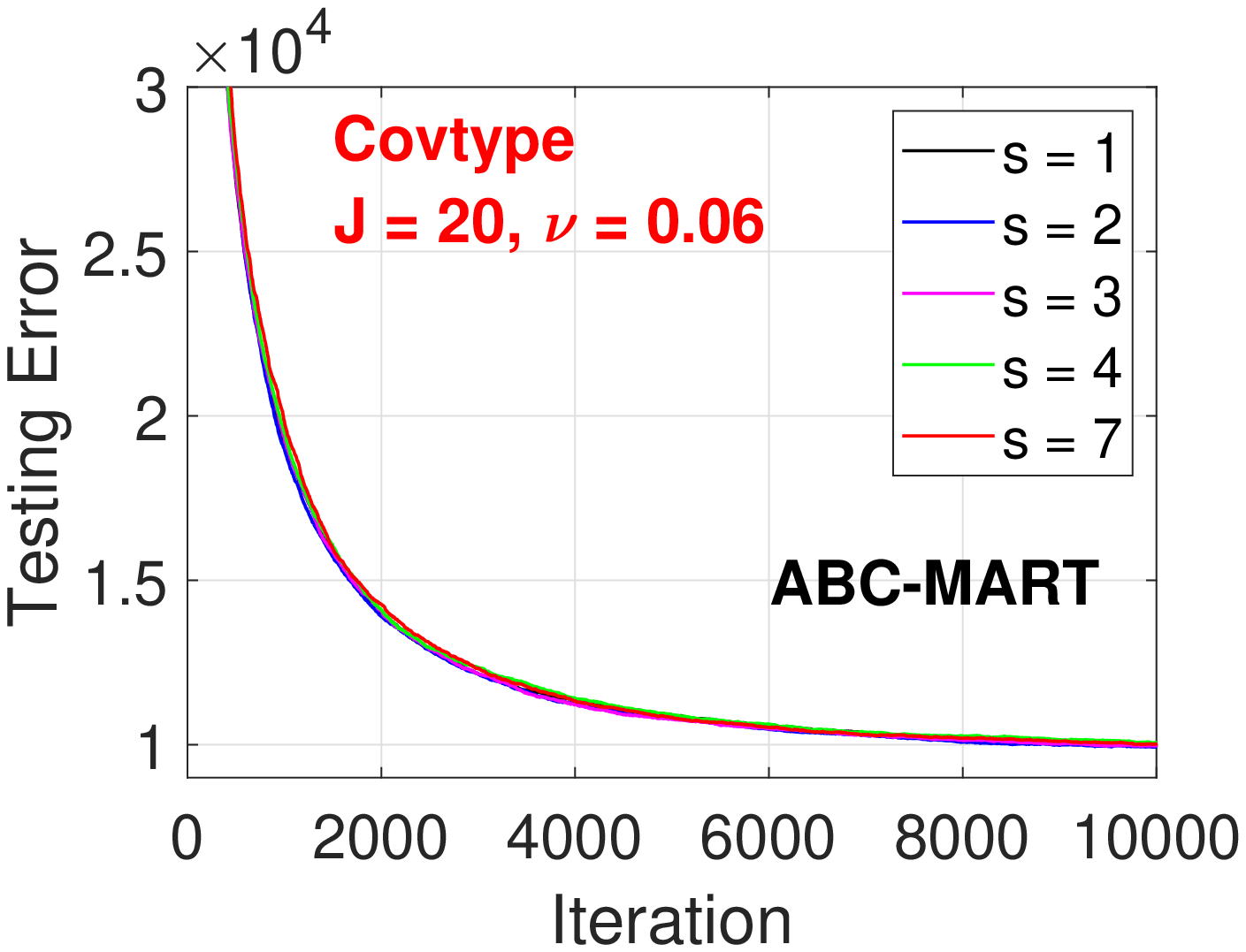}
    \includegraphics[width=2.4in]{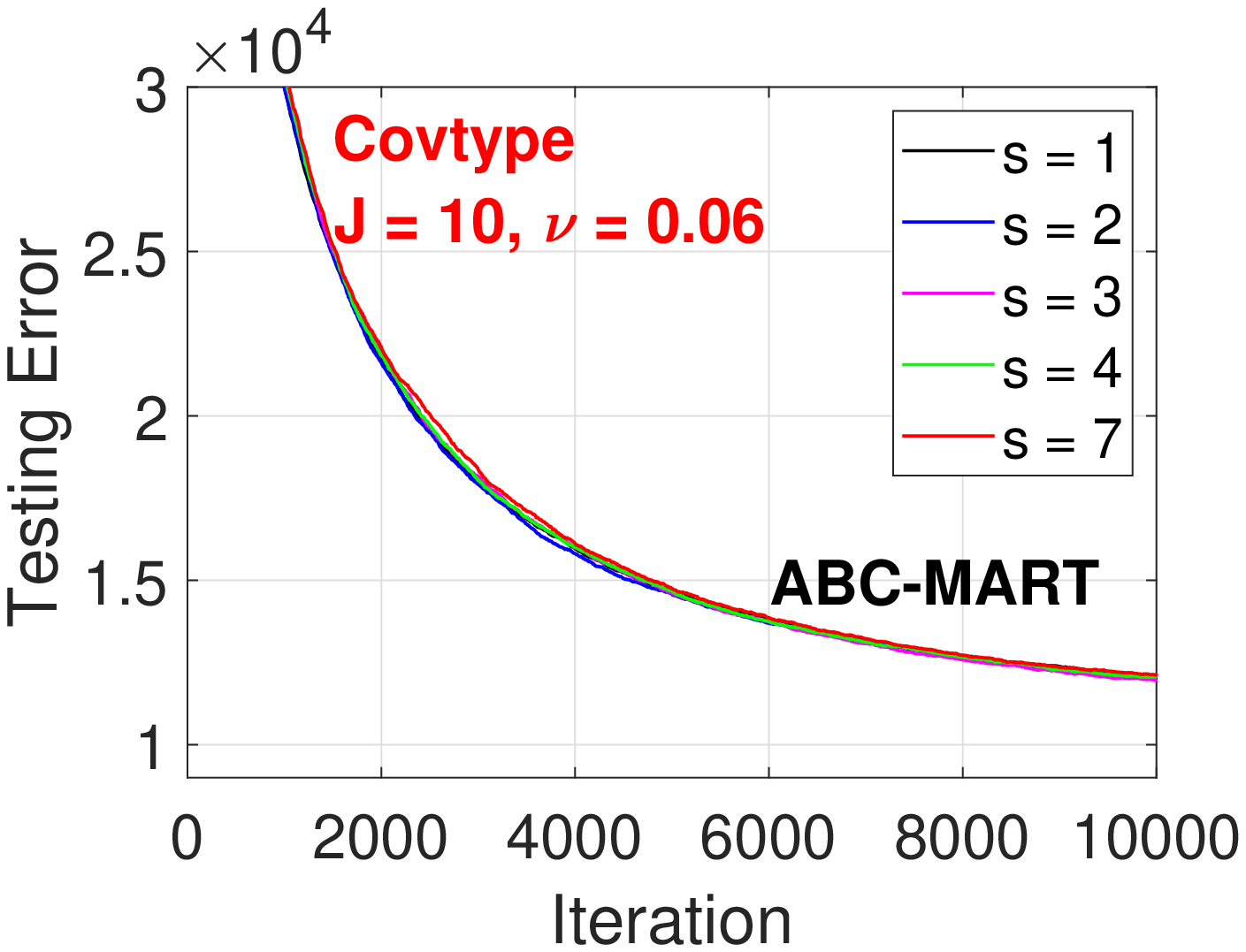}
}

\mbox{
    \includegraphics[width=2.4in]{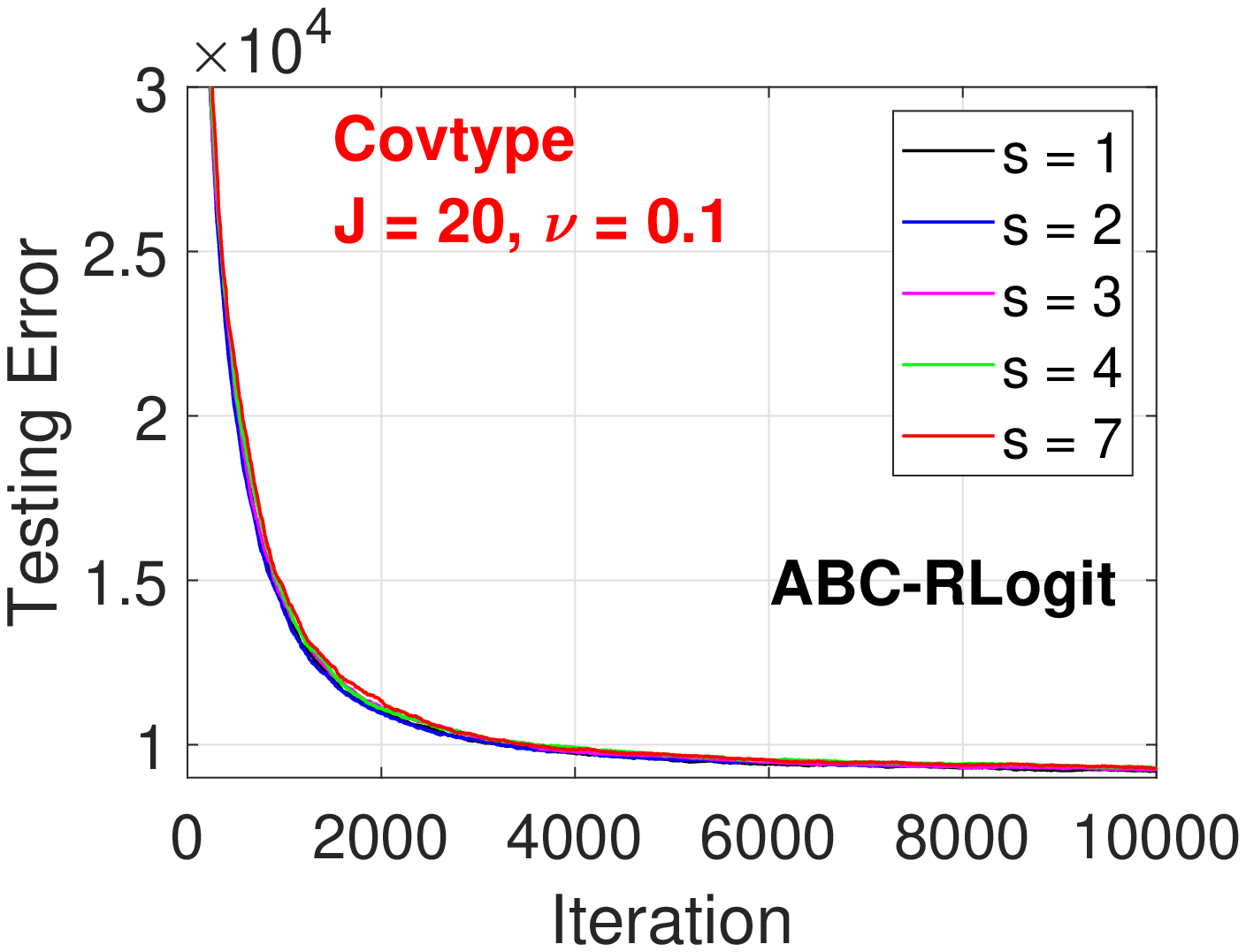}
    \includegraphics[width=2.4in]{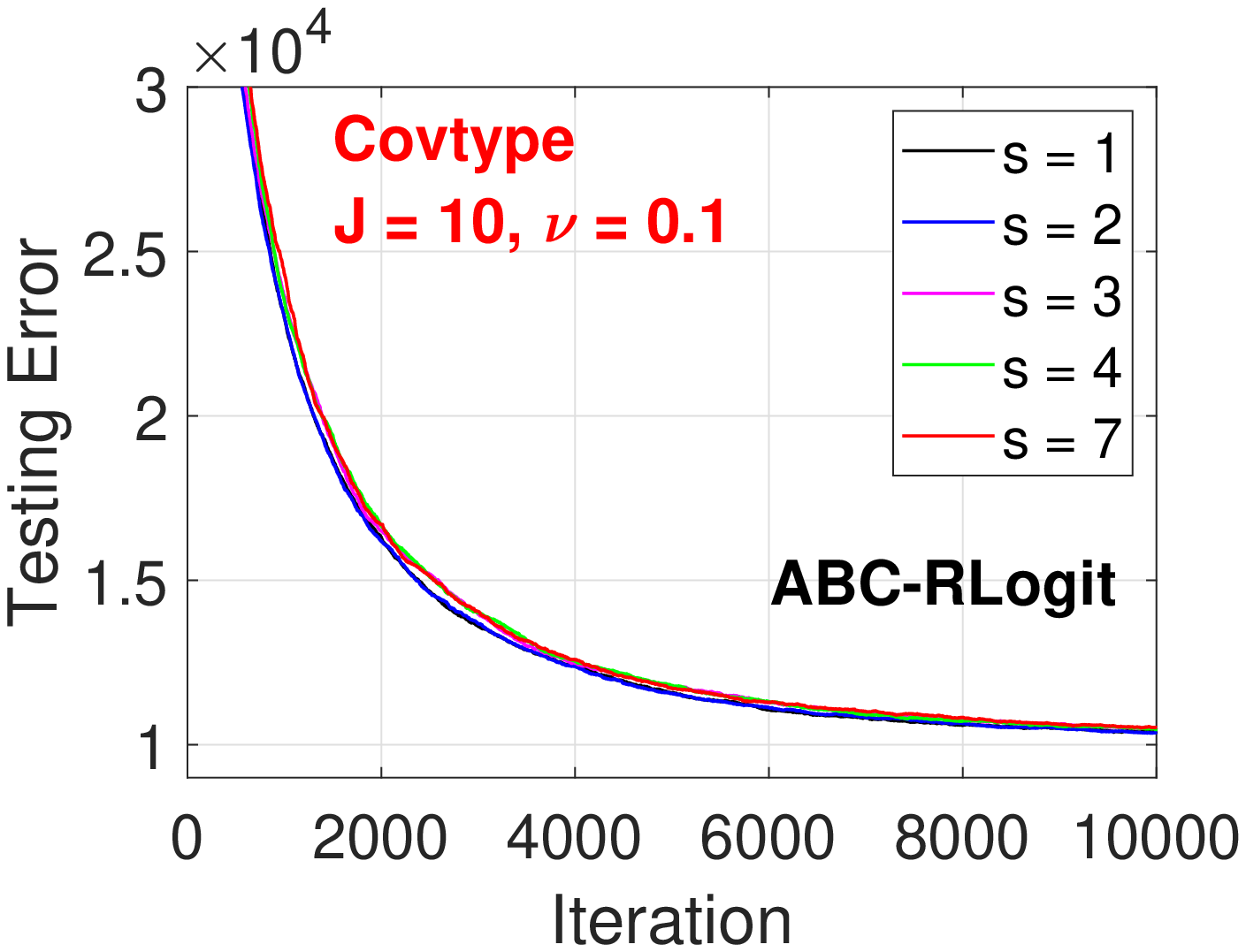}
}

\mbox{
    \includegraphics[width=2.4in]{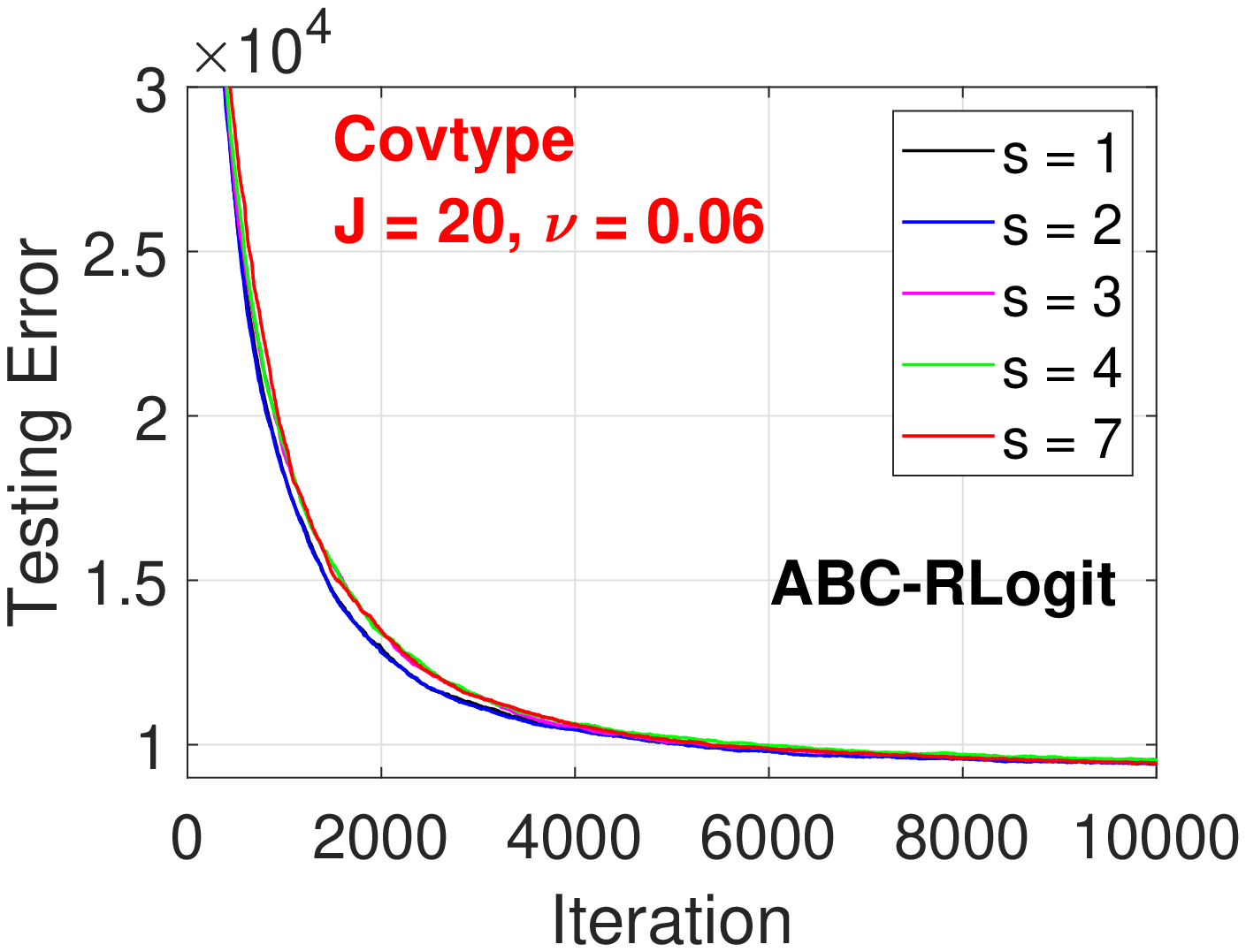}
    \includegraphics[width=2.4in]{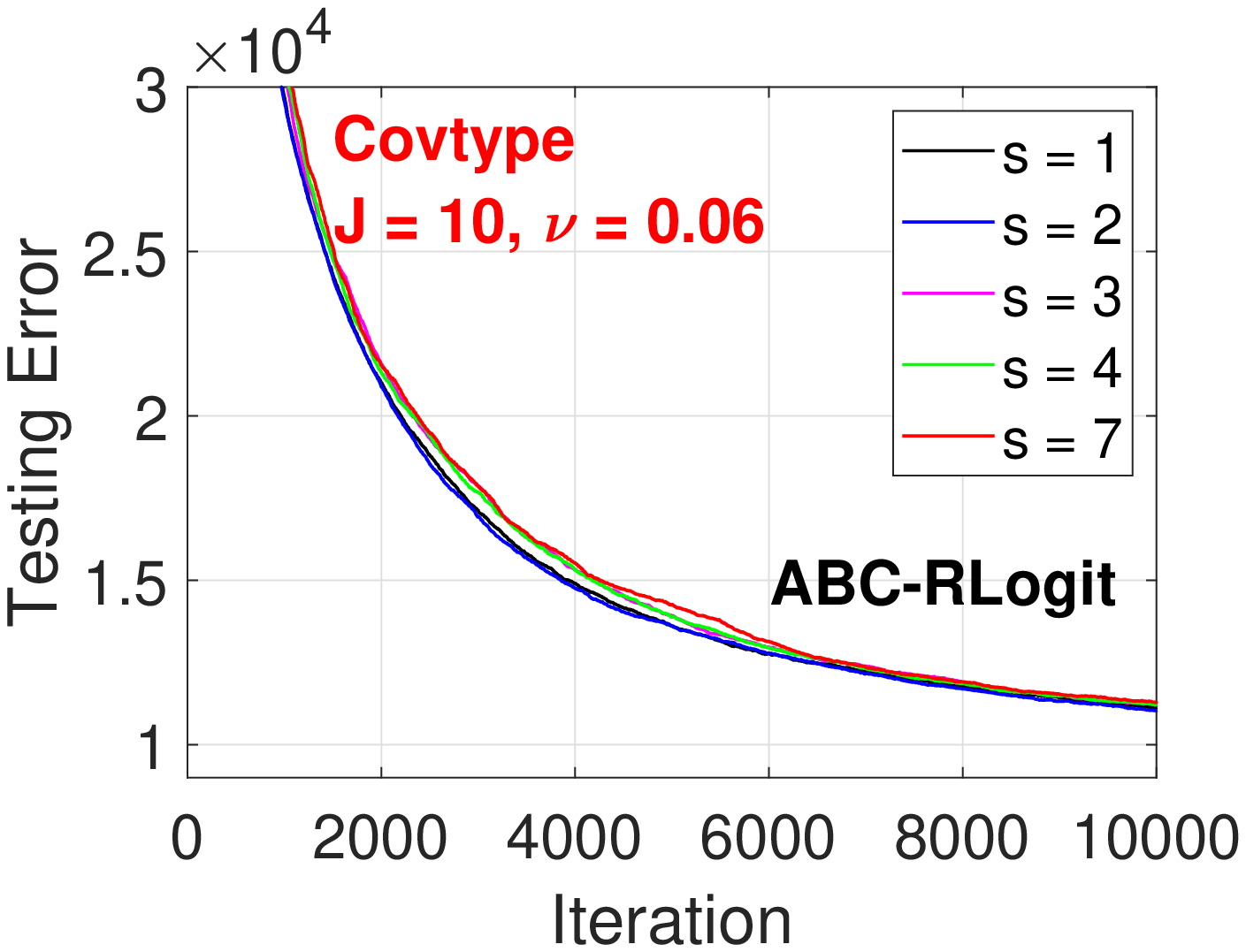}
}
\end{center}

\vspace{-0.1in}

\caption{{\em Covertype} dataset. Test classification errors based on the ``$s$-worst classes'' search strategy for both ABC-MART and ABC-RobustLogitBoost, for $s\in\{1,2,3,4,7\}$.   }\label{fig:Covertype_s}
\end{figure}

\begin{figure}[h]
\begin{center}
\mbox{
    \includegraphics[width=2.4in]{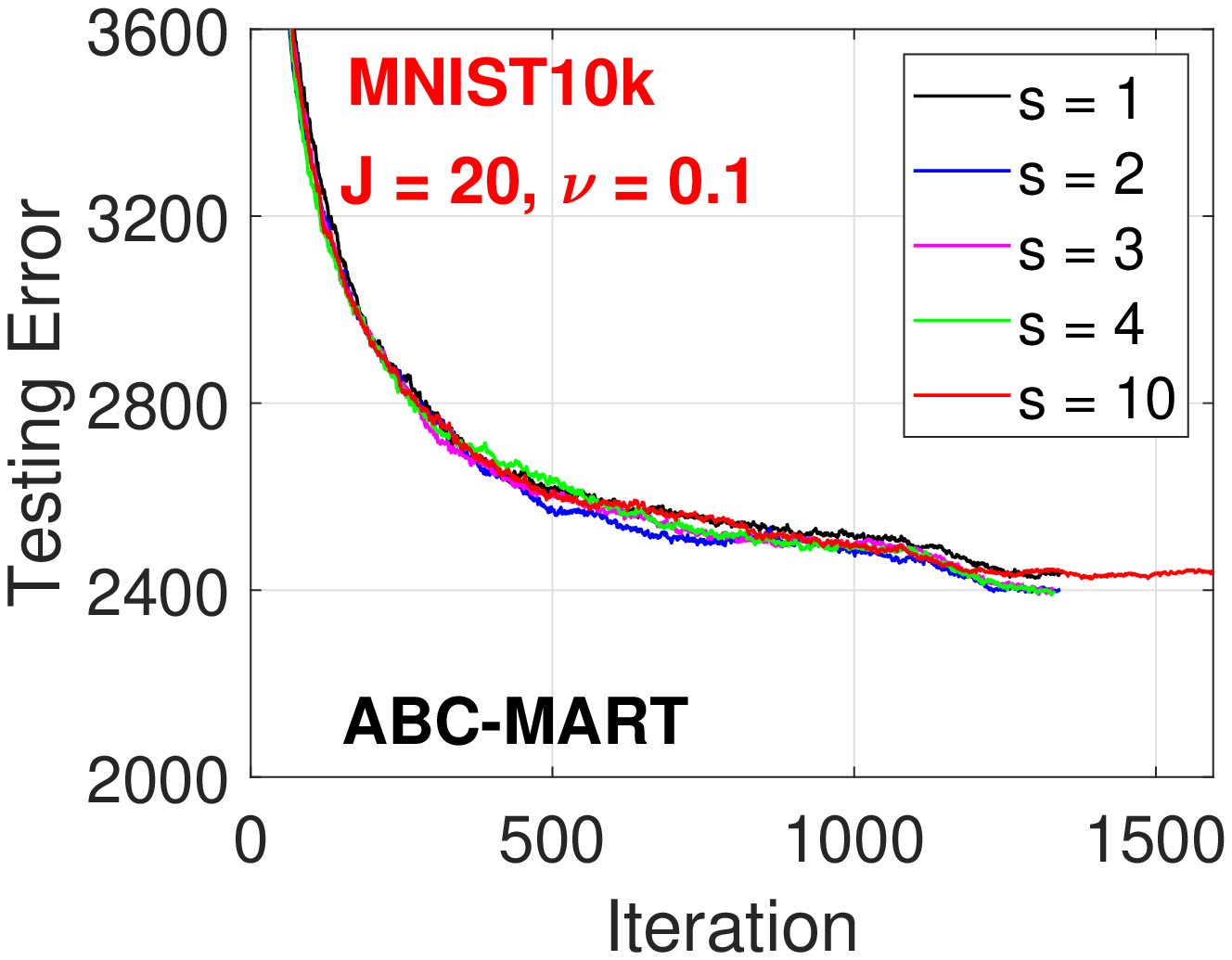}
    \includegraphics[width=2.4in]{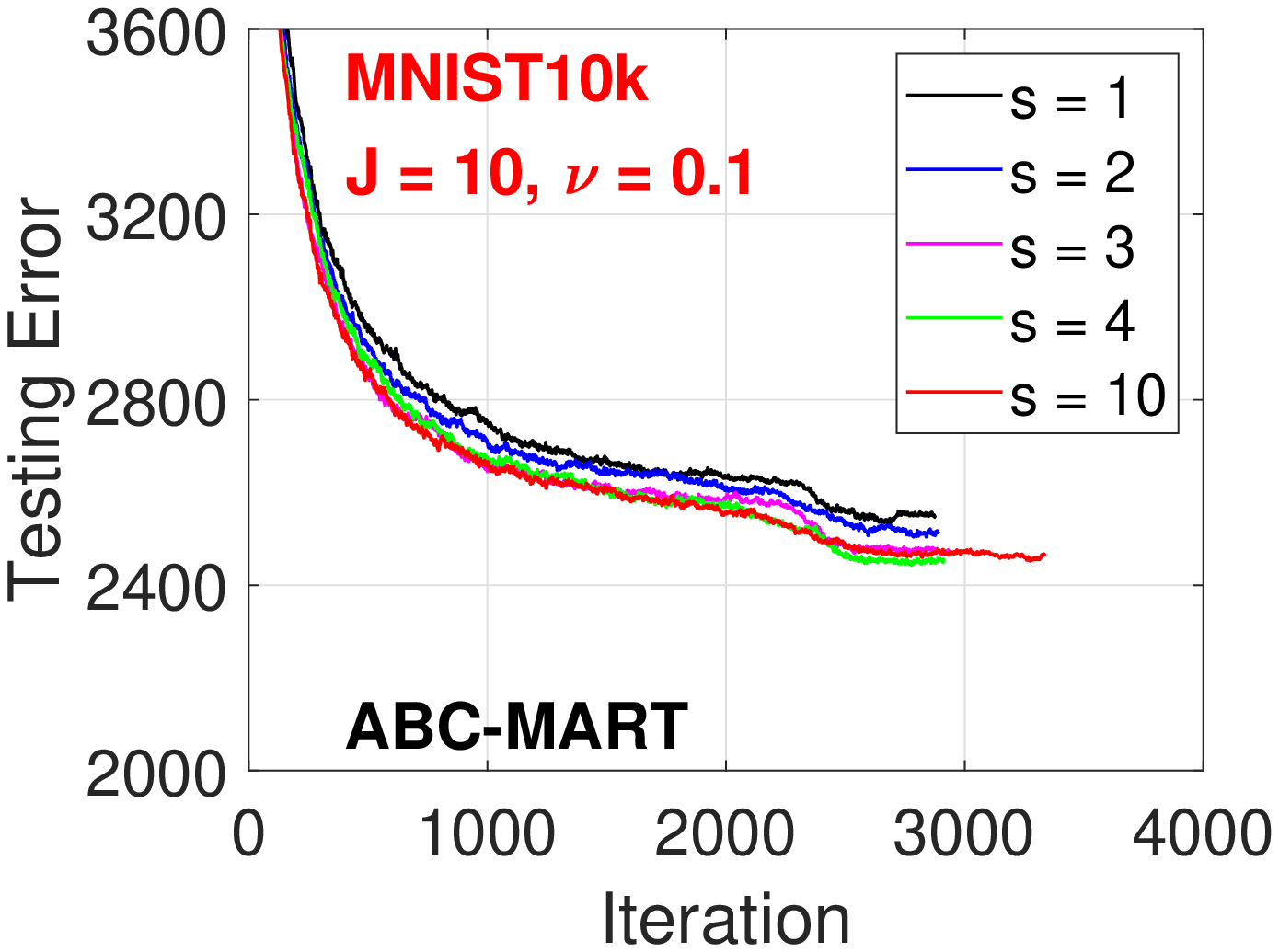}
}
\mbox{
    \includegraphics[width=2.4in]{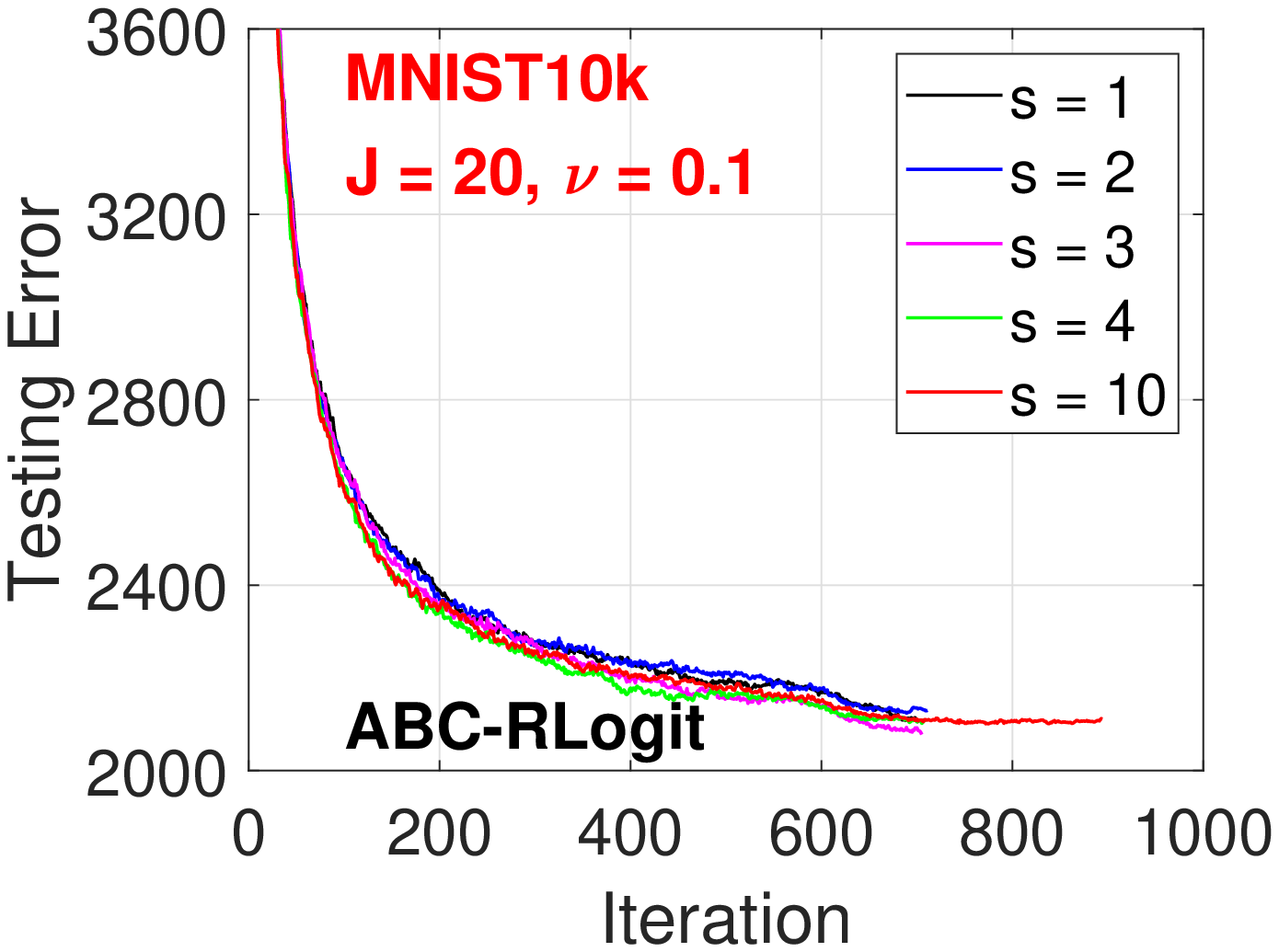}
    \includegraphics[width=2.4in]{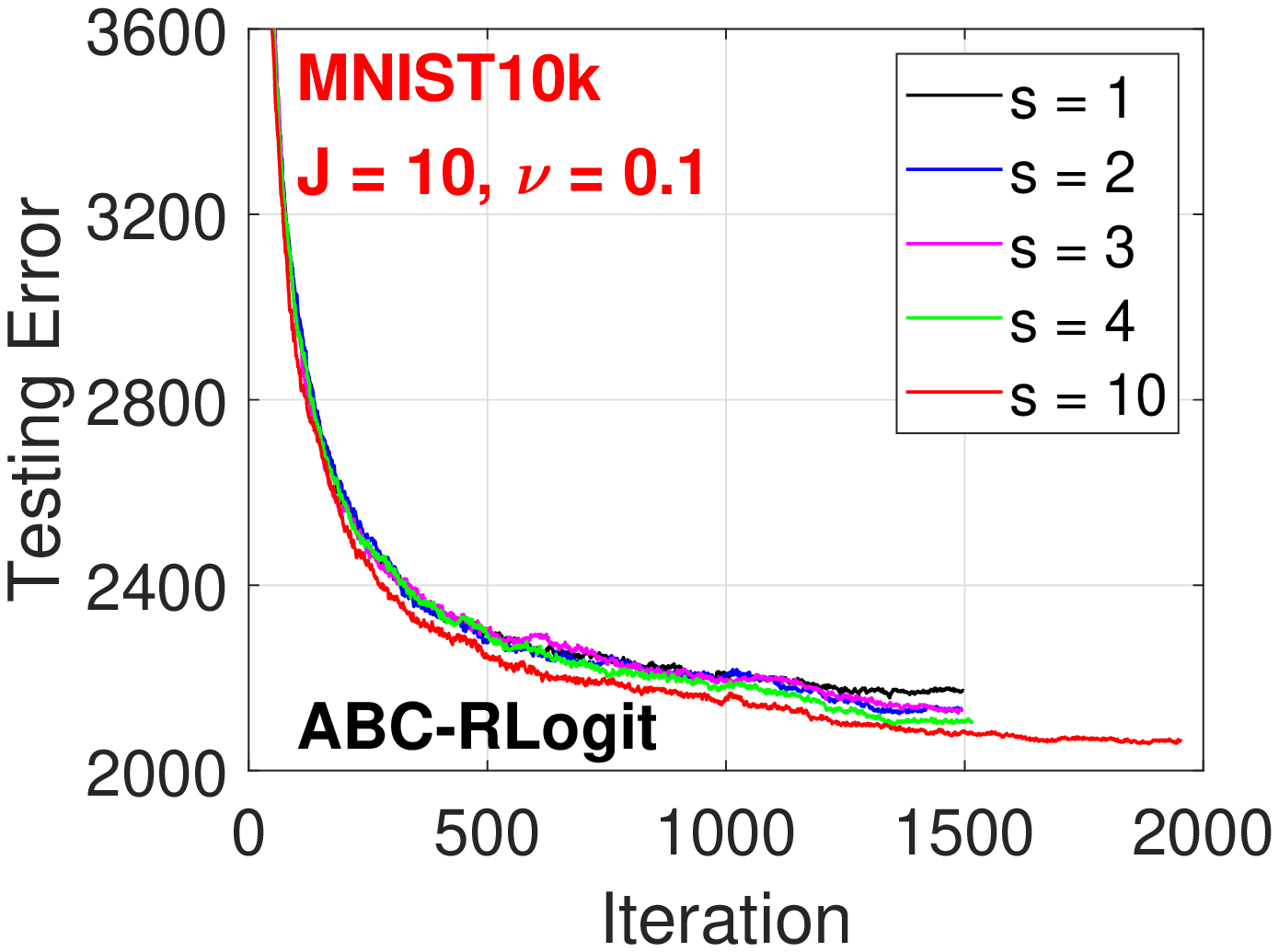}
}

\mbox{
    \includegraphics[width=2.4in]{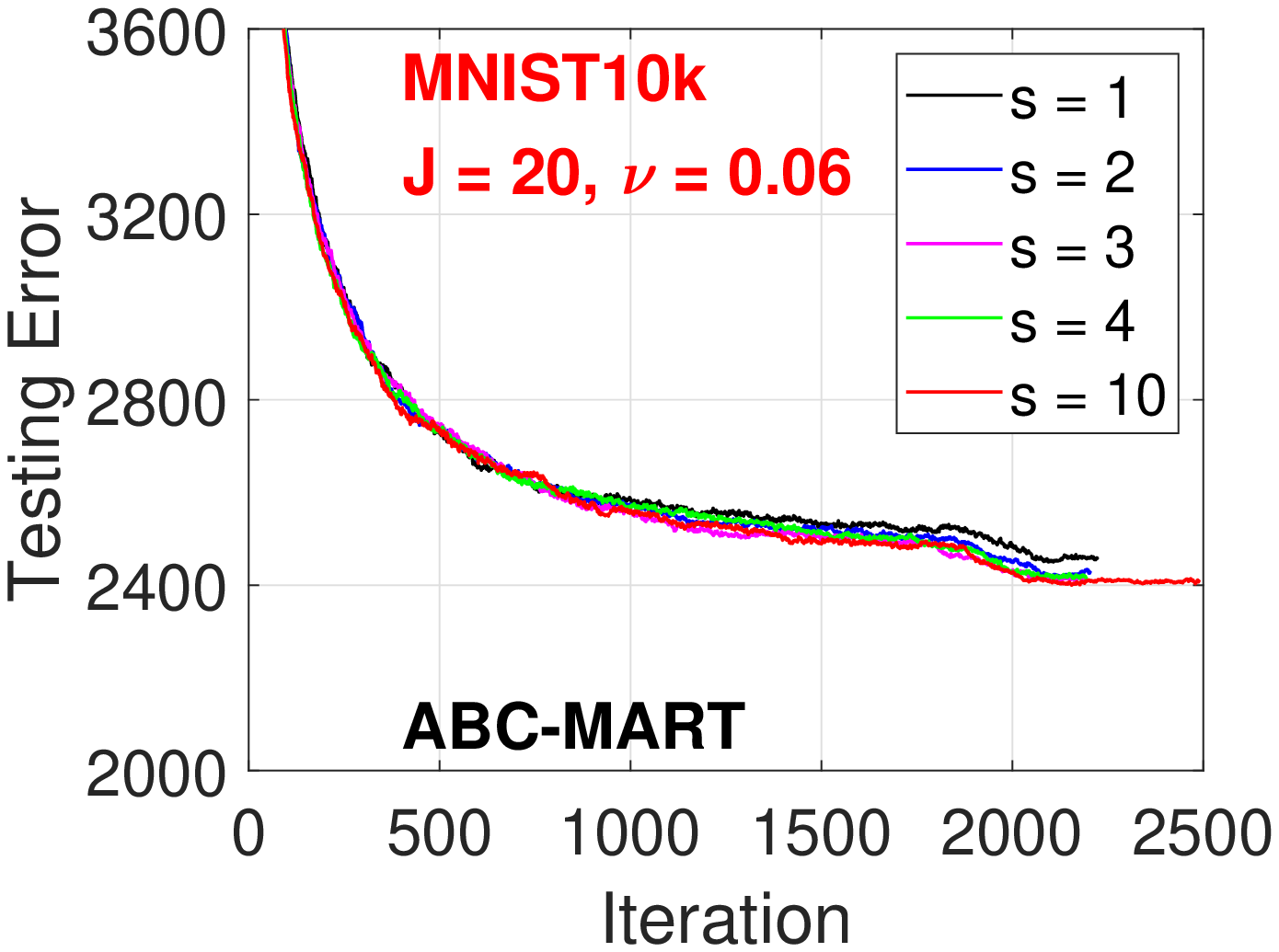}
    \includegraphics[width=2.4in]{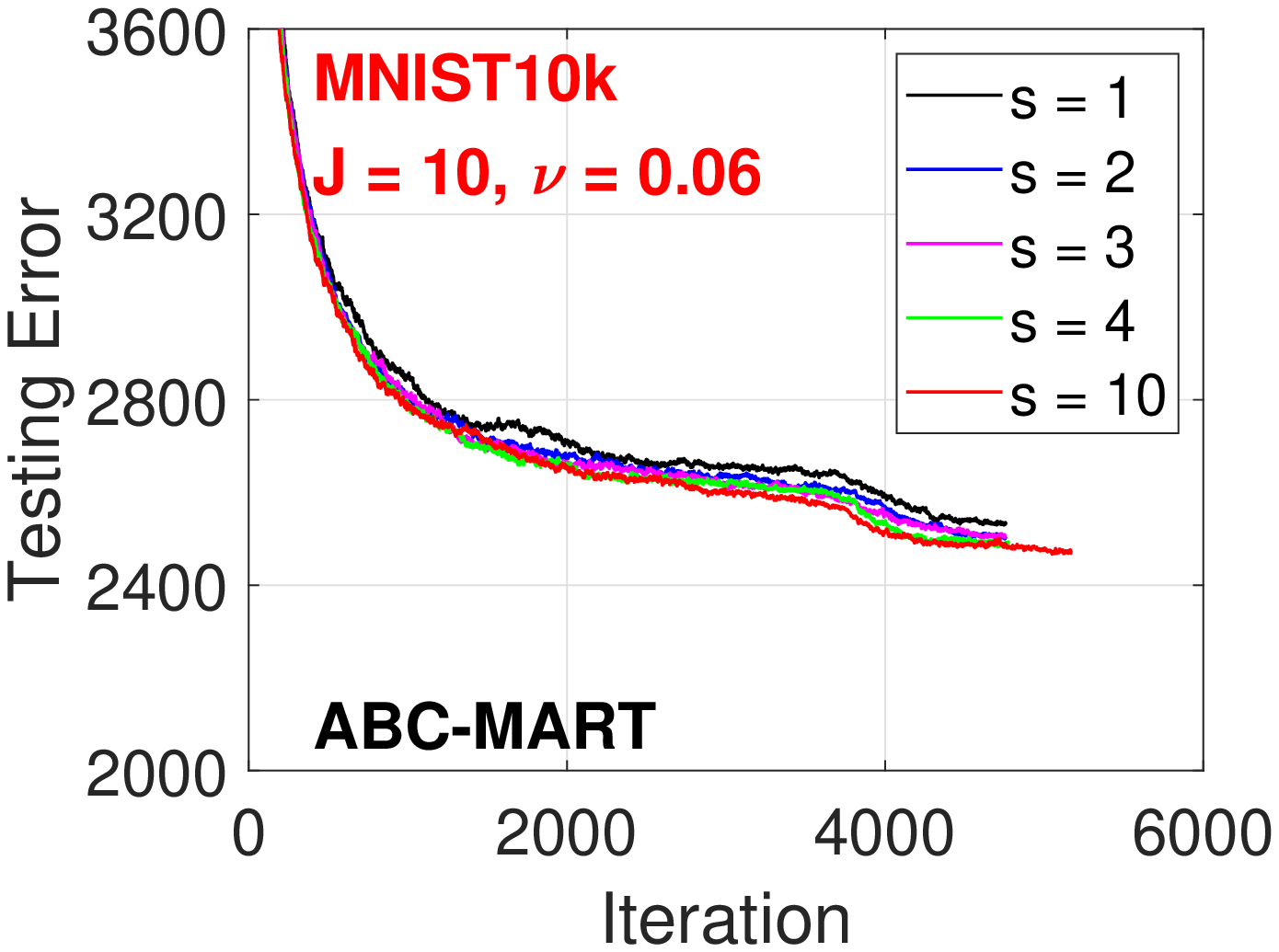}
}
\mbox{
    \includegraphics[width=2.4in]{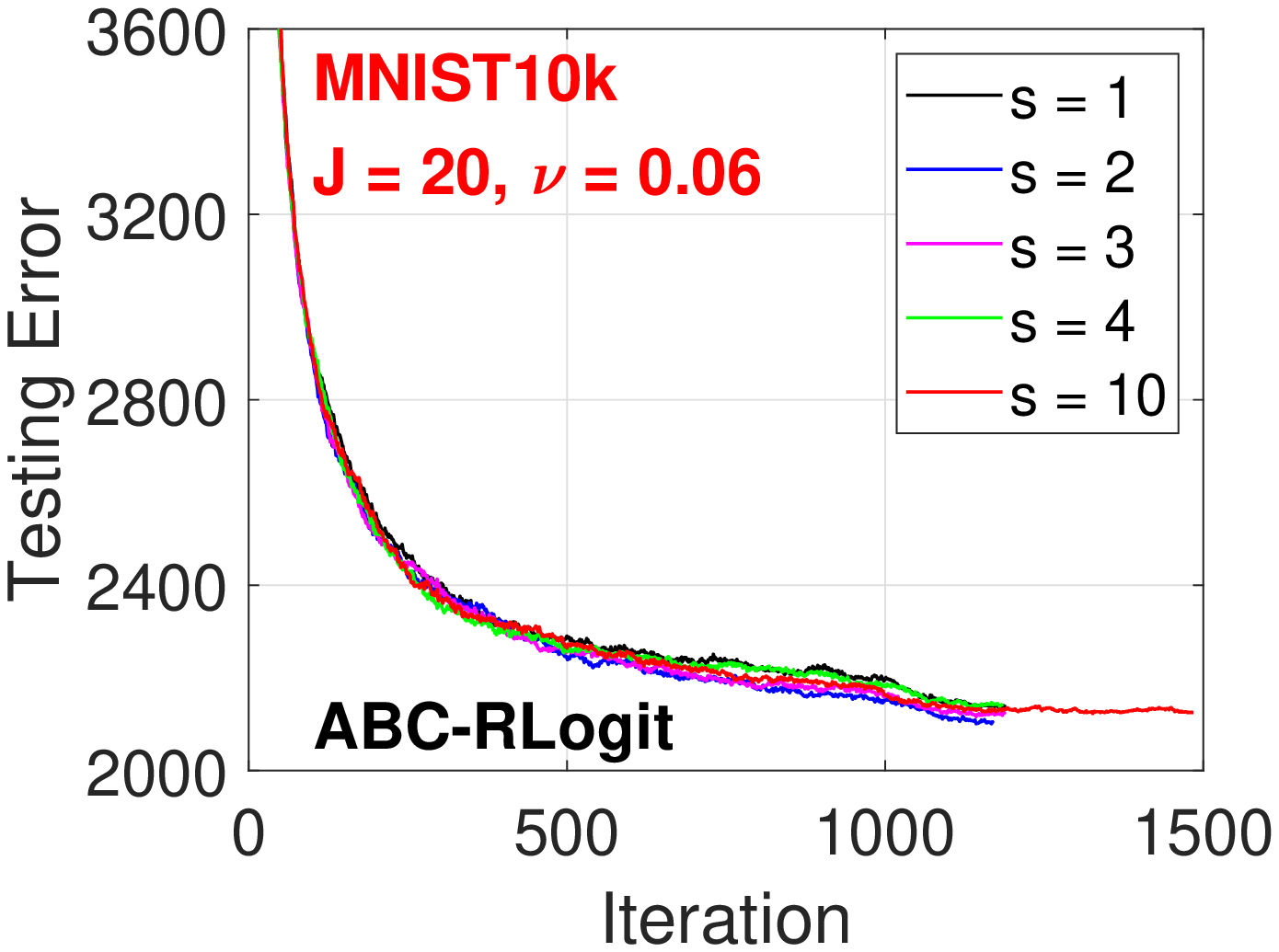}
    \includegraphics[width=2.4in]{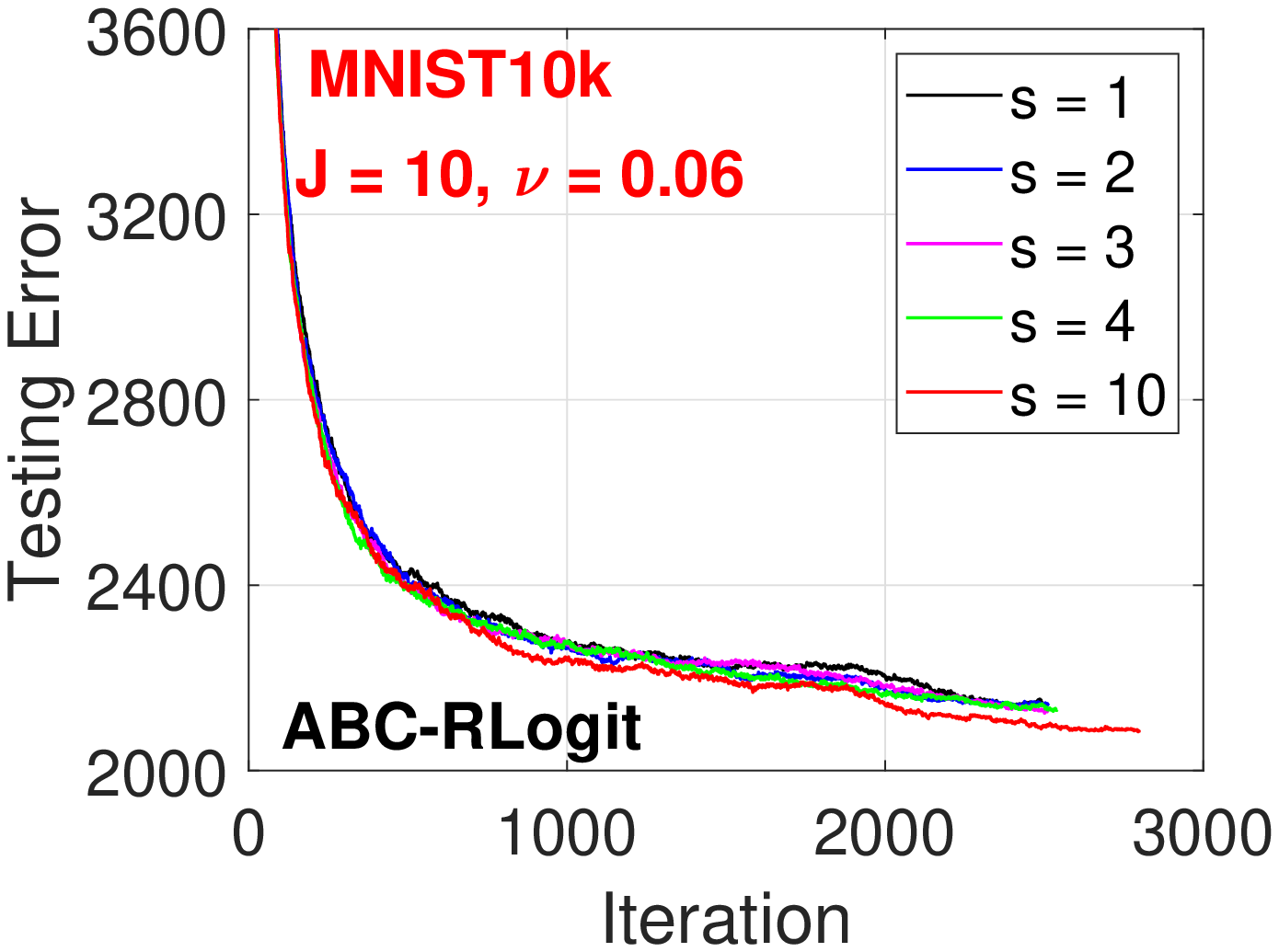}
}
\end{center}
\caption{{\em Mnist10k} dataset. Test classification errors based on the ``$s$-worst classes'' search strategy for both ABC-MART and ABC-RobustLogitBoost, for $s\in\{1,2,3,4,10\}$.
}\label{fig_Mnist10k_s}
\end{figure}

\begin{figure}[h]
\begin{center}
\mbox{
    \includegraphics[width=2.4in]{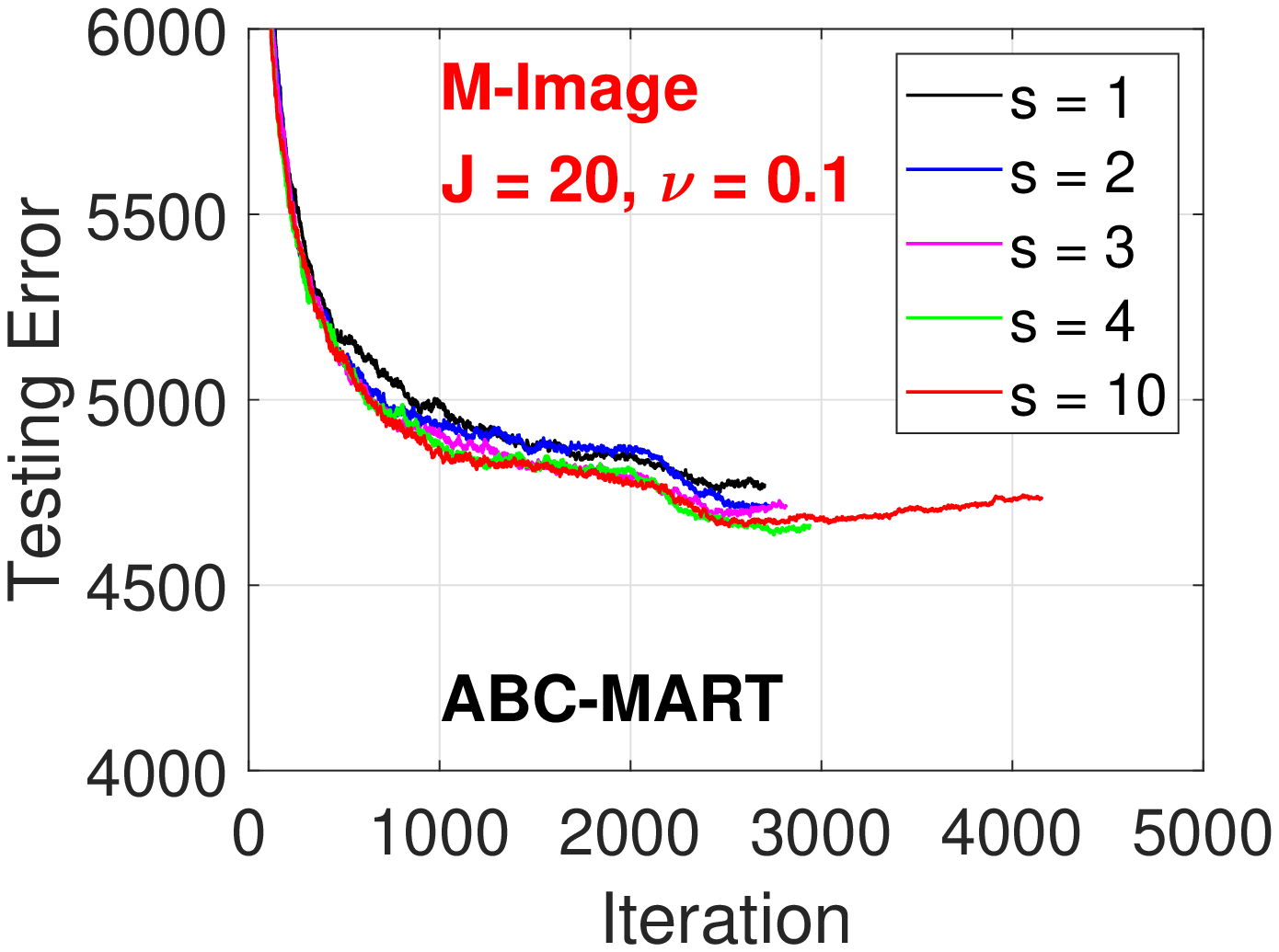}\hspace{-0.1in}
    \includegraphics[width=2.4in]{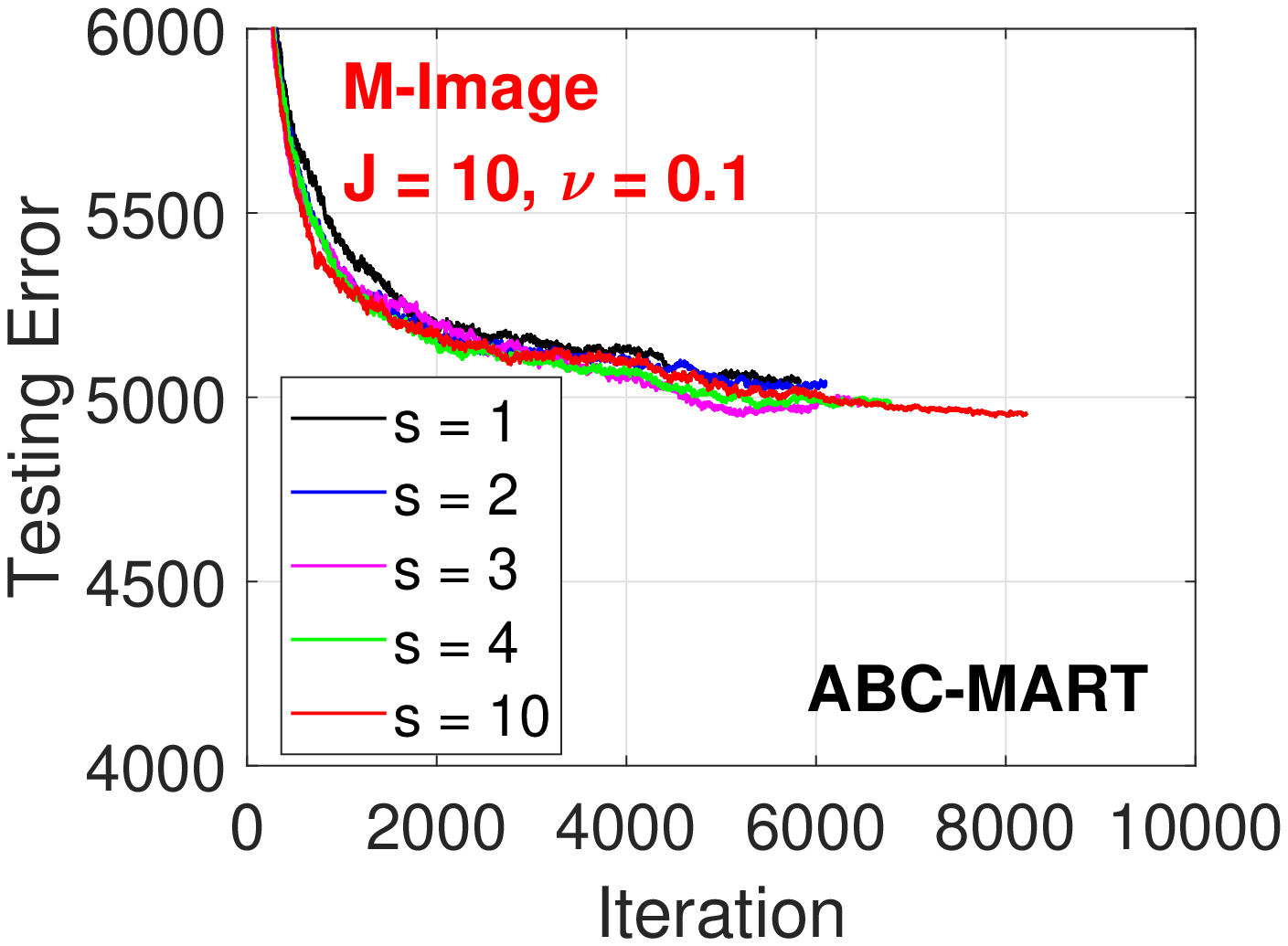}
}

\mbox{
    \includegraphics[width=2.4in]{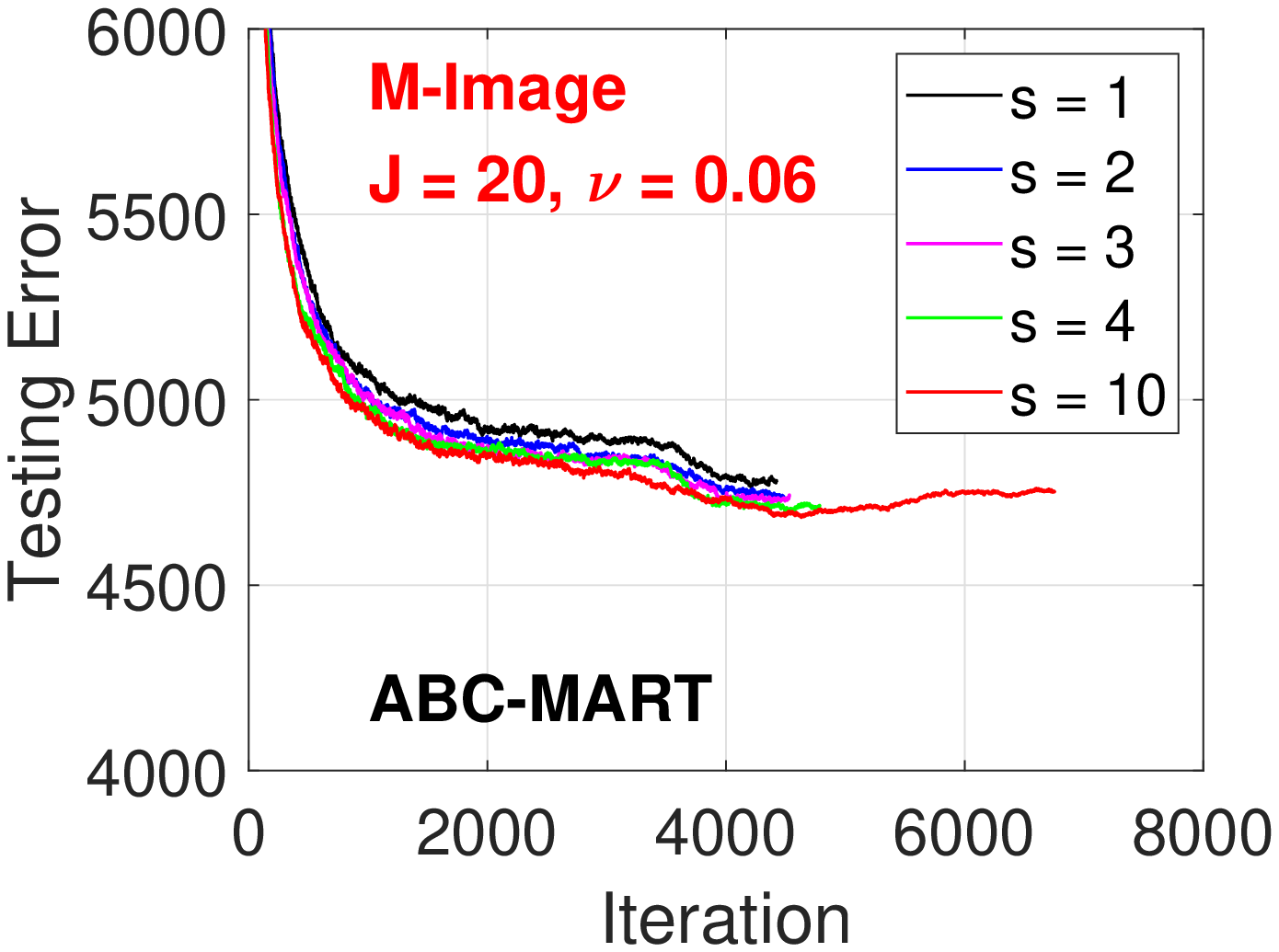}
    \includegraphics[width=2.4in]{fig/M-Image/M-Image-Test-J20v006_abcmart_s.eps}
}

\mbox{
    \includegraphics[width=2.4in]{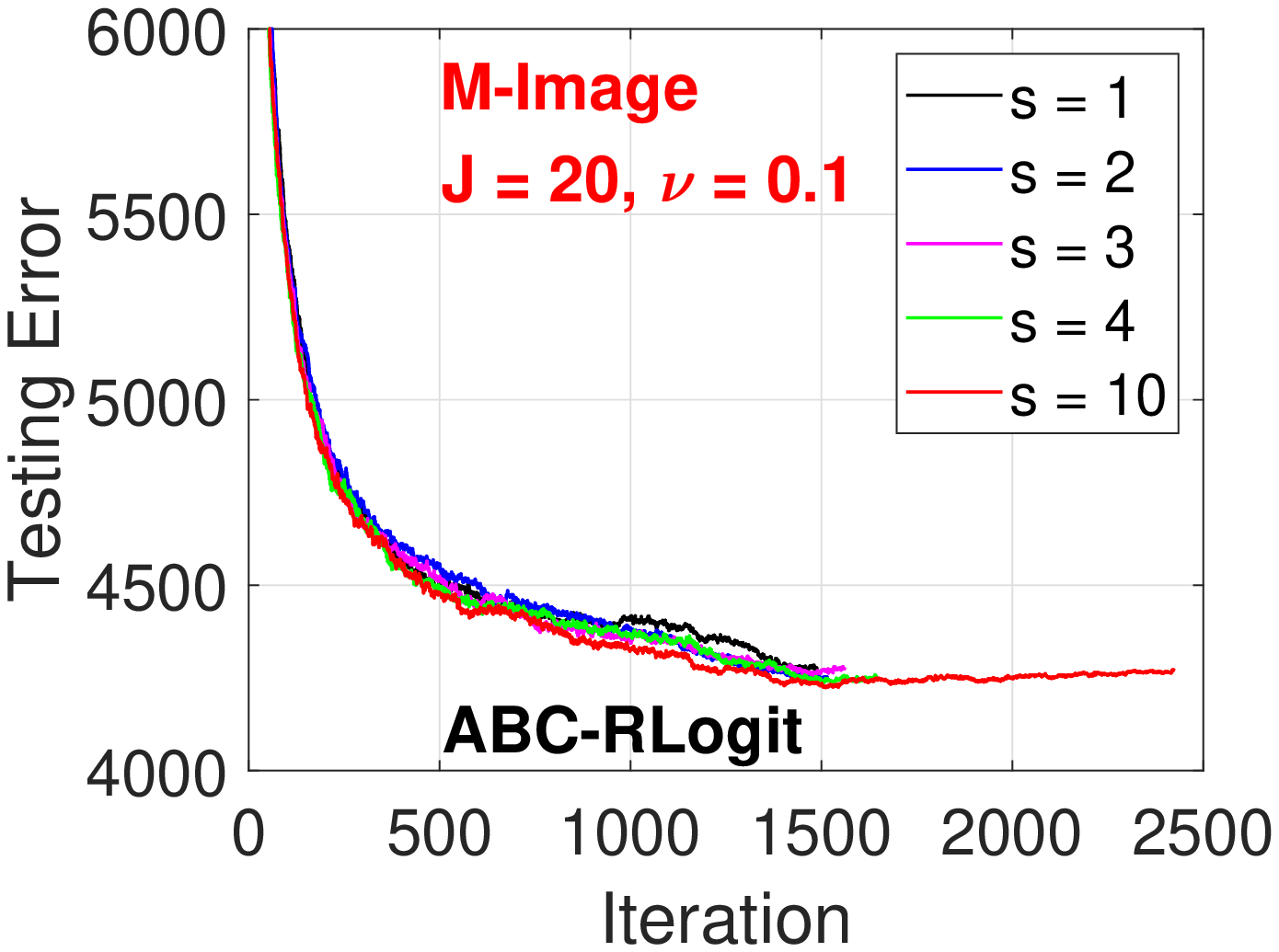}
    \includegraphics[width=2.4in]{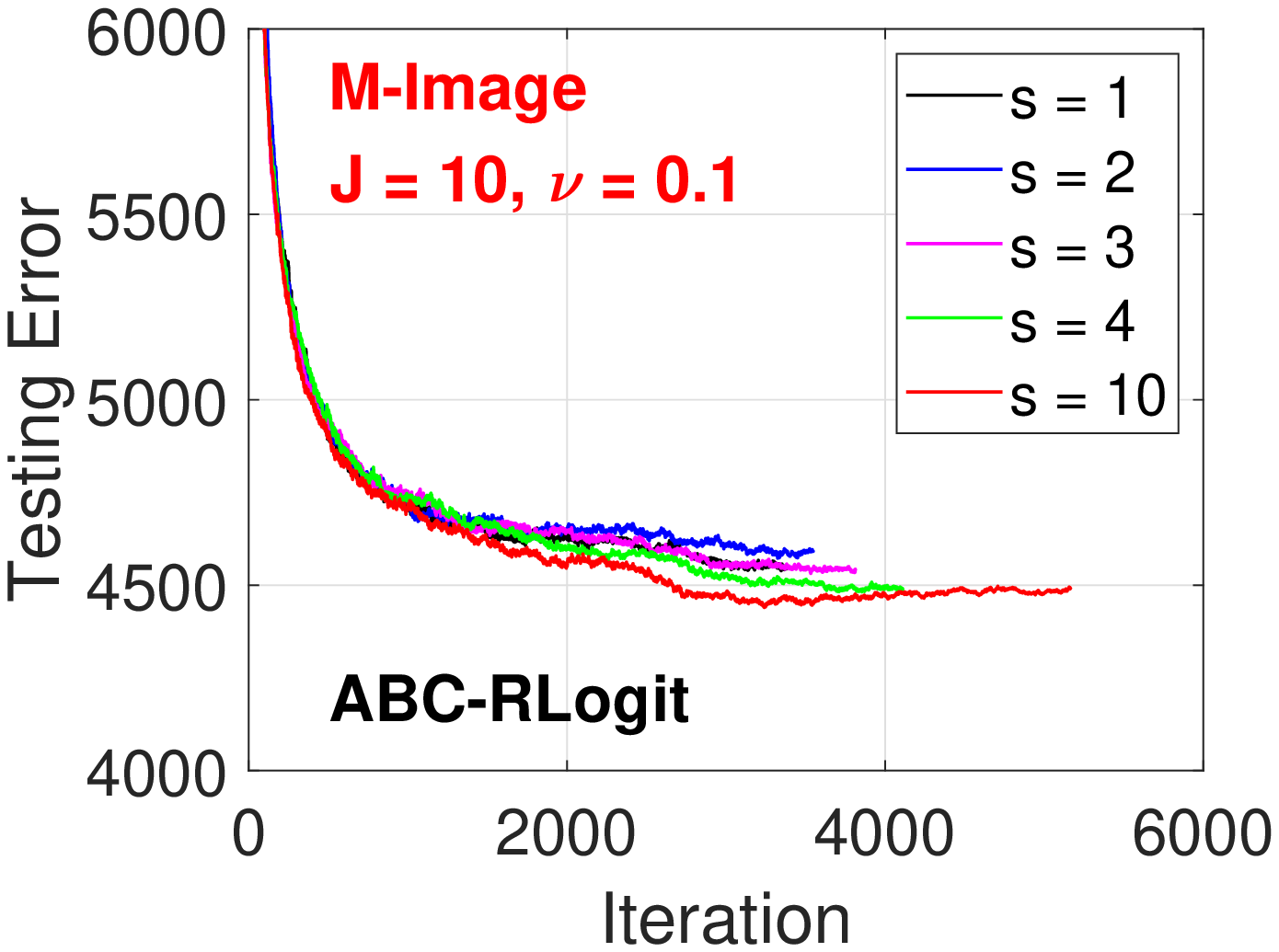}
}

\mbox{
    \includegraphics[width=2.4in]{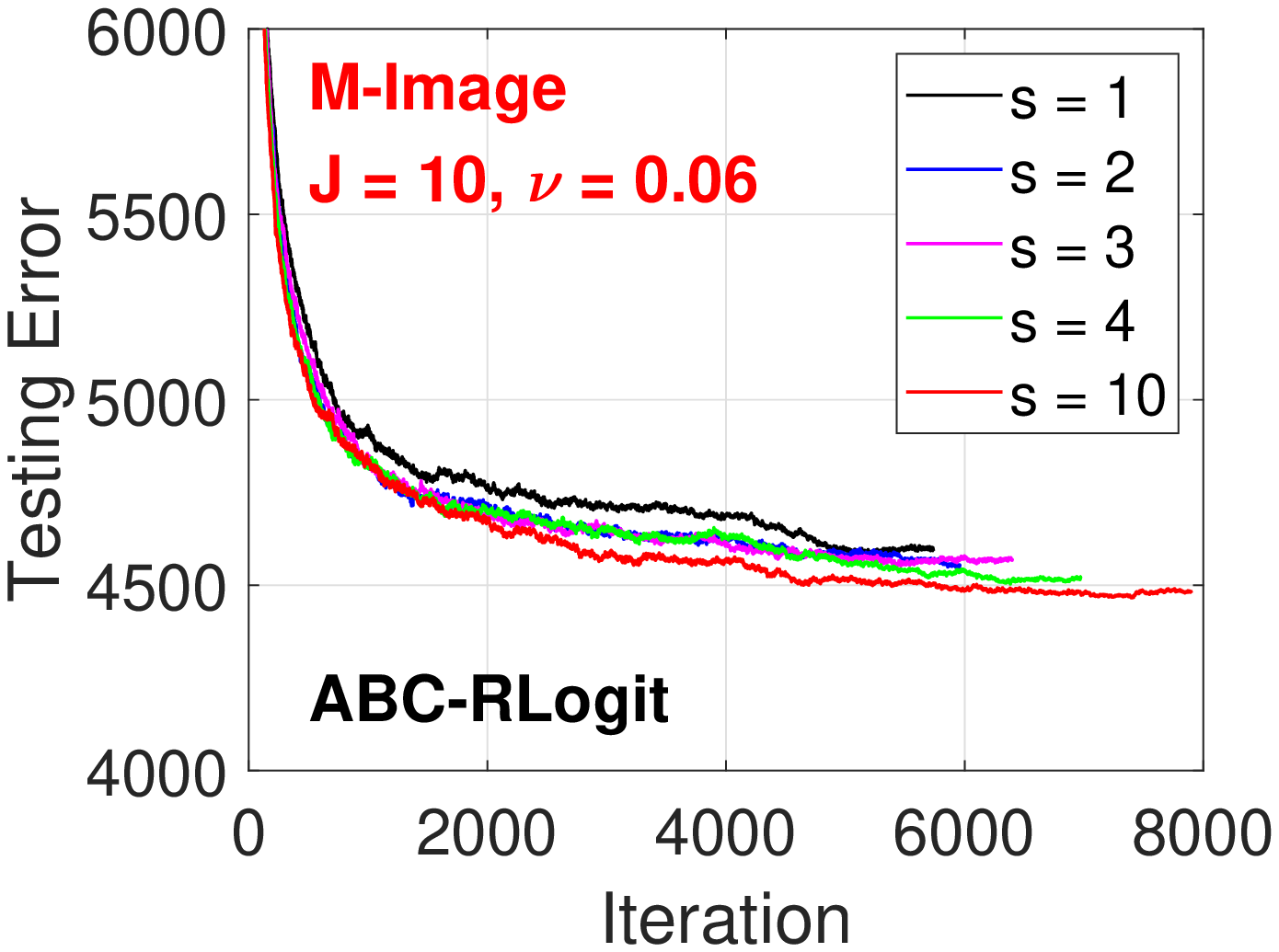}
    \includegraphics[width=2.4in]{fig/M-Image/M-Image-Test-J10v006_abclogit_s.eps}
}

\end{center}
\caption{{\em M-Image} dataset. Test classification errors based on the ``$s$-worst classes'' search strategy for both ABC-MART and ABC-RobustLogitBoost, for $s\in\{1,2,3,4,10\}$.   }\label{fig:M-Image_s}
\end{figure}

\begin{figure}[h]
\begin{center}
\mbox{
    \includegraphics[width=2.4in]{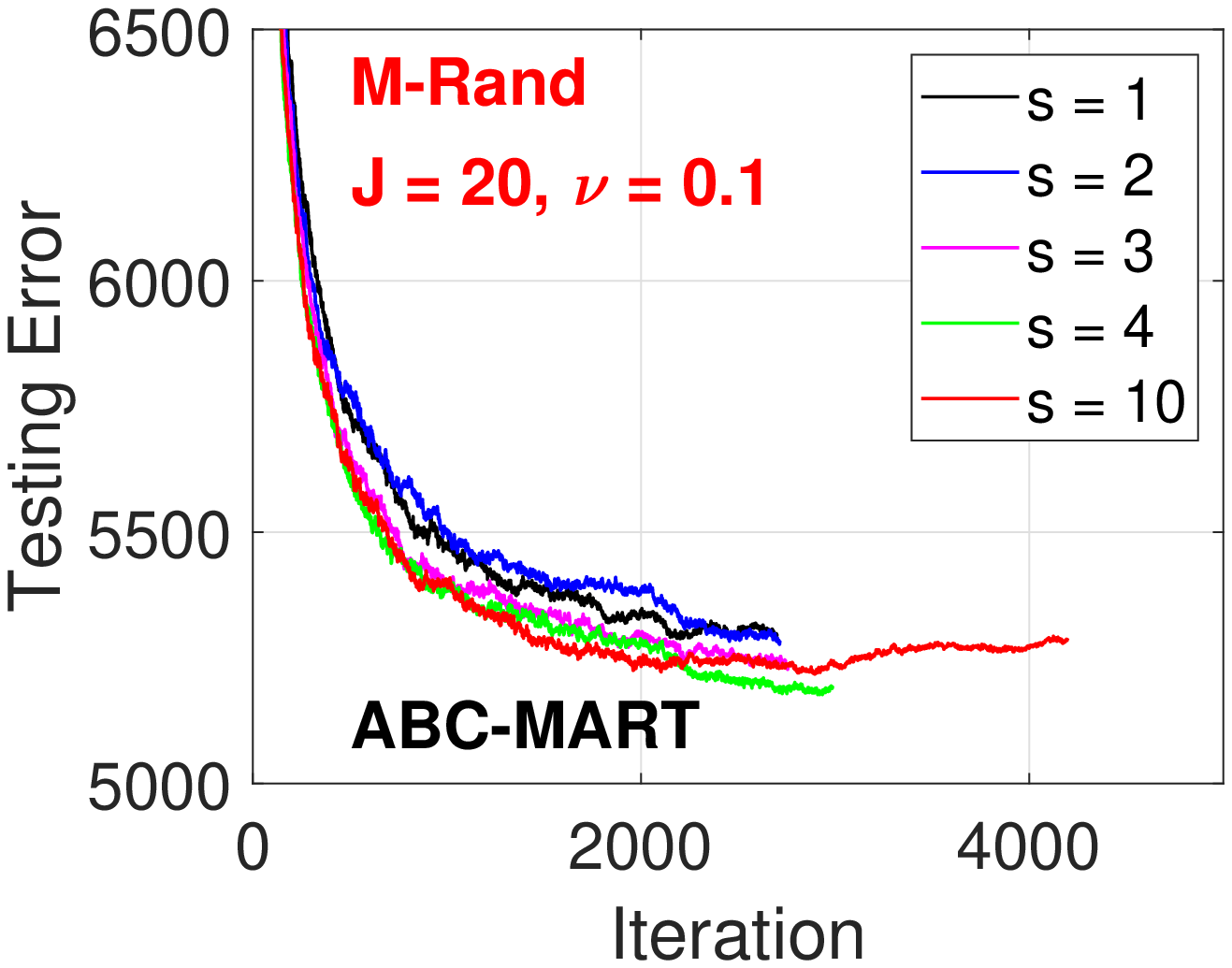}
    \includegraphics[width=2.4in]{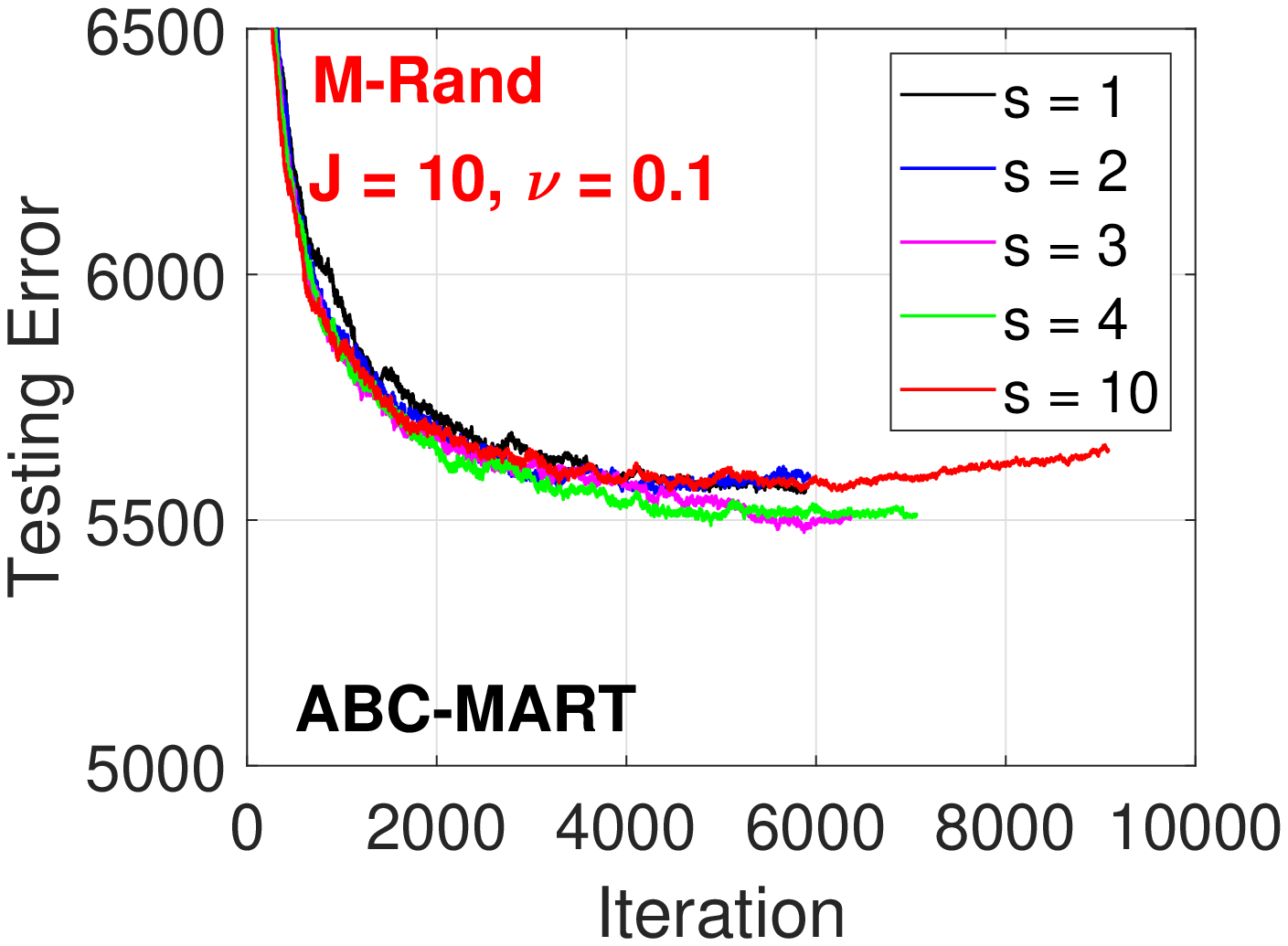}
}

\mbox{
    \includegraphics[width=2.4in]{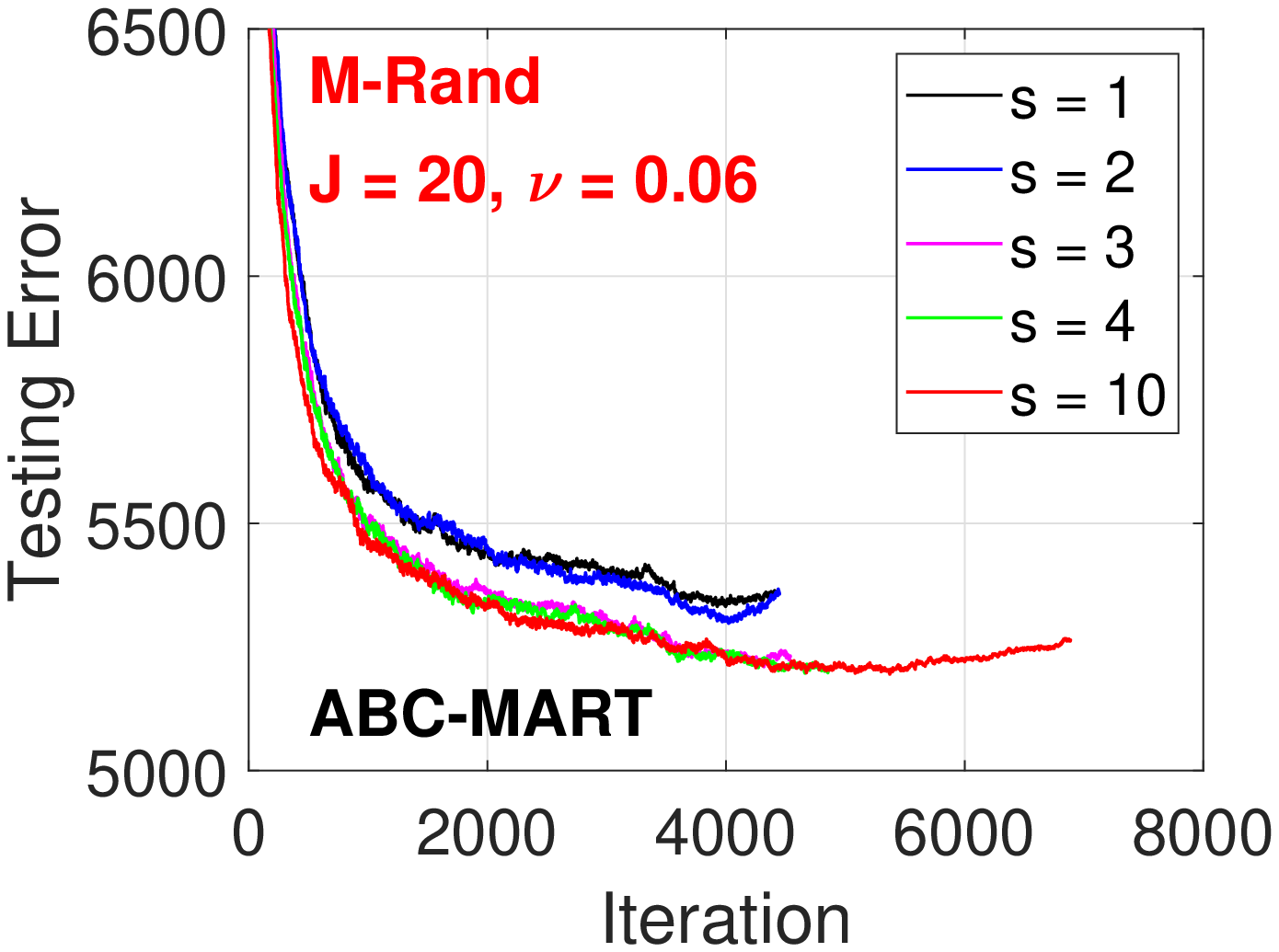}
    \includegraphics[width=2.4in]{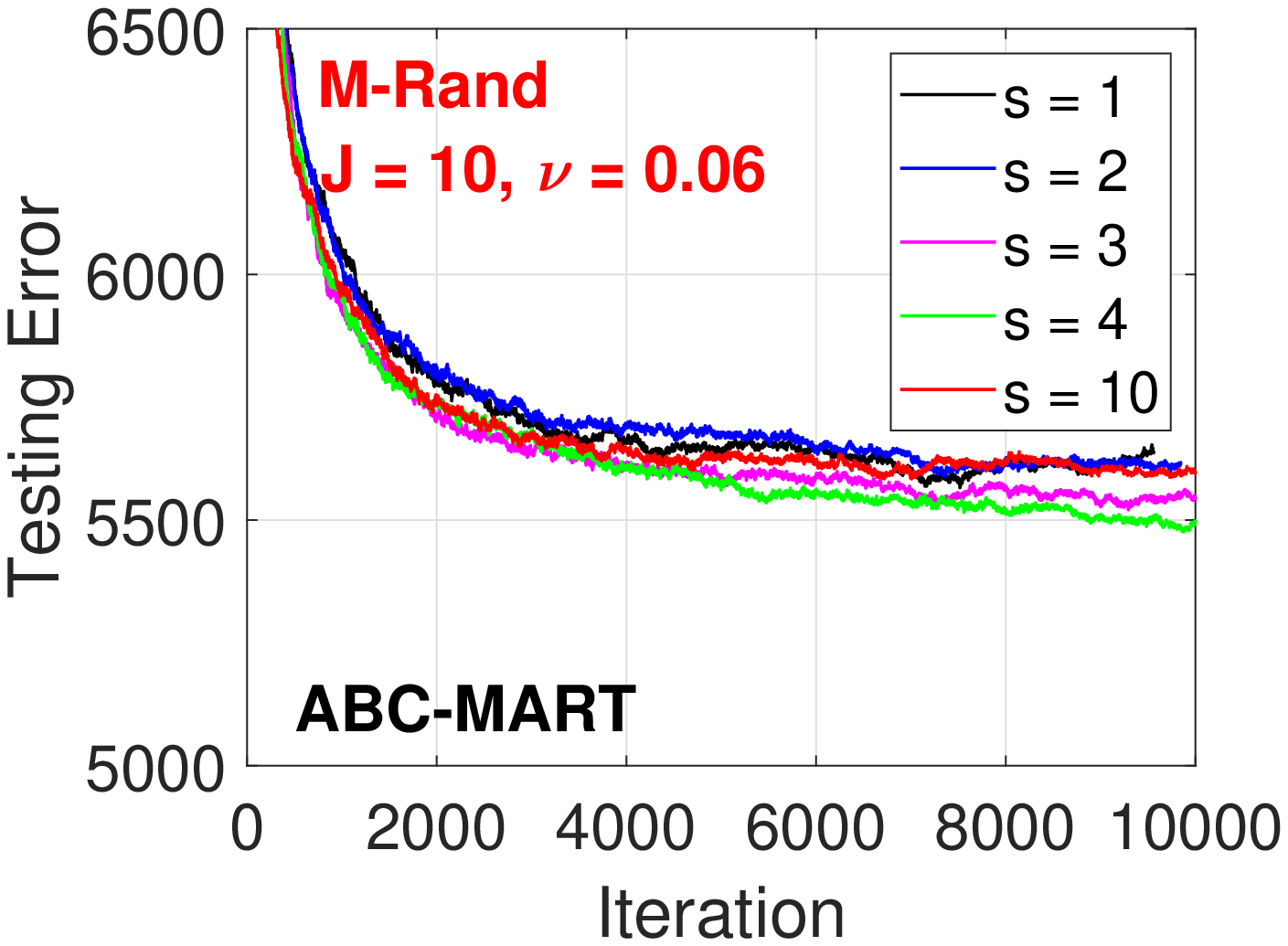}
}

\mbox{
    \includegraphics[width=2.4in]{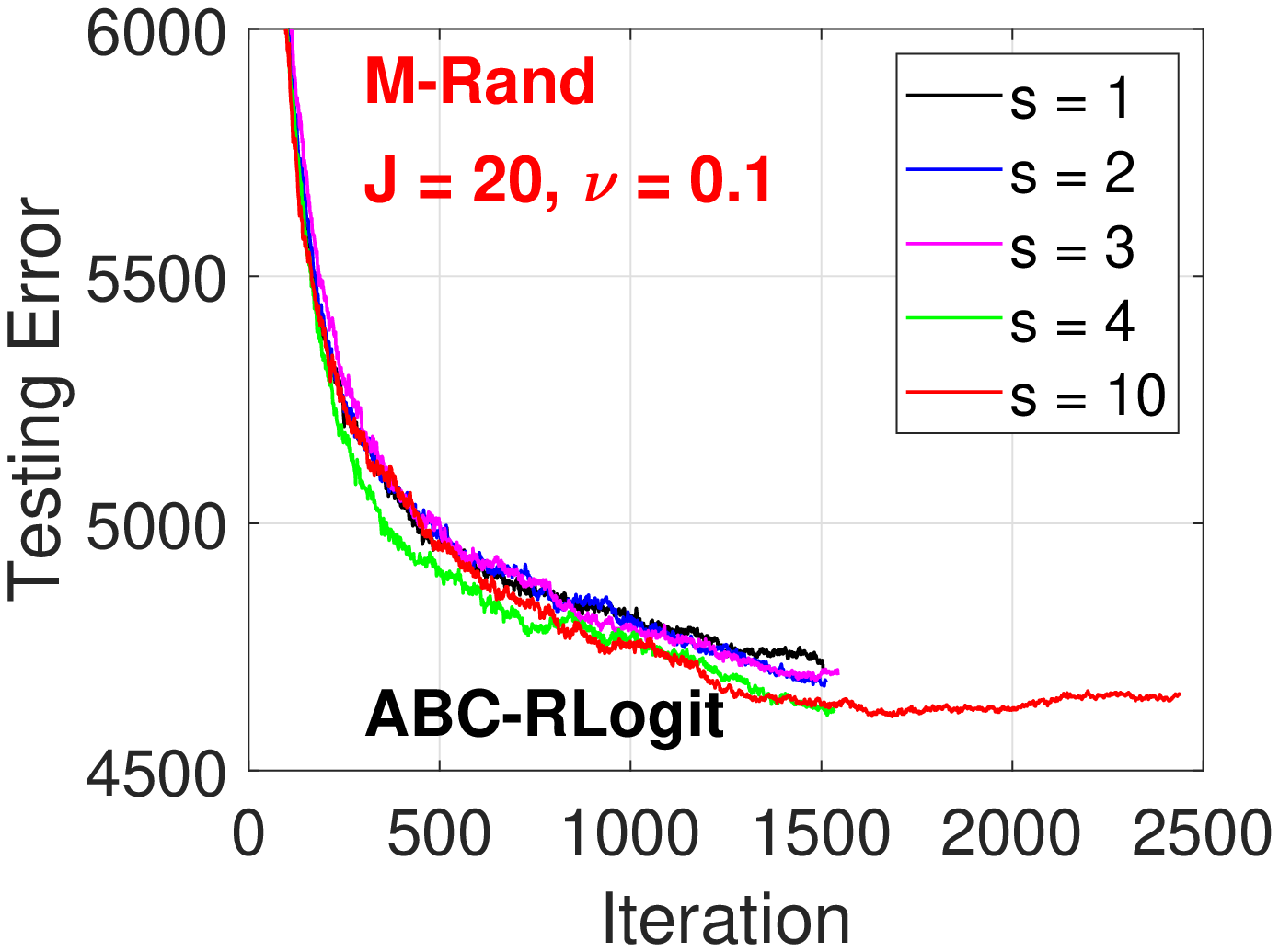}
    \includegraphics[width=2.4in]{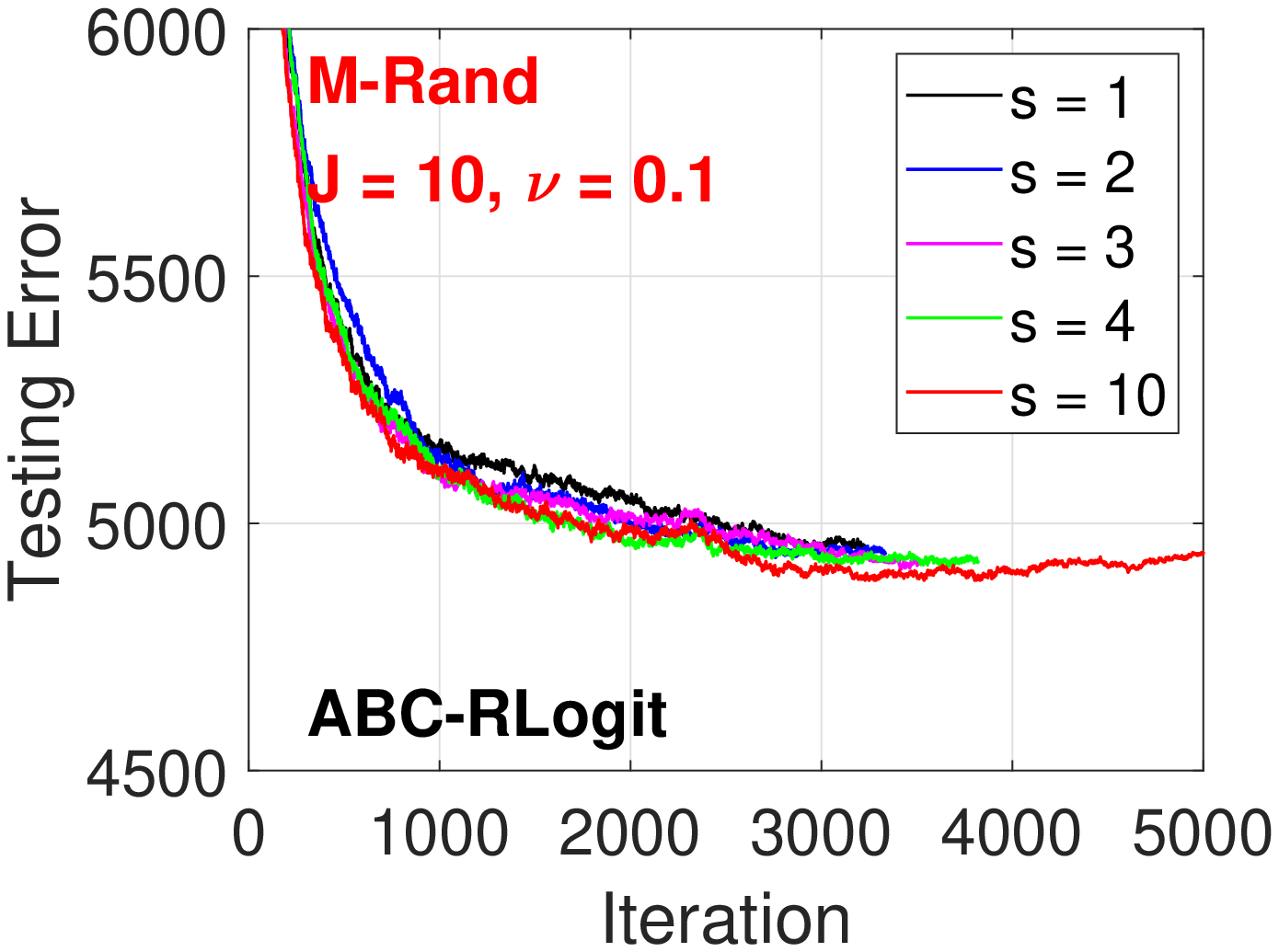}
}

\mbox{
    \includegraphics[width=2.4in]{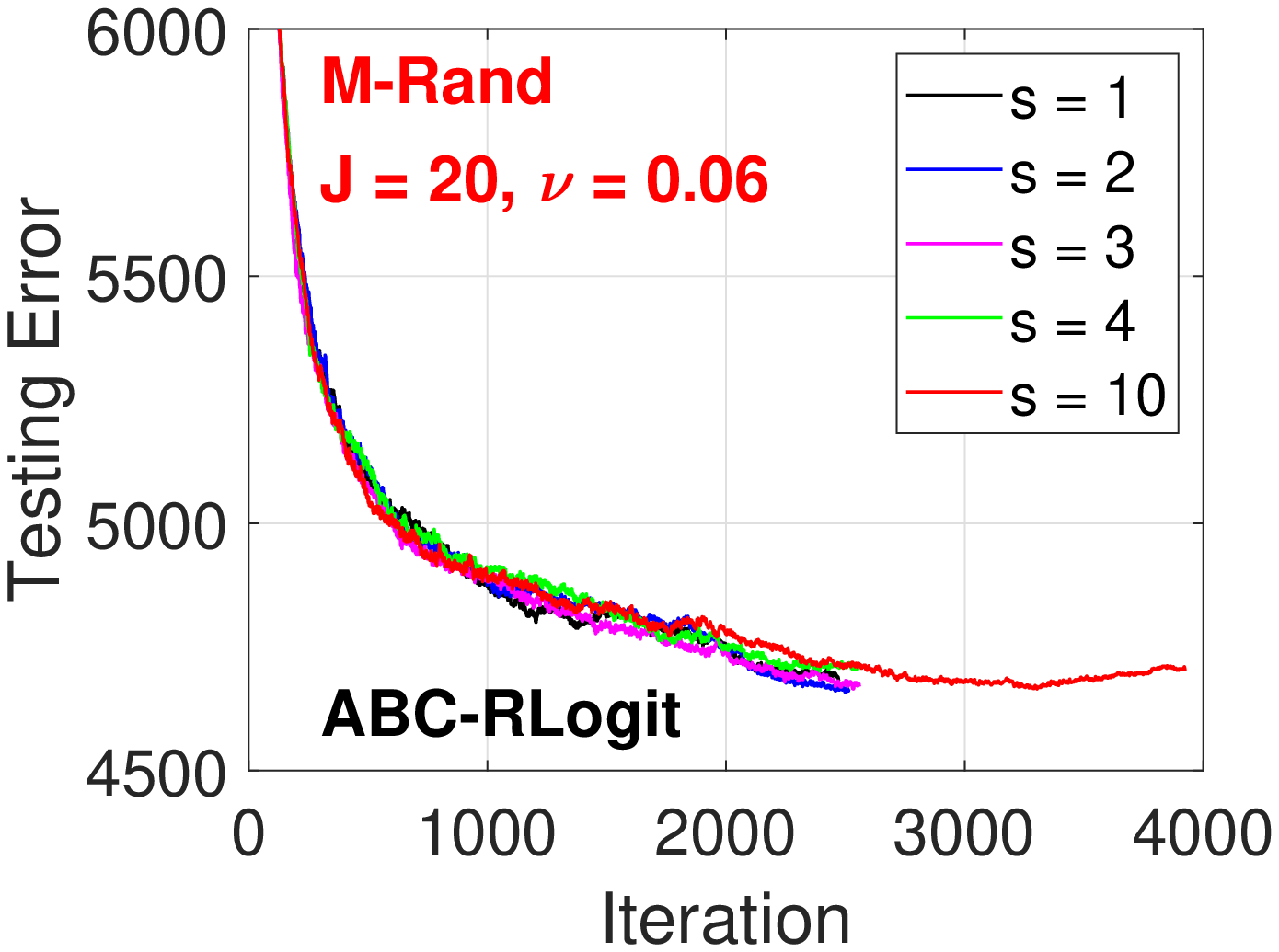}
    \includegraphics[width=2.4in]{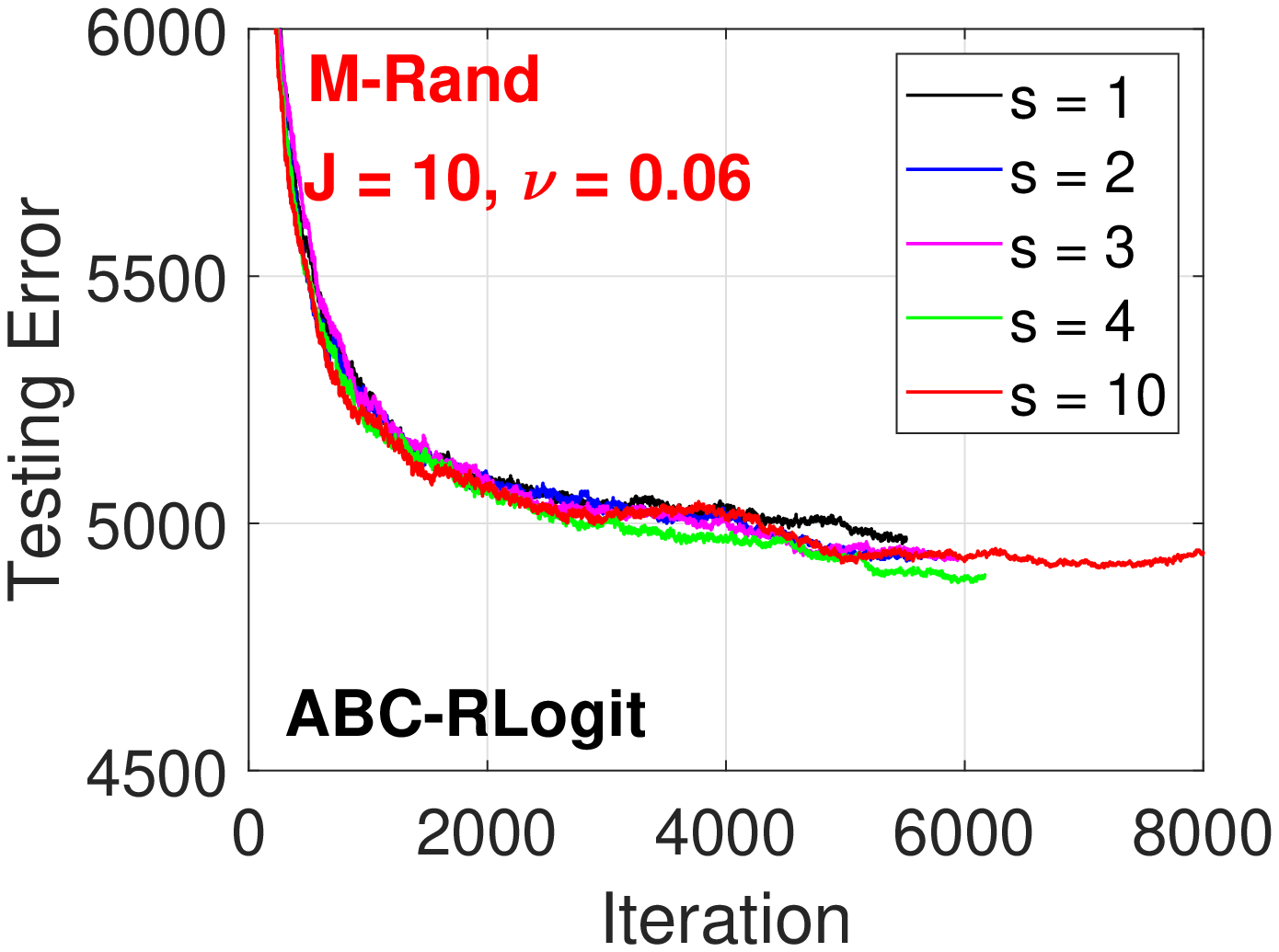}
}
\end{center}
\caption{{\em M-Rand} dataset. Test classification errors based on the ``$s$-worst classes'' search strategy for both ABC-MART and ABC-RobustLogitBoost, for $s\in\{1,2,3,4,10\}$.   }\label{fig:M-Rand_s}
\end{figure}

\begin{figure}[h]
\begin{center}
\mbox{
    \includegraphics[width=2.4in]{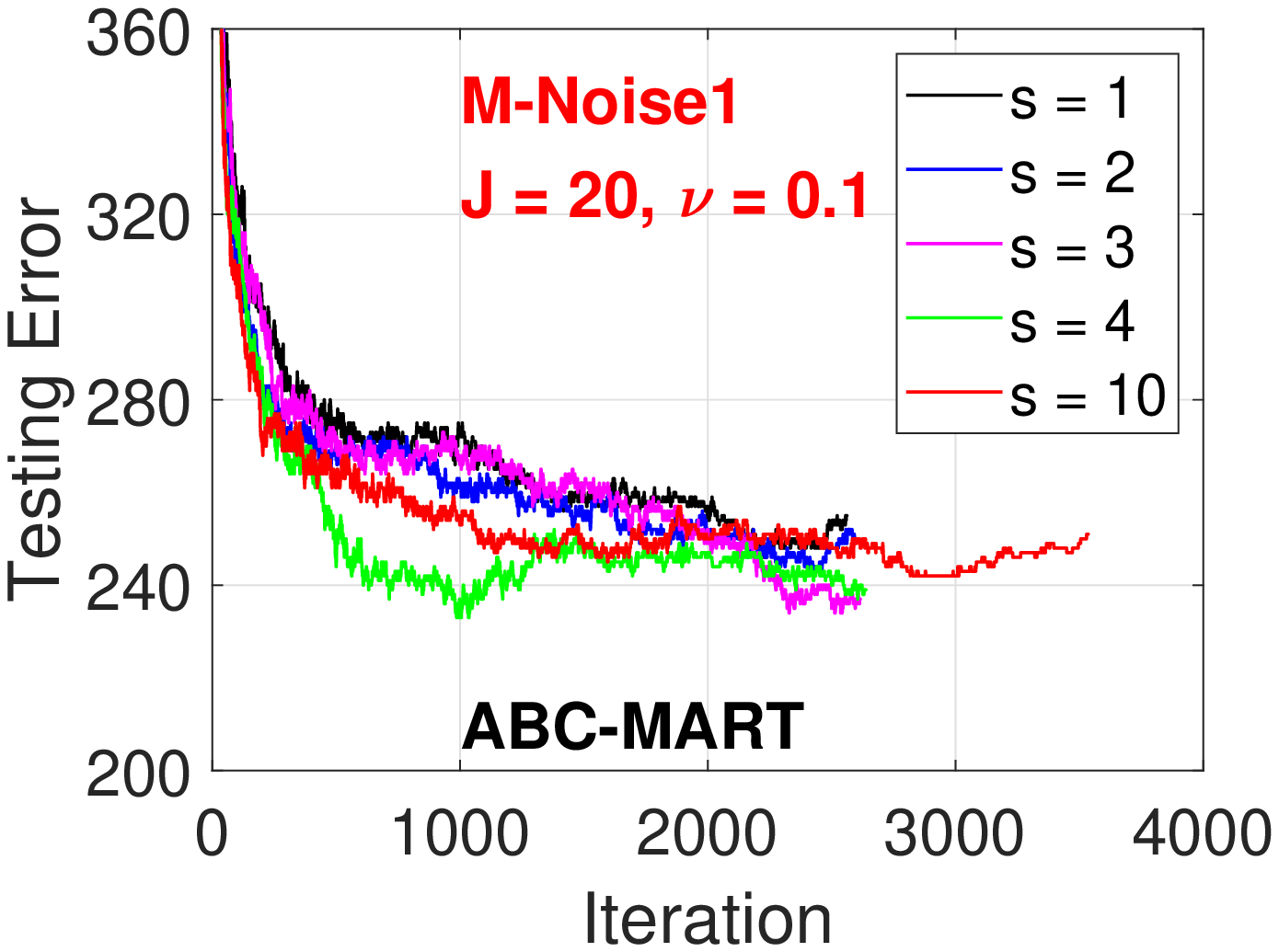}
    \includegraphics[width=2.4in]{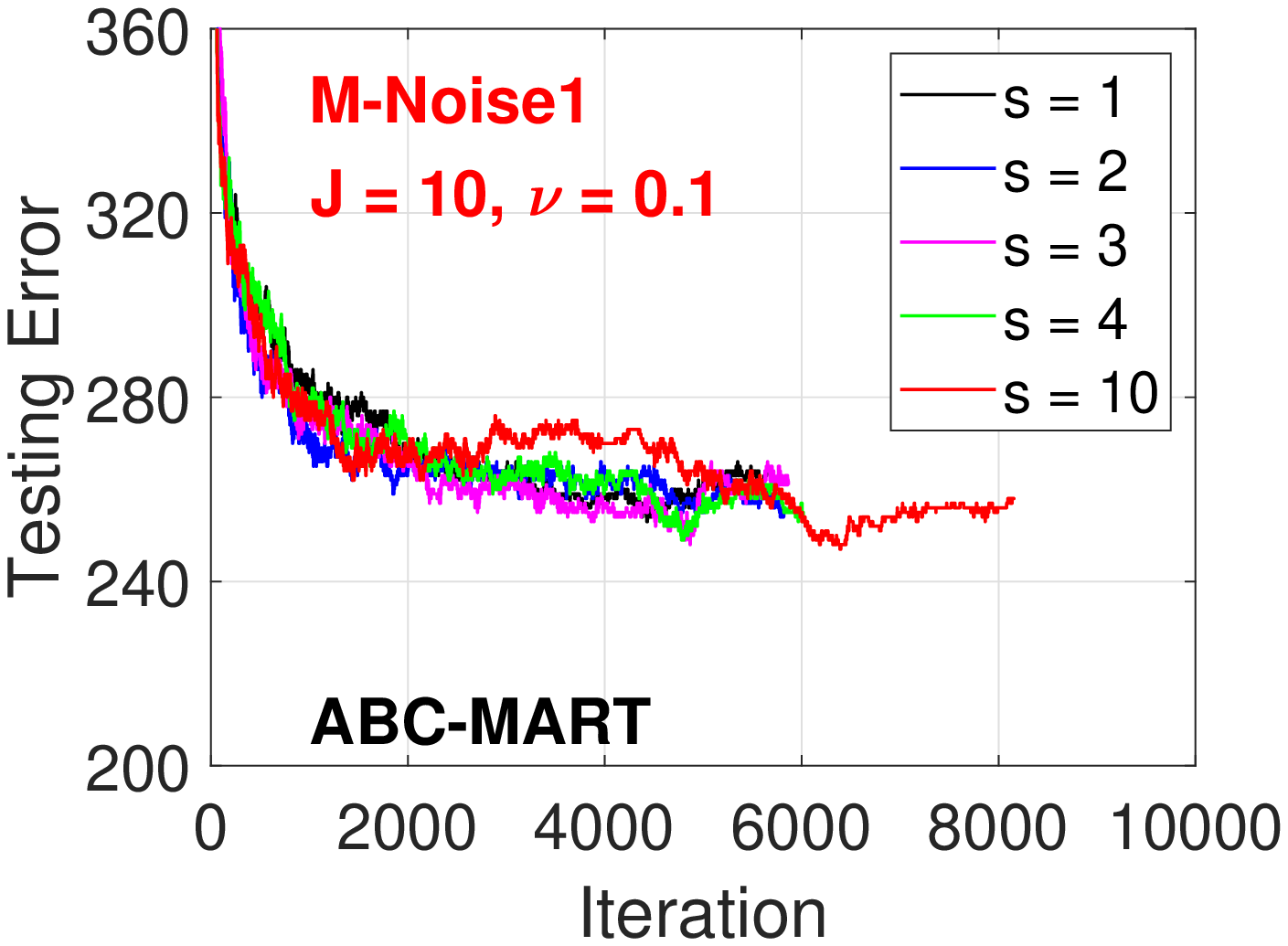}
}

\mbox{
    \includegraphics[width=2.4in]{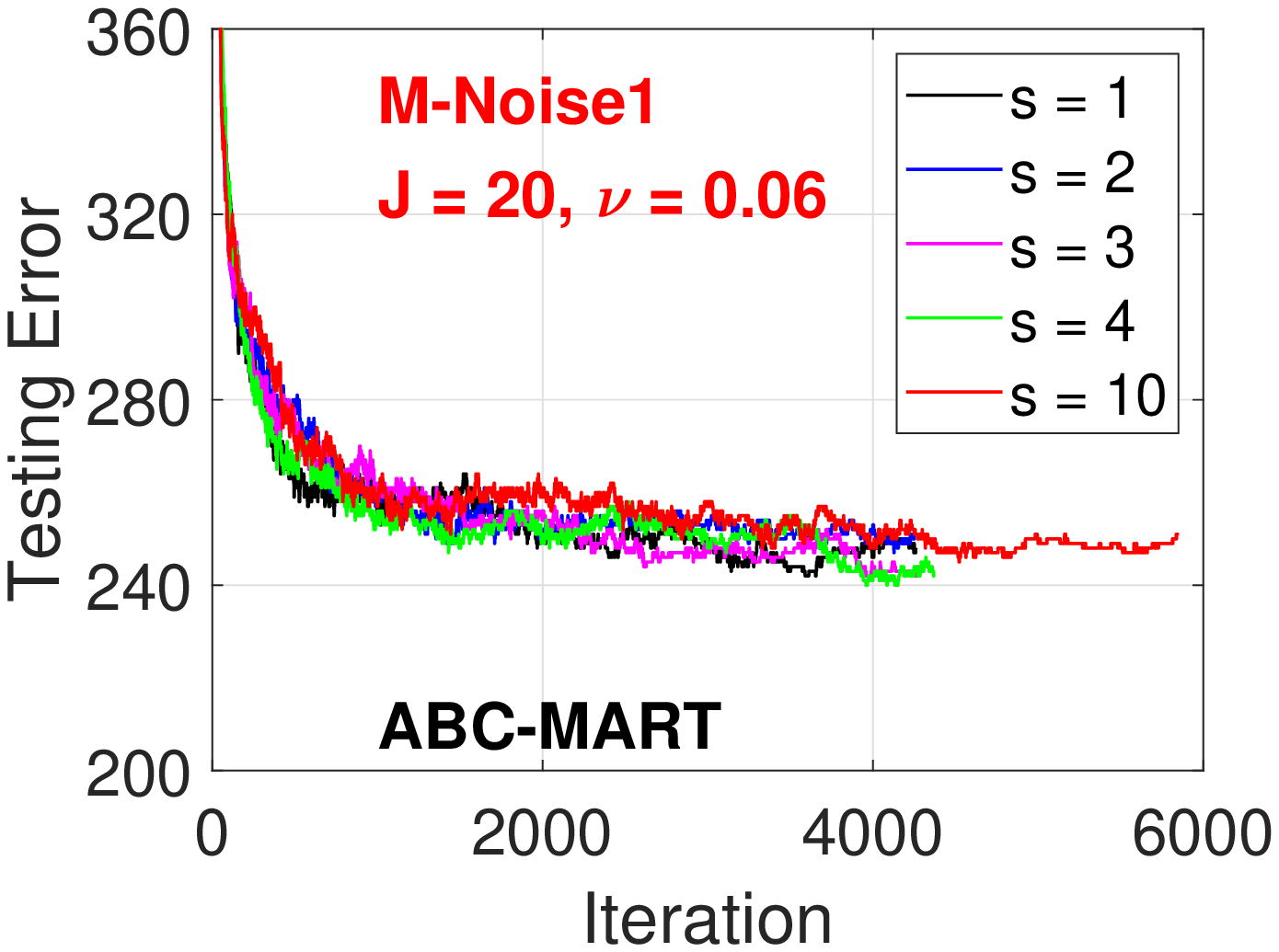}
    \includegraphics[width=2.4in]{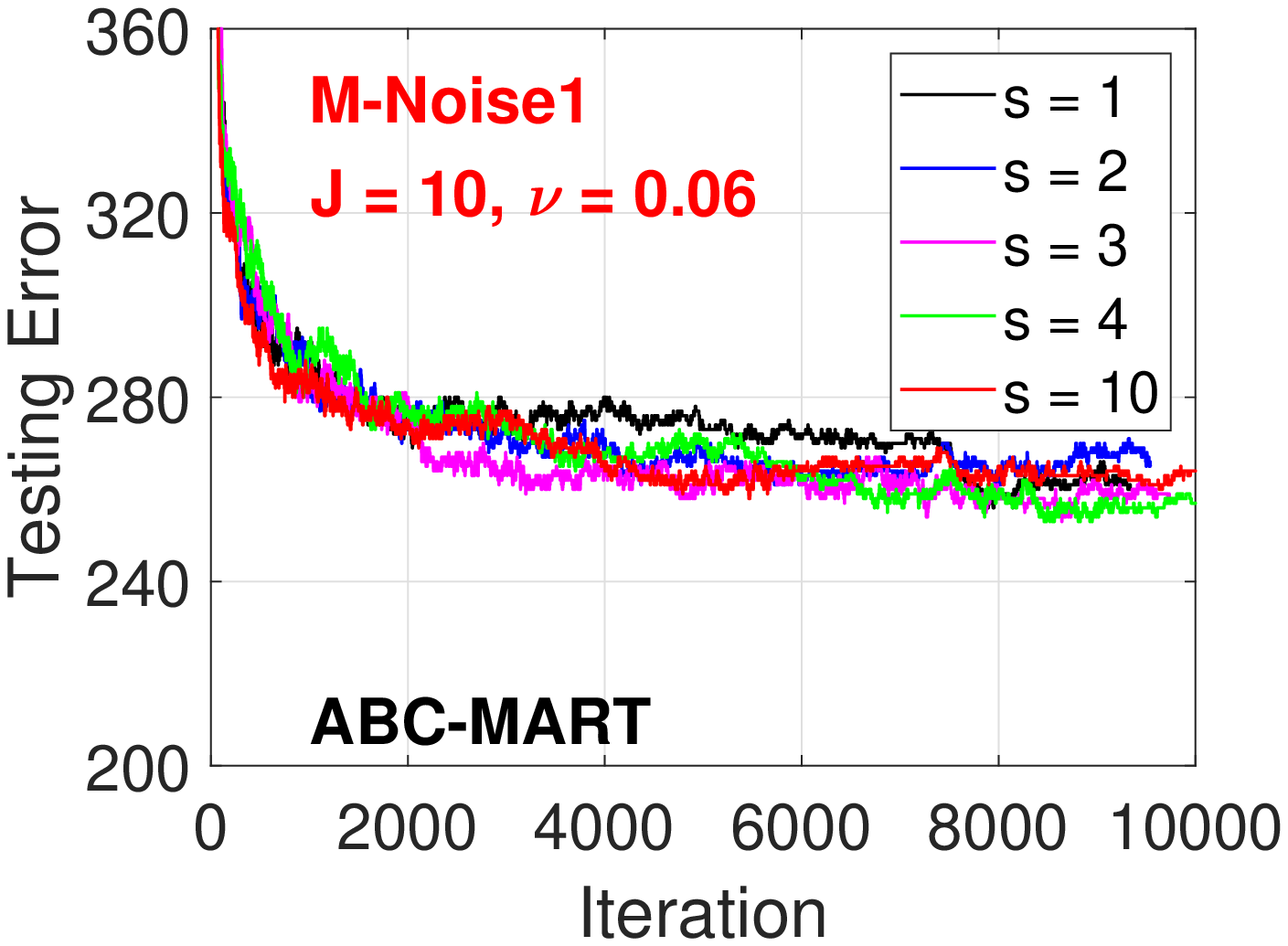}
}

\mbox{
    \includegraphics[width=2.4in]{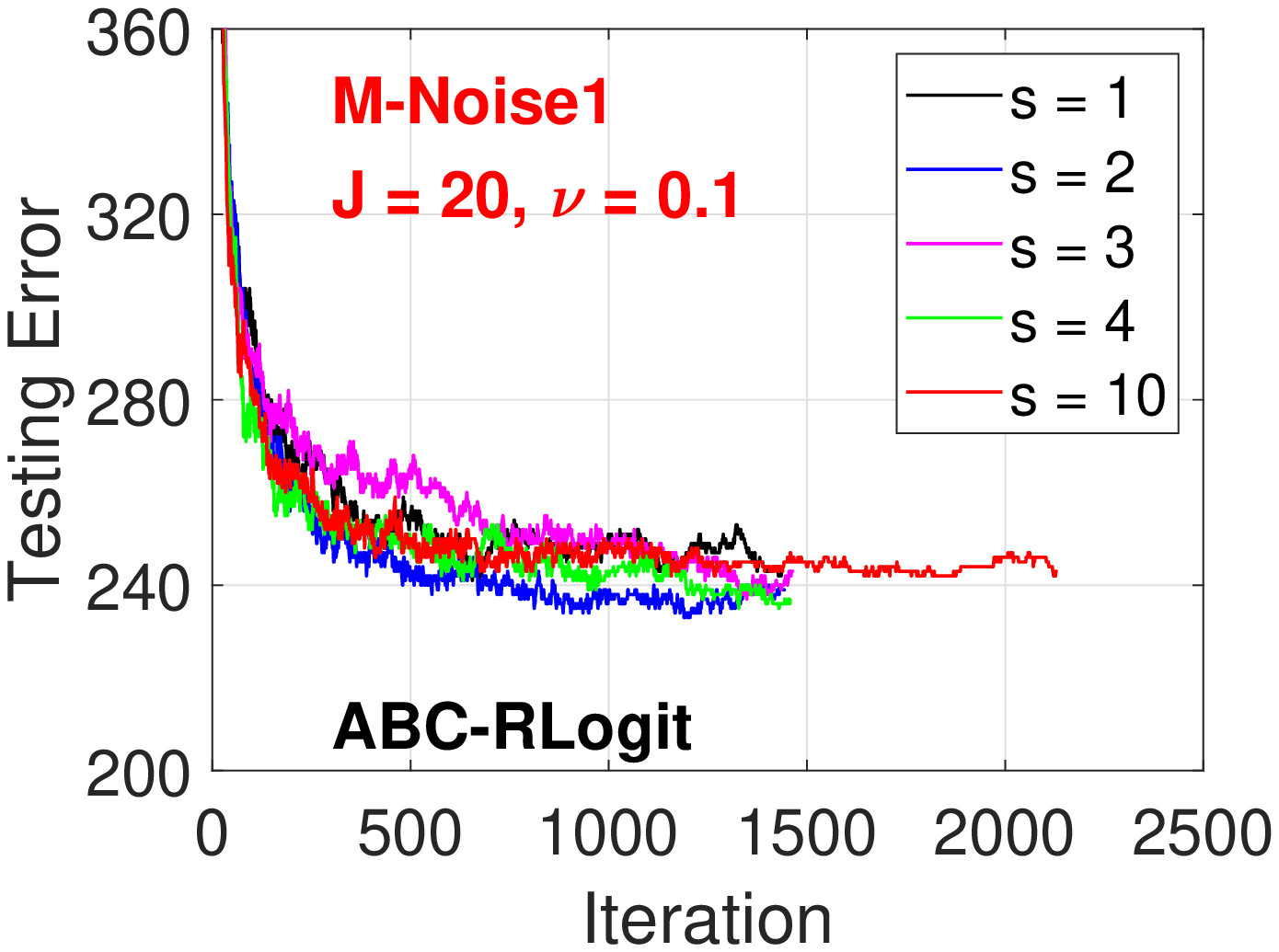}
    \includegraphics[width=2.4in]{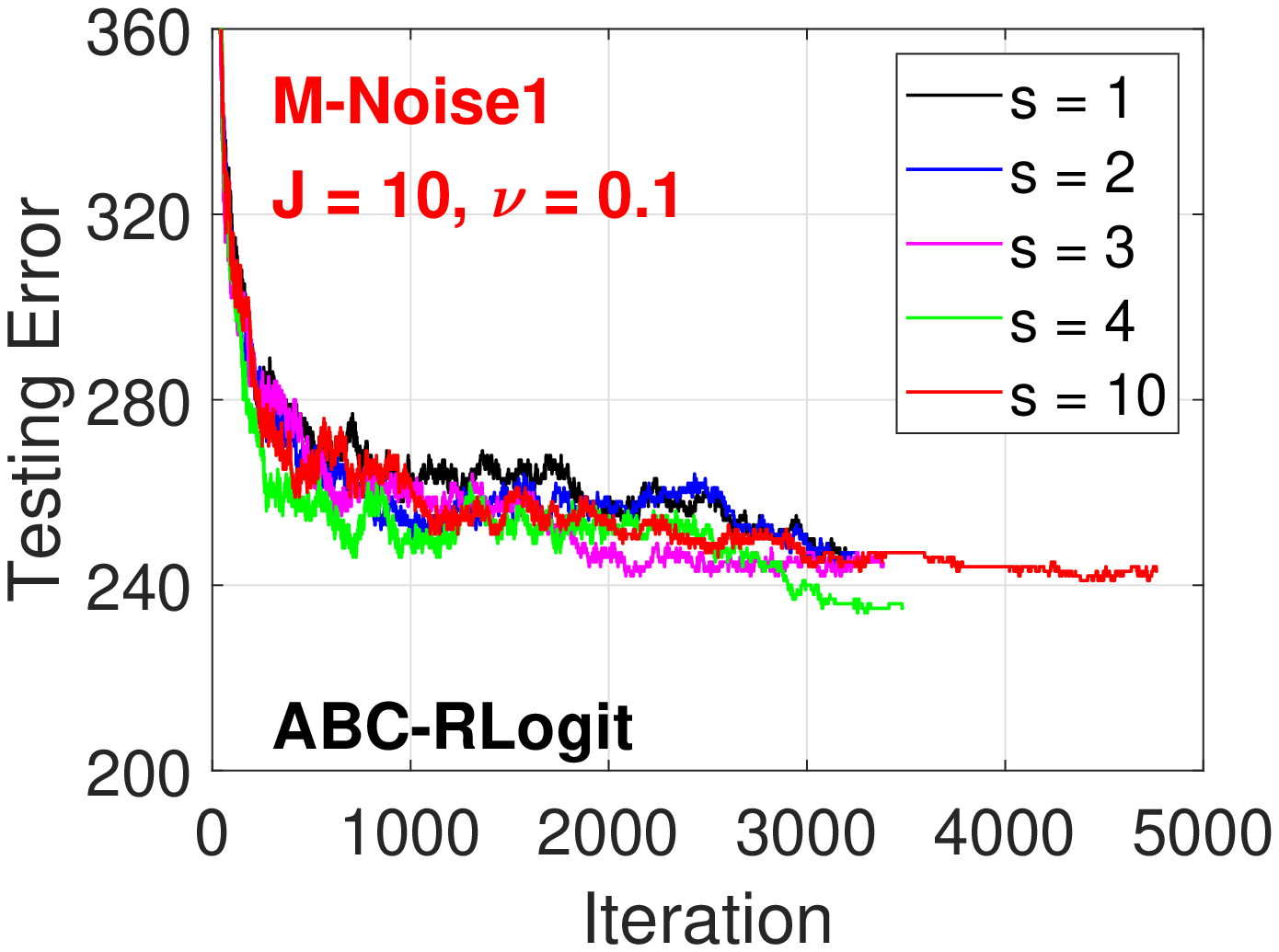}
}

\mbox{
    \includegraphics[width=2.4in]{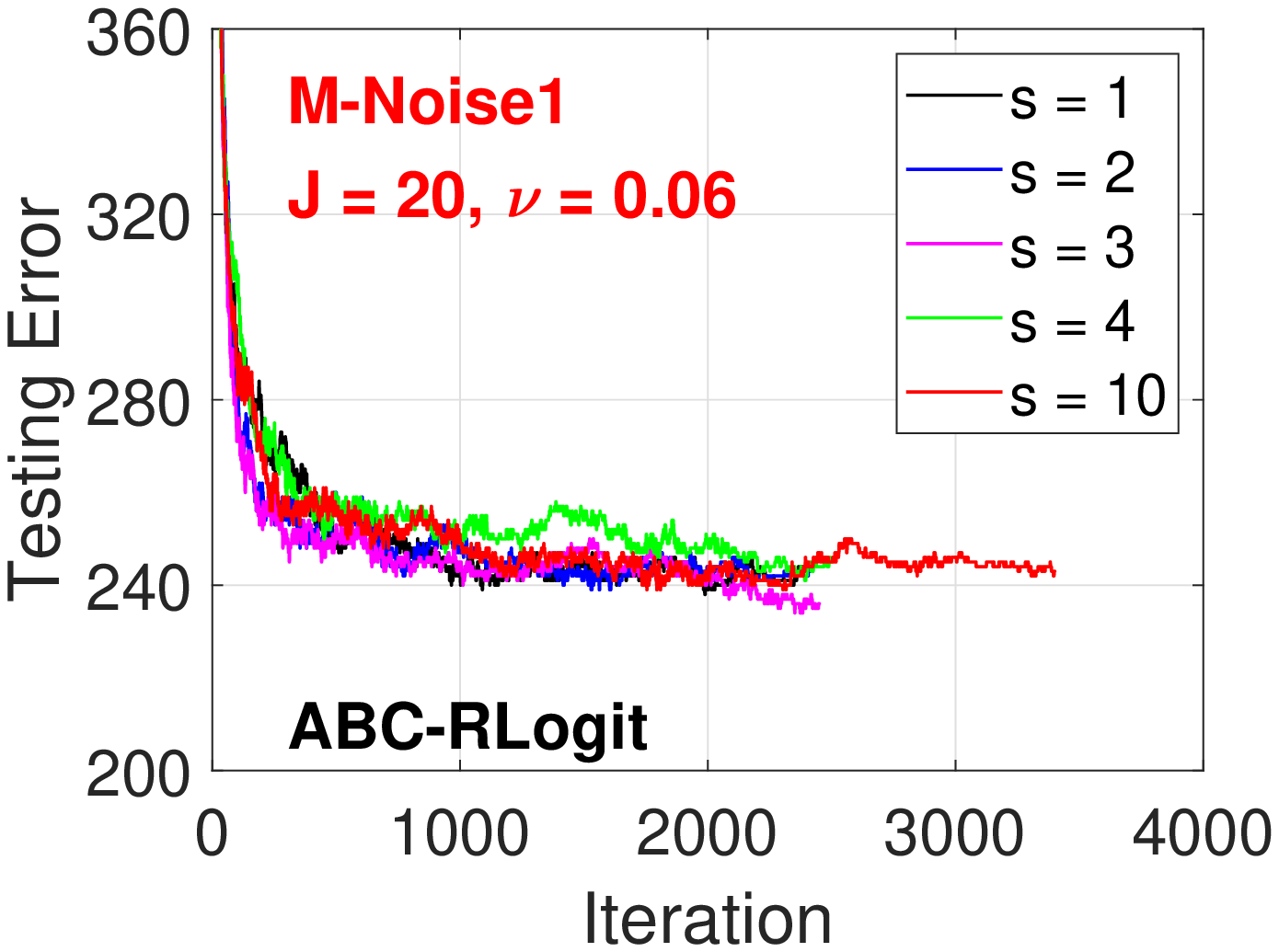}
    \includegraphics[width=2.4in]{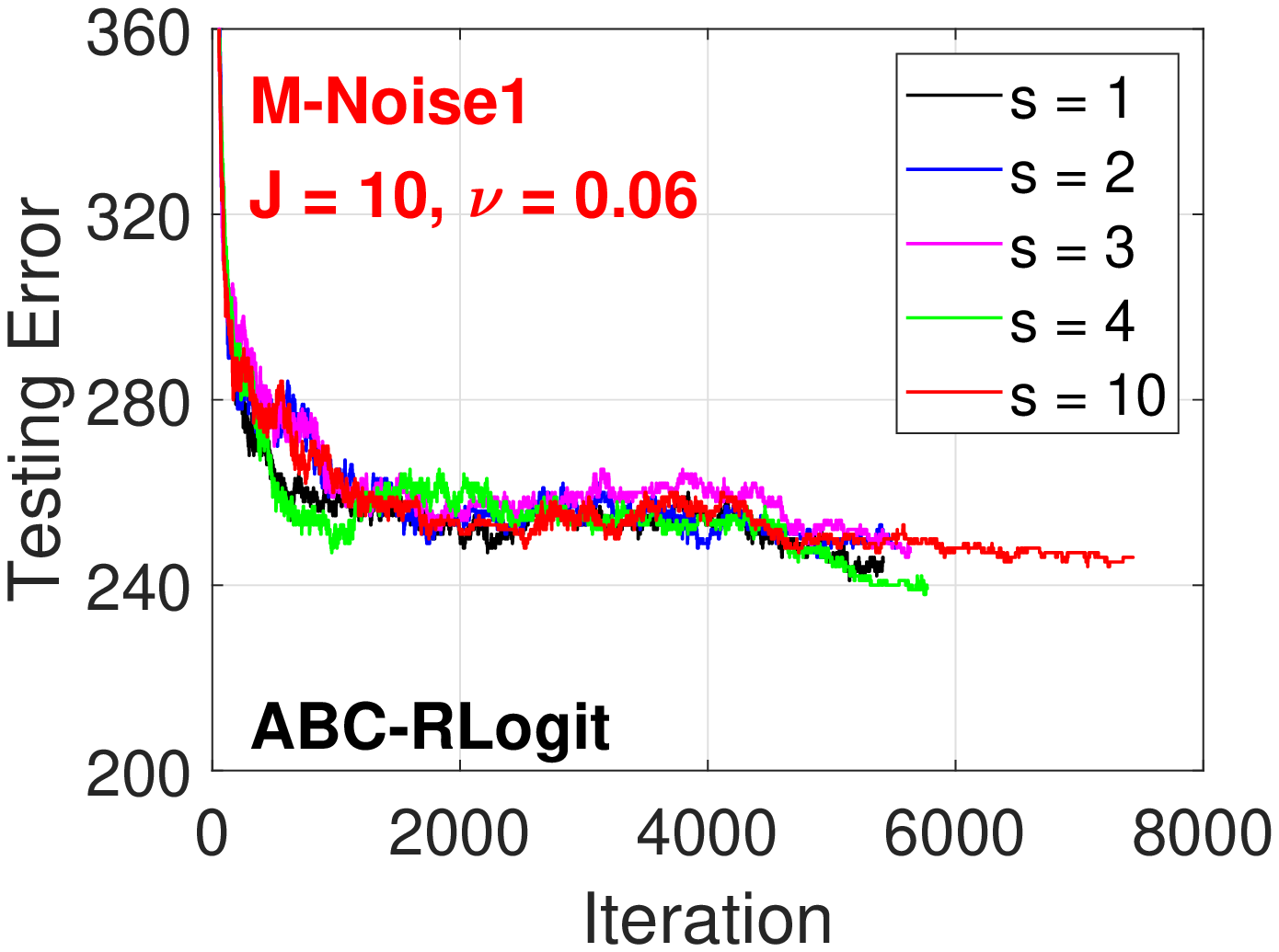}
}
\end{center}
\caption{{\em M-Noise1} dataset. Test classification errors based on the ``$s$-worst classes search strategy for both ABC-MART and ABC-RobustLogitBoost, for $s\in\{1,2,3,4,10\}$.   }\label{fig:M-Noise1_s}
\end{figure}

\begin{figure}[h]
\begin{center}
\mbox{
    \includegraphics[width=2.4in]{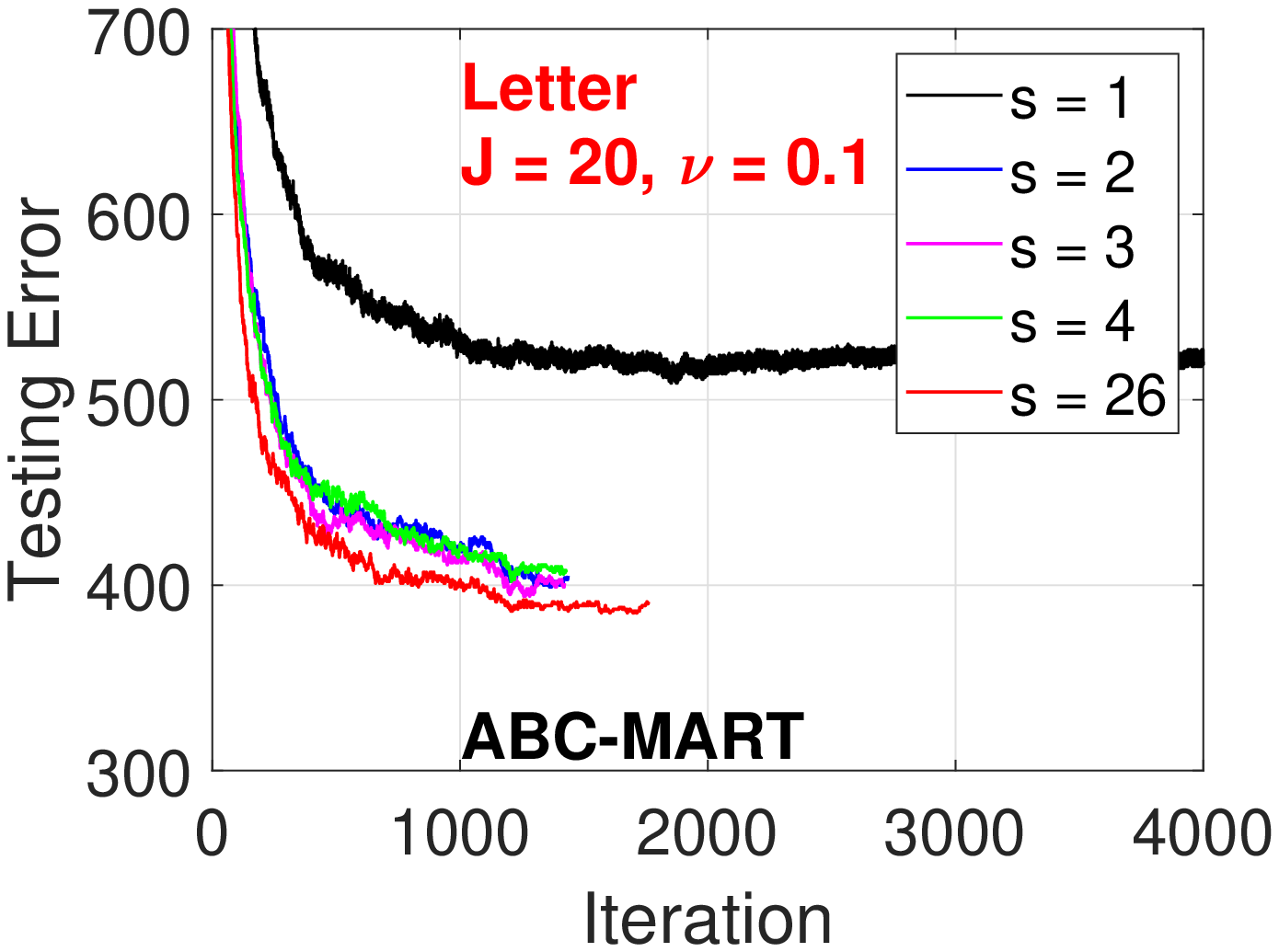}
    \includegraphics[width=2.4in]{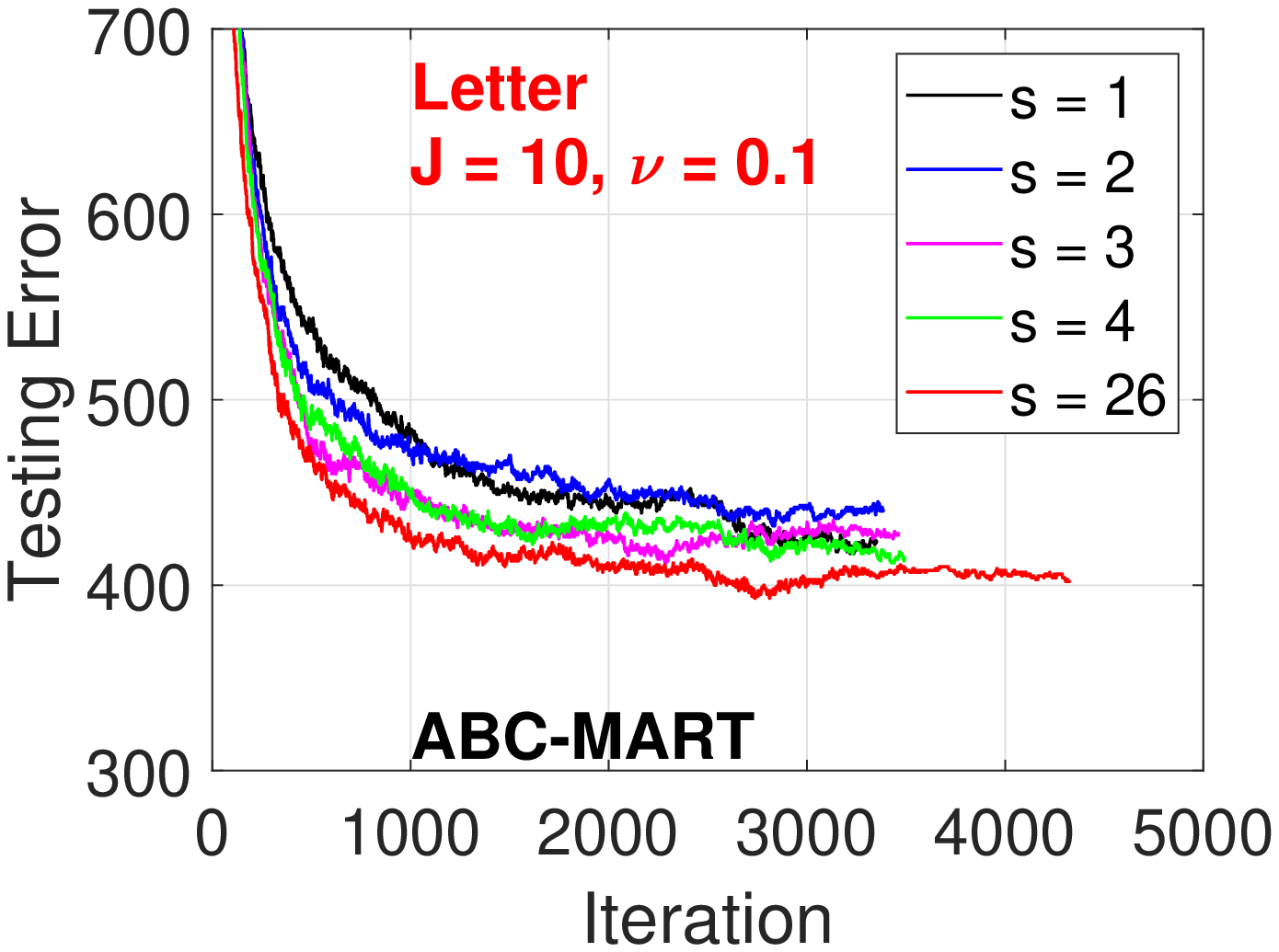}
}

\mbox{
    \includegraphics[width=2.4in]{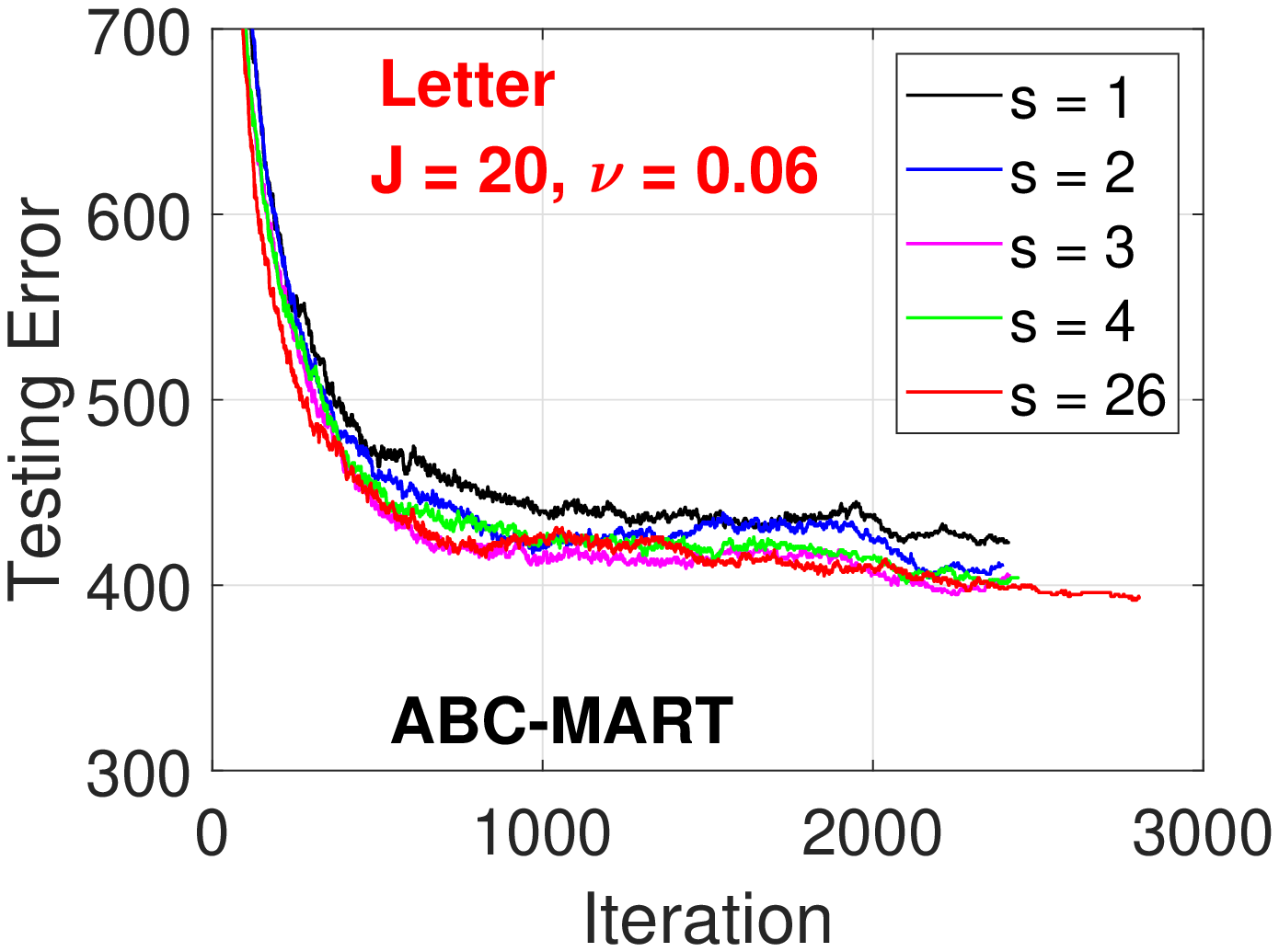}
    \includegraphics[width=2.4in]{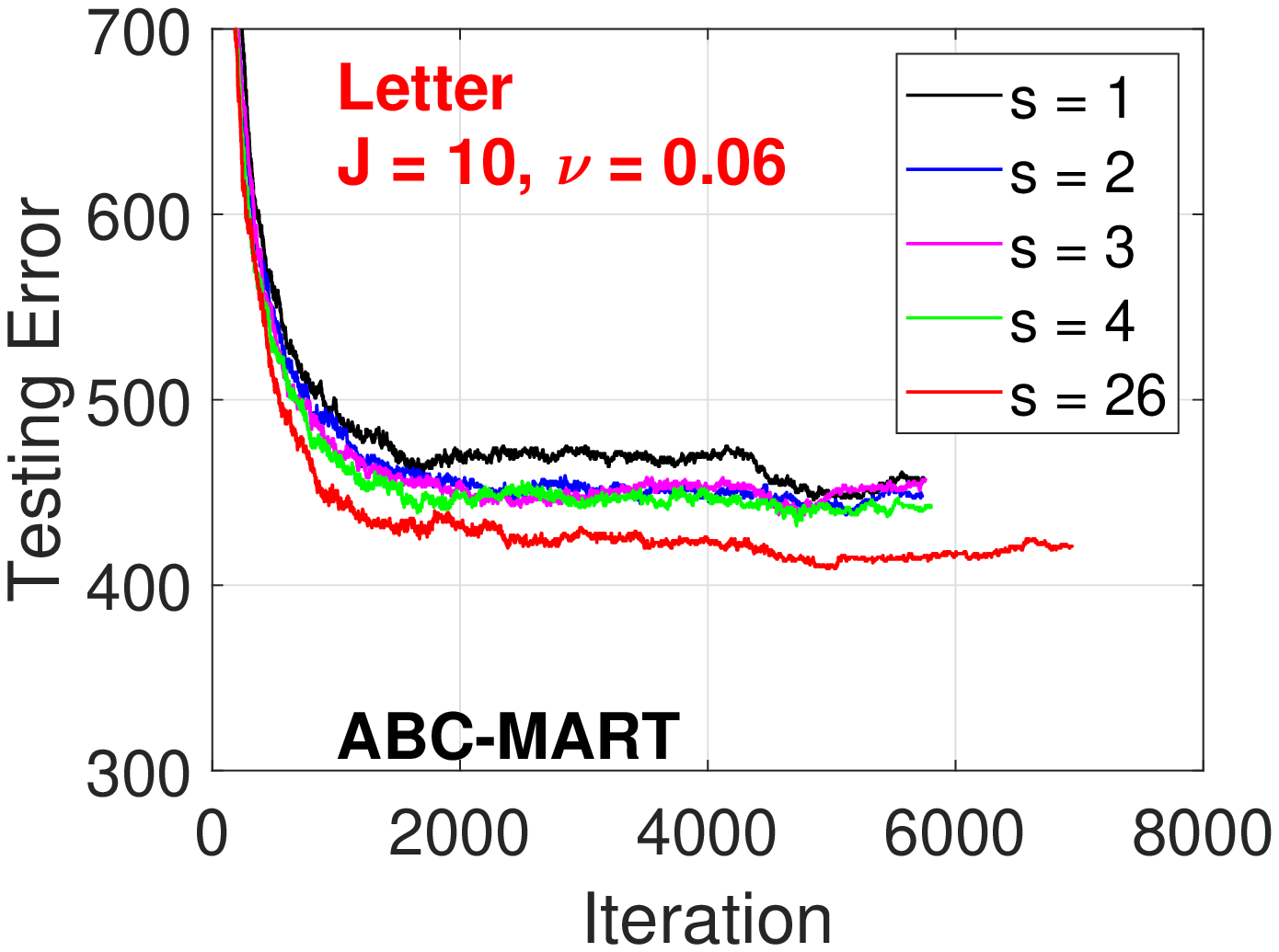}
}

\mbox{
    \includegraphics[width=2.4in]{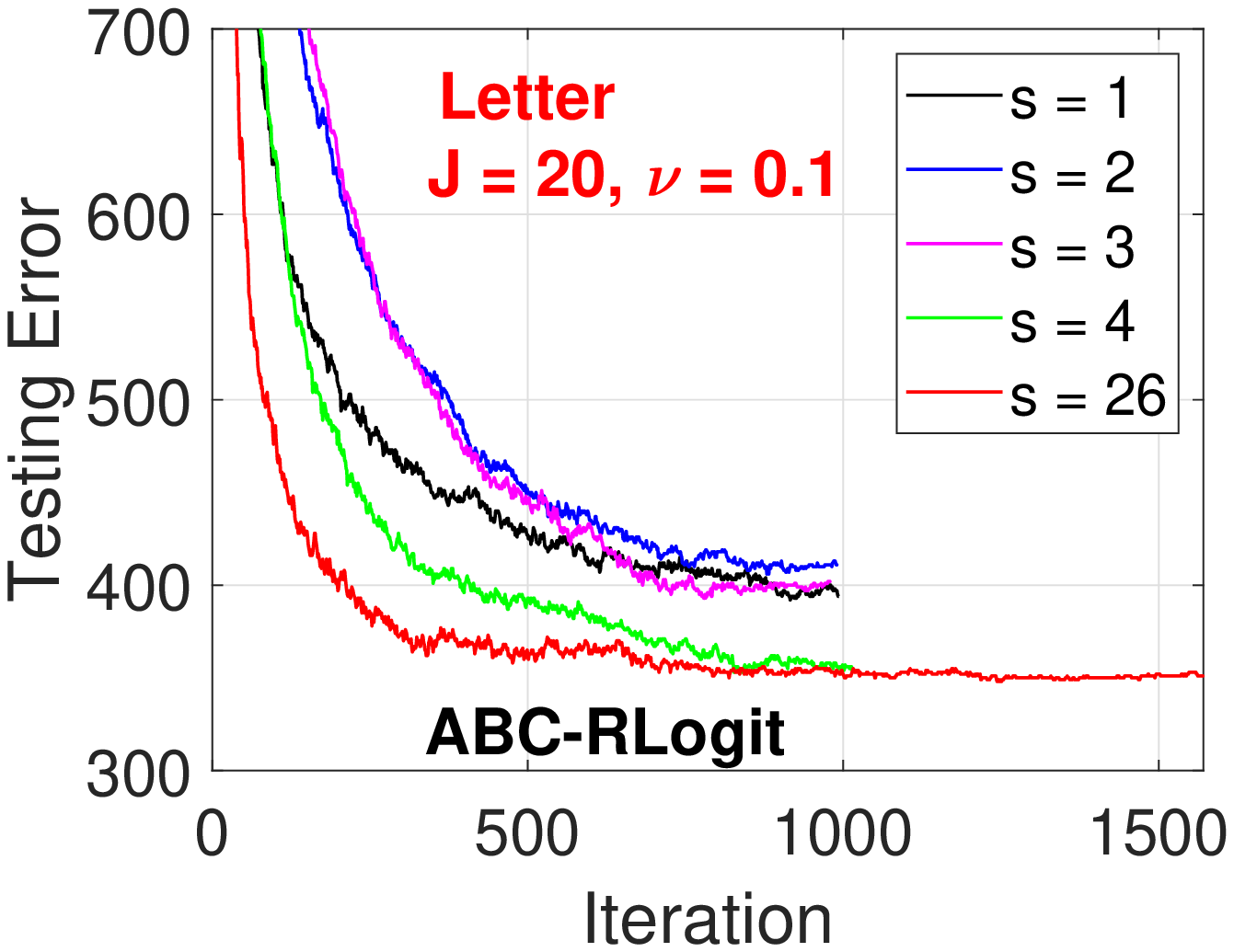}
    \includegraphics[width=2.4in]{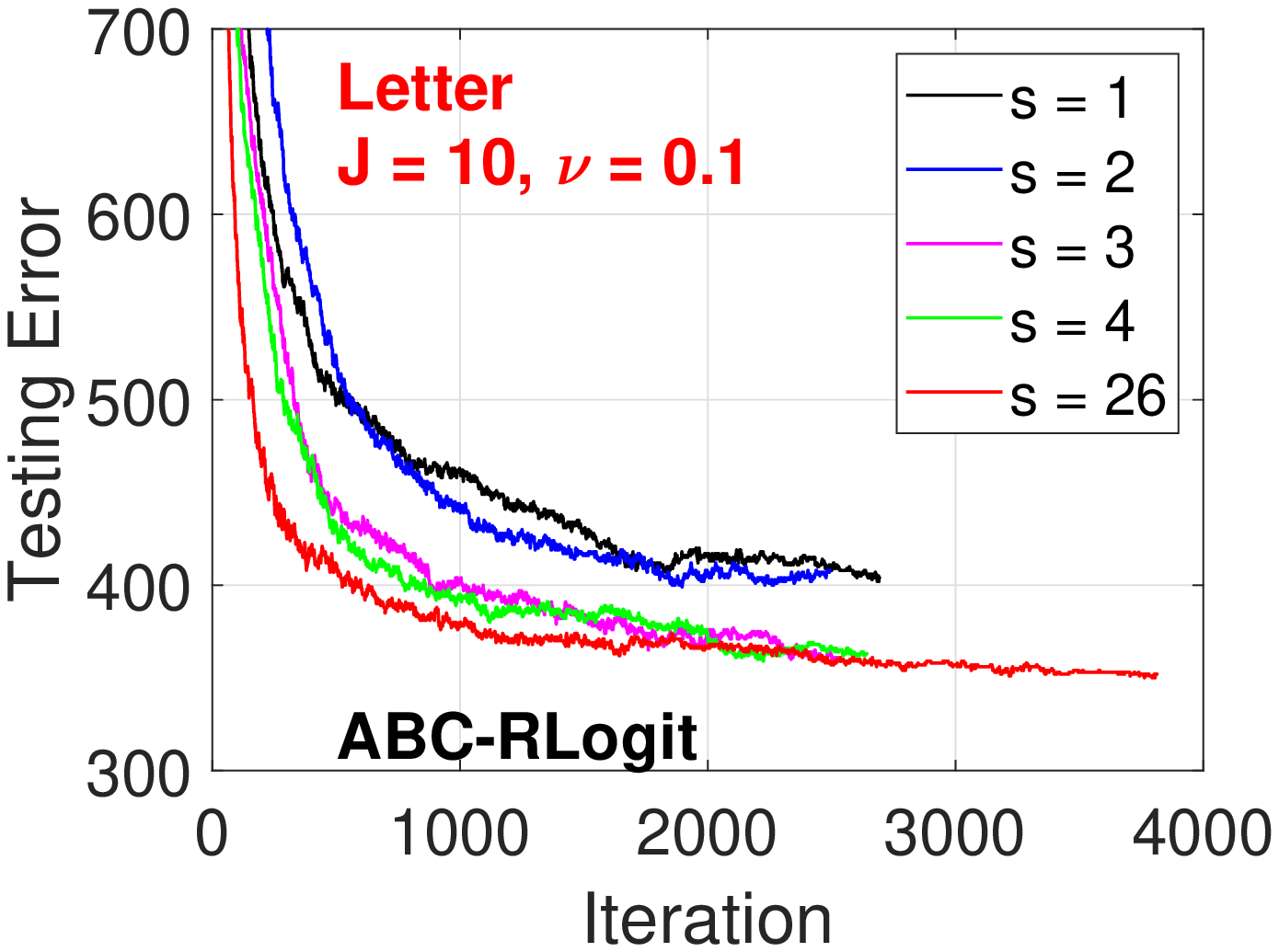}
}

\mbox{
    \includegraphics[width=2.4in]{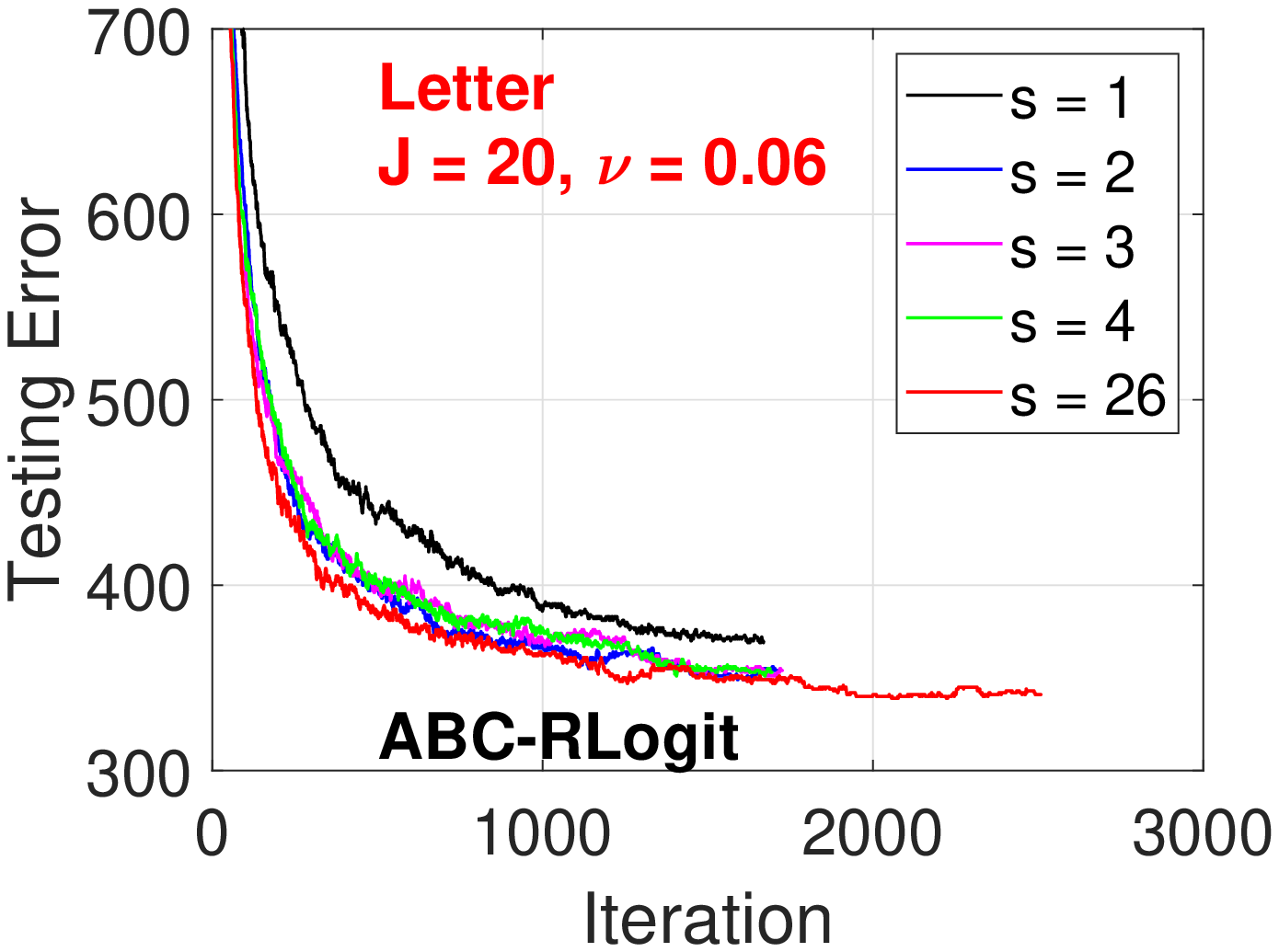}
    \includegraphics[width=2.4in]{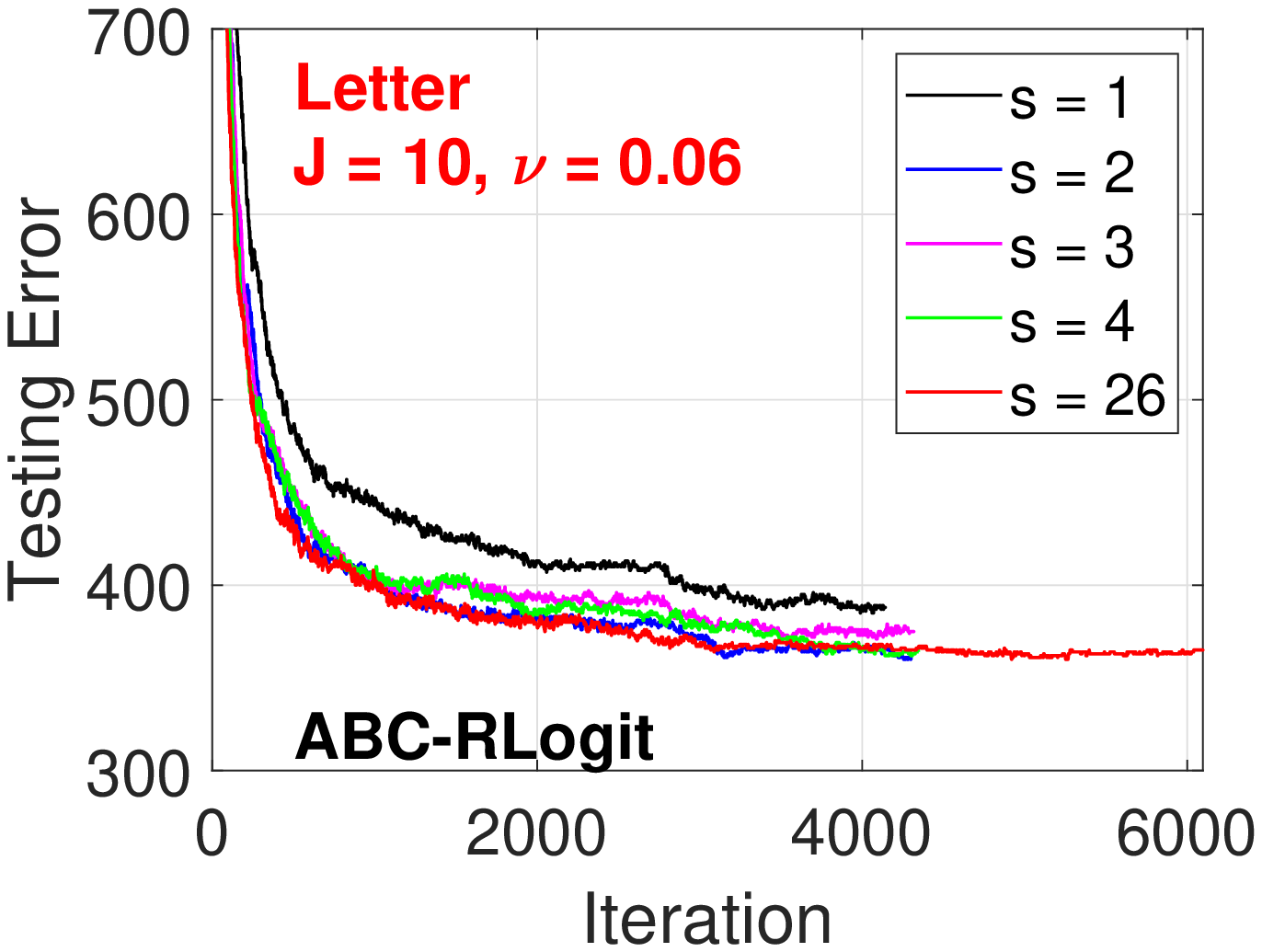}
}
\end{center}

\caption{{\em Letter} dataset. Test classification errors based on the ``$s$-worst classes'' search strategy for both ABC-MART and ABC-RobustLogitBoost, for $s\in\{1,2,3,4,26\}$.  We can see the ``catastrophic failure'' with ABC-MART, $J=20$, and $\nu=0.1$.  }\label{fig:Letter10k_s}
\end{figure}

\begin{figure}[h]
\begin{center}
\mbox{
    \includegraphics[width=2.2in]{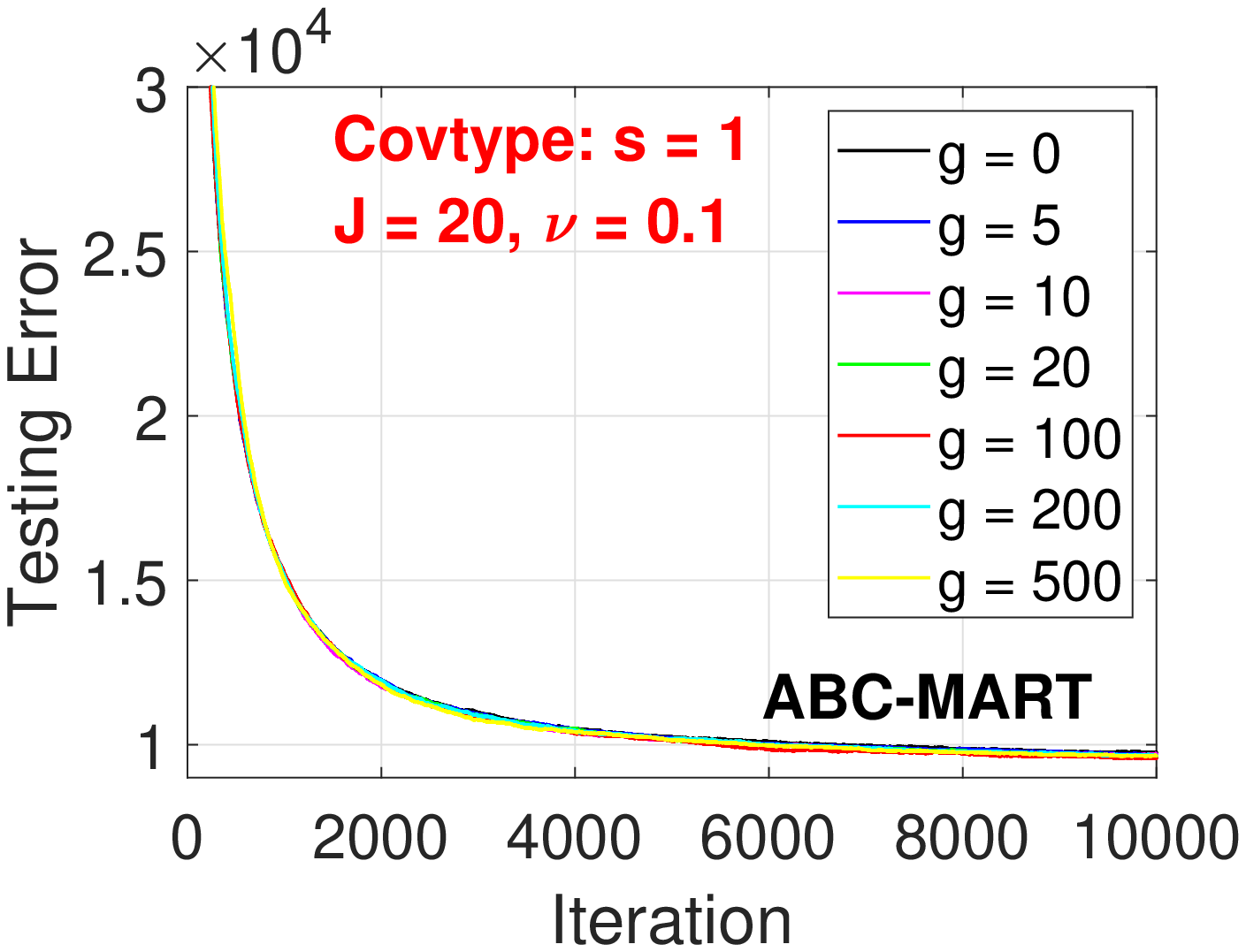}
    \includegraphics[width=2.2in]{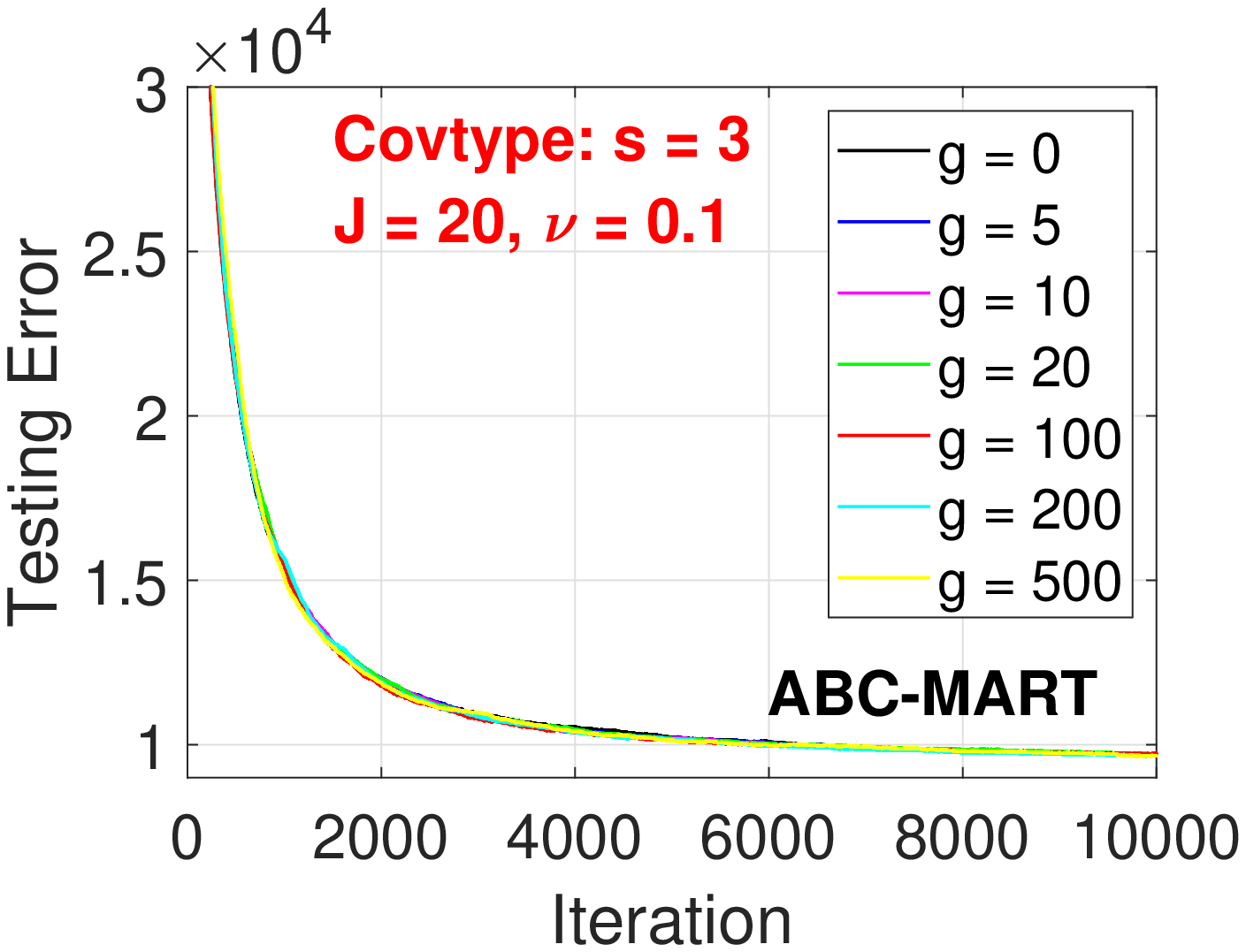}
    \includegraphics[width=2.2in]{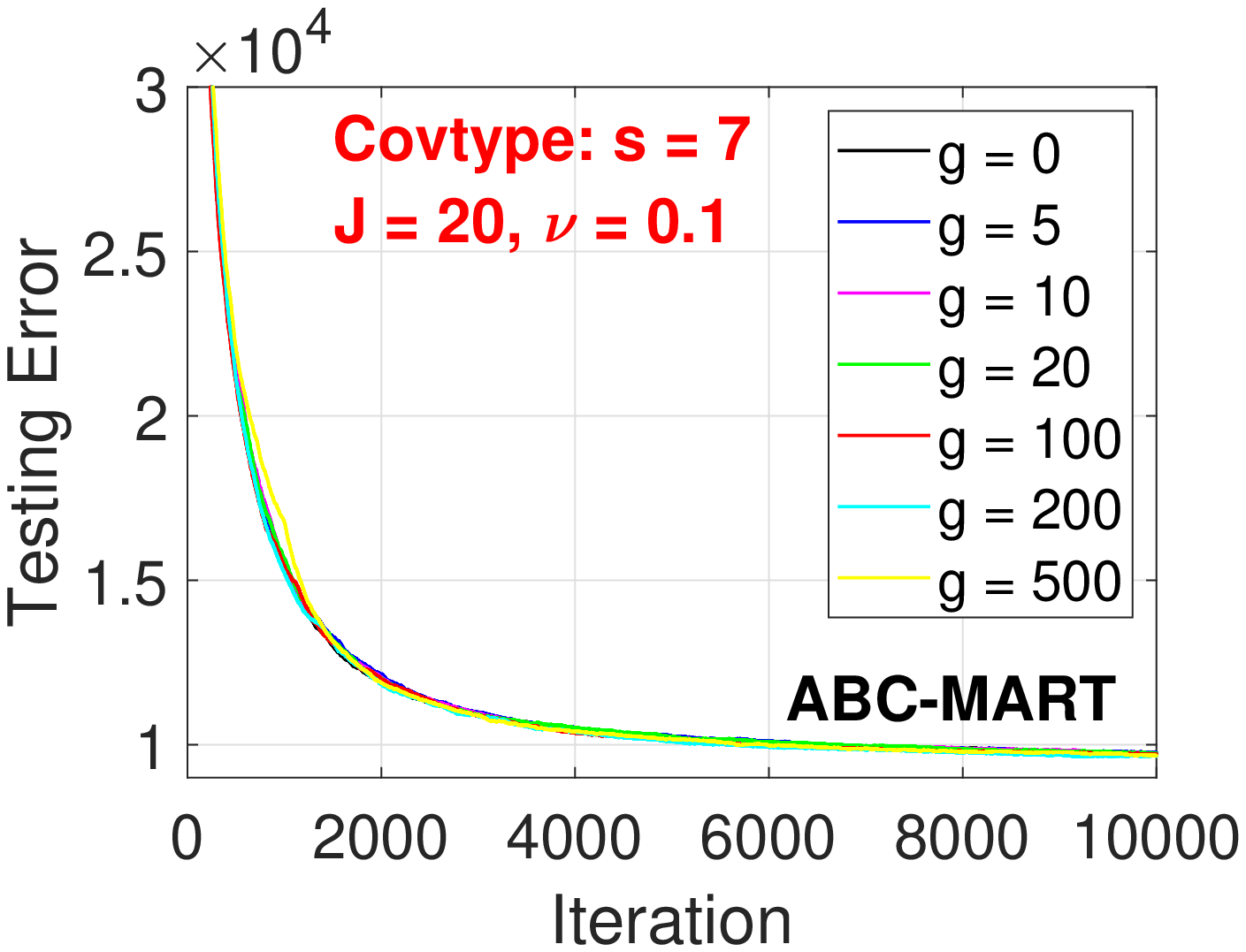}
}

\mbox{
    \includegraphics[width=2.2in]{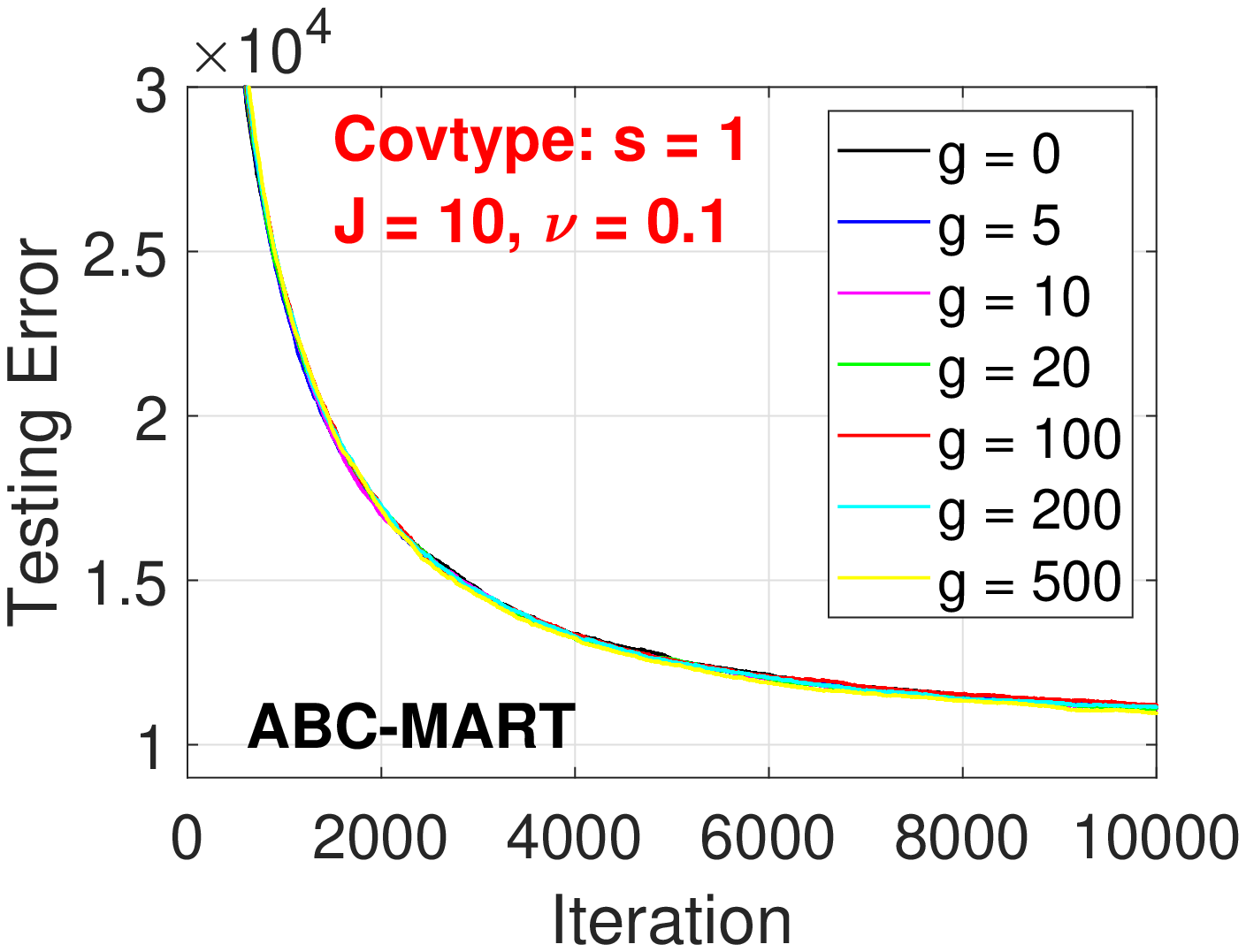}
    \includegraphics[width=2.2in]{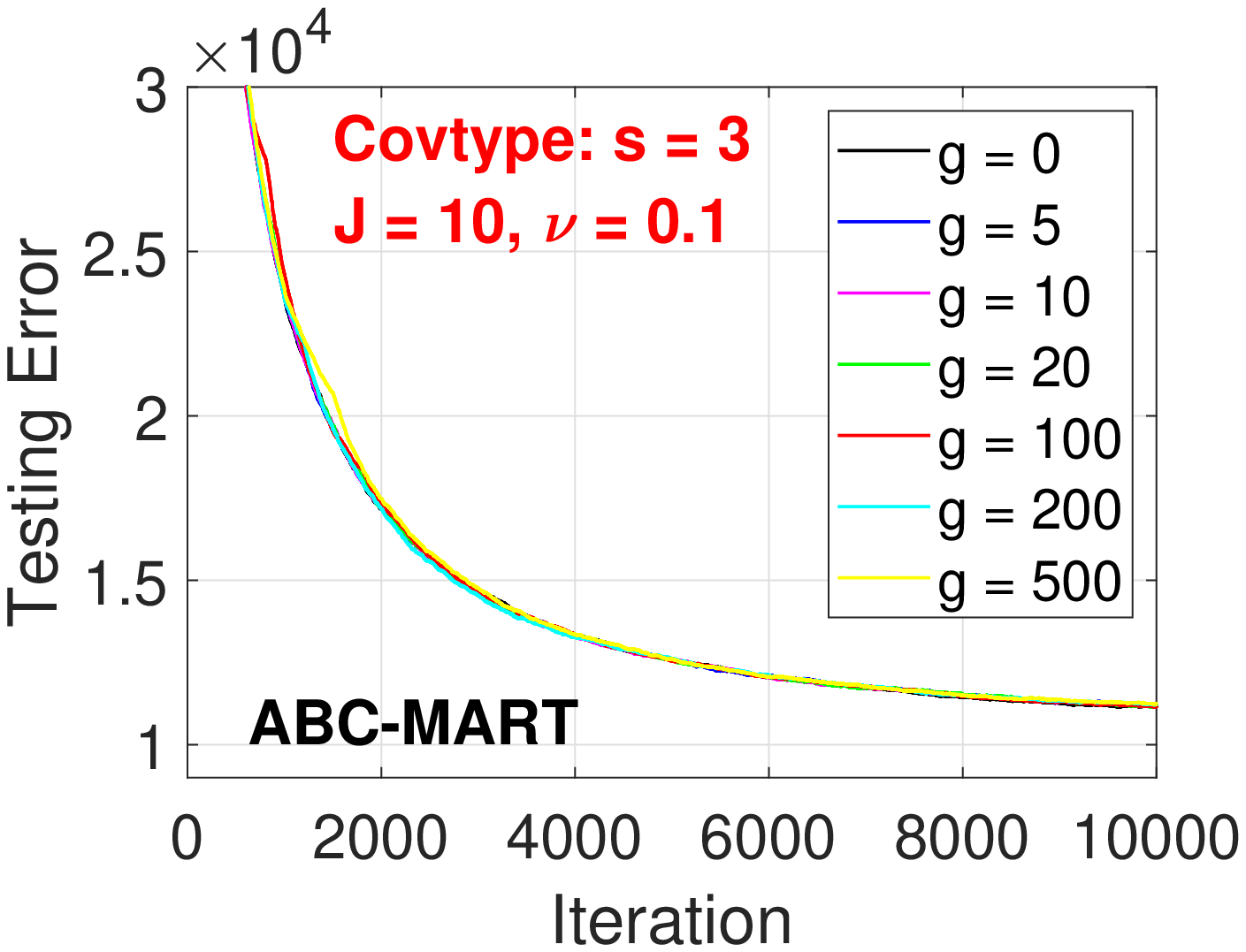}
    \includegraphics[width=2.2in]{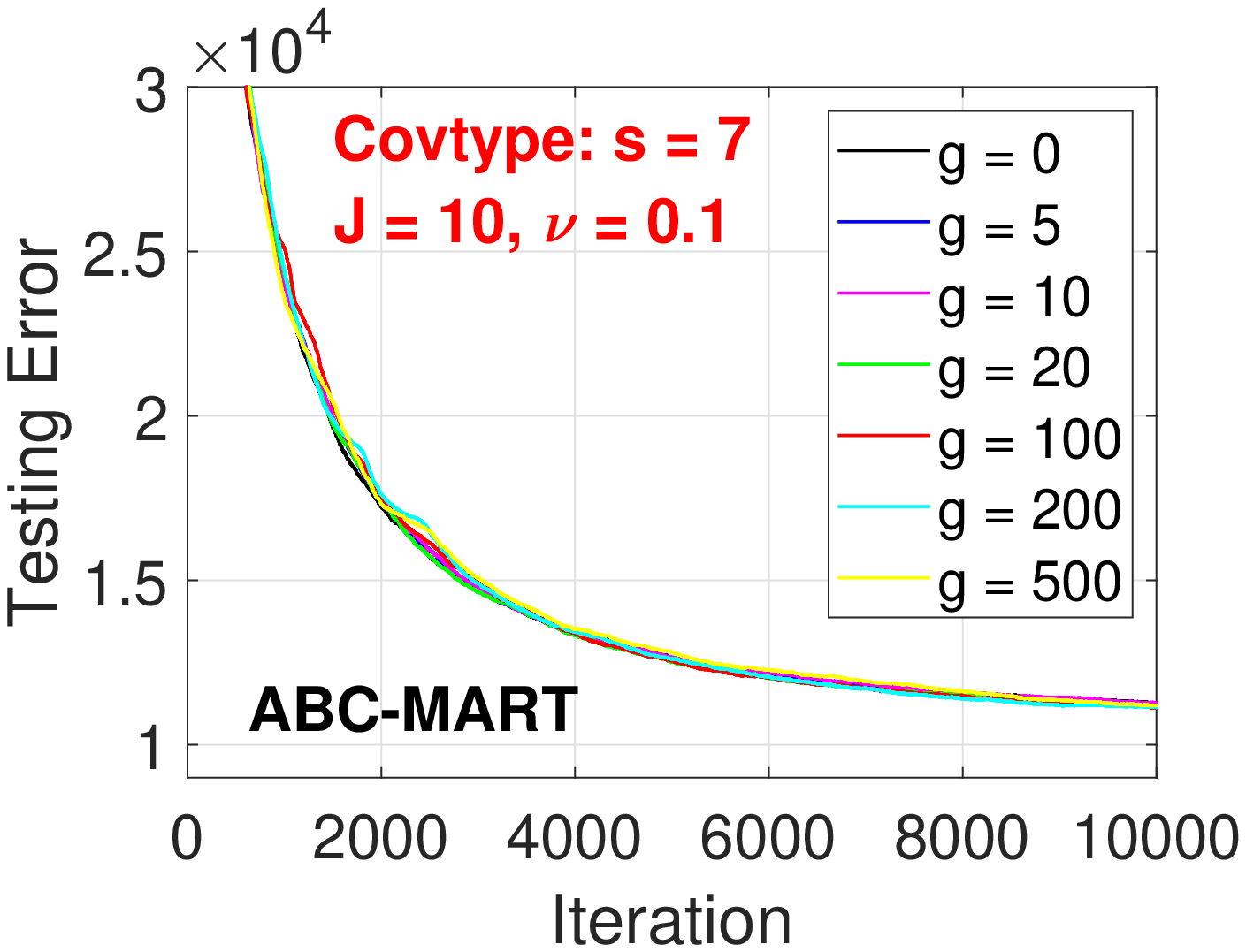}
}

\mbox{
    \includegraphics[width=2.2in]{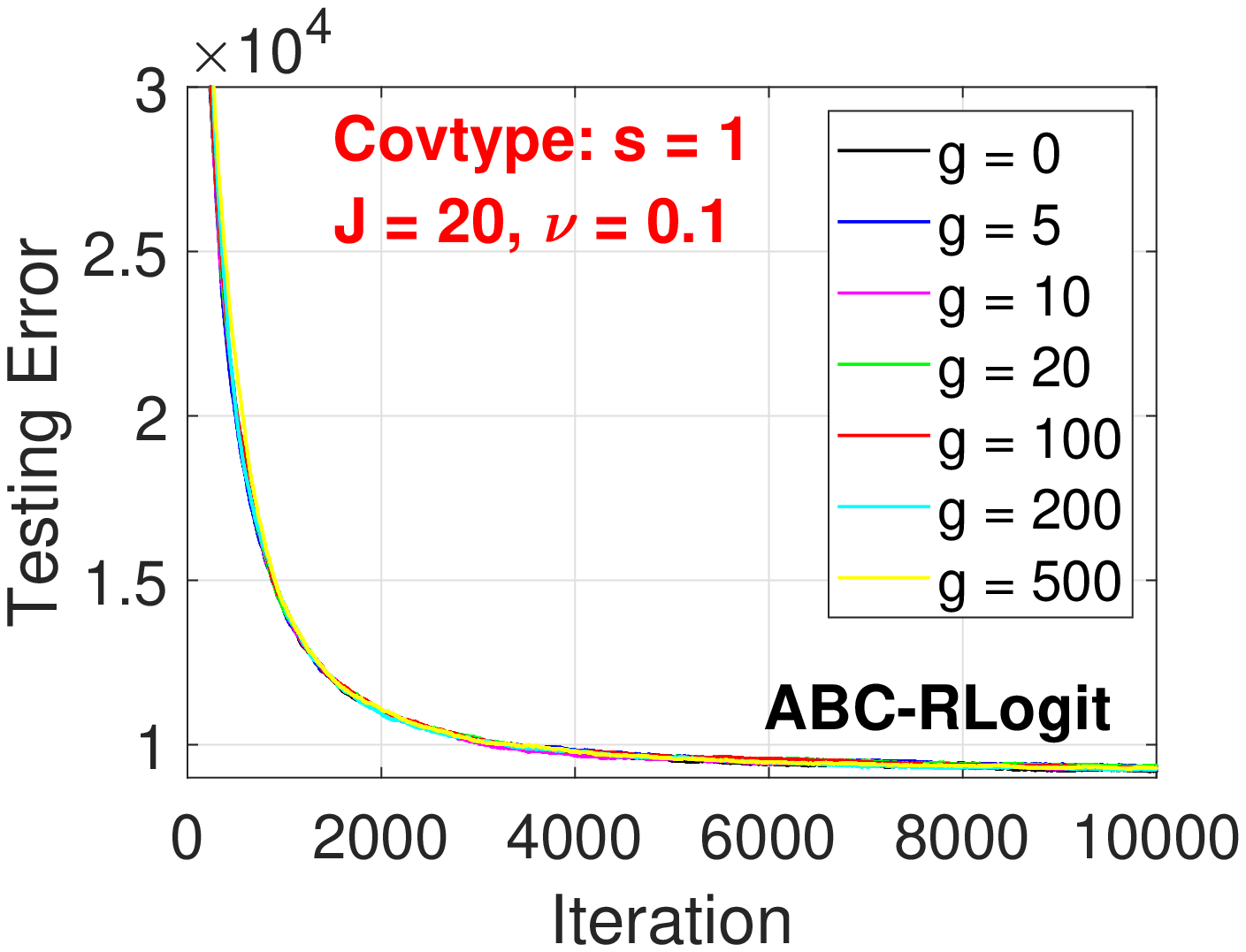}
    \includegraphics[width=2.2in]{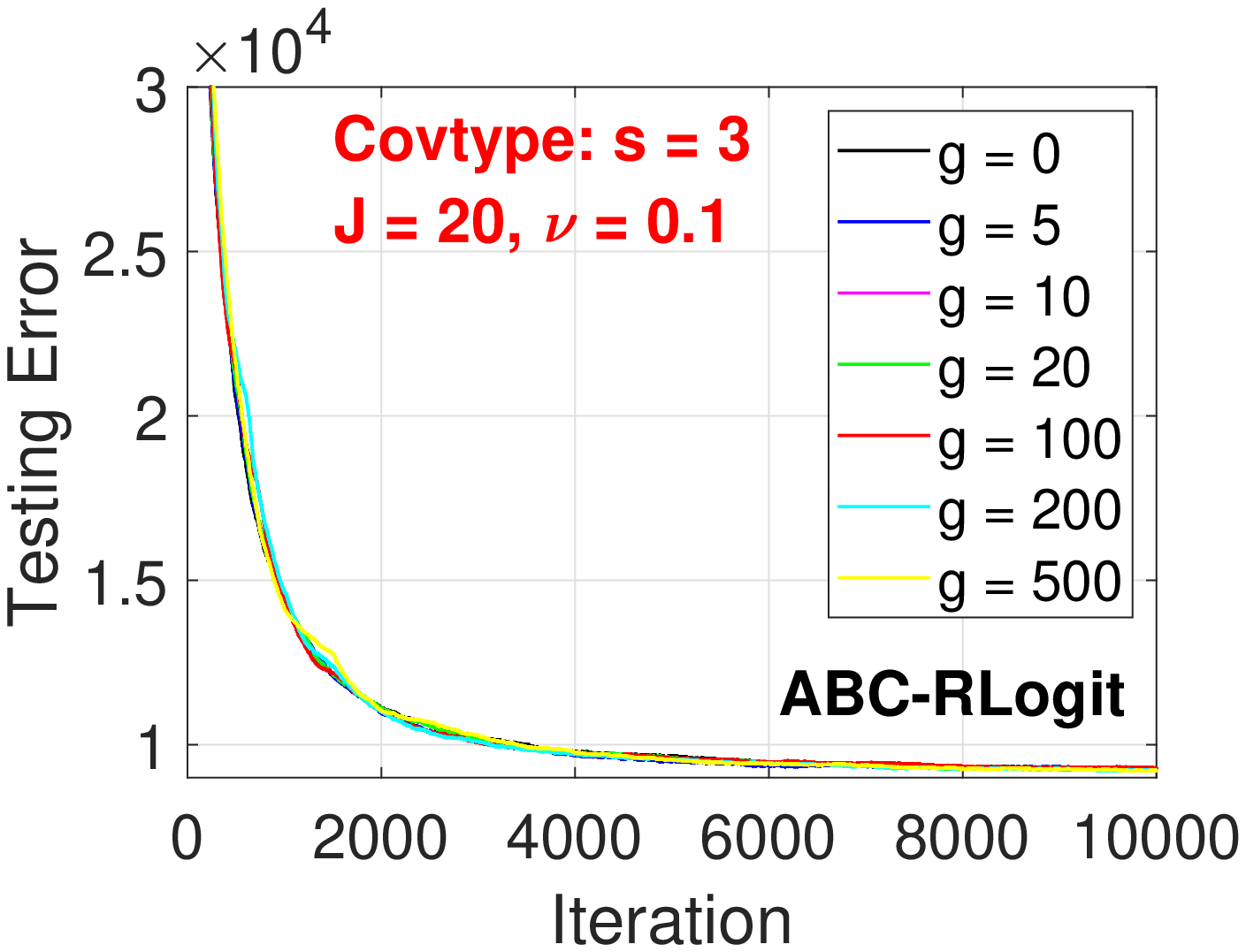}
    \includegraphics[width=2.2in]{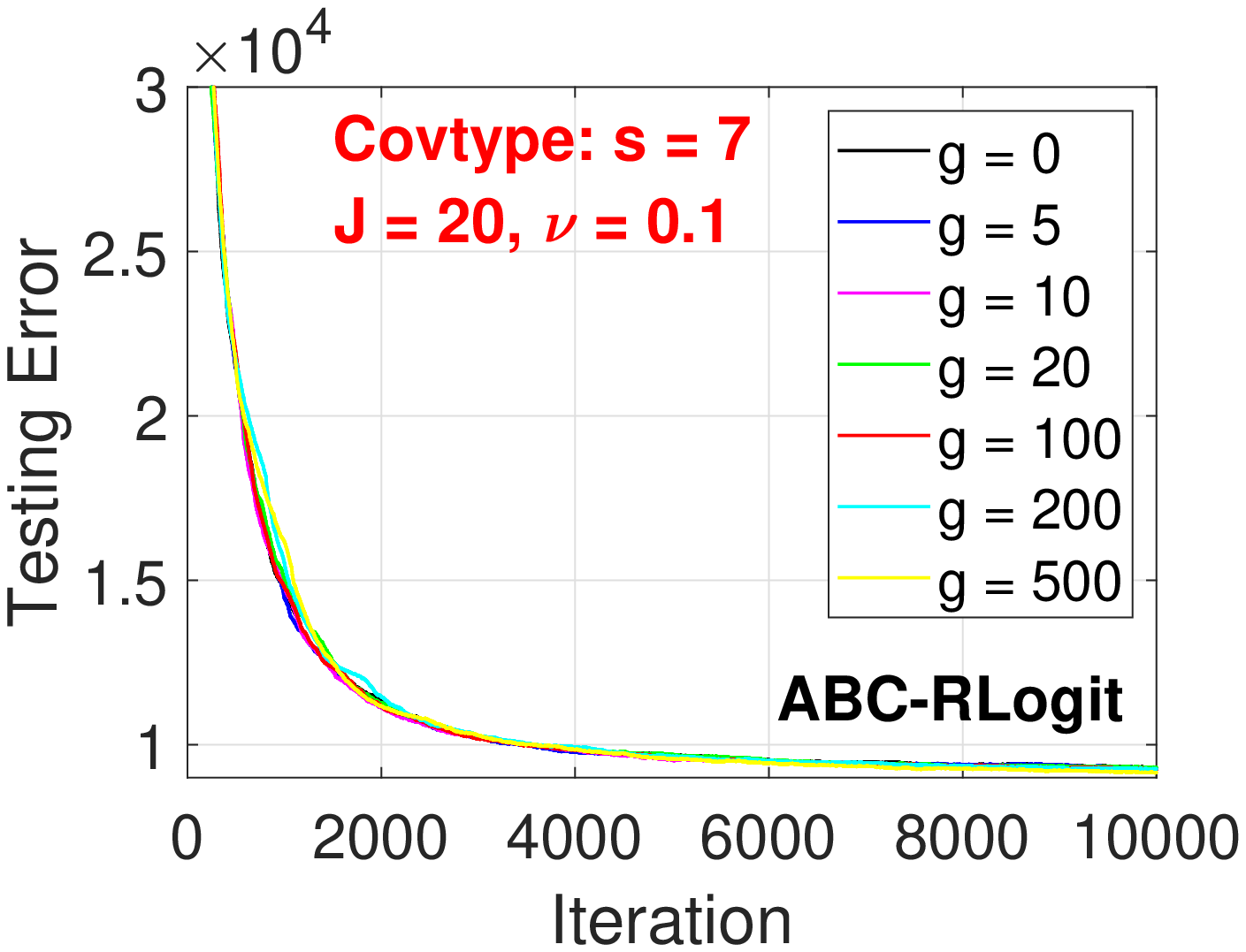}
}

\mbox{
    \includegraphics[width=2.2in]{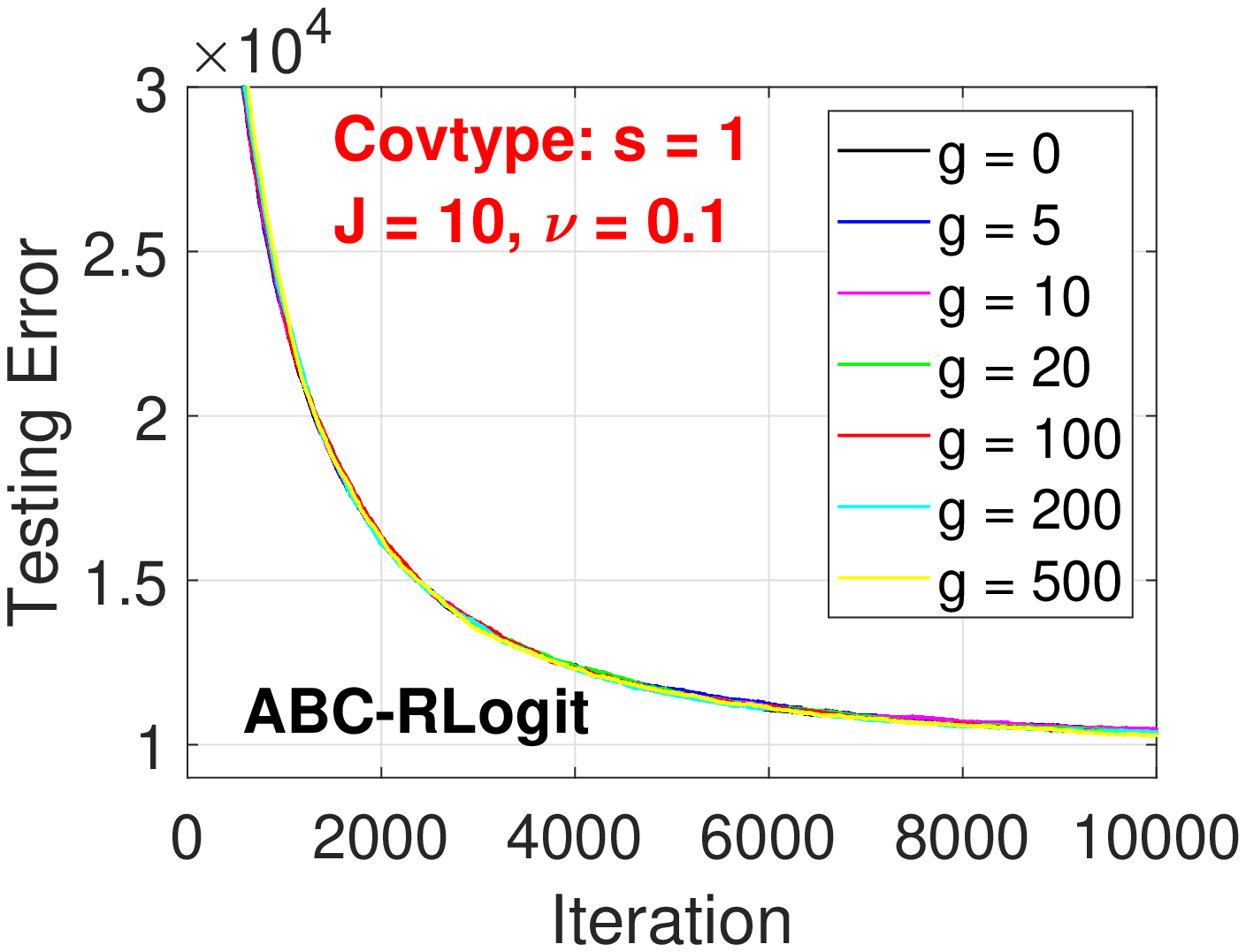}
    \includegraphics[width=2.2in]{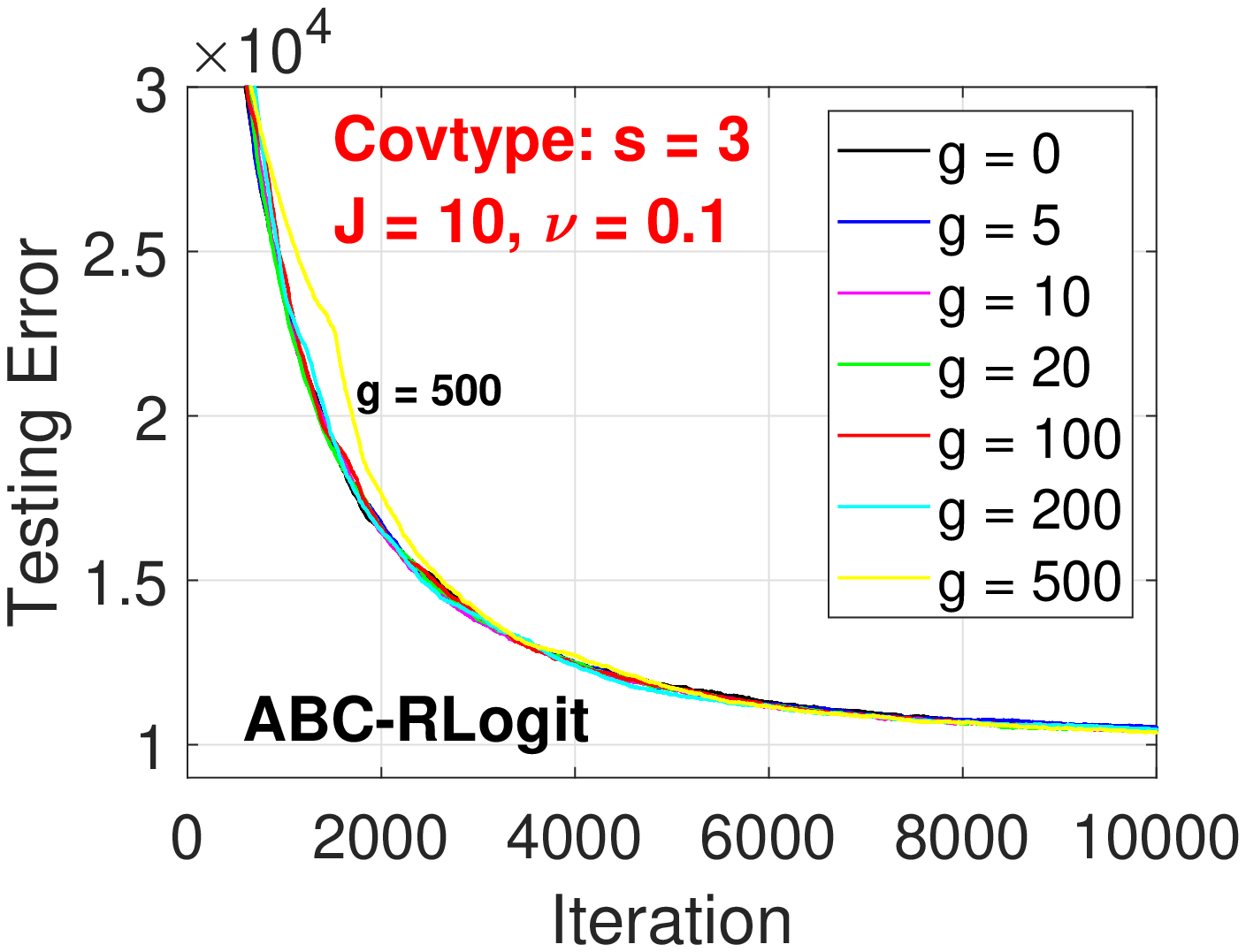}
    \includegraphics[width=2.2in]{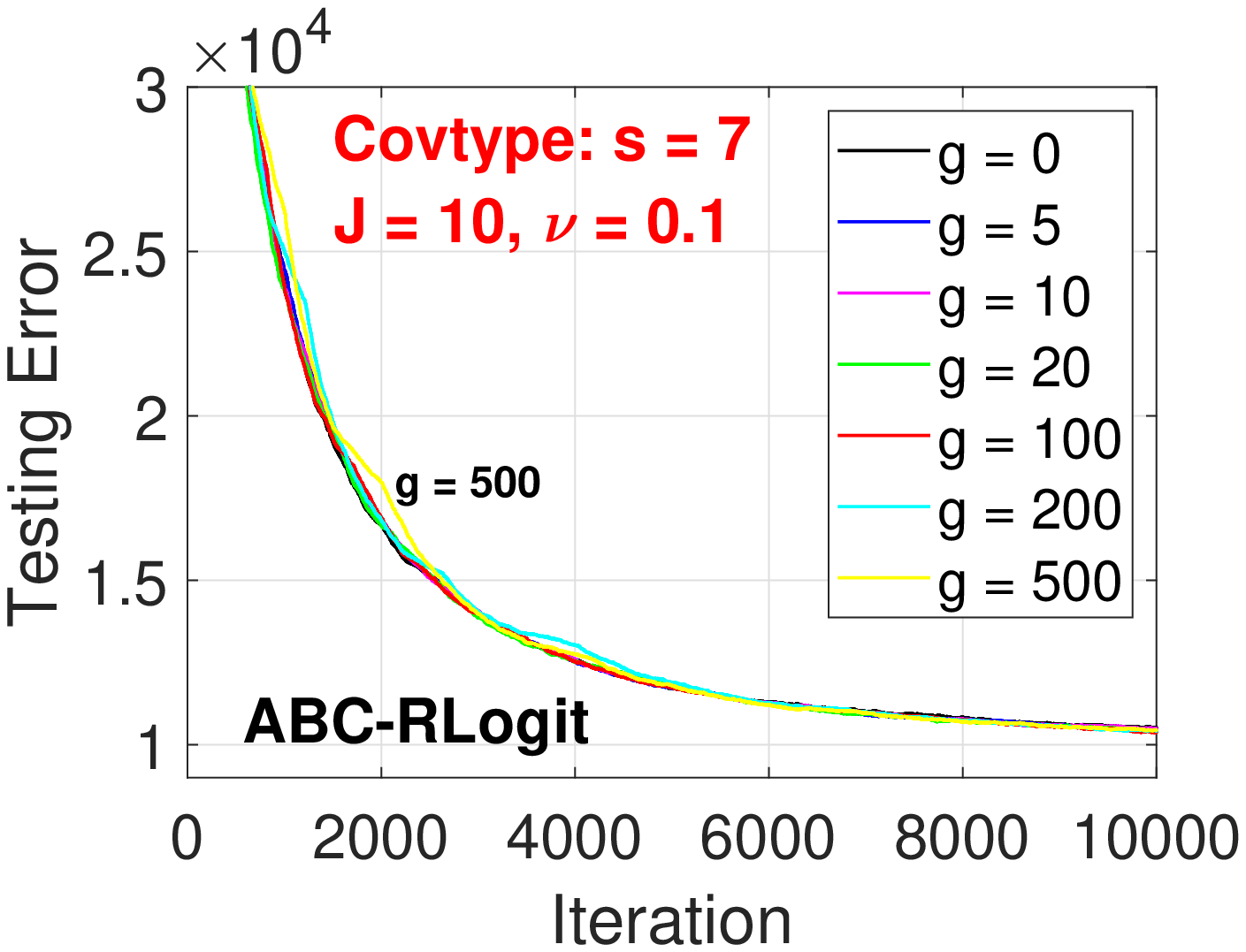}
}

\end{center}

\vspace{-0.1in}

\caption{{\em Covertype} dataset. Test classification errors of ABC-MART and ABC-RobustLogitBoost with parameters $s$ and $g$. We present results for  $g\in\{0,5,10,20,100,200,500\}$ and three selected $\in\{1,3,7\}$.   }\label{fig:Covertype_sg}
\end{figure}

\begin{figure}[h]
\begin{center}
\mbox{
    \includegraphics[width=2.2in]{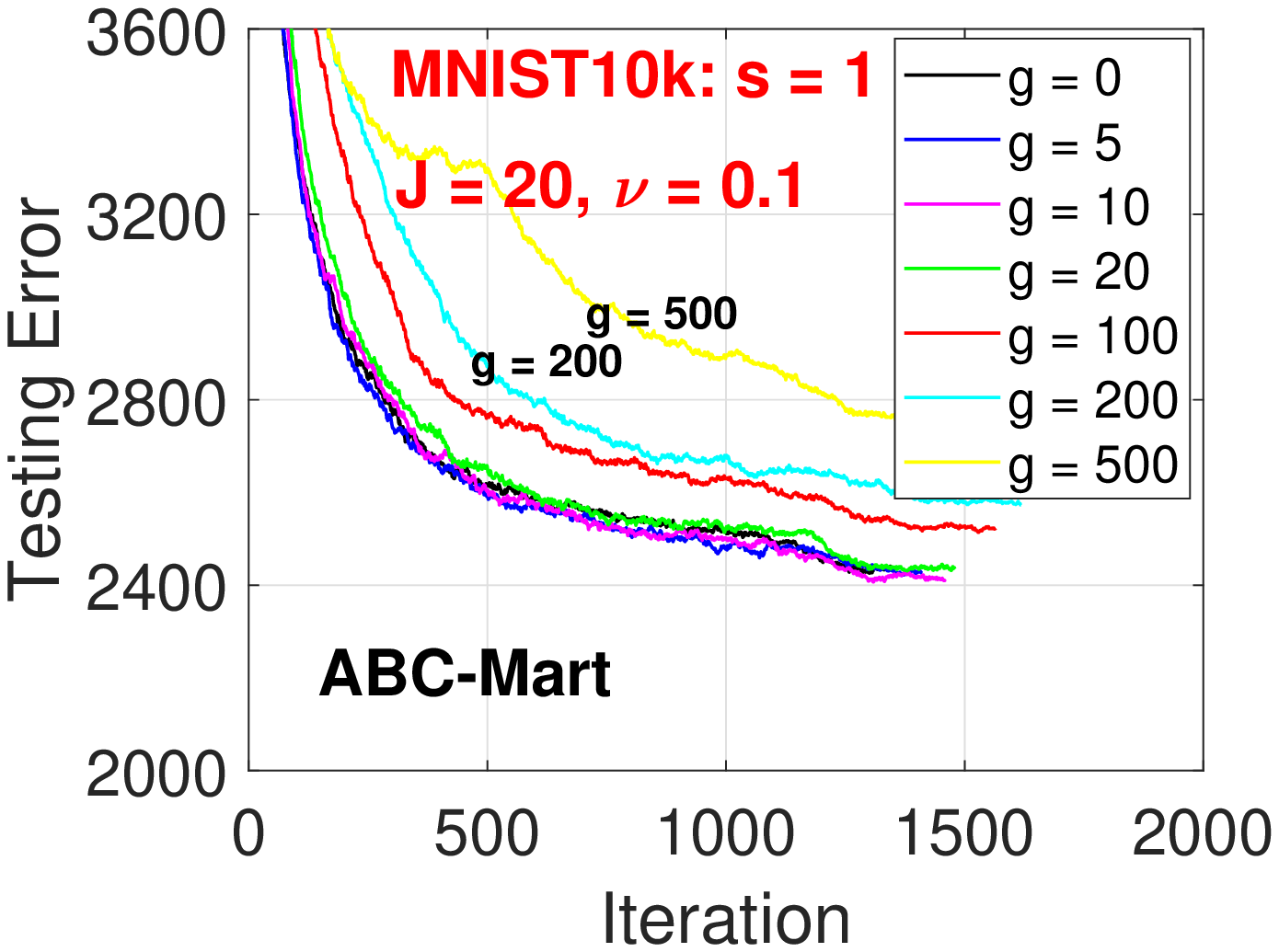}
    \includegraphics[width=2.2in]{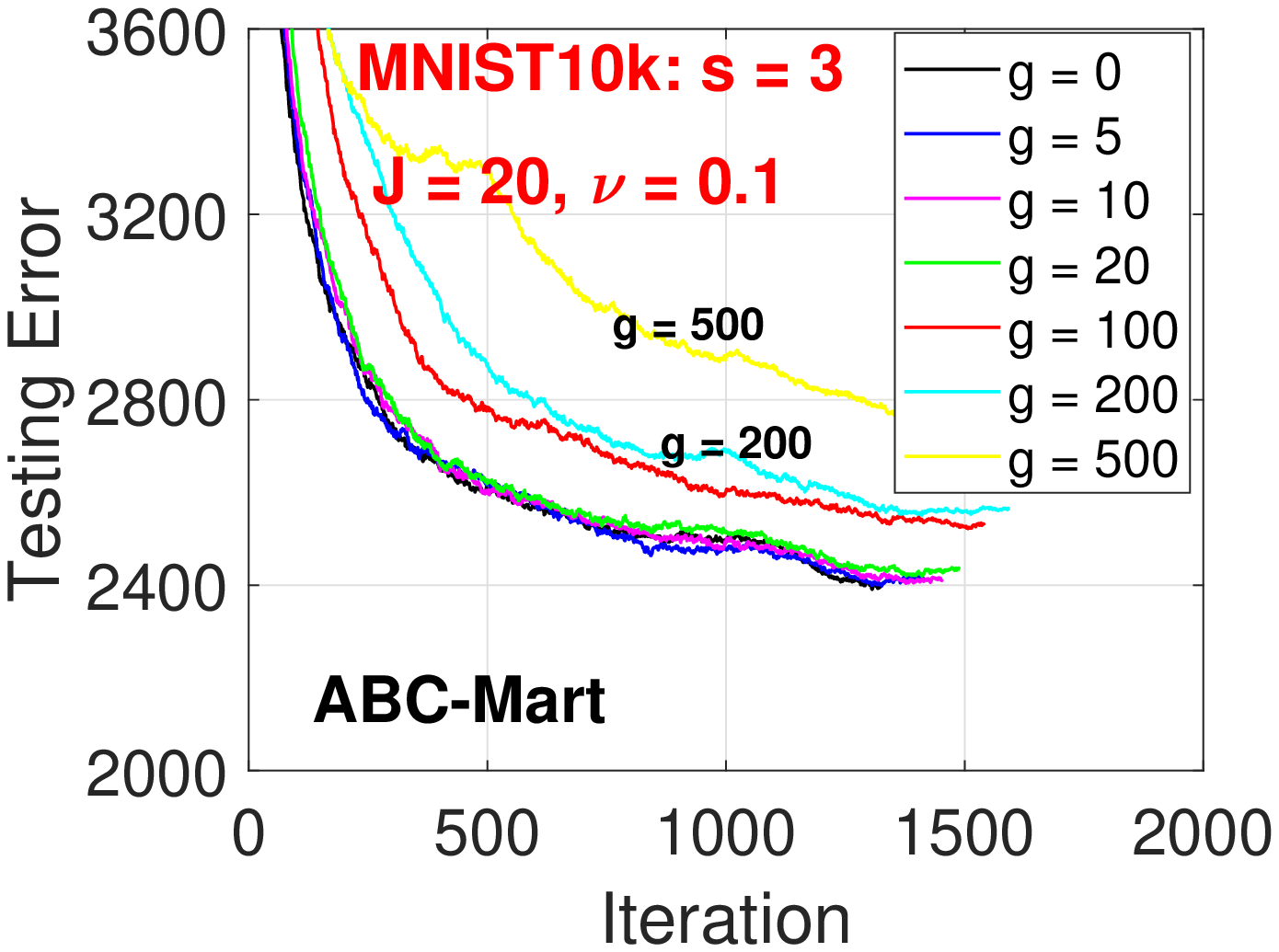}
    \includegraphics[width=2.2in]{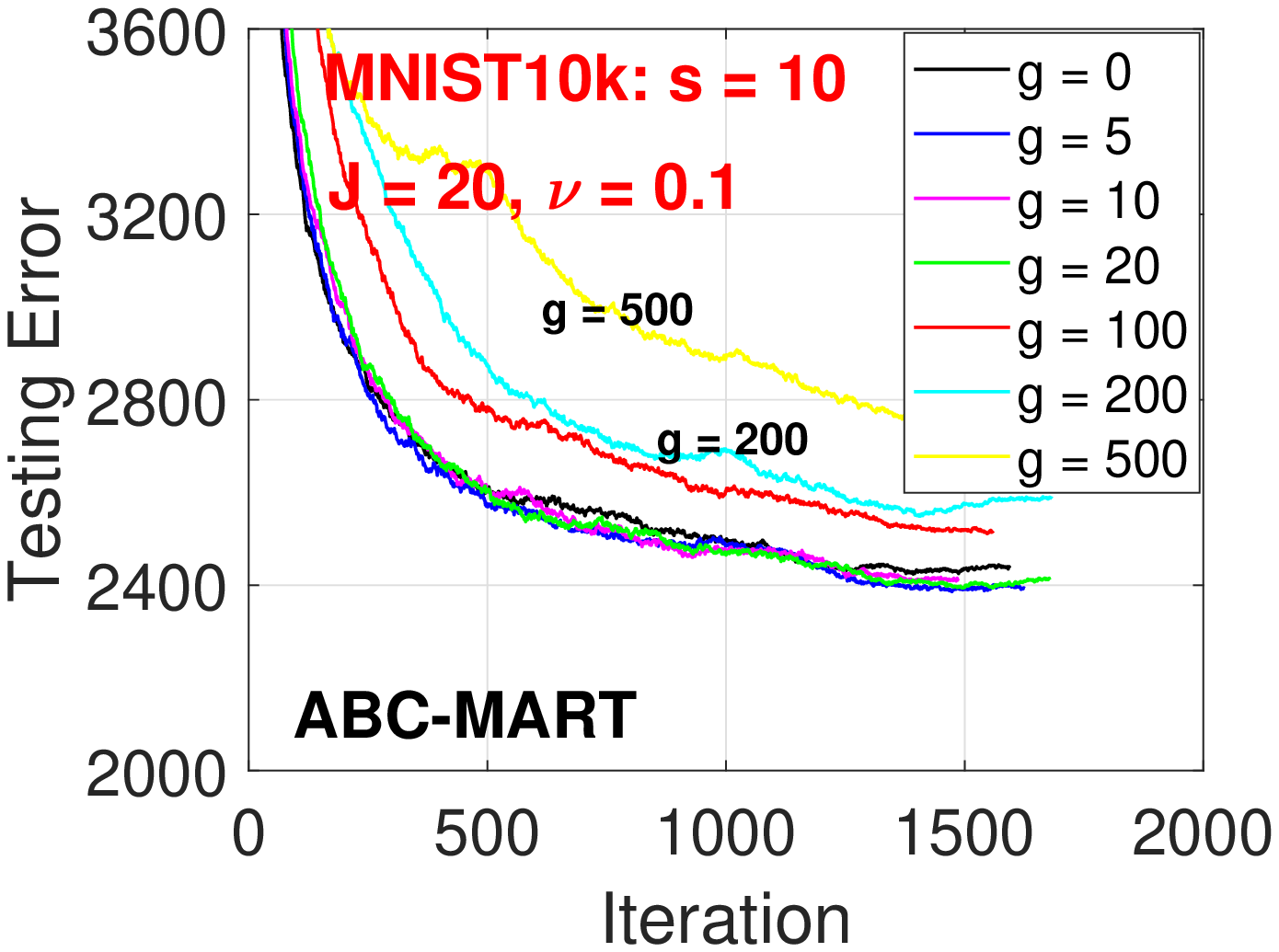}
}

\mbox{
    \includegraphics[width=2.2in]{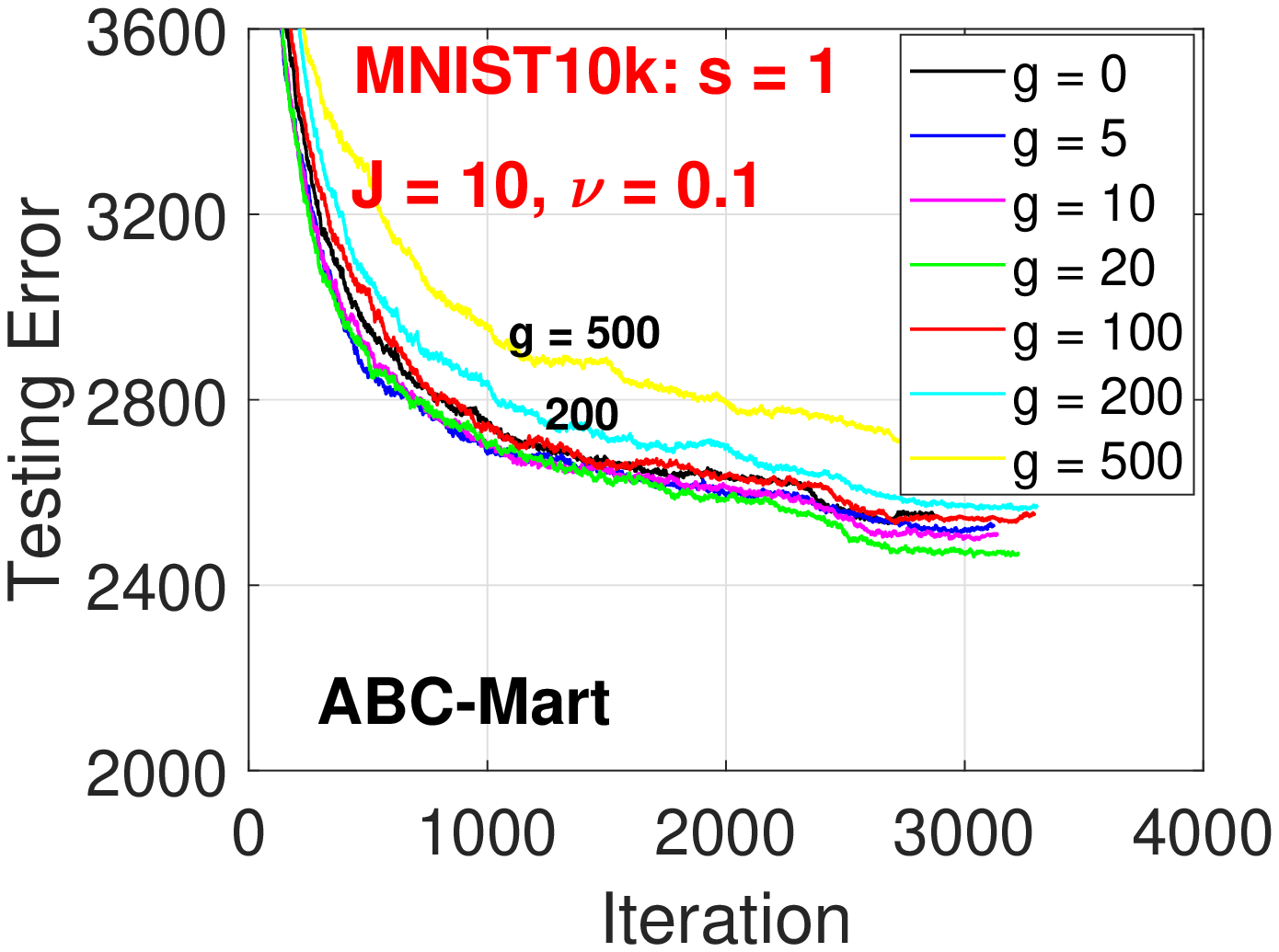}
    \includegraphics[width=2.2in]{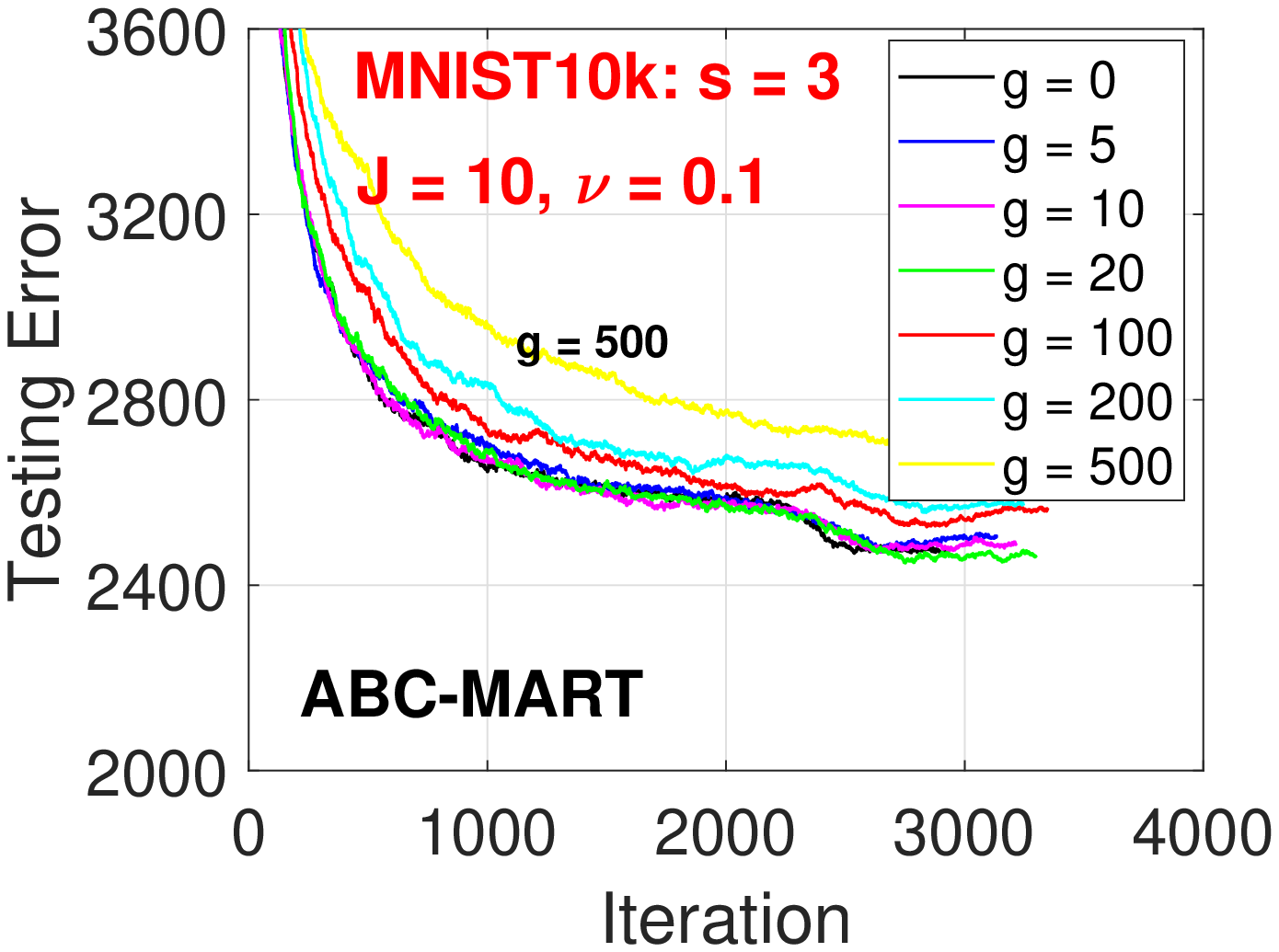}
    \includegraphics[width=2.2in]{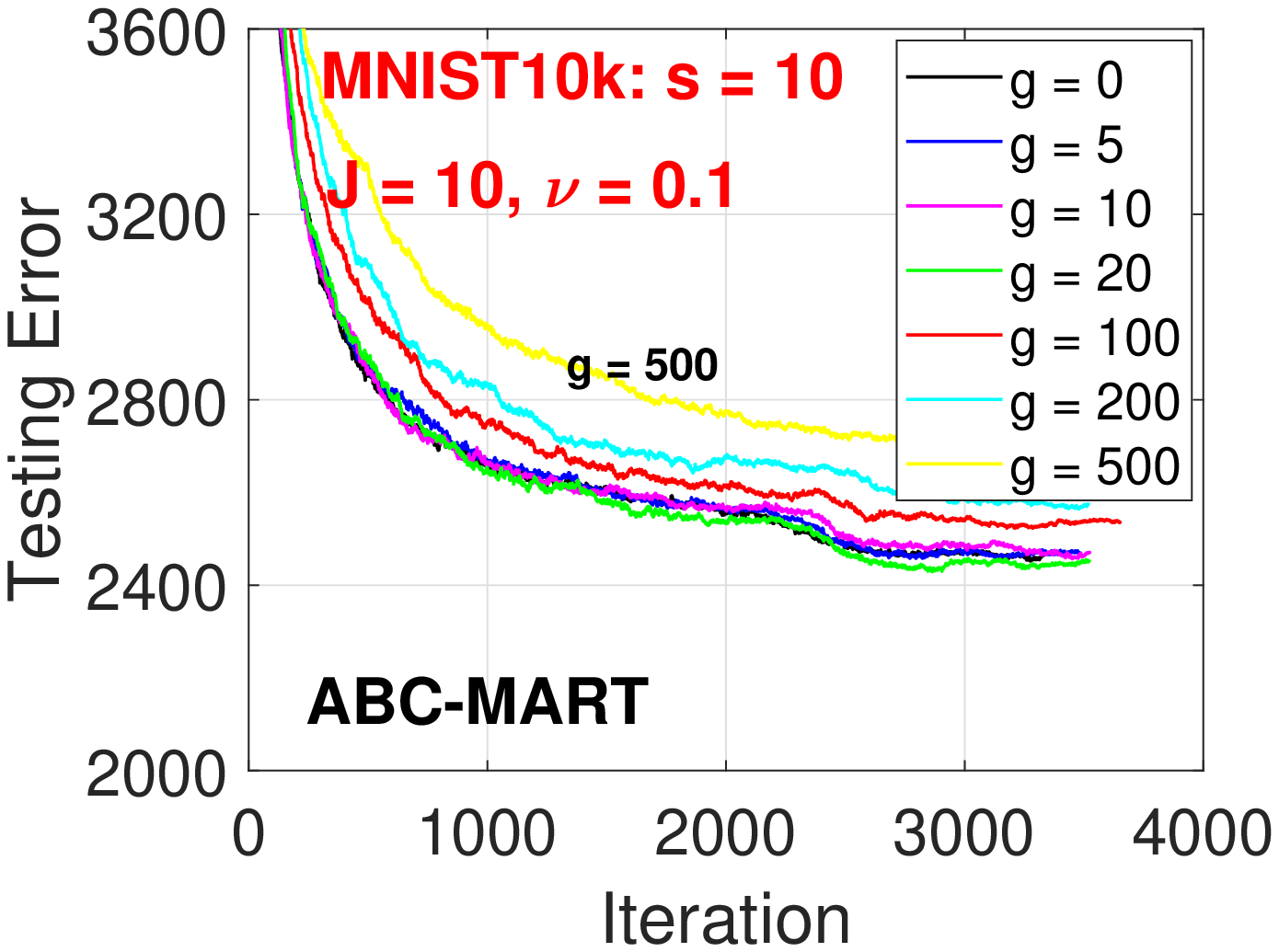}
}

\mbox{
    \includegraphics[width=2.2in]{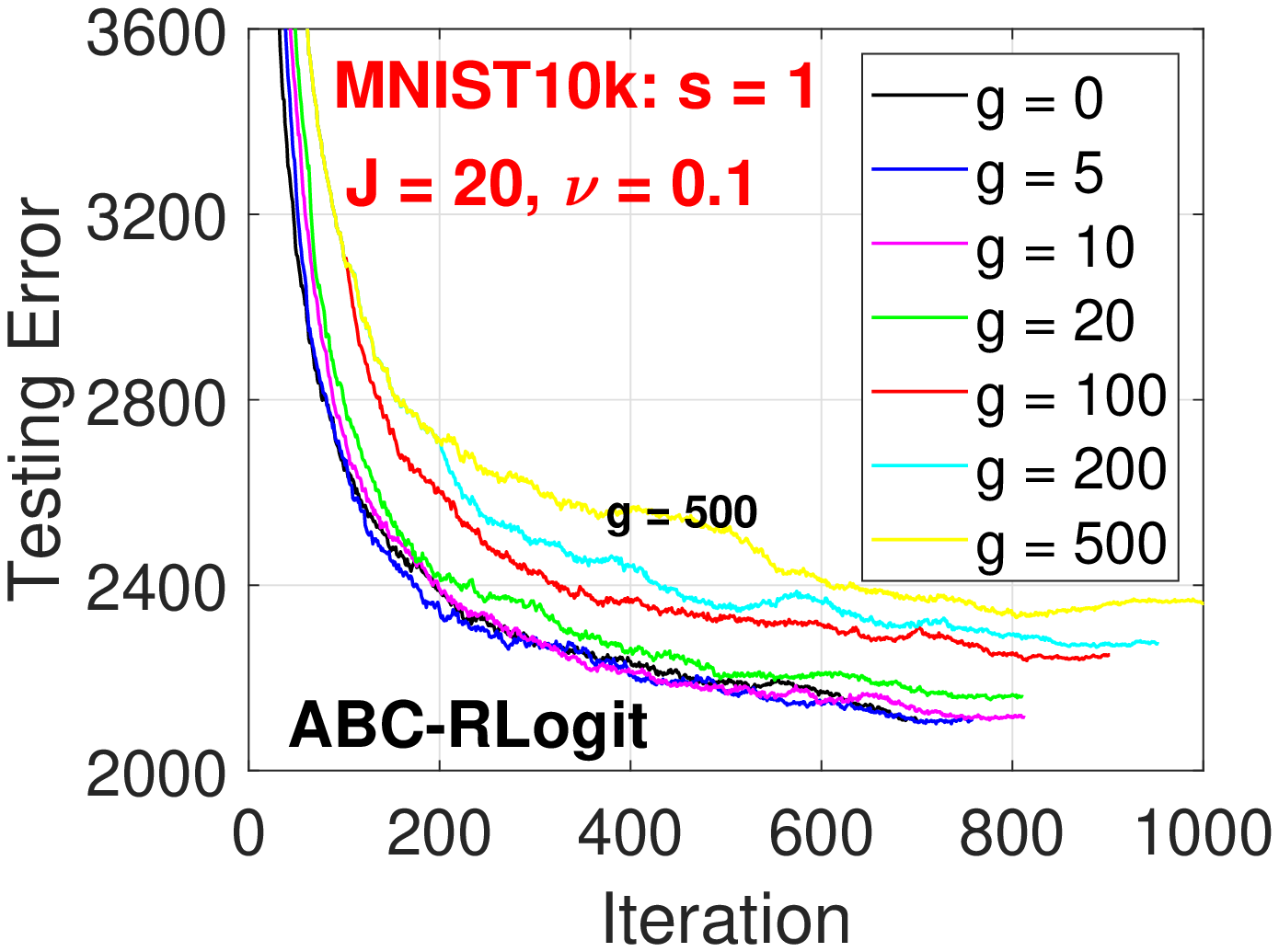}
    \includegraphics[width=2.2in]{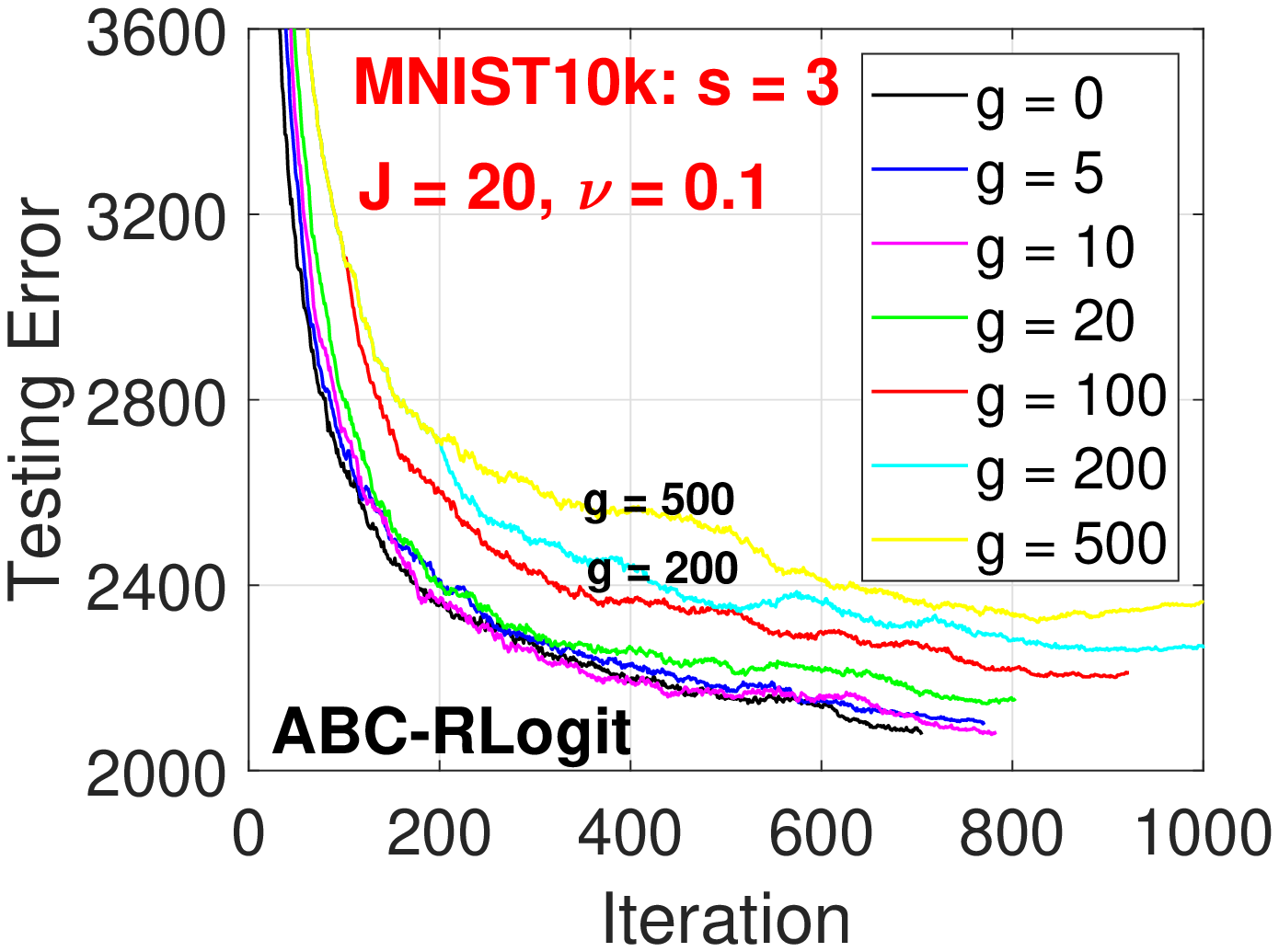}
    \includegraphics[width=2.2in]{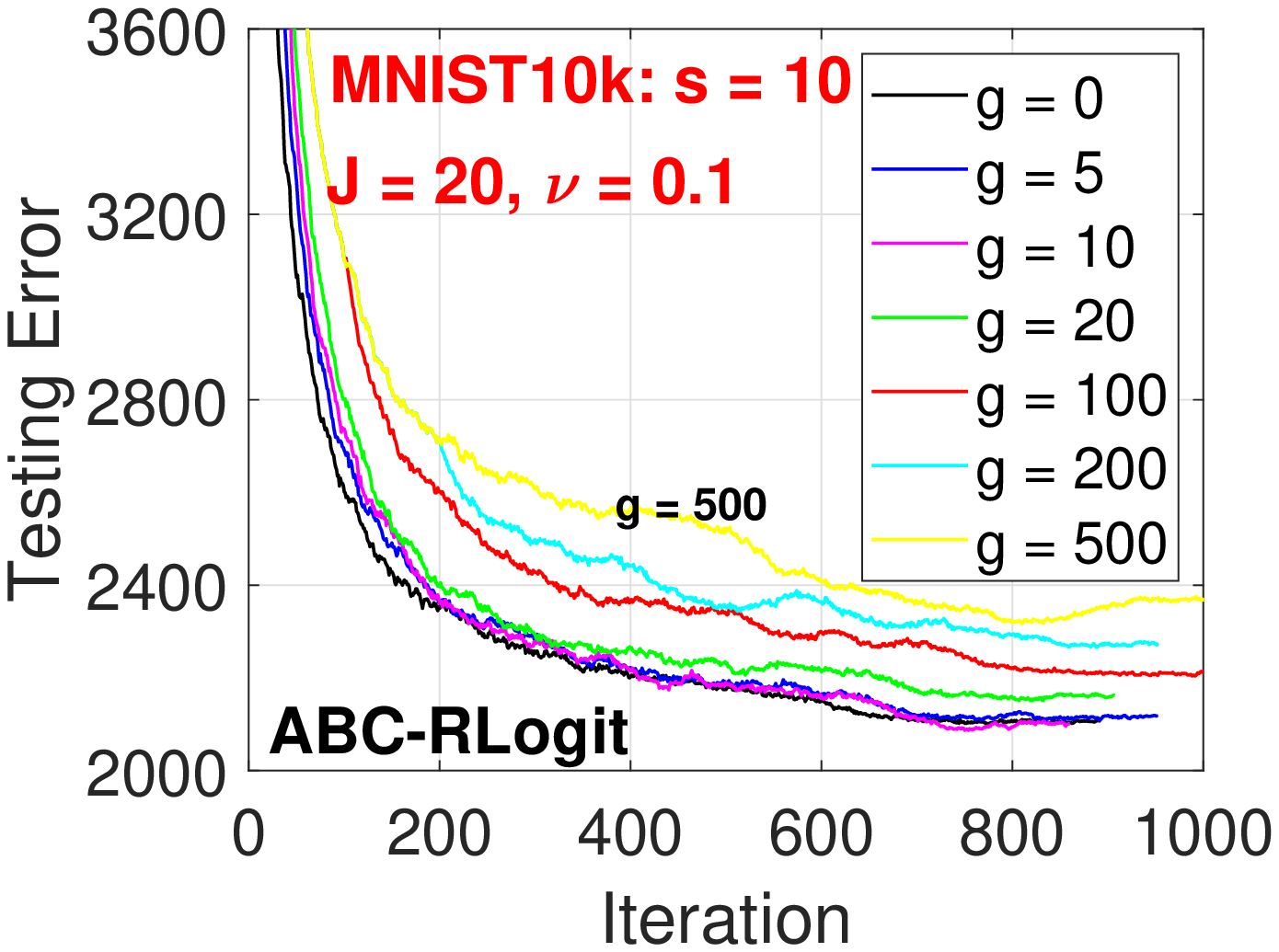}
}

\mbox{
    \includegraphics[width=2.2in]{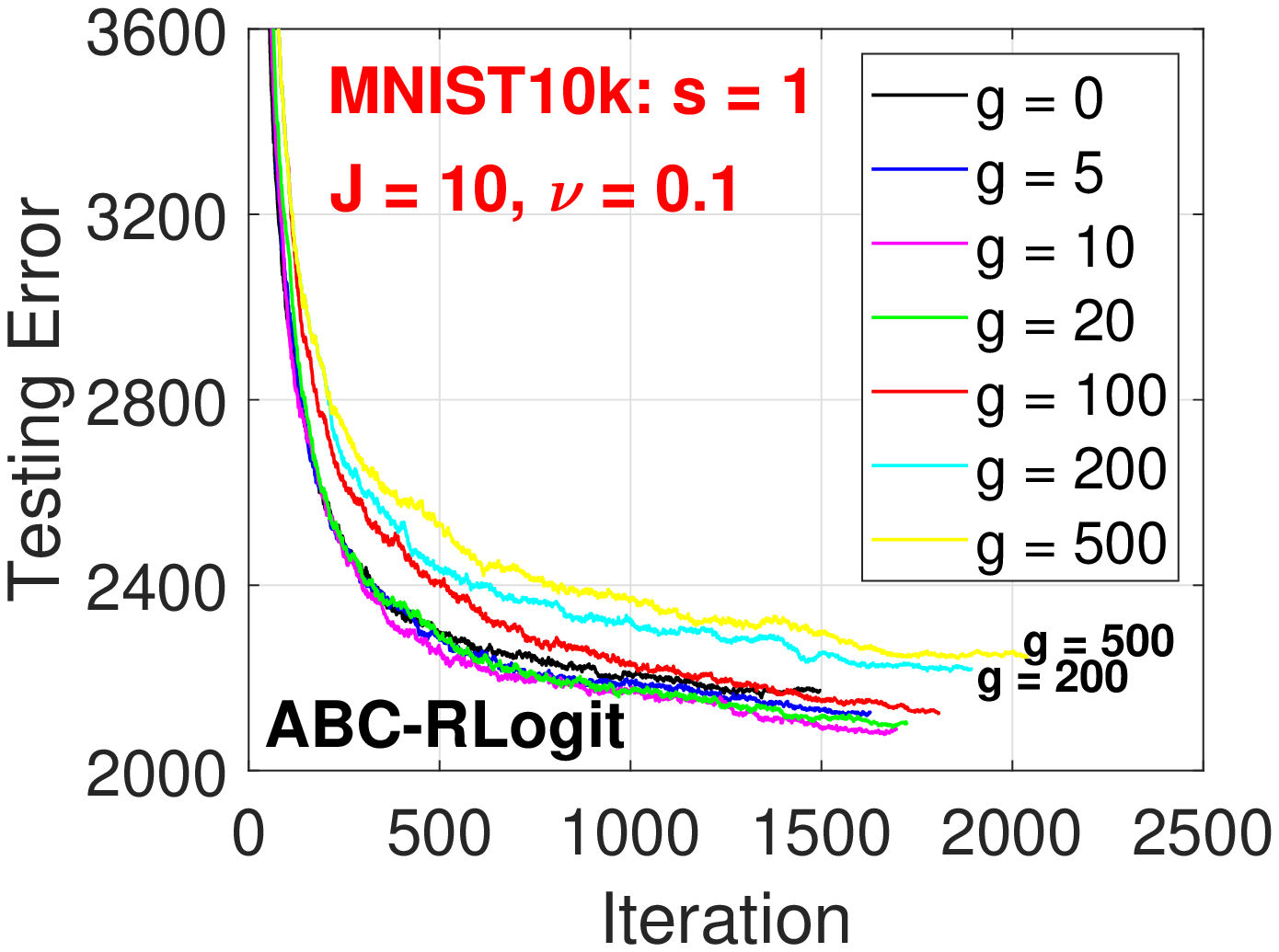}
    \includegraphics[width=2.2in]{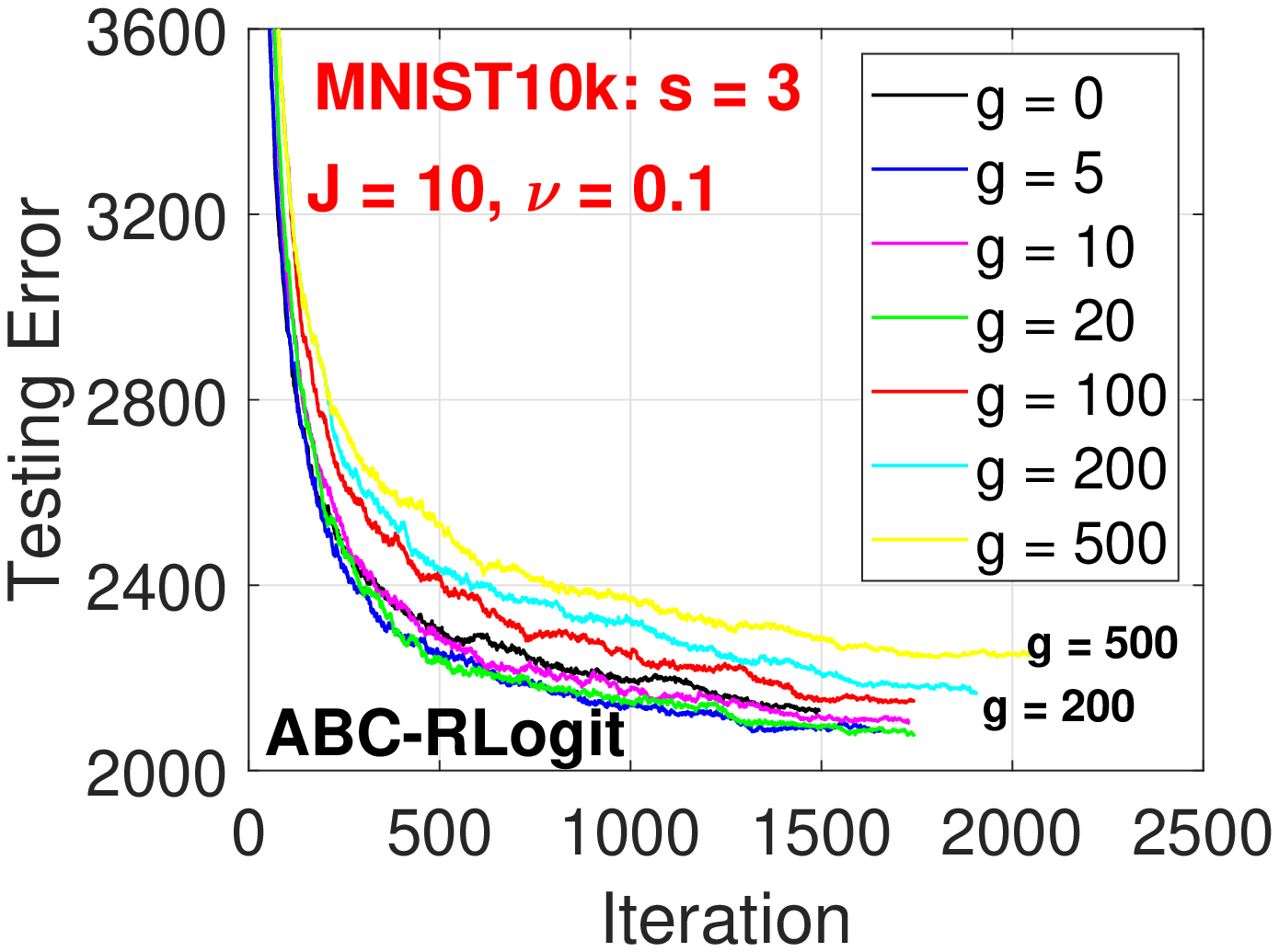}
    \includegraphics[width=2.2in]{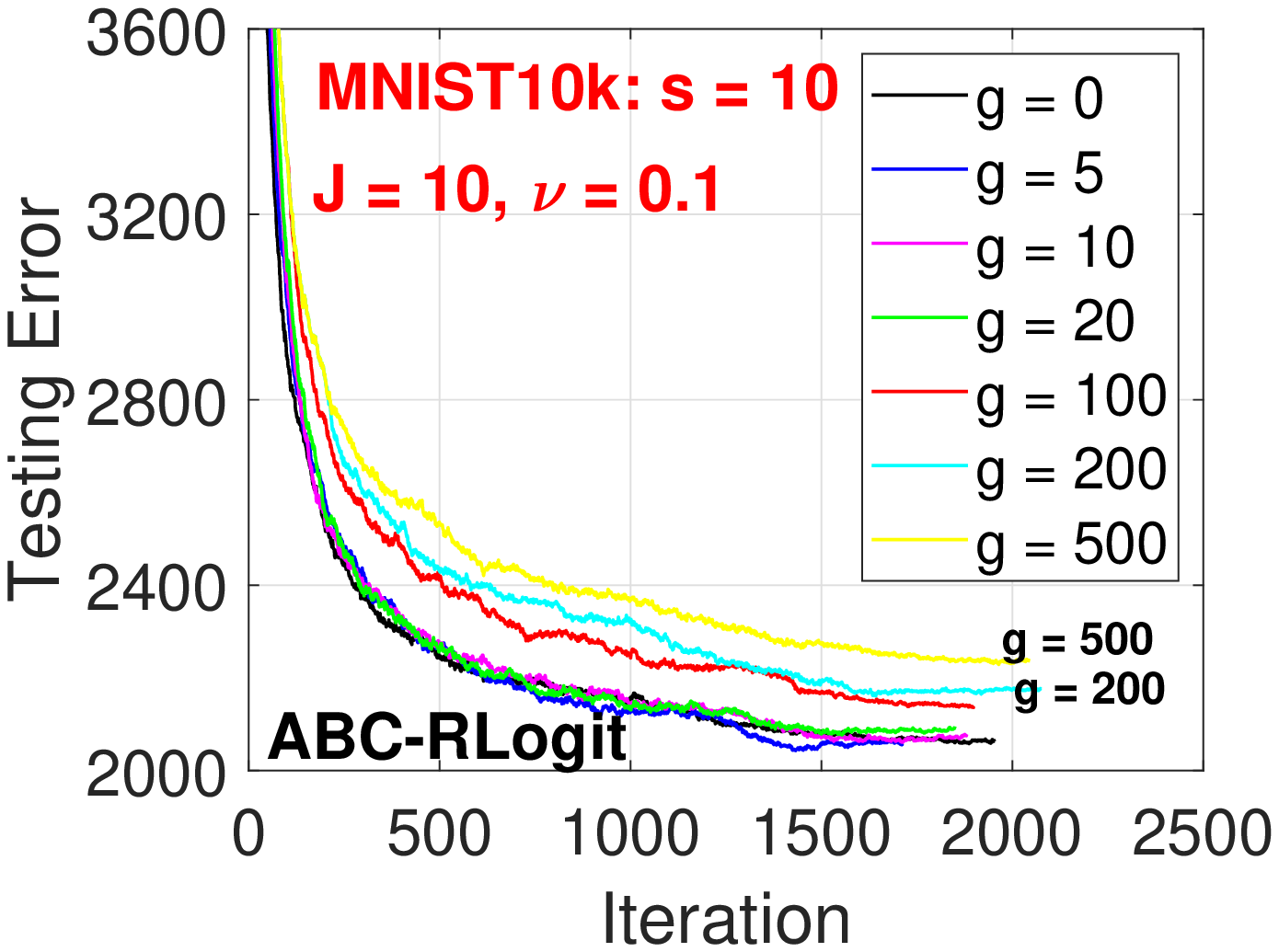}
}

\end{center}

\vspace{-0.1in}

\caption{{\em Mnist10k} dataset. Test classification errors of ABC-MART and ABC-RobustLogitBoost with parameters $s$ and $g$. We present results for  $g\in\{0,5,10,20,100,200,500\}$ and three selected $\in\{1,3,10\}$.
}\label{fig:Mnist10k_sg}
\end{figure}

\begin{figure}[h]
\begin{center}
\mbox{
    \includegraphics[width=2.2in]{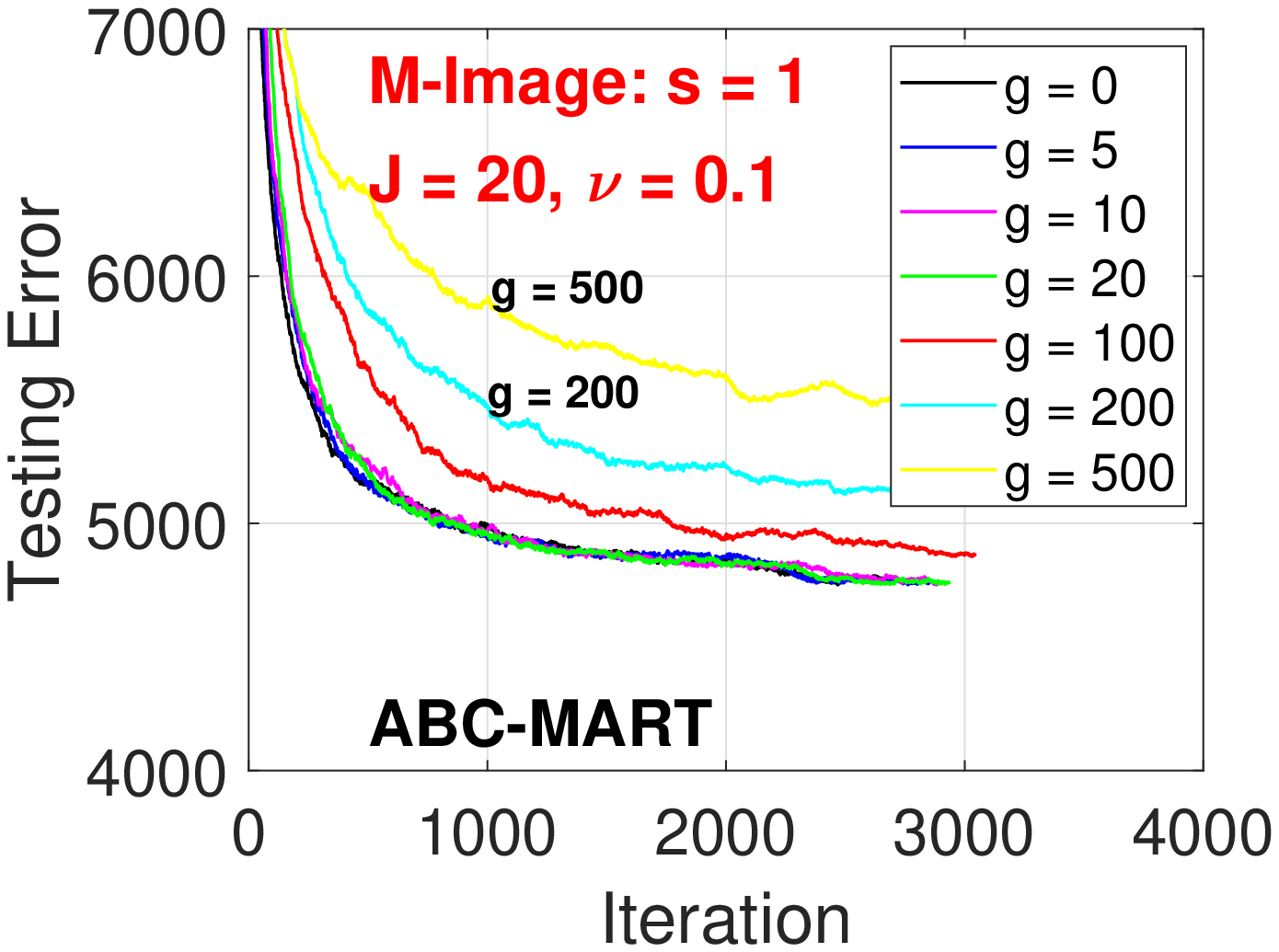}
    \includegraphics[width=2.2in]{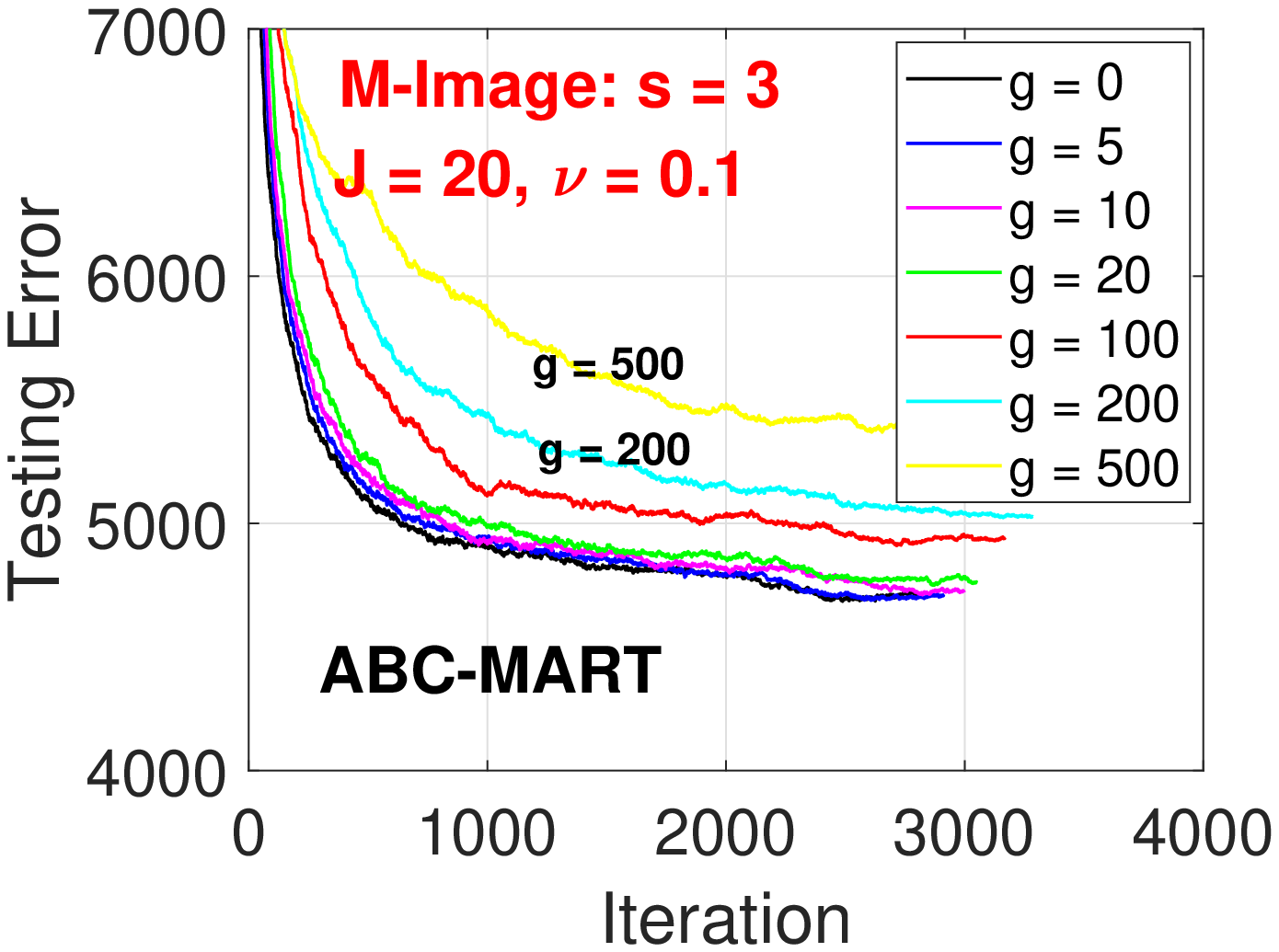}
    \includegraphics[width=2.2in]{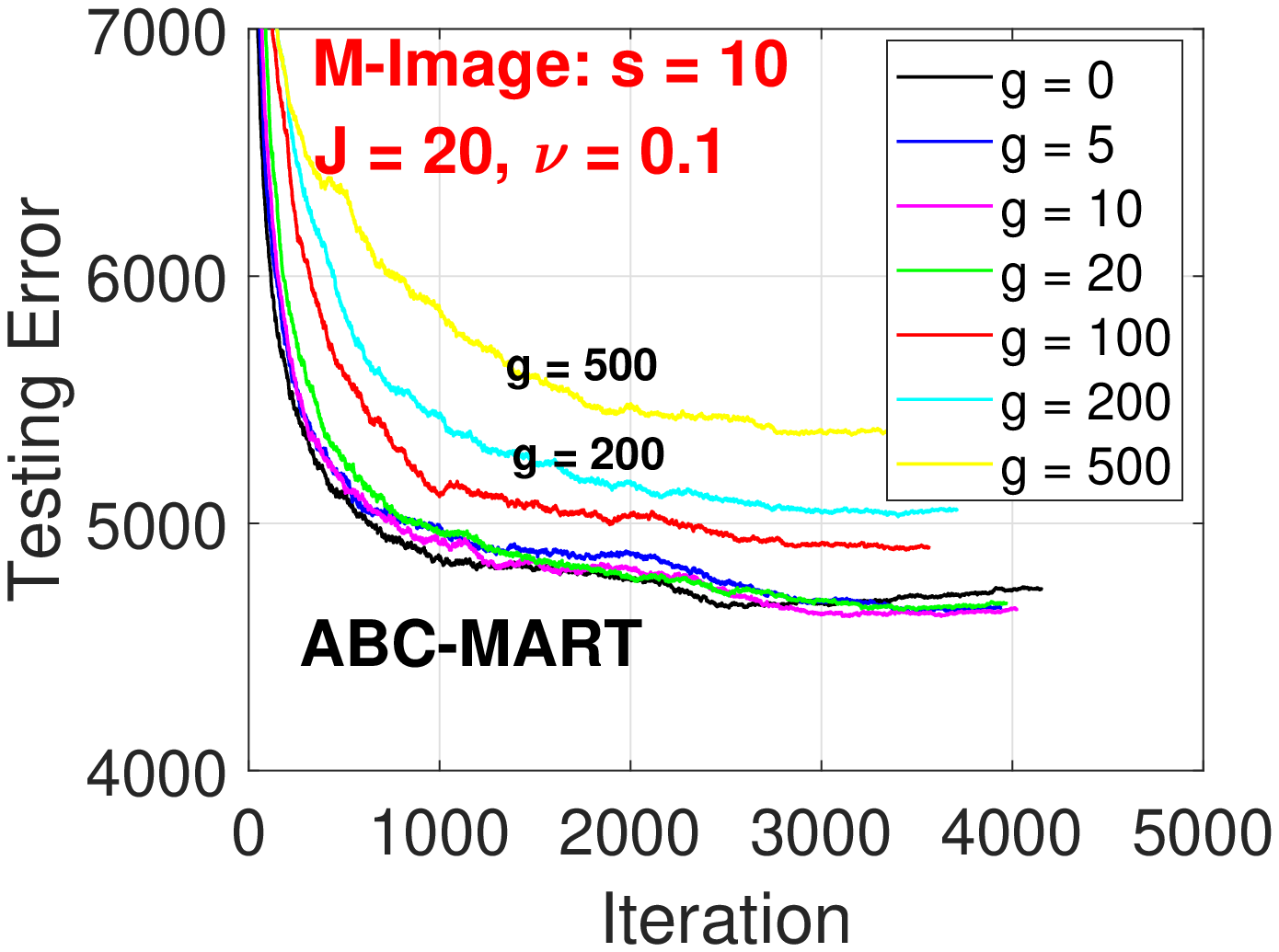}
}

\mbox{
    \includegraphics[width=2.2in]{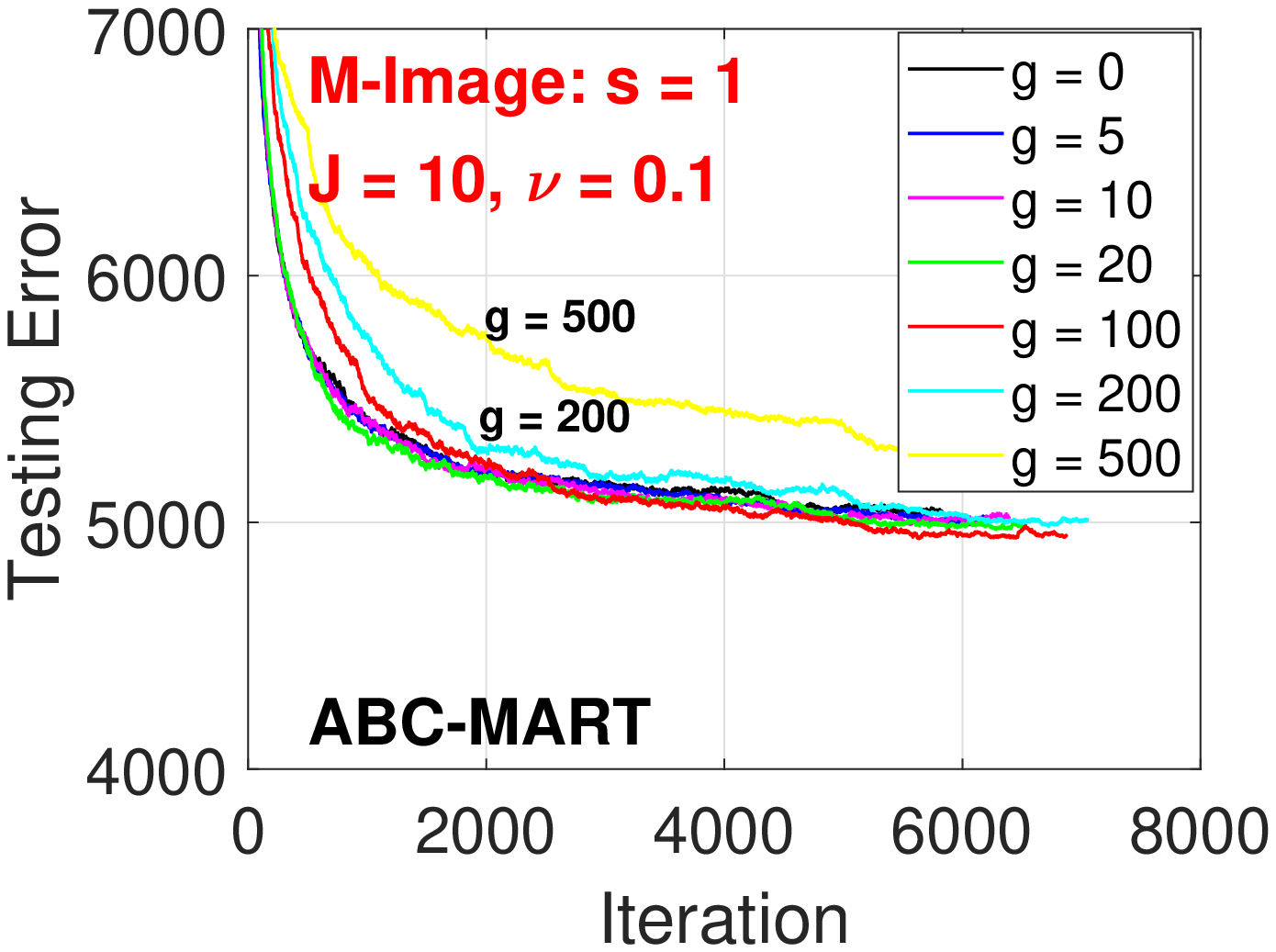}
    \includegraphics[width=2.2in]{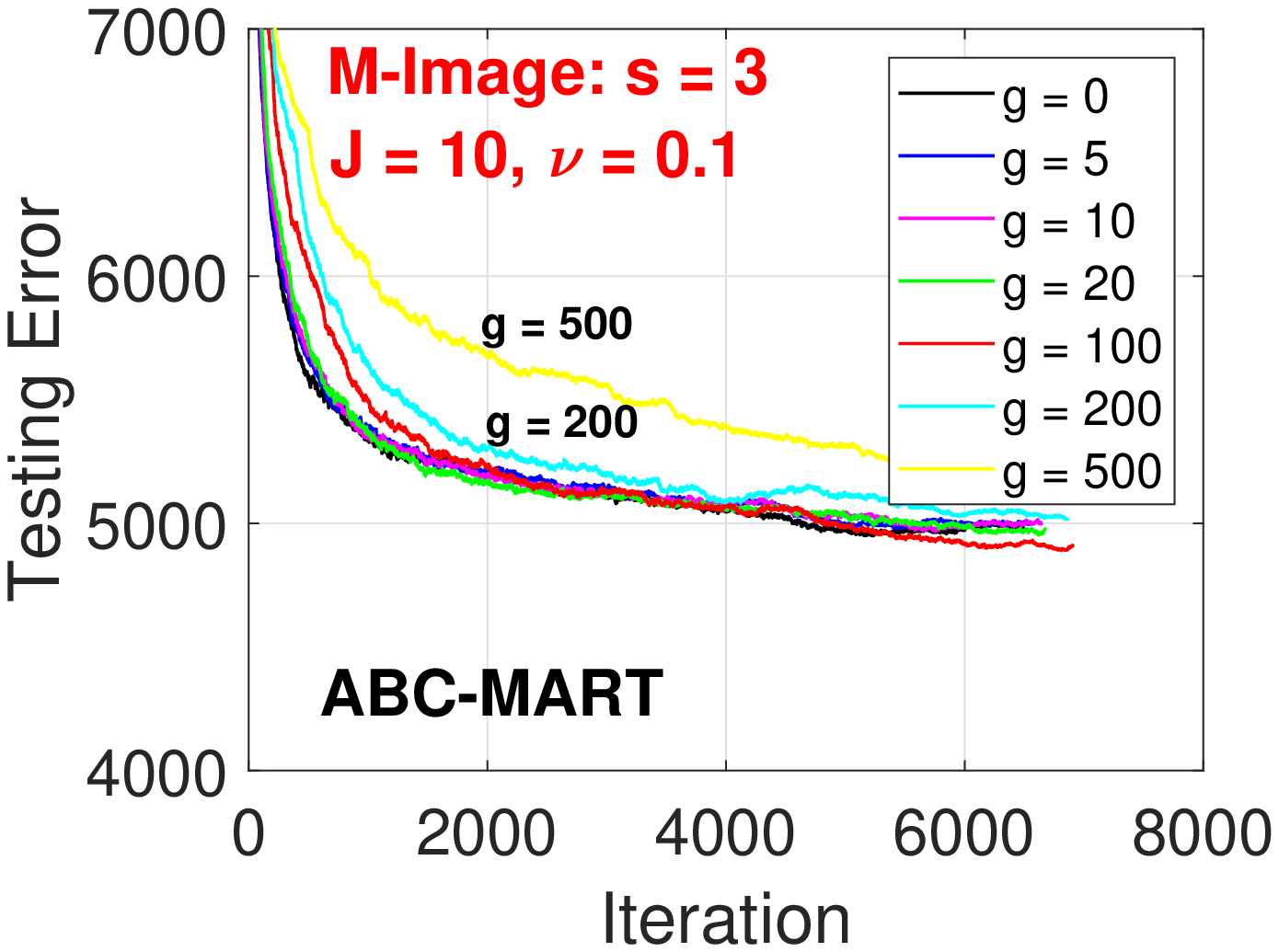}
    \includegraphics[width=2.2in]{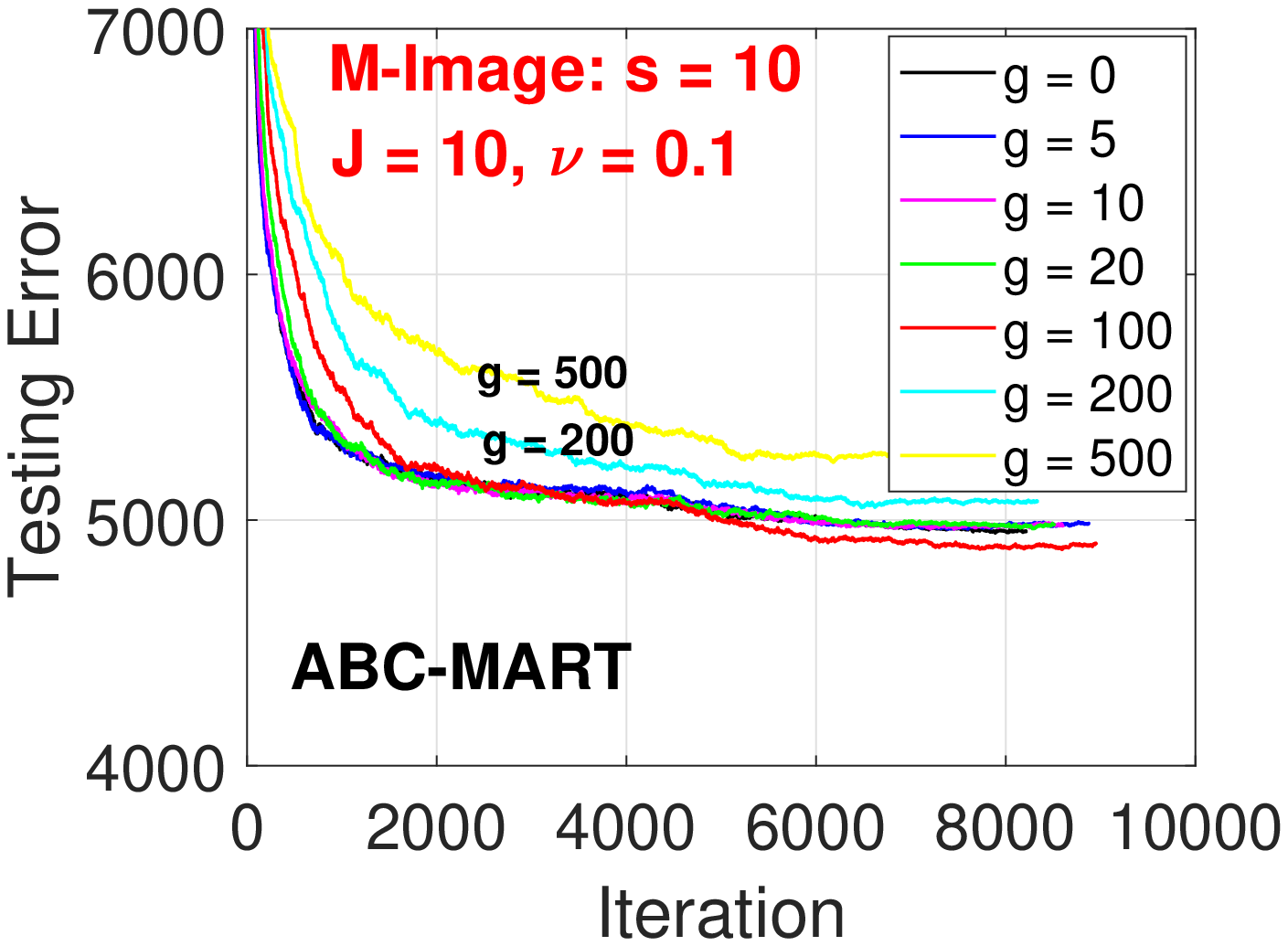}
}

\mbox{
    \includegraphics[width=2.2in]{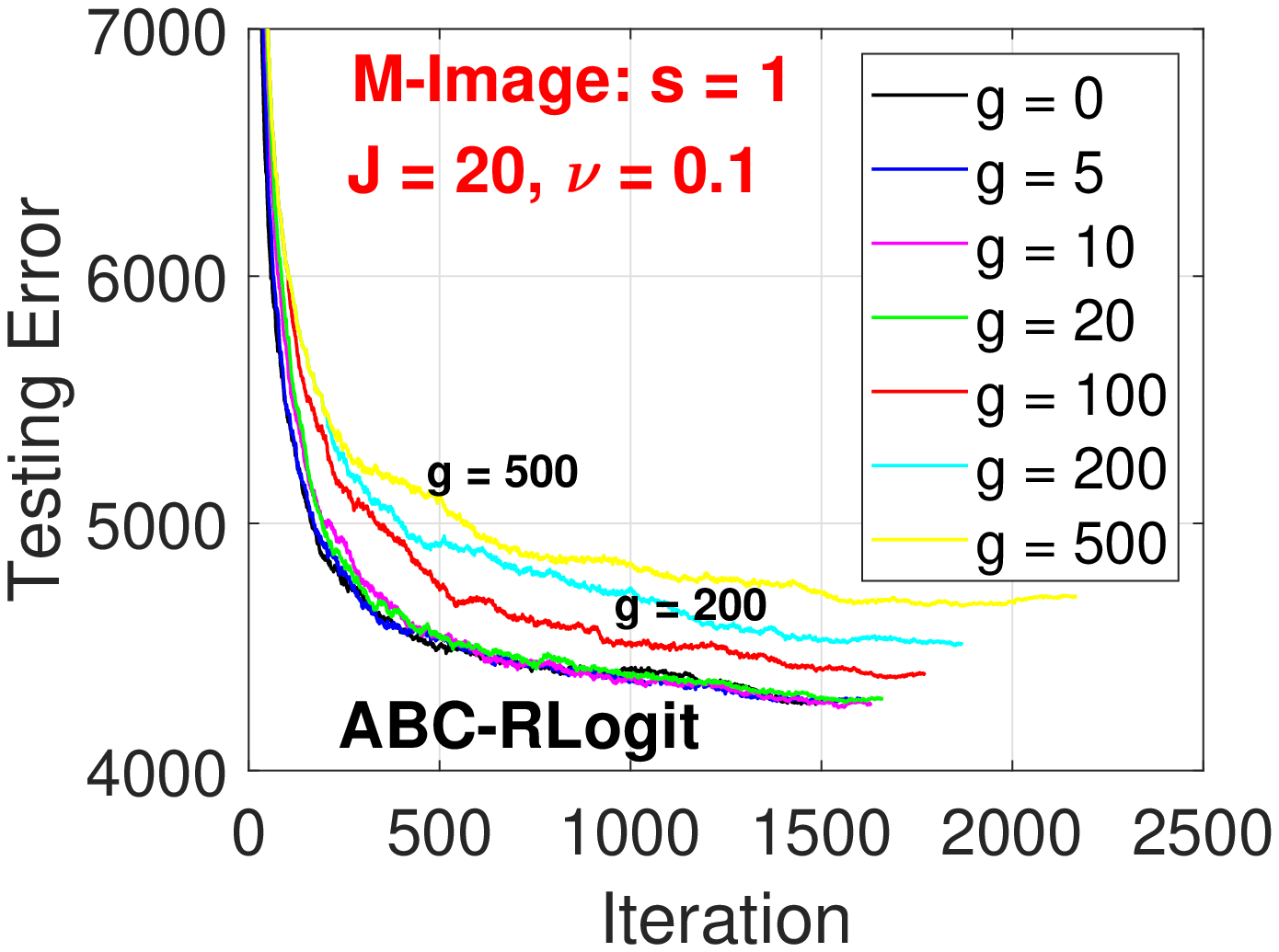}
    \includegraphics[width=2.2in]{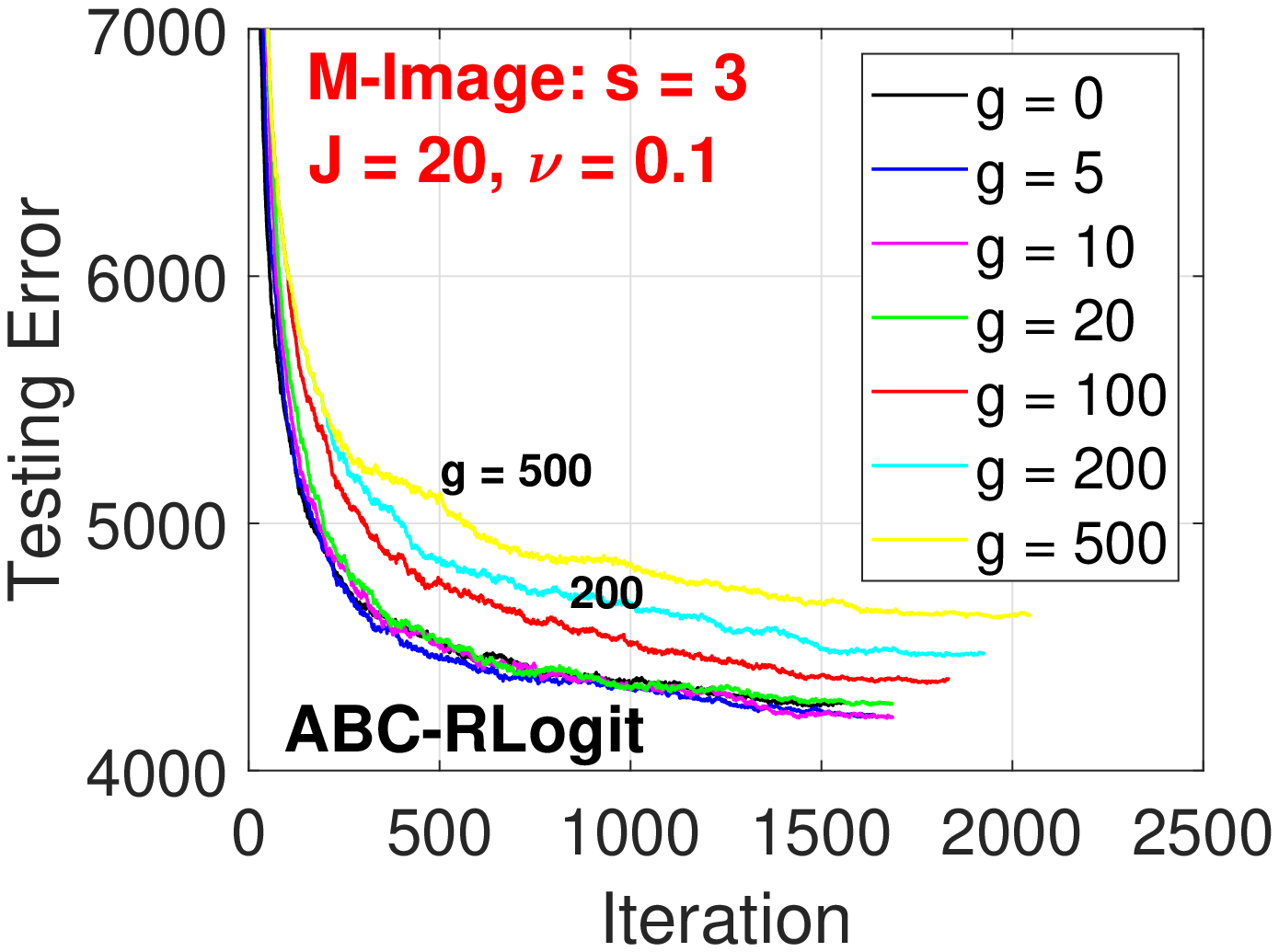}
    \includegraphics[width=2.2in]{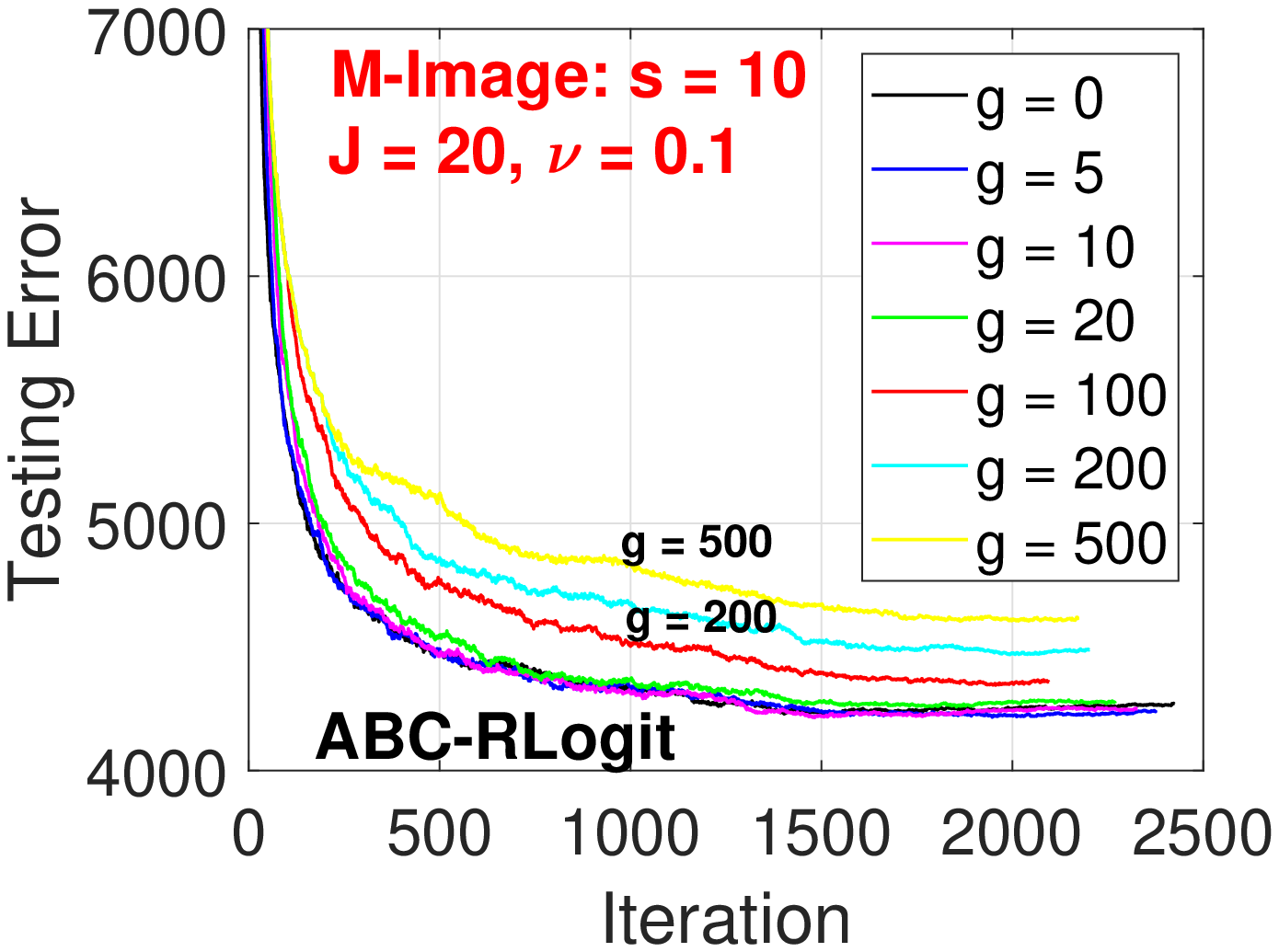}
}

\mbox{
    \includegraphics[width=2.2in]{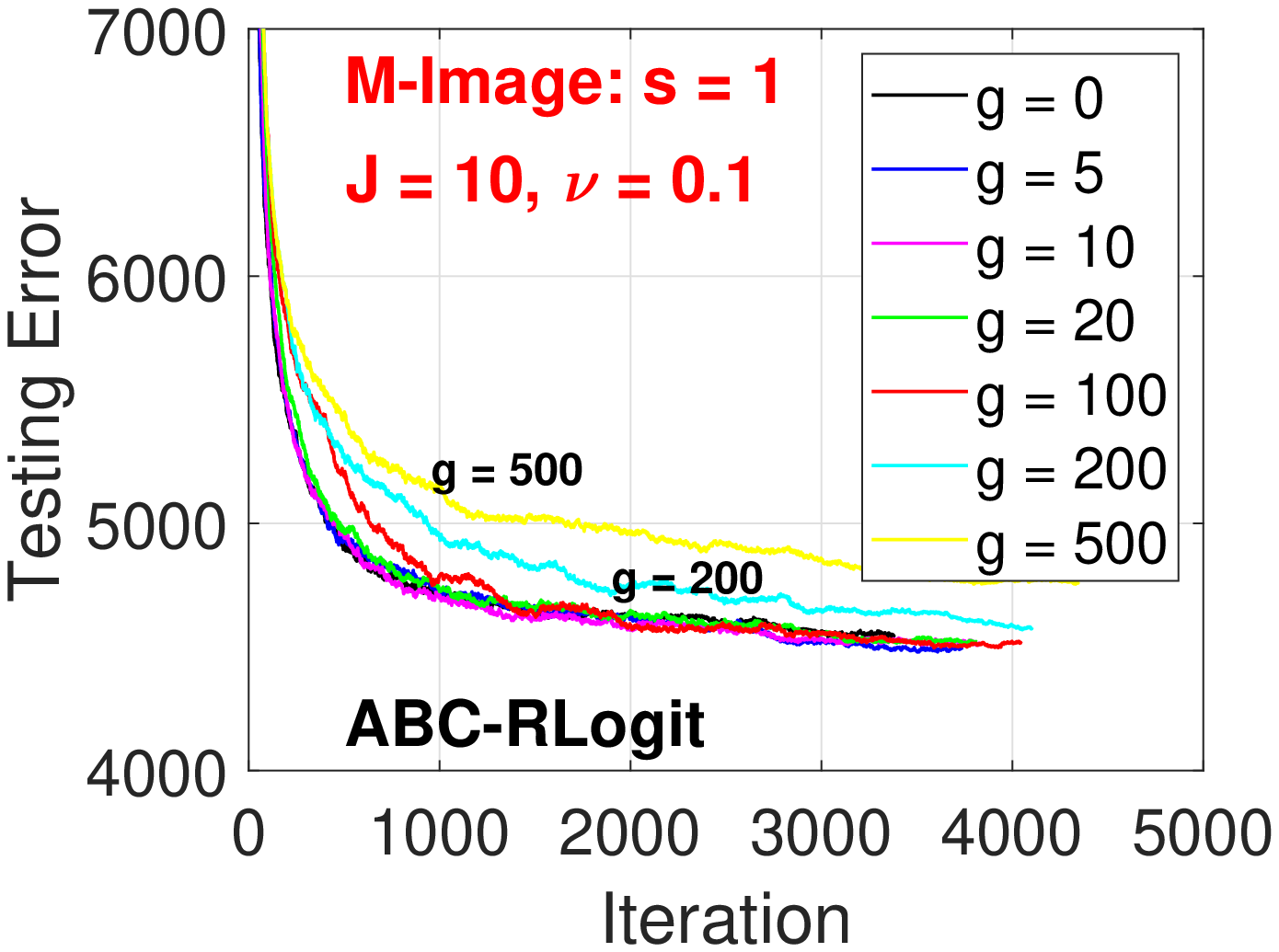}
    \includegraphics[width=2.2in]{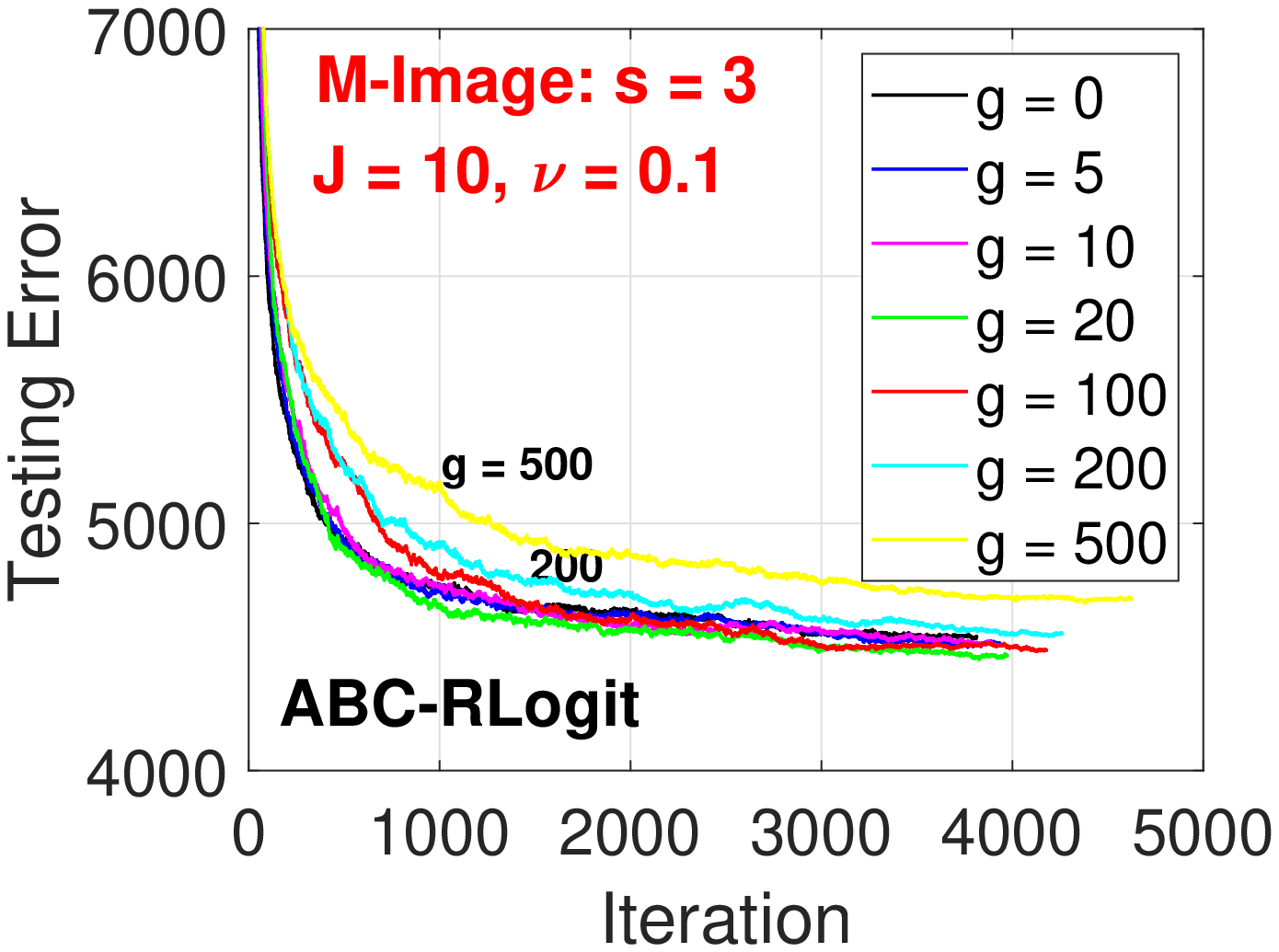}
    \includegraphics[width=2.2in]{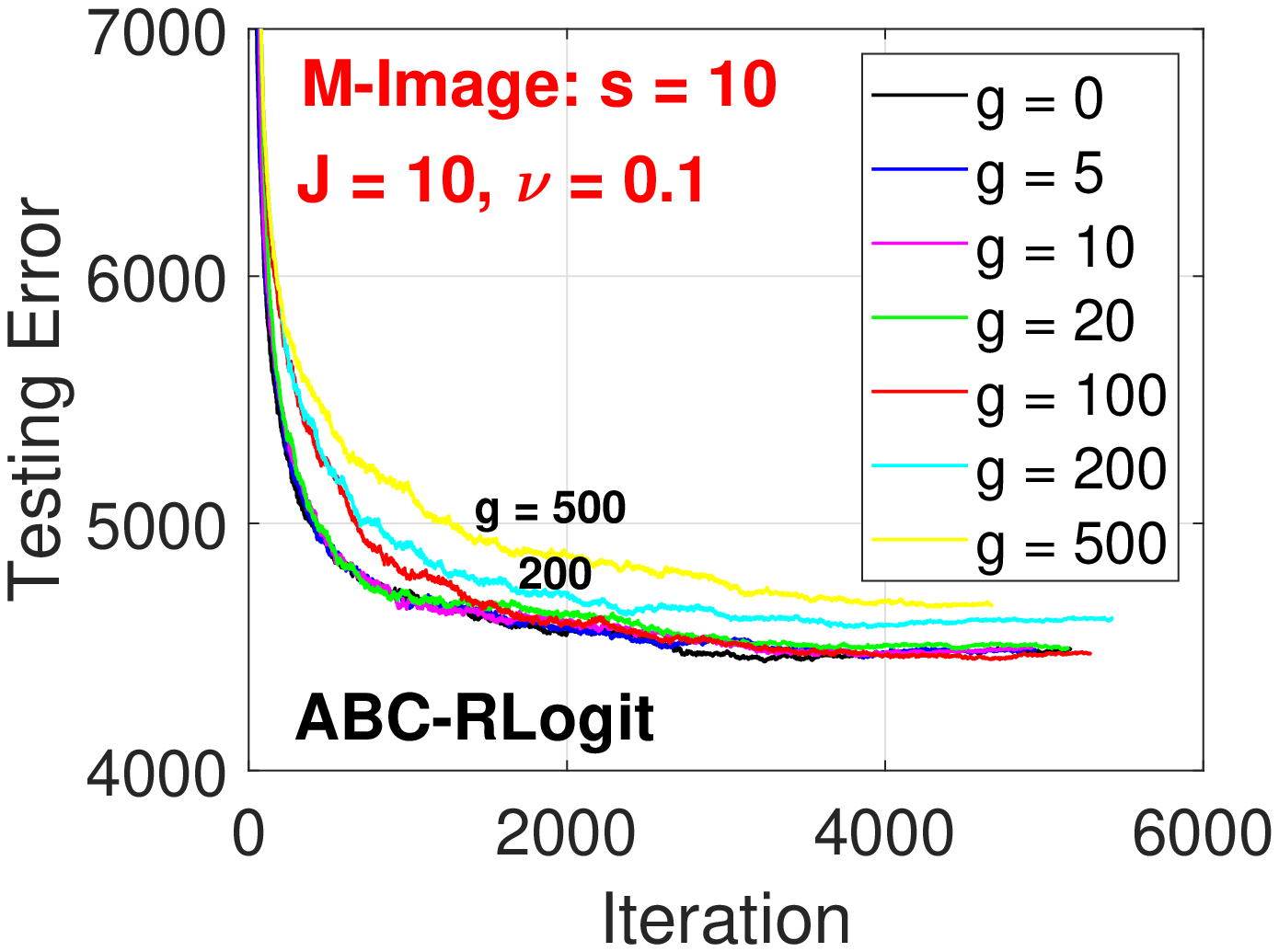}
}

\end{center}

\vspace{-0.1in}

\caption{{\em M-Image} dataset. Test classification errors of ABC-MART and ABC-RobustLogitBoost with parameters $s$ and $g$. We present results for  $g\in\{0,5,10,20,100,200,500\}$ and three selected $\in\{1,3,10\}$.
}\label{fig:M-Image_sg}
\end{figure}

\begin{figure}[h]
\begin{center}
\mbox{
    \includegraphics[width=2.2in]{fig/M-Rand/M-Rand-Test-J20v01_abcmart_s1g.eps}
    \includegraphics[width=2.2in]{fig/M-Rand/M-Rand-Test-J20v01_abcmart_s3g.eps}
    \includegraphics[width=2.2in]{fig/M-Rand/M-Rand-Test-J20v01_abcmart_s10g.eps}
}

\mbox{
    \includegraphics[width=2.2in]{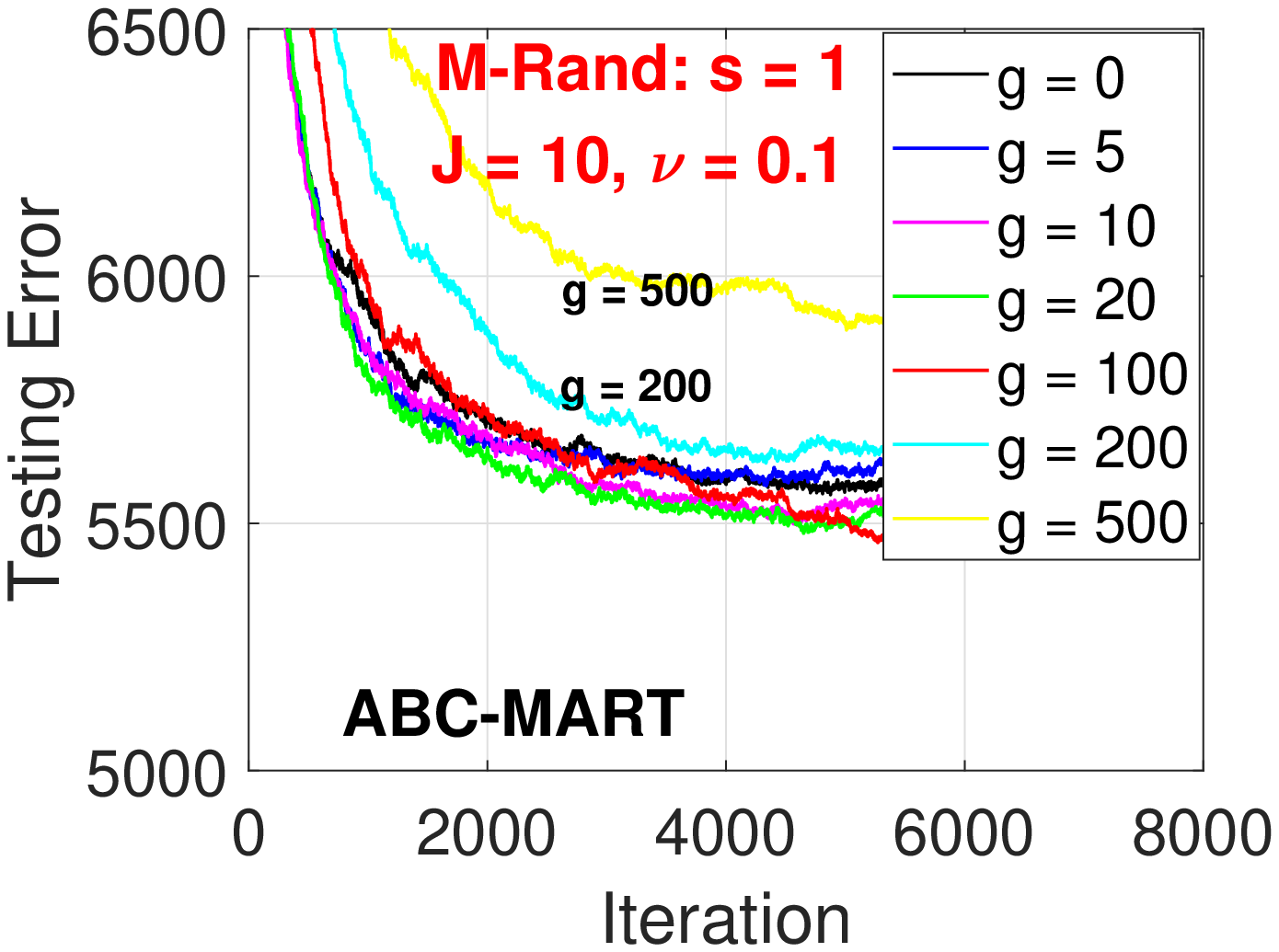}
    \includegraphics[width=2.2in]{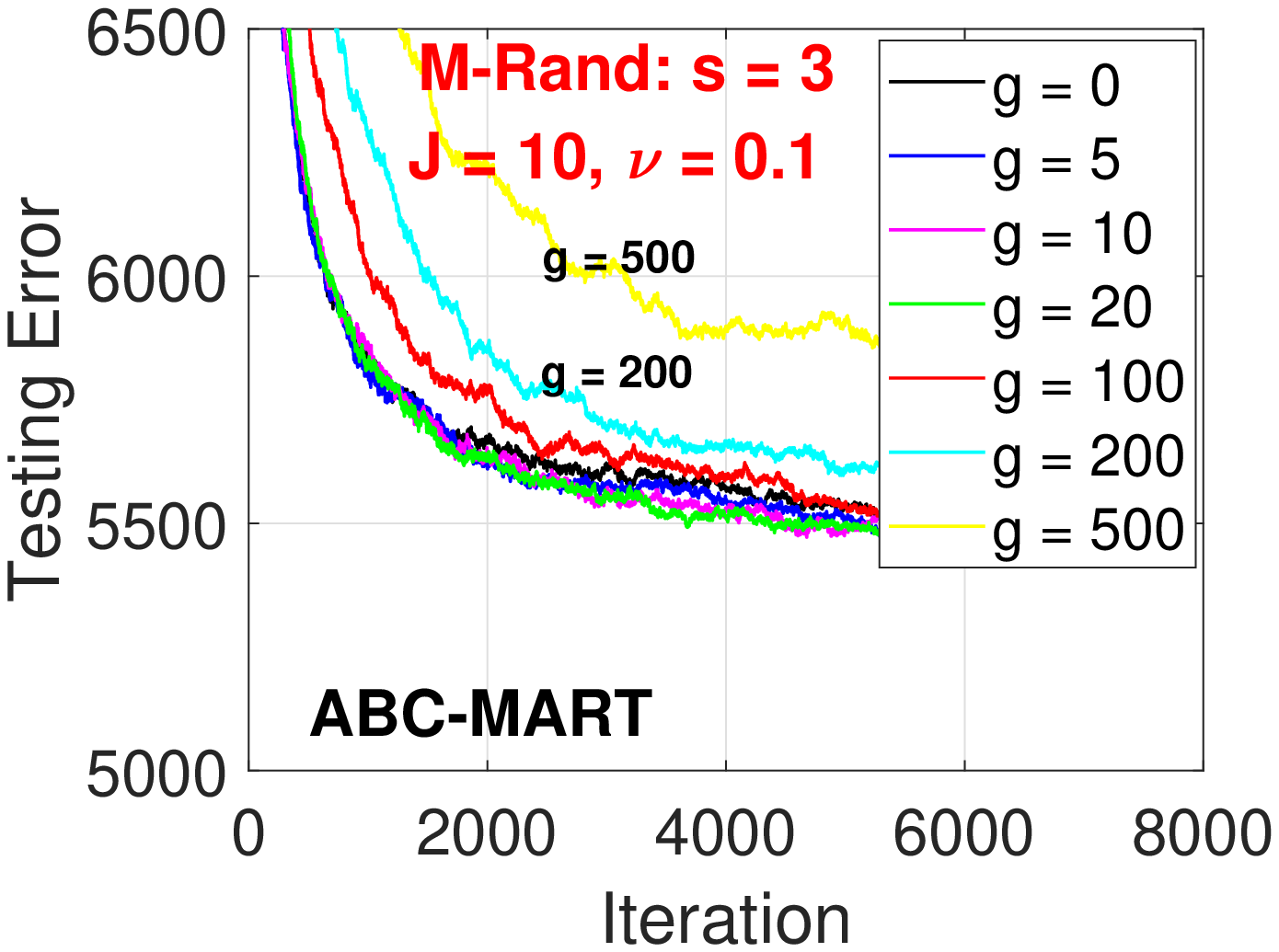}
    \includegraphics[width=2.2in]{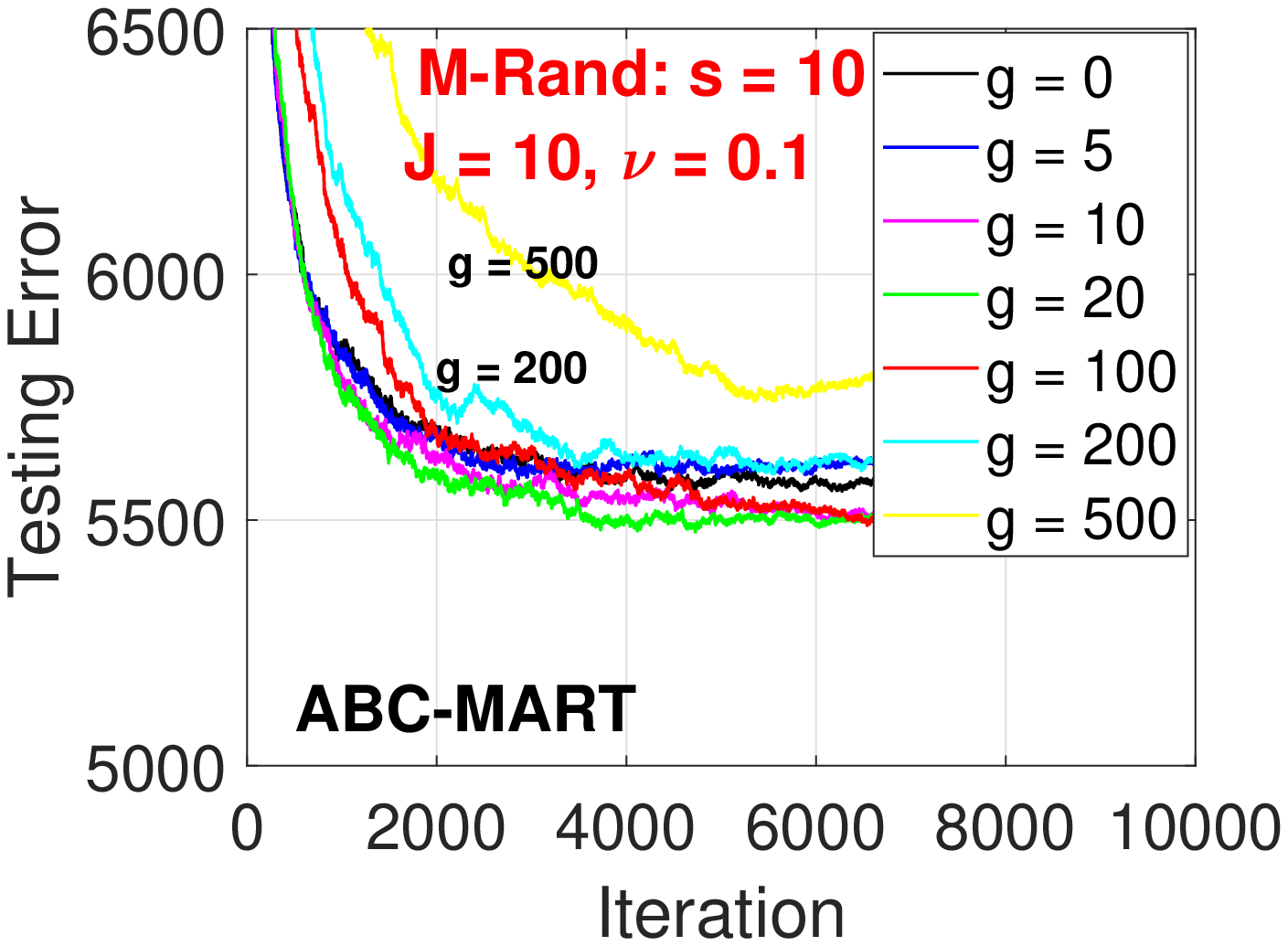}
}

\mbox{
    \includegraphics[width=2.2in]{fig/M-Rand/M-Rand-Test-J20v01_abclogit_s1g.eps}
    \includegraphics[width=2.2in]{fig/M-Rand/M-Rand-Test-J20v01_abclogit_s3g.eps}
    \includegraphics[width=2.2in]{fig/M-Rand/M-Rand-Test-J20v01_abclogit_s10g.eps}
}

\mbox{
    \includegraphics[width=2.2in]{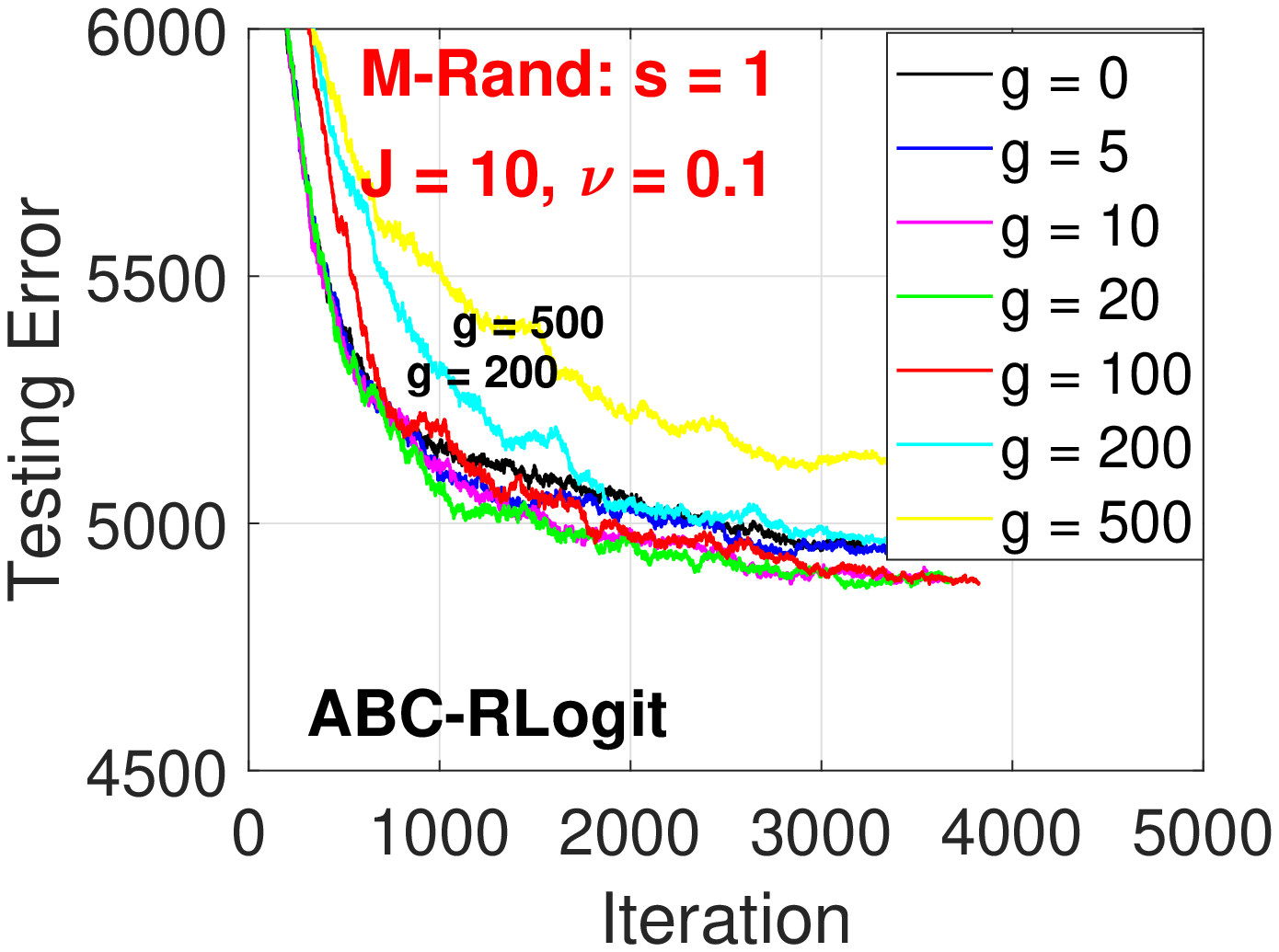}
    \includegraphics[width=2.2in]{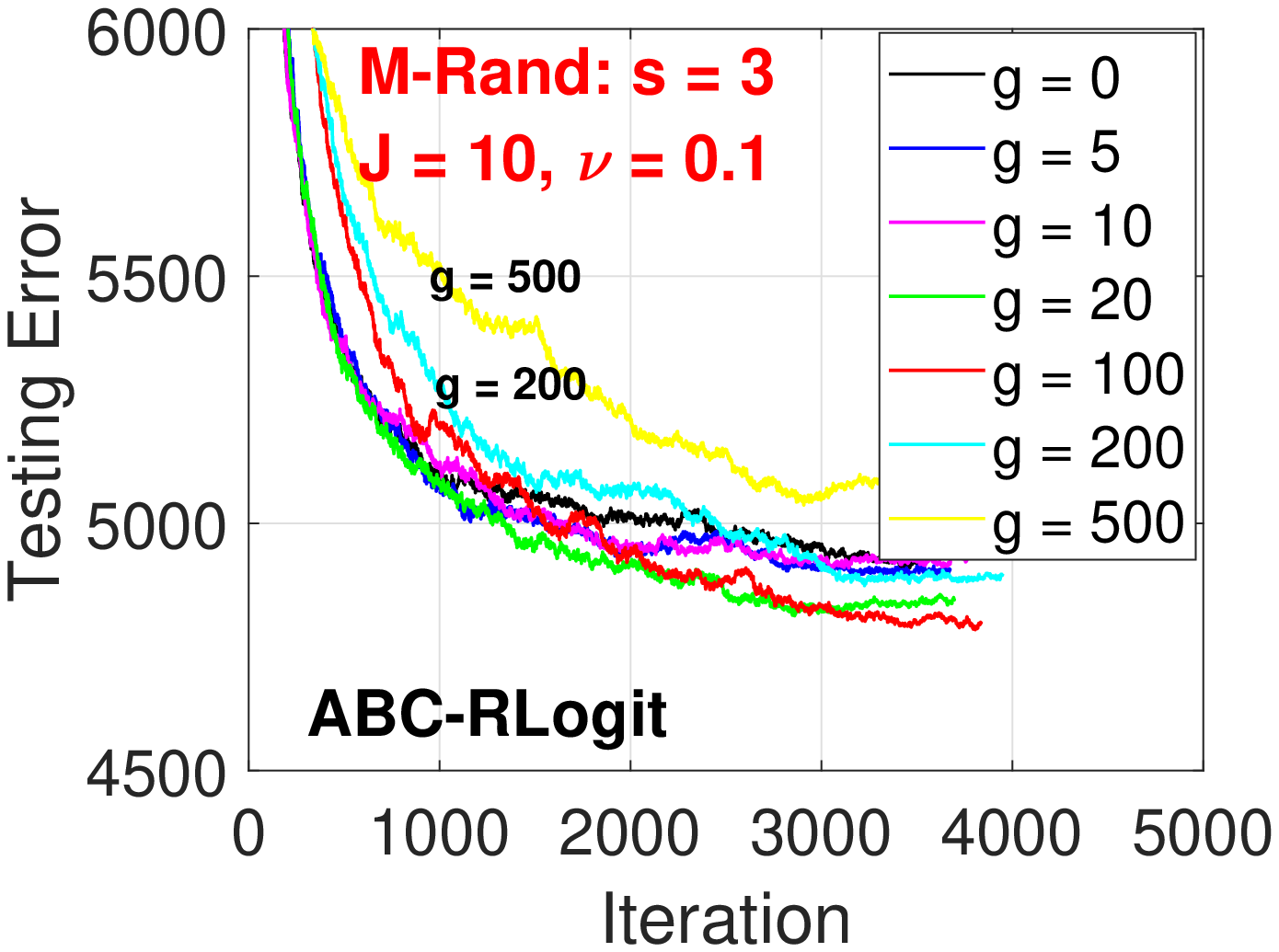}
    \includegraphics[width=2.2in]{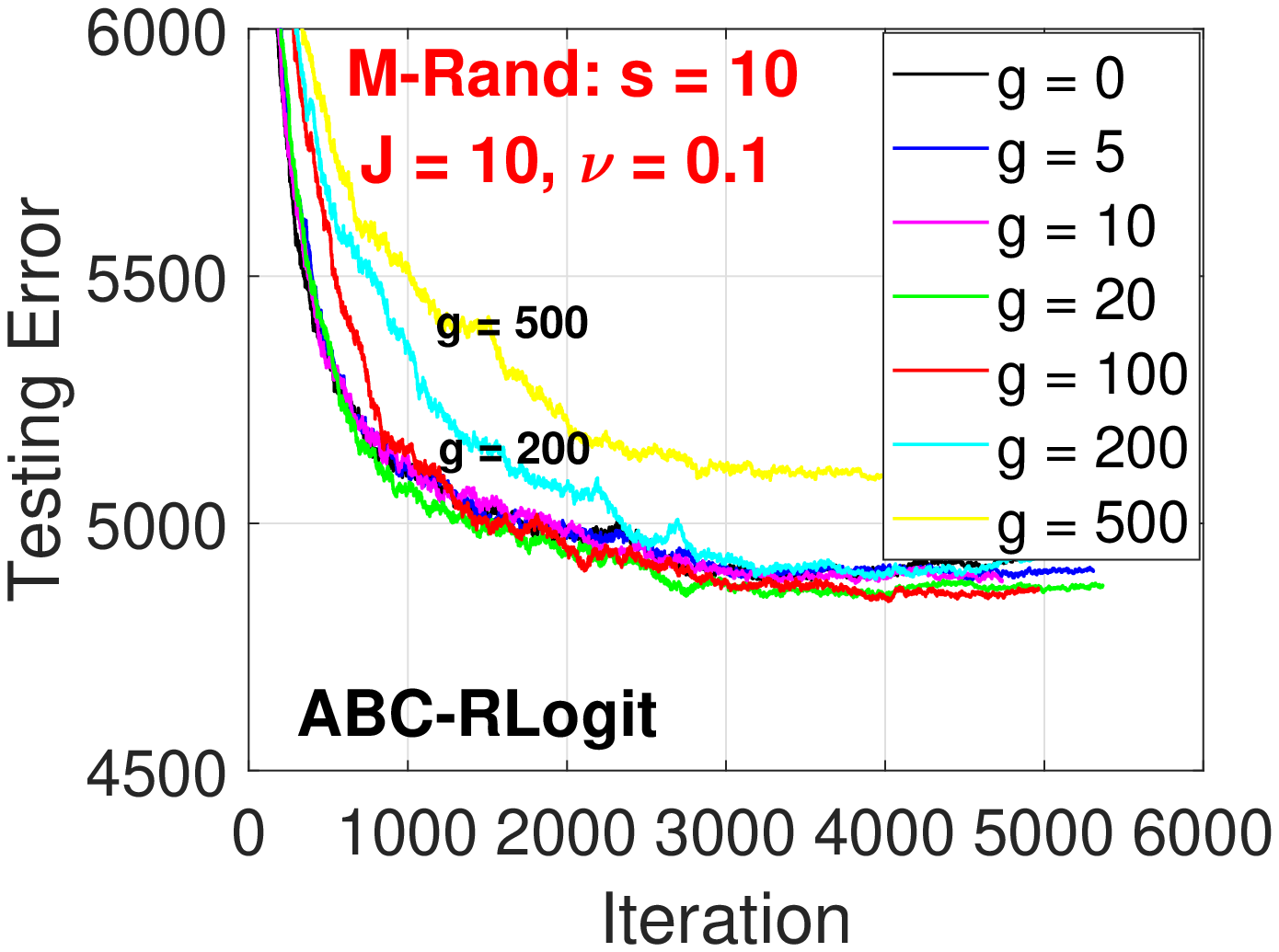}
}

\end{center}

\vspace{-0.1in}

\caption{{\em M-Rand} dataset. Test classification errors of ABC-MART and ABC-RobustLogitBoost with parameters $s$ and $g$. We present results for  $g\in\{0,5,10,20,100,200,500\}$ and three selected $\in\{1,3,10\}$.   }\label{fig:M-Rand_sg}
\end{figure}

\begin{figure}[h]
\begin{center}
\mbox{
    \includegraphics[width=2.2in]{fig/Letter10k/Letter10k-Test-J20v01_abcmart_s1g.eps}
    \includegraphics[width=2.2in]{fig/Letter10k/Letter10k-Test-J20v01_abcmart_s3g.eps}
    \includegraphics[width=2.2in]{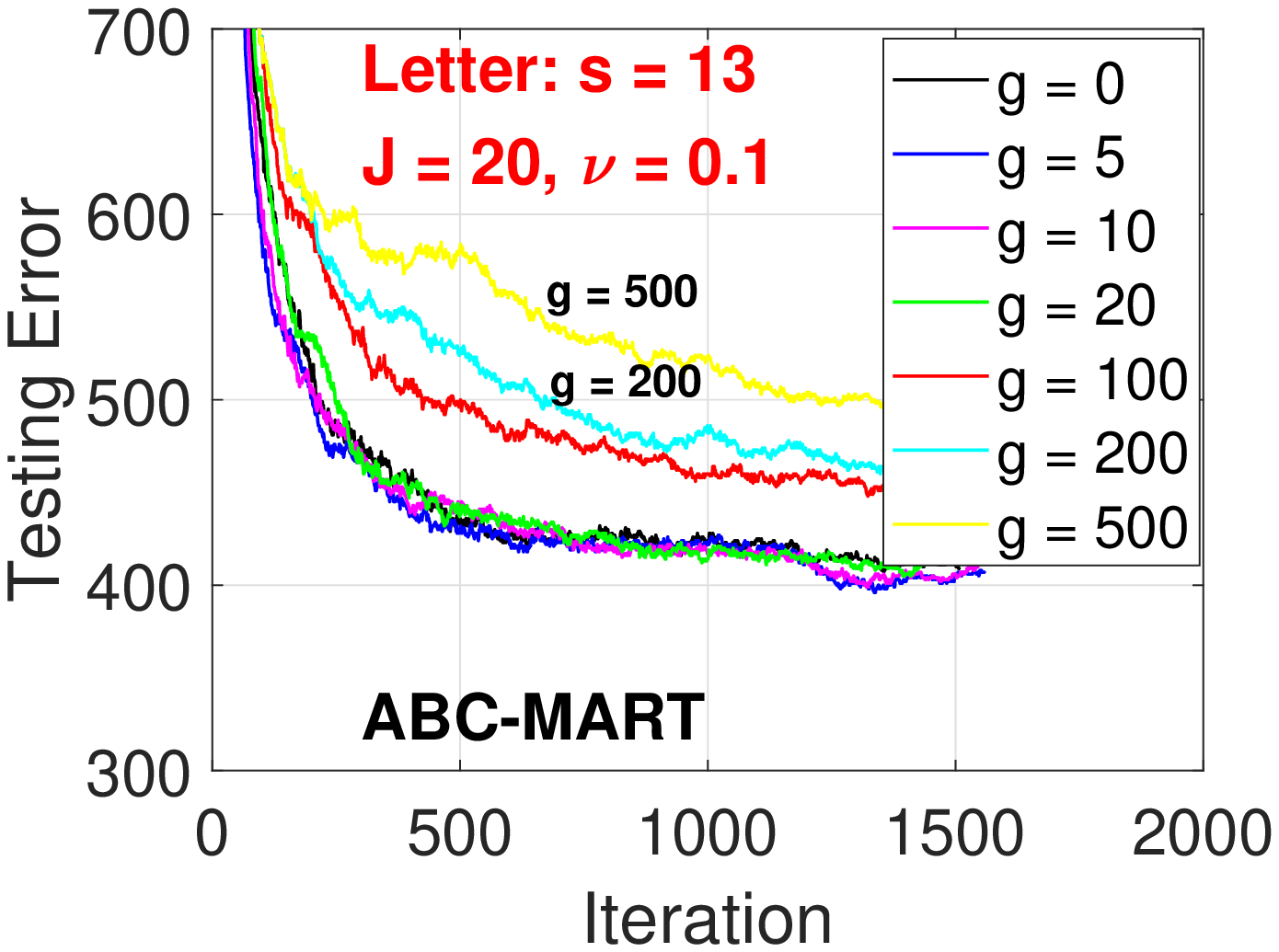}
}

\mbox{
    \includegraphics[width=2.2in]{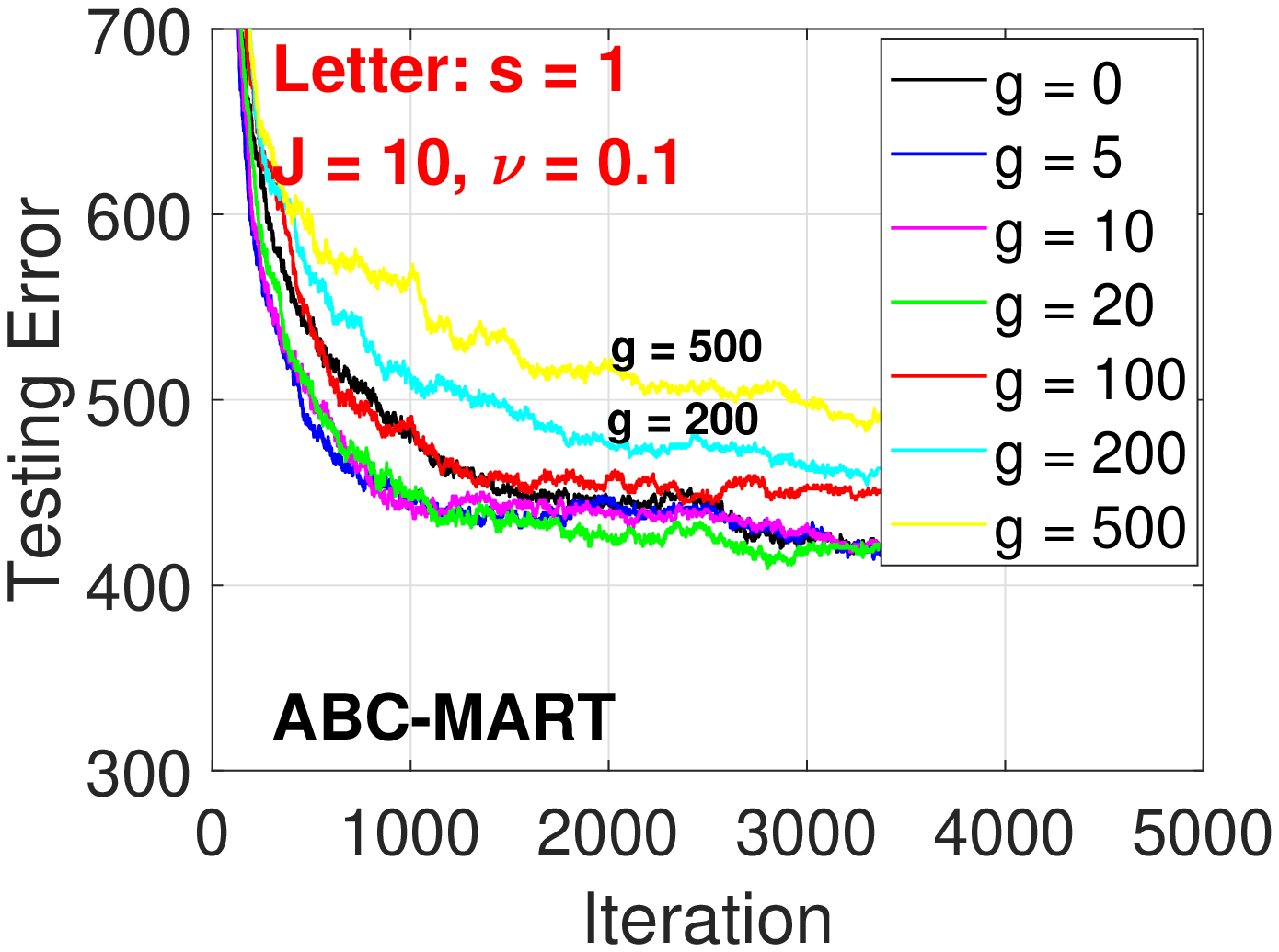}
    \includegraphics[width=2.2in]{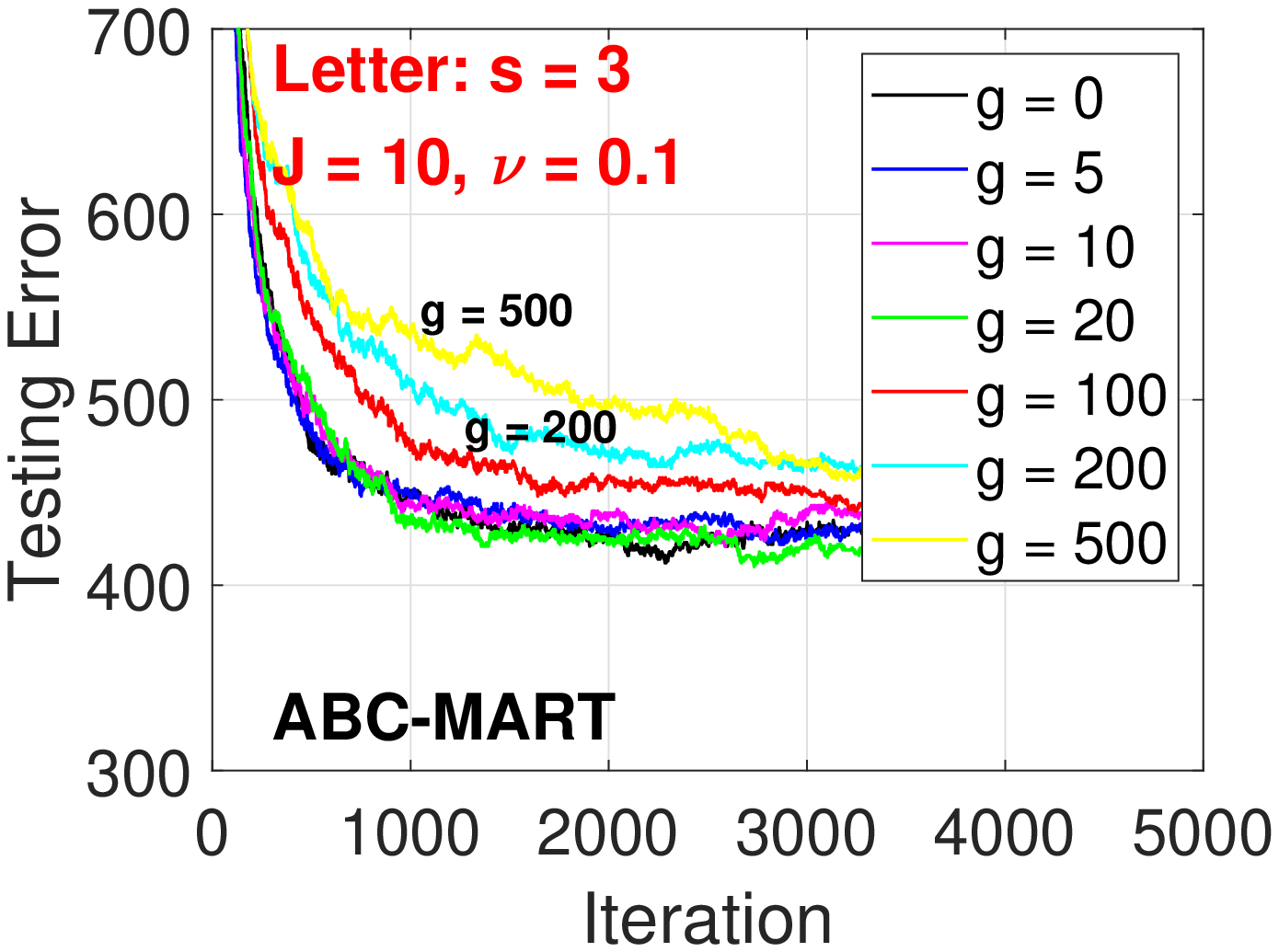}
    \includegraphics[width=2.2in]{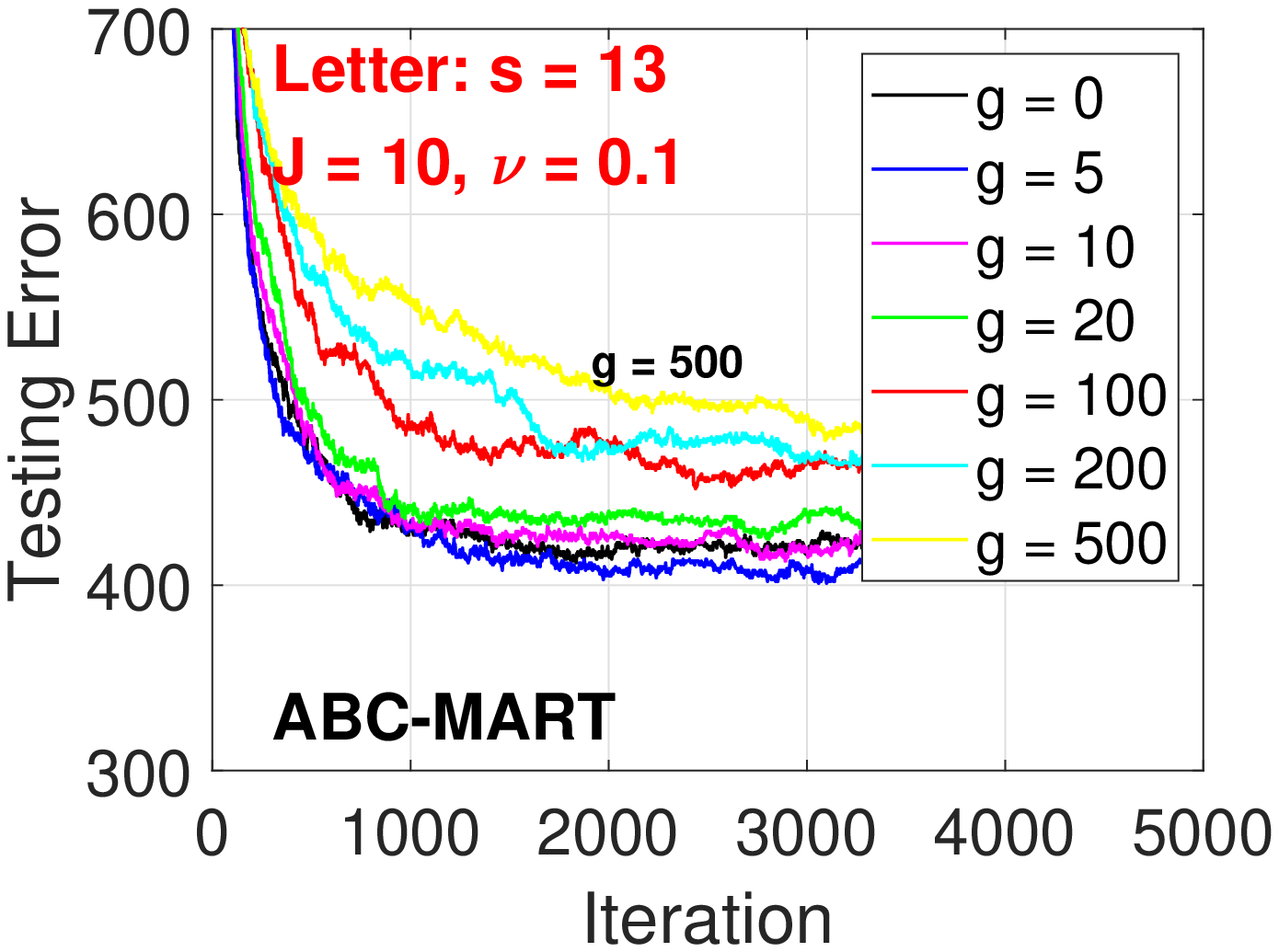}
}

\mbox{
    \includegraphics[width=2.2in]{fig/Letter10k/Letter10k-Test-J20v01_abclogit_s1g.eps}
    \includegraphics[width=2.2in]{fig/Letter10k/Letter10k-Test-J20v01_abclogit_s3g.eps}
    \includegraphics[width=2.2in]{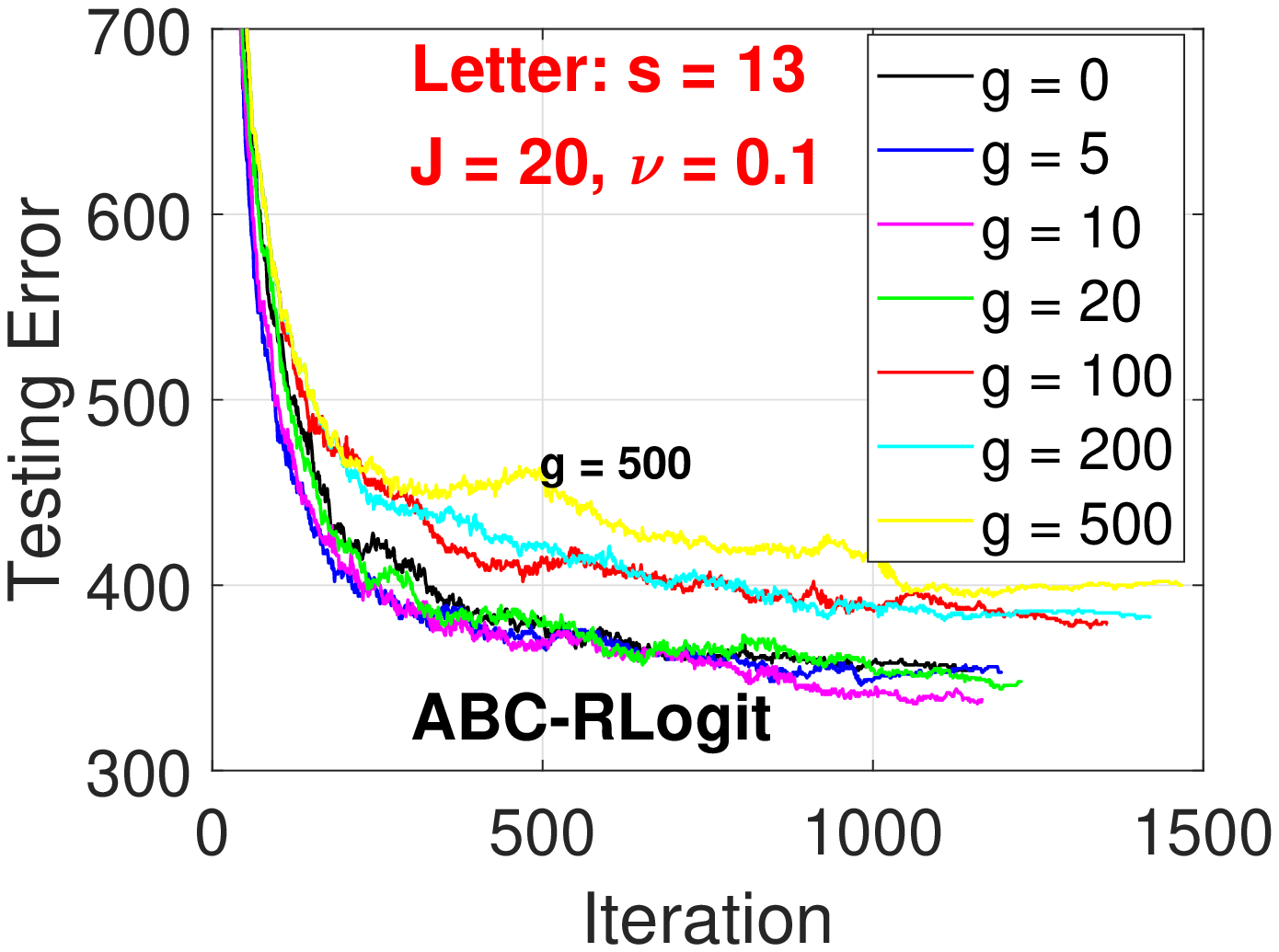}
}

\mbox{
    \includegraphics[width=2.2in]{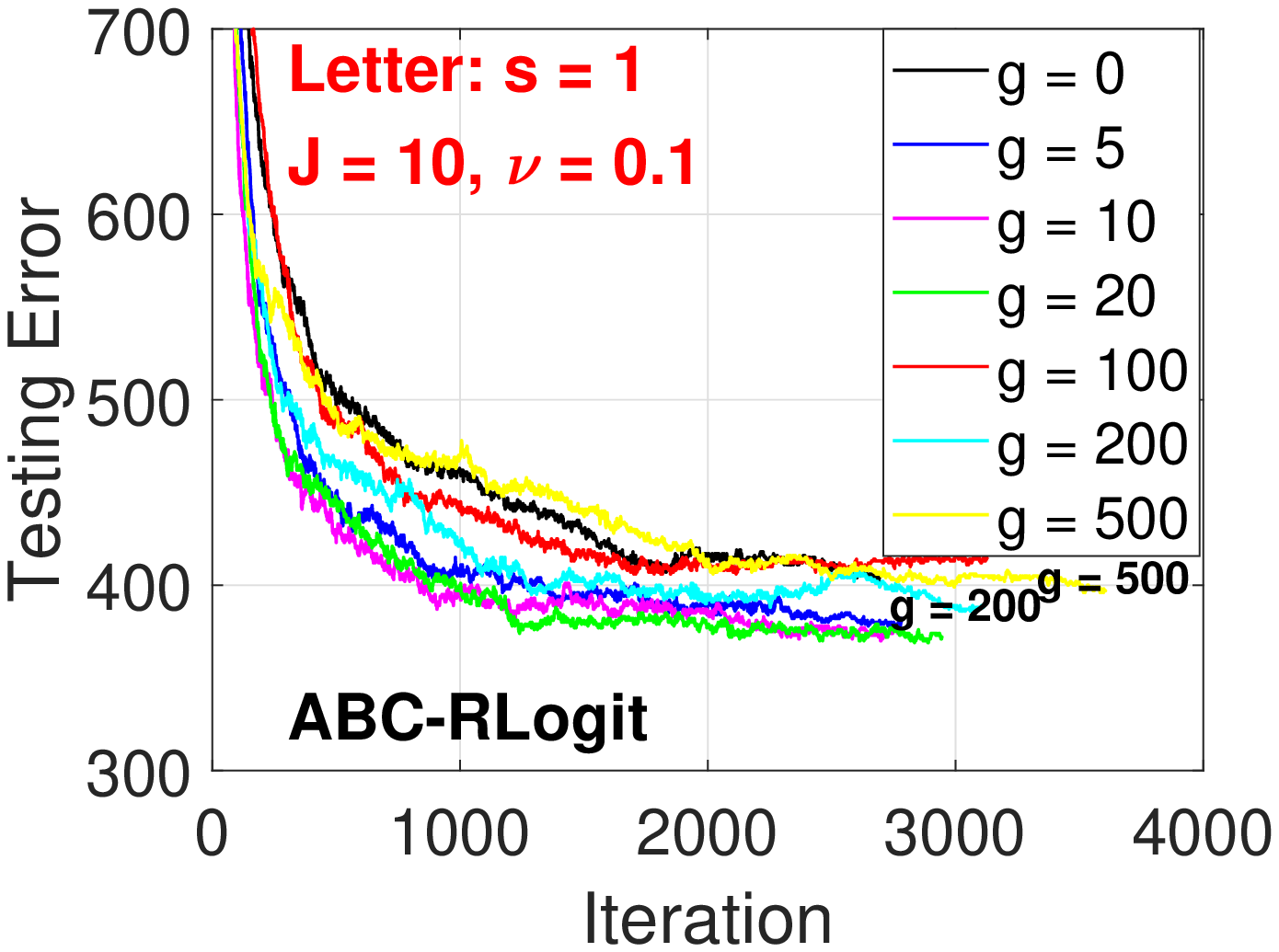}
    \includegraphics[width=2.2in]{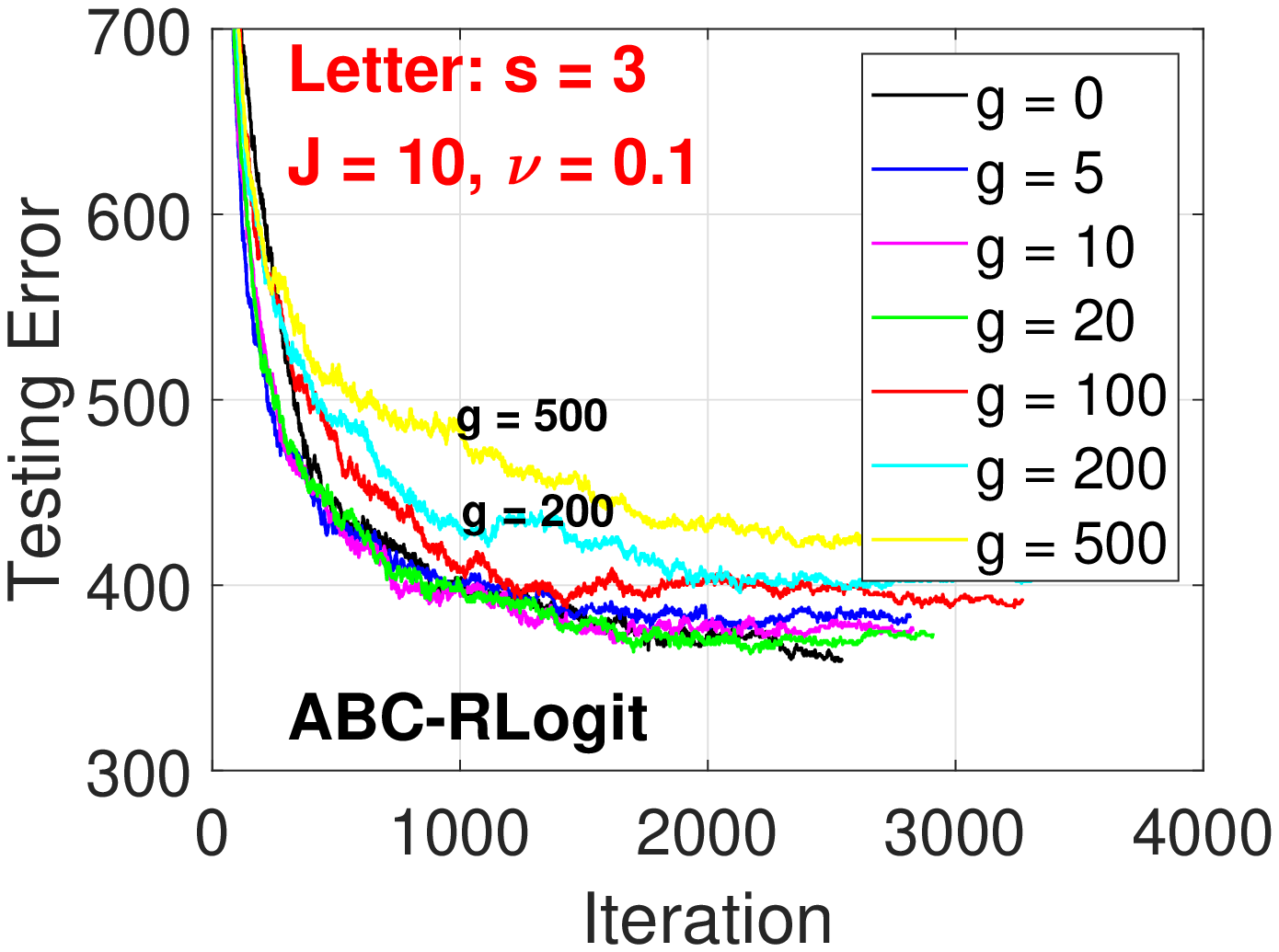}
    \includegraphics[width=2.2in]{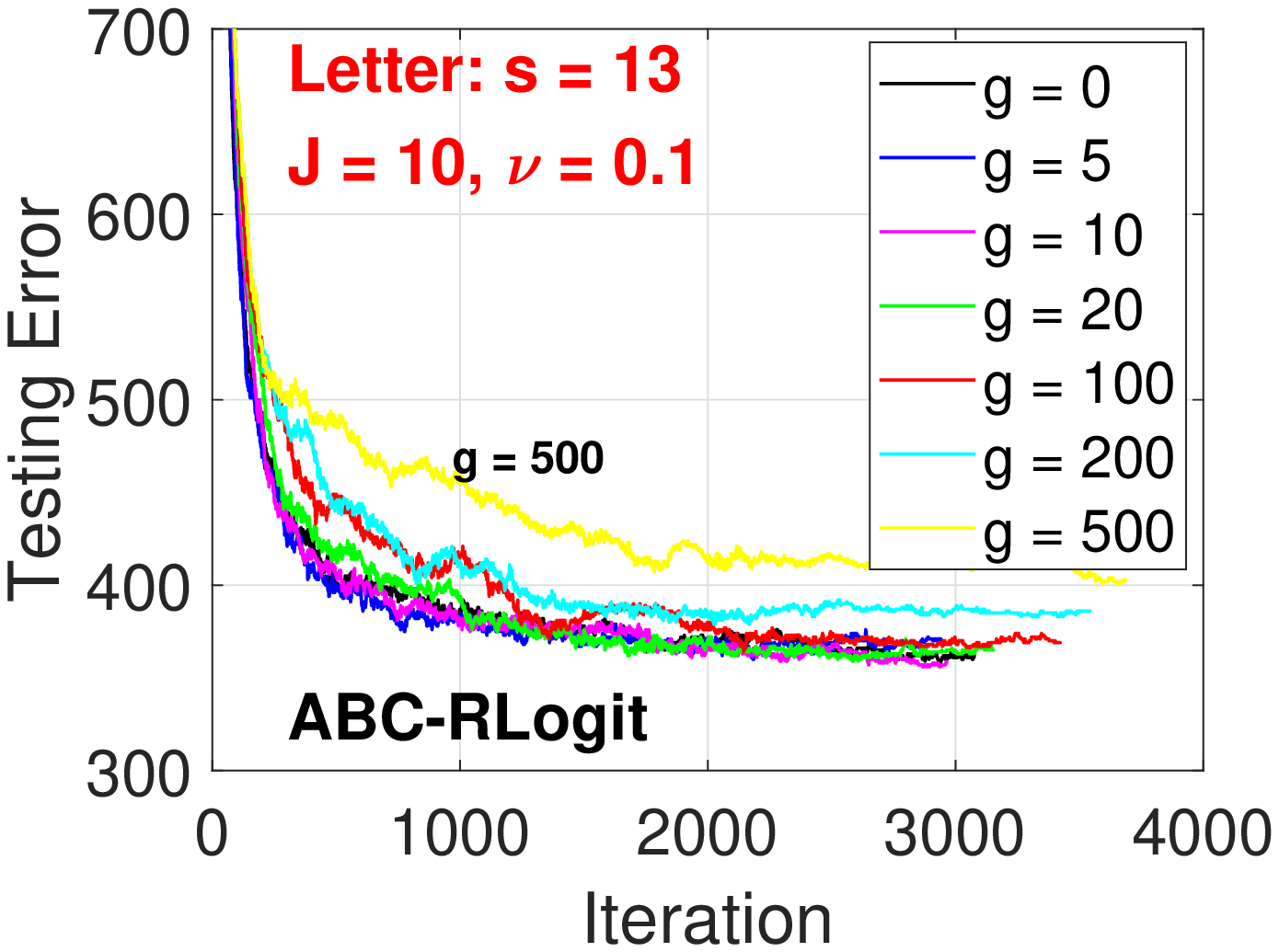}
}

\end{center}

\vspace{-0.1in}

\caption{{\em Letter} dataset. Test classification errors of ABC-MART and ABC-RobustLogitBoost with parameters $s$ and $g$. We present results for  $g\in\{0,5,10,20,100,200,500\}$ and three selected $\in\{1,3,13\}$.   }\label{fig:Letter10k_sg}
\end{figure}

To not overwhelm readers with a large number of figures, we first present a selected small set of experiments in Figure~\ref{fig:M-Rand-Letter10k} for {\em M-Rand} and {\em Letter} datasets, for $s\in\{1, 3, K\}$ and $g\in\{0, 5, 10, 20, 100, 200, 500\}$. We provide more detailed experiments in Figures~\ref{fig:Covertype_s} to~\ref{fig:Letter10k_s}, for $g=0$ and $s=1, 2, 3, ..., K$. Then in Figures~\ref{fig:Covertype_sg} to~\ref{fig:Letter10k_sg}, we present, for selected $s$ values, the results with a series of $g\in\{0, 5, 10, 20,100, 200,500\}$. We summarize the findings from the experimental results as follows:
\begin{itemize}
\item When $g=0$, i.e., we conduct search at every iteration, using the ``exhaustive search'' strategy (i.e., $s=K$) produces better results than using the ``worst class'' strategy (i.e., $s=1$), but the difference is typically not too big. However, in some cases, we observe ``catastrophic failures'' with the ``worst class'' strategy, for example, ABC-MART on the {\em Letter} dataset with $J=20$ and $\nu=0.1$. Interestingly, the best accuracy does not necessarily occur at $s=K$. In these 6 datasets, it looks $s=2\sim 4$ might be a good initial choice.
\item ABC-Boost algorithms can typically tolerate fairly large gaps (i.e., large $g$ values). In almost all cases using $g=10$ produces results which are (close to) the best. For some datasets, the algorithms can tolerate $g=100$ or even larger. Interestingly, for ABC-MART on the {\em Letter} dataset, using $s=1$ and $g>0$ actually avoids the ``catastrophic failure'' (at $s=1$ and $g=0$). This might be a good example to illustrate that sometimes doing less work might lead to better outcomes.
\item In general, ABC-RobustLogitBoost is less sensitive to parameters $s$ and $g$, compared to ABC-MART.
\end{itemize}

\begin{algorithm}[h]{\small
$F_{i,k} = 0$,\ \  $p_{i,k} = \frac{1}{K}$, \ \ \ $k = 0$ to  $K-1$, \ $i = 1$ to $N$ \\
$L_{\textit{prev}}^{(k)} = \sum_{i=1}^{N}{1_{y_i = k}}$, \ \ \ $k = 0$ to $K - 1$\\
For $m=1$ to $M$ Do\\
\hspace{0.1in} If $(m - 1) \textit{ mod } (g + 1) = 0$ Then\\
\hspace{0.2in}    $L_{\textit{prev}}^{(k'_0)}, L_{\textit{prev}}^{(k'_1)}, \dots, L_{\textit{prev}}^{(k'_{K-1})} = \text{Sort} \  L_{\textit{prev}}^{(k)}, \ k = 0$ to $K-1$ decreasingly.\\
\hspace{0.2in}    $\textit{search\_classes} = \{k'_0, k'_1, \dots, k'_{s-1}\}$\\
\hspace{0.1in} Else\\
\hspace{0.2in} $\textit{search\_classes} = \{B(m - 1)\}$\\
\hspace{0.1in} End\\
\hspace{0.1in}    For $b \in \textit{search\_classes}$, Do\\
\hspace{0.2in}    For $k=0$ to $K-1$, $k\neq b$, Do\\
\hspace{0.3in}  $\left\{R_{j,k,m}\right\}_{j=1}^J = J$-terminal
node regression tree from  $\{-(r_{i,b} - p_{i,b}) +  (r_{i,k} - p_{i,k}), \ \ \mathbf{x}_{i}\}_{i=1}^N$  with weights $p_{i,b}(1-p_{i,b})+p_{i,k}(1-p_{i,k})+2p_{i,b}p_{i,k}$, using the tree split gain formula Eq.~\eqref{eqn:logit_gain}.
 \\
 \hspace{0.3in}  $\beta_{j,k,m} = \frac{ \sum_{\mathbf{x}_i \in
  R_{j,k,m}} -(r_{i,b} - p_{i,b}) + (r_{i,k} - p_{i,k})  }{ \sum_{\mathbf{x}_i\in
  R_{j,k,m}} p_{i,b}(1-p_{i,b})+ p_{i,k}\left(1-p_{i,k}\right) + 2p_{i,b}p_{i,k} }$ \\
\hspace{0.3in}  $G_{i,k,b} = F_{i,k} +
\nu\sum_{j=1}^J\beta_{j,k,m}1_{\mathbf{x}_i\in R_{j,k,m}}$ \\
 \hspace{0.2in} End\\
\hspace{0.2in} $G_{i,b,b} = - \sum_{k\neq b} G_{i,k,b}$ \\
\hspace{0.2in}  $q_{i,k} = \exp(G_{i,k,b})/\sum_{s=0}^{K-1}\exp(G_{i,s,b})$ \\
\hspace{0.2in} $L^{(b)} = -\sum_{i=1}^N \sum_{k=0}^{K-1} r_{i,k}\log\left(q_{i,k}\right)$\\
\hspace{0.1in} End\\
\hspace{0.1in} $B(m) = \underset{b}{\text{argmin}} \  \ L^{(b)}$\\
\hspace{0.1in} $F_{i,k} = G_{i,k,B(m)}$\\
\hspace{0.1in}  $p_{i,k} = \exp(F_{i,k})/\sum_{s=0}^{K-1}\exp(F_{i,s})$ \\
End}
\caption{Fast-ABC-RobustLogitBoost using the ``$s$-worst classes'' search strategy and the ``gap‘’ strategy (with parameter $g$) for the base class. }
\label{alg:fast-abc-LogitBoost}
\end{algorithm}

In summary, the experimental results on these datasets verify that Fast ABC-Boost with parameters $(s,g)$ works well. $s$ typically can be small (e.g., $s=2\sim 4$) and $g$ often can be pretty large (e.g., $g=10\sim 20$).  When $s=2$ and $g=10$, the cost of ABC-Boost is about $1.2KM$, which is only slightly larger than $KM$.

\newpage\clearpage

\newpage\clearpage

\newpage\clearpage

\newpage\clearpage

\subsection{Fast ABC-Boost with a ``Warm-up'' Stage }

Lastly, we would like introduce another idea by allowing a ``warm-up'' stage before training ABC-Boost. That is, we train MART or Robust LogitBoost in the first $w$ iterations and we shift to ABC-Boost starting at the $(w+1)$-th iteration.  This idea is also quite natural. Since MART and Robust LogitBoost are numerically stable, we might as well first obtain a fairly reasonable model before we use ABC-Boost for better accuracy.  With parameters $w$, $g$, and $w$, the computational cost becomes $O\left(Kw+s(K-1)\frac{1}{g+1}(M-w) + (K-1)\frac{g}{g+1}(M-w)\right)$.

\begin{figure}[h]
\begin{center}
\mbox{
    \includegraphics[width=2.2in]{fig/Letter10k/Letter10k-Test-J20v01_abcmart_s1g.eps}
    \includegraphics[width=2.2in]{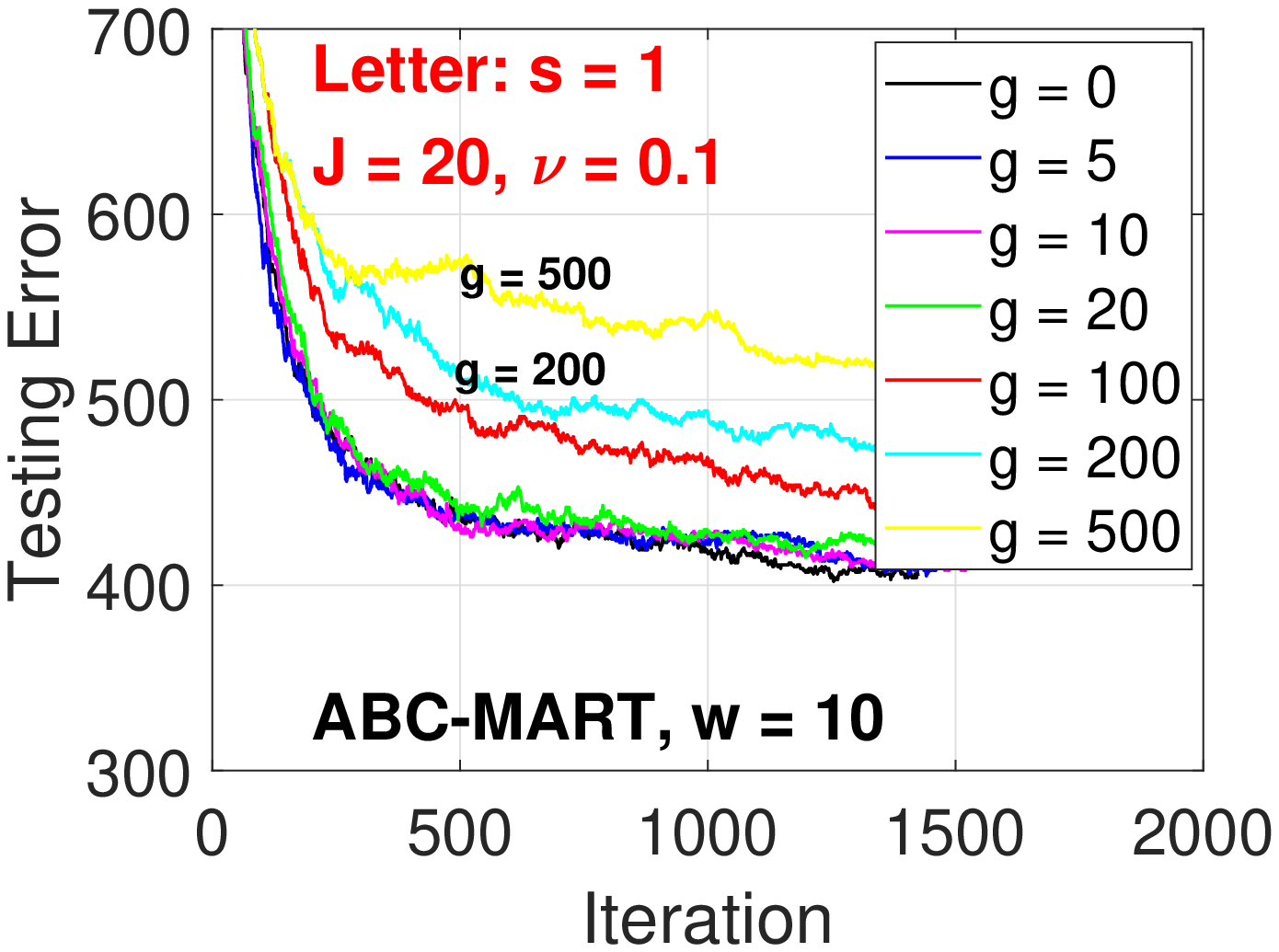}
    \includegraphics[width=2.2in]{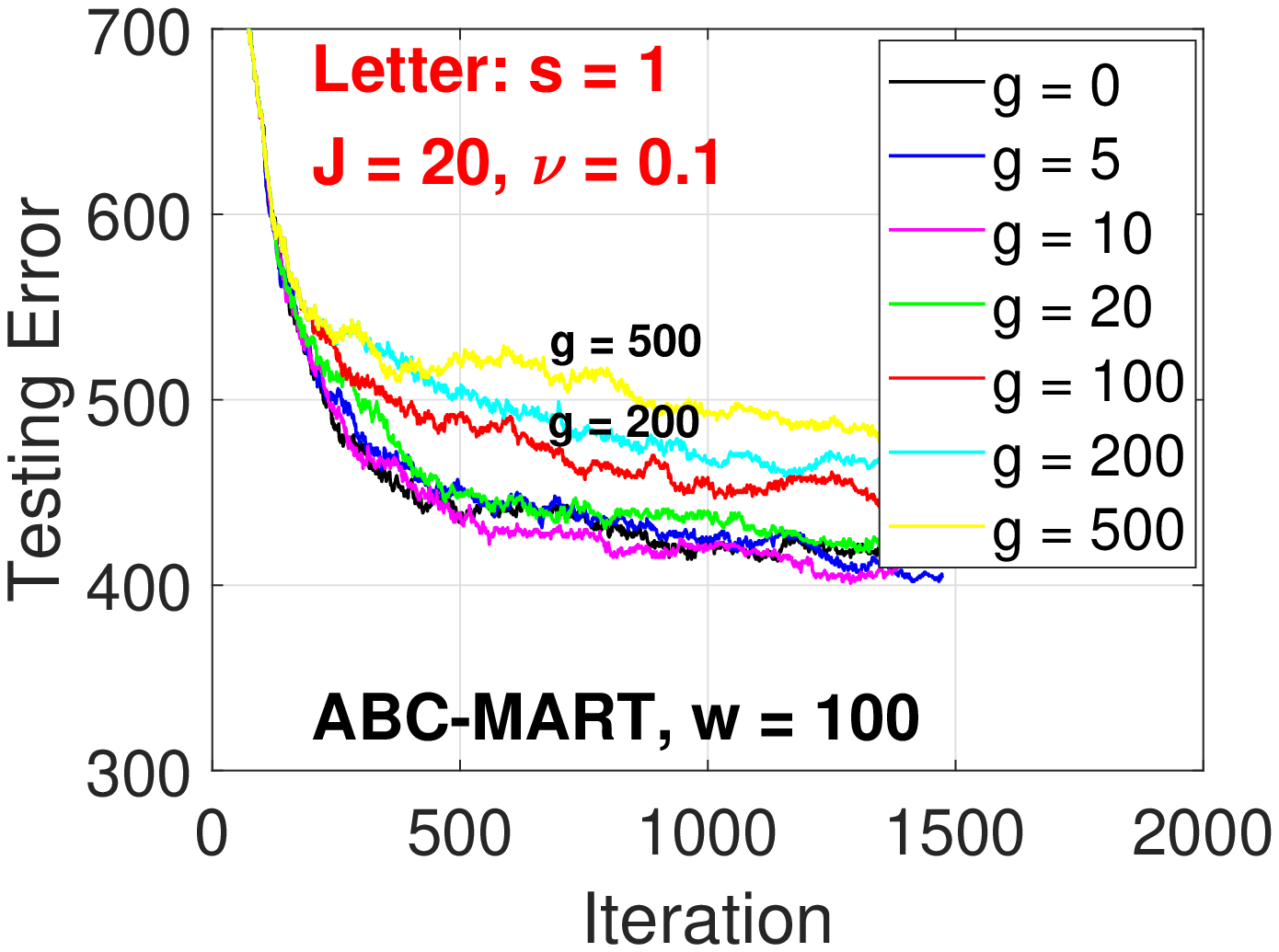}
}

\mbox{
    \includegraphics[width=2.2in]{fig/Letter10k/Letter10k-Test-J20v01_abcmart_s3g.eps}
    \includegraphics[width=2.2in]{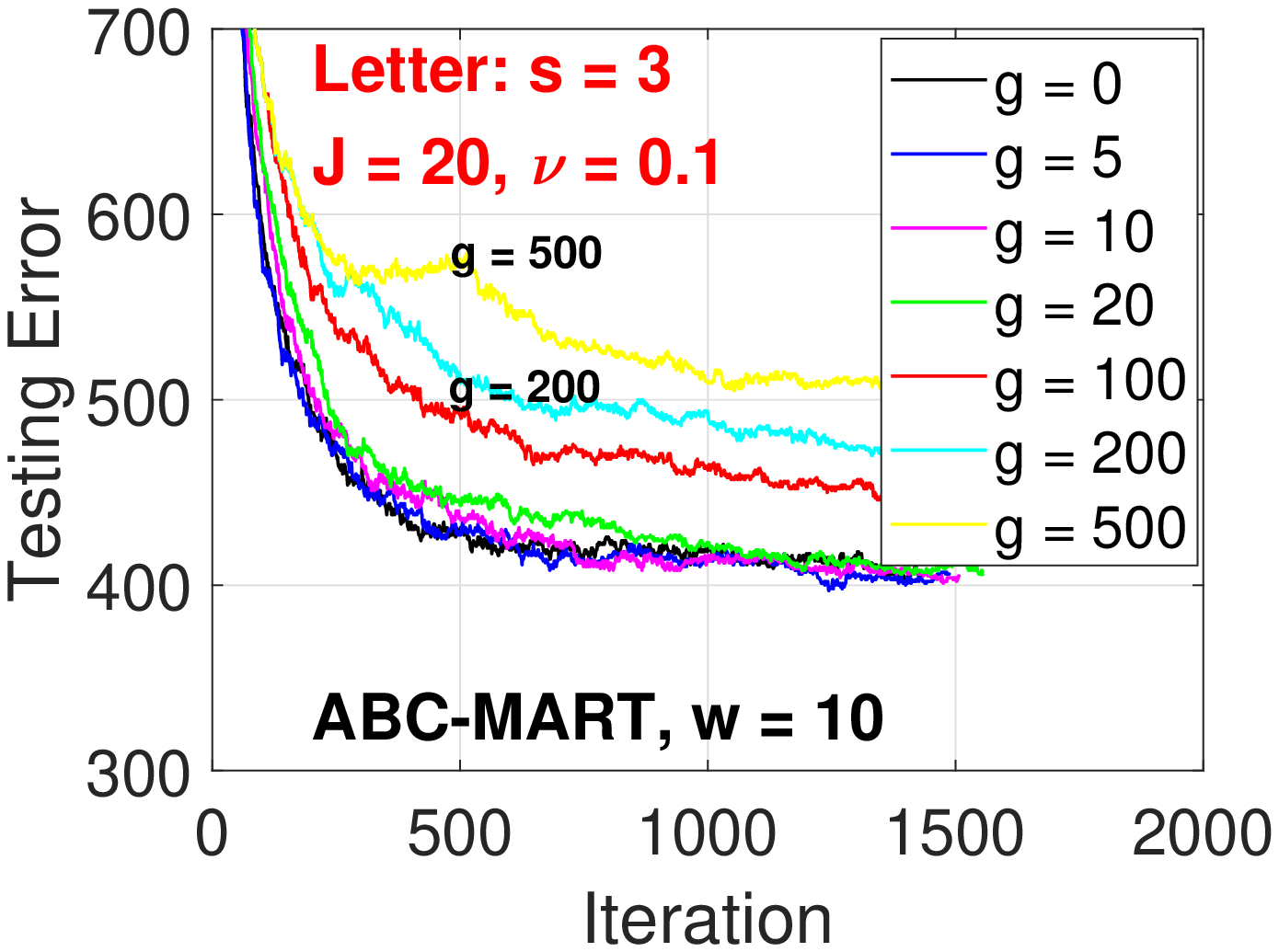}
    \includegraphics[width=2.2in]{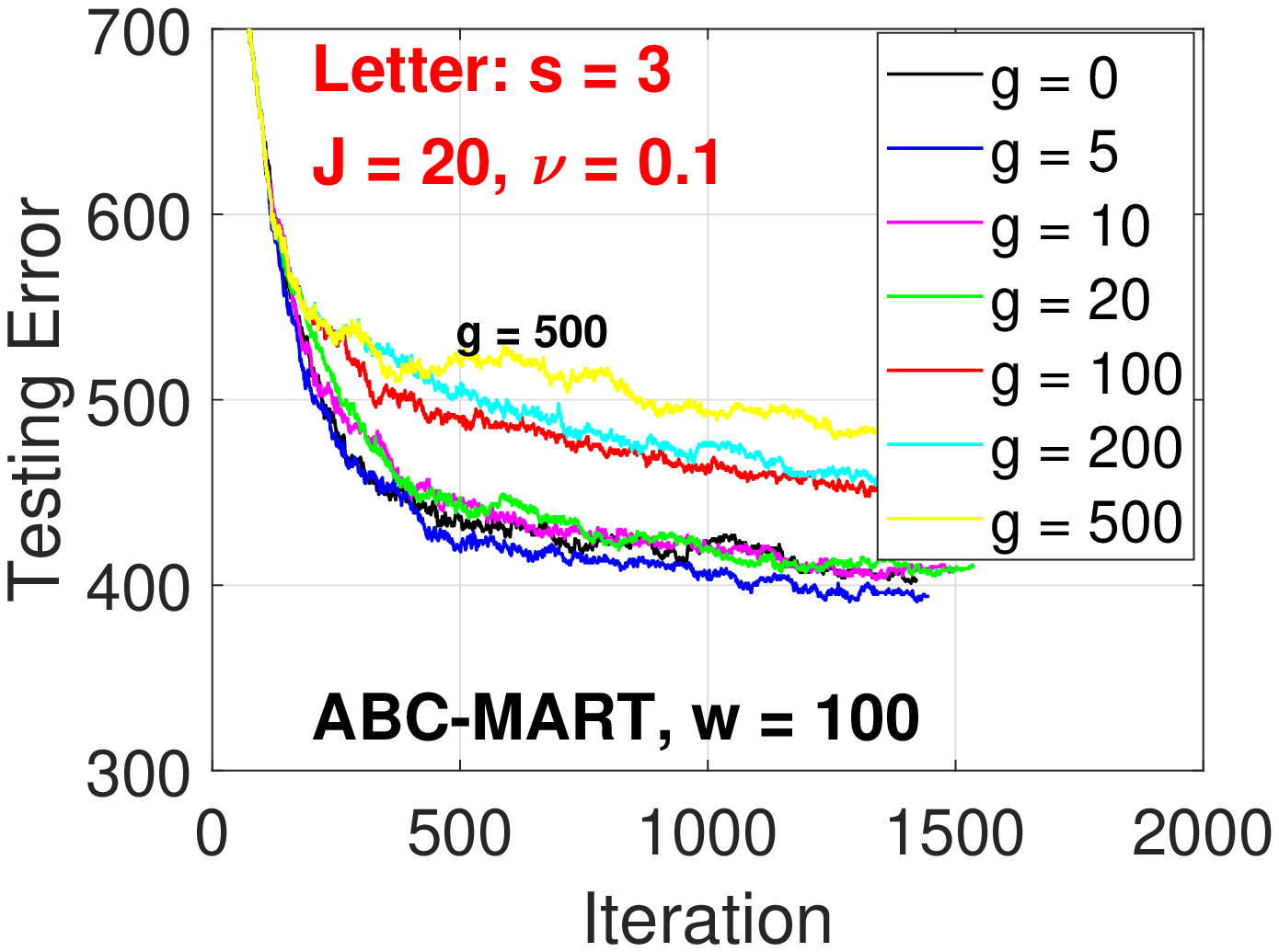}
}

\mbox{
    \includegraphics[width=2.2in]{fig/Letter10k/Letter10k-Test-J20v01_abclogit_s1g.eps}
    \includegraphics[width=2.2in]{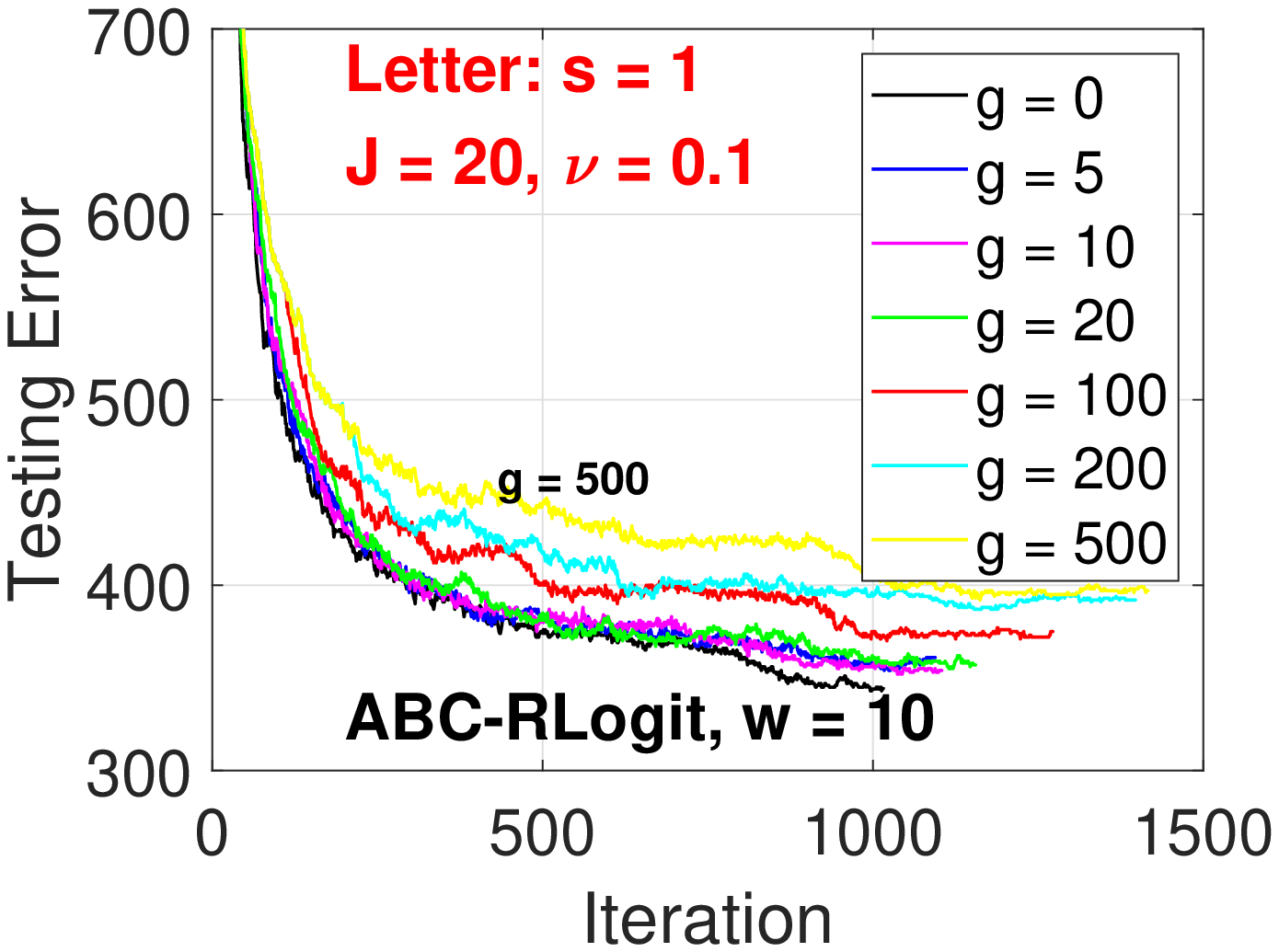}
    \includegraphics[width=2.2in]{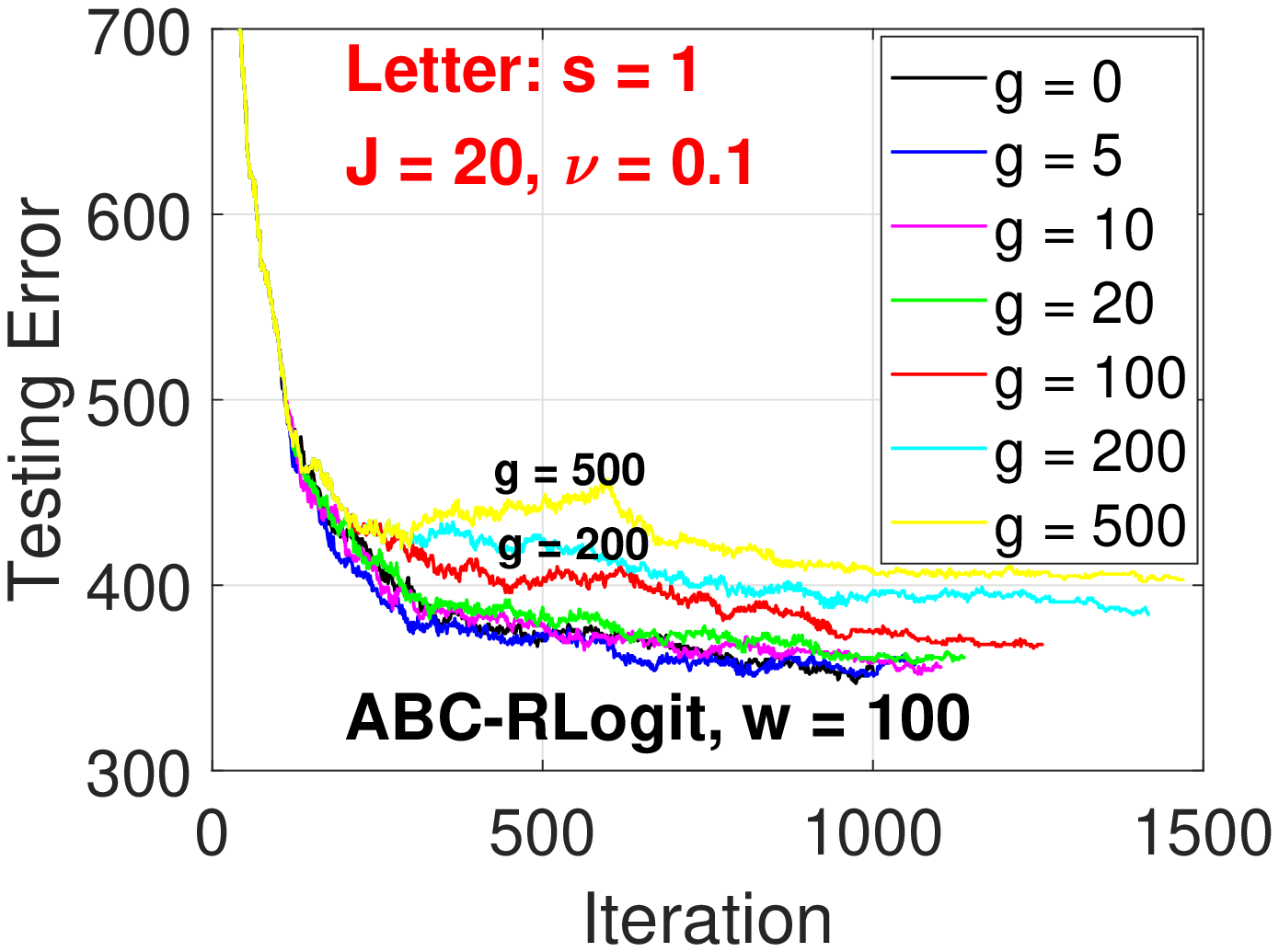}
}

\mbox{
    \includegraphics[width=2.2in]{fig/Letter10k/Letter10k-Test-J20v01_abclogit_s3g.eps}
    \includegraphics[width=2.2in]{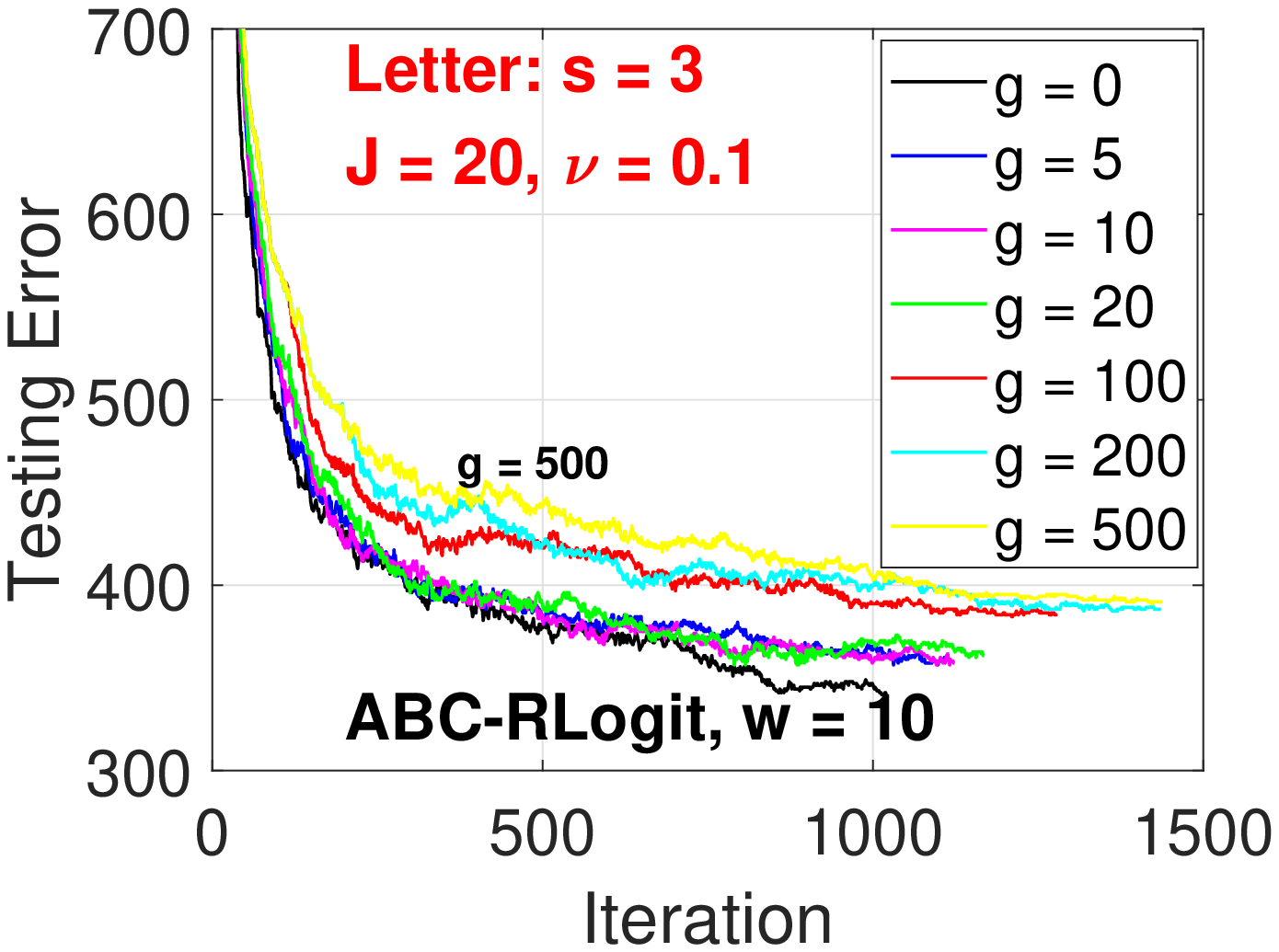}
    \includegraphics[width=2.2in]{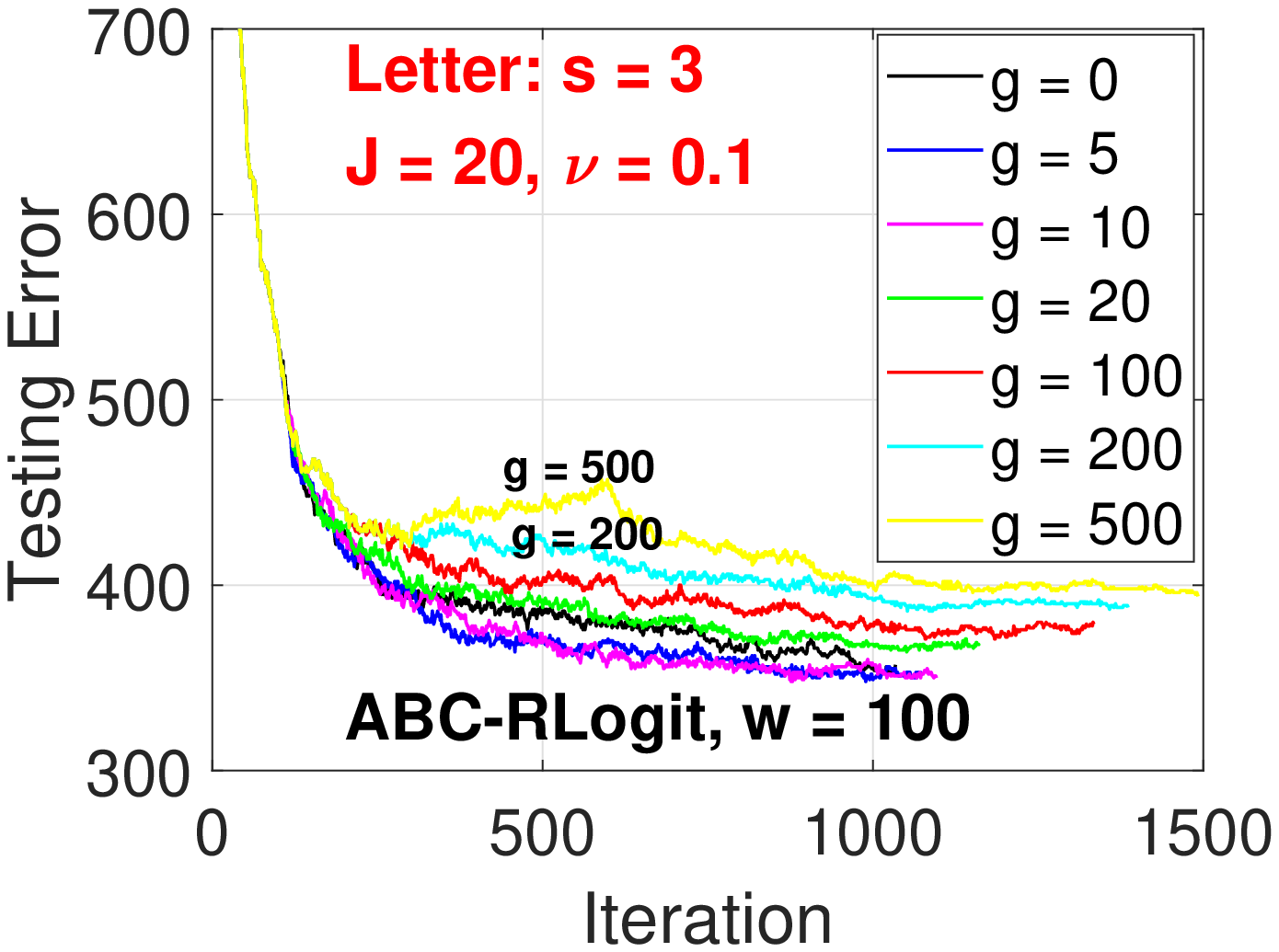}
}

\end{center}

\vspace{-0.1in}

\caption{{\em Letter} dataset. ABC-Boost with an initial ``warm-up'' stage for $w$ iterations, for $w=100$ (right), $w=10$ (middle), and $w=0$ (left). We can see that allowing a ``warm-up'' stage ($w>0$) provides another mechanism to avoid ``catastrophic failure'', even just with $w=10$. }\label{fig:Letter10k_sgw}
\end{figure}

Figure~\ref{fig:Letter10k_sgw} presents the experimental results on the {\em Letter} dataset, for $w=0$ (left column), $w=10$ (middle column), and $w=100$ (right column), and $s\in\{1,3\}$. One thing we can see if that this ``warm-up'' stage provides a mechanism to avoid ``catastrophic failures''. When $s=1$, $g=0$, $J=20$, $\nu=0.1$, using $w=0$ leads to ``catastrophic failure'' but using $w=10$ or $w=100$ successfully avoids the bad performance case.

\section{Conclusion}

\noindent It has been for more than 10 years since the initial development of ABC-Boost~\citep{Proc:ABC_ICML09,Proc:ABC_UAI10} and Robust LogitBoost~\citep{Proc:ABC_UAI10}. The idea of Robust LogitBoost, i.e., the tree-split gain formula using the second-order information~\eqref{eqn:logit_gain}, has been widely adopted by popular tree platforms. However, the idea of ABC-Boost for improving multi-class classification does not seem be broadly adopted. We think one of the main reasons might be due to the lack of an efficient and reliable strategy for choosing the base class. The ``exhaustive search'' strategy as published in~\citet{Proc:ABC_ICML09,Proc:ABC_UAI10} is too expensive, but the ``worst class'' strategy as in the original 2008 technical report~\citep{li2008adaptive} is usually less accurate than the ``exhaustive search'' strategy, and sometimes even exhibits ``catastrophic failures''.\\

\noindent In this paper, we propose a unified framework for Fast ABC-Boost by introducing three parameters: $s$, $g$, and $w$.  In the ``warm-up'' stage, we first conduct $w$ boosting iterations using MART or Robust LogitBoost and then shift to ABC-Boost starting at the $(w+1)$-th iteration. With ABC-Boost, we only search for the base class at every $g+1$ iterations. When we do need to search for the base class, we only search within the ``$s$-worst classes''. All those efforts aim at substantially reducing the computational complexity of ABC-Boost while maintaining the good accuracy. In our experiments, it appears that $s=2\sim 4$, $g=10\sim 20$, $w=10\sim 100$ might be a good initial choice of these parameters. They can be treated as  tunable parameters and it is not surprising that we might obtain even better accuracy at some combinations of $(s,g,w)$ than using the most expensive ``exhaustive search'' strategy (i.e., $s=K$, $g=0$, and $w=0$). \\

\noindent With properly chosen parameters, ABC-Boost can be computationally efficient. The training cost of ABC-Boost with parameters $(s,g,w)$ would be $O\left(Kw+s(K-1)\frac{1}{g+1}(M-w) + (K-1)\frac{g}{g+1}(M-w)\right)$, which is not necessarily much larger than $O(KM)$, the cost of MART or Robust LogitBoost. In fact, when $K$ is small (e.g., $K=3$ or $5$), the cost of ABC-Boost might be even smaller. Also, we have always assumed $M$ is the same for all boosting procedures. In our experiments, however, we observe that typically the training loss of ABC-Boost converges to zero noticeably faster.\\

\noindent In summary, for multi-class classification tasks using boosting and trees, we would recommend ABC-RobustLogitBoost with parameters initially chosen from $s=2\sim 4$, $g=10\sim 20$, and $w=10\sim 100$.

\newpage\clearpage

\bibliographystyle{plainnat}
\bibliography{standard}

\end{document}